\documentclass{article} %

\usepackage{etoolbox}       %
\newtoggle{arxiv}%
\toggletrue{arxiv}%
\iftoggle{arxiv}{
    \usepackage[parfill]{parskip}

    \usepackage[%
        a4paper,
        inner=25mm,
        outer=25mm,%
        top=25mm,
        bottom=25mm,
        marginparsep=5mm,
        marginparwidth=40mm,
    ]{geometry}

    \newlength{\defbaselineskip}
    \setlength{\defbaselineskip}{\baselineskip}
    \setlength{\parskip}{6pt}%
    \setlength{\parindent}{0pt}%

}{
    \usepackage{iclr2027_conference,times}
}

\usepackage{hyperref}
\usepackage{url}
\usepackage{natbib}

\usepackage[utf8]{inputenc} %
\usepackage[T1]{fontenc}    %

\usepackage[dvipsnames,table]{xcolor} %

\usepackage{graphicx}       %
\usepackage{booktabs}       %
\usepackage{multicol}       %
\usepackage{multirow}       %
\usepackage{makecell}       %
\usepackage{colortbl}       %
\usepackage{caption}        %
\usepackage{subcaption}     %
\usepackage{wrapfig}        %
\usepackage{adjustbox}      %
\usepackage{tikz}           %
\usepackage{titletoc}       %
\usepackage{enumitem}

\usepackage{mathtools}
\usepackage{amsmath,amsthm} %
\usepackage{amsfonts}       %
\usepackage{amssymb}        %
\usepackage{nicefrac}       %
\usepackage{microtype}      %
\usepackage{relsize}

\usepackage[capitalize,nameinlink]{cleveref}  %

\usepackage{comment}        %
\usepackage{array}

\usepackage{doi}            %
\usepackage[normalem]{ulem} %
\usepackage{placeins}       %
\usepackage[detect-all]{siunitx} %

\usepackage{bbding}         %
\definecolor{cold}{HTML}{008acd}
\def\frozen{\raisebox{-.15ex}{\color{cold}\SnowflakeChevron}}
\DeclareUnicodeCharacter{2744}{\frozen}%

\usepackage{pifont}         %
\def\cmark{\ding{51}} %
\def\xmark{\ding{55}}
\definecolor{vlightgray}{gray}{0.87}

\newcommand{\keptmark}{{\color{ForestGreen}\cmark}}   %
\newcommand{\partmark}{{\color{Goldenrod!60!black}\cmark}}     %
\newcommand{\lostmark}{{\color{BrickRed}\xmark}}      %
\def\correctmark{{\color{ForestGreen}\cmark{}}}

\def\incorrectmark{{\color{red}\xmark{}}}

\definecolor{tblue}{HTML}{1F77B4}
\definecolor{torange}{HTML}{FF7F0E}
\definecolor{tgreen}{HTML}{2CA02C}
\definecolor{tred}{HTML}{FF0000}

\definecolor{linkcolor}{HTML}{991408}  %
\definecolor{citecolor}{HTML}{2E7E2A}  %
\definecolor{filecolor}{HTML}{131877}  %
\definecolor{menucolor}{HTML}{727500}  %
\definecolor{runcolor} {HTML}{137776}  %
\definecolor{urlcolor} {HTML}{0a2bbf}  %
\definecolor{linkblue}{HTML}{0057B8}%

\newcommand{\appref}[1]{\hyperref[#1]{Appendix~\ref*{#1}}}%

\crefname{section}{Sec.}{Sections}
\crefname{appendix}{Appx.}{Appendices}
\renewcommand*{\appref}[1]{\hyperref[#1]{Appx.~\ref*{#1}}}

\def\Snospace~{\S{}}%

\newcommand{\mypara}[1]{\noindent\textbf{#1}}

\newcommand{\hreffoot}[2]{\href{#1}{#2}\footnote{\url{#1}}}

\newcommand{\best}[1]{\ifmmode\mathbf{#1}\else\textbf{#1}\fi}
\newcommand{\secbest}[1]{\underline{#1}}
\newcommand{\runnerup}[1]{\secbest{#1}}
\newcommand{\na}[1]{{\color{gray}N/A}}
\newcommand{\dup}[1]{\textcolor{ForestGreen}{#1}}
\newcommand{\ddown}[1]{\textcolor{BrickRed}{#1}}
\newcommand{\dzero}[1]{\textcolor{gray}{#1}}

\newcommand{\hider}[1]{}

\usepackage{amsmath,amsfonts,bm}

\def\eqref#1{equation~\ref{#1}}

\def\floor#1{\lfloor #1 \rfloor}
\def\1{\bm{1}}

\DeclareMathAlphabet{\mathsfit}{\encodingdefault}{\sfdefault}{m}{sl}
\SetMathAlphabet{\mathsfit}{bold}{\encodingdefault}{\sfdefault}{bx}{n}

\iftoggle{arxiv}{
    \hypersetup{colorlinks=true, allcolors=linkblue}
    \captionsetup{font=small,labelfont=it}

    \RequirePackage[T1]{fontenc}
    \RequirePackage[tt=false, type1=true]{libertine}
    \RequirePackage[varqu]{zi4}
    \RequirePackage[libertine]{newtxmath}

    \renewcommand{\And}{\and}

    \expandafter\patchcmd\csname @maketitle\endcsname{\large}{\normalsize}{}{%
        \PackageWarning{main}{Author-size patch to maketitle failed}}
}{}

\title{SpireBench: Evaluating Long-Horizon Strategic Play of Agentic~Models with \texorpdfstring{\emph{Slay the Spire}}{Slay the Spire}}

\author{%
\textbf{Scott C. Lowe}\\
Vector Institute\\
Toronto, Canada \\
\texttt{scott.lowe@vectorinstitute.ai} \\
\And
\textbf{Sander W. Keemink}\\
Department of Machine Learning and Neural Computing \\
Donders Institute for Brain, Cognition, and Behaviour \\
Radboud University \\
Nijmegen, Netherlands \\
\texttt{sander.keemink@donders.ru.nl}
}

\iftoggle{arxiv}{
    \usepackage{fancyhdr}
    \pagestyle{fancy}
    \AtBeginDocument{\edef\noheaderpagenum{\IfRefUndefinedExpandable{fig:learning}{0}{\getpagerefnumber{fig:learning}}}}
    \newcommand{\unlessnoheaderpage}[2]{\ifnum\value{page}=\noheaderpagenum\relax#2\else#1\fi}
    \fancyhead[L]{\unlessnoheaderpage{SpireBench}{}}
    \fancyhead[R]{\unlessnoheaderpage{Lowe, et al. (2026)}{}}
    \let\origheadrule\headrule
    \renewcommand{\headrule}{\unlessnoheaderpage{\origheadrule}{}}
}{}

\date{}

\usepackage{refcount} %

\begin{document}

\maketitle

\begin{abstract}
Agentic large language models now tackle complex tasks, yet strategic reasoning over long horizons is under-studied: most agentic benchmarks are short and episodic, and performance on long tasks is hard to dissect into stored knowledge, reasoning, and luck. We introduce SpireBench, in which models play the complete commercial roguelike deck-builder \emph{Slay the Spire}, where players build their future action space by drafting cards into a deck over the course of the game. Across 20 large language models (LLMs) at the game's highest difficulty, frontier models beat heuristic baselines, matching the average human player but far behind experts. This gap widens as the run lengthens: the strongest models win the first act as often as experts, then fall behind. What separates strong models from weak ones is their short-term decision making; all misjudge risk over the medium-term, preserving hit points rather than pursuing larger rewards that would let them grow stronger. Our puzzle suite shows the same in isolation: accuracy falls as a decision's payoff is further into the future, even when outcomes are certain. A self-consistent renaming of every game term shows that part of their performance rests on prior knowledge of the game, and models do not reliably improve from the memories they write across runs, which are often wrong and overly cautious. %
Our harness is the first to give a model the complete original game.
Overall, SpireBench offers an unsaturated long-horizon benchmark with tools to locate where and why agents fail.
\end{abstract}

\section{Introduction}

Large language model (LLM) progress has been rapid, with high scores on tests of a wide range of skills associated with general intelligence \citep{stanfordhai2026index}, and LLMs increasingly act as autonomous agents over extended periods \citep{brohi2025agentic}. However, single-step competence does not necessarily translate into reliable performance over many sequential decisions, a shortfall termed the \emph{horizon gap} \citep{chen2026horizongap}. Performance on long tasks is also hard to dissect: does a model rely on stored knowledge or on reasoning~\citep{Lowe2024_System2Nigh}, and can it store accurate long-term memories and use them effectively? How is it impacted by stochasticity in its environment?

Prior work answers parts of these separately: models fail to act on rules they can state~\citep{paglieri2024balrog}, recall the games they play~\citep{cui2025tales}, and gain from memory on short, checkable problems~\citep{suzgun2026dynamic}. But agentic benchmarks \citep{yehudai2025survey} are mostly short-horizon, game benchmarks rarely hand a model every decision a player makes, and reported memory gains \citep{shinn2023reflexion, zhao2024expel} come where the agent can test each lesson, against executed code or a per-attempt success signal. To solve these shortcomings, we introduce SpireBench, built on the commercial game \emph{Slay the Spire} (StS). A single run of StS gives one delayed, noisy outcome for hundreds of decisions; SpireBench assesses the model's ability to act and learn over multiple long-horizon runs.

\begin{figure}[t]
\centering
\includegraphics[width=0.95\linewidth]{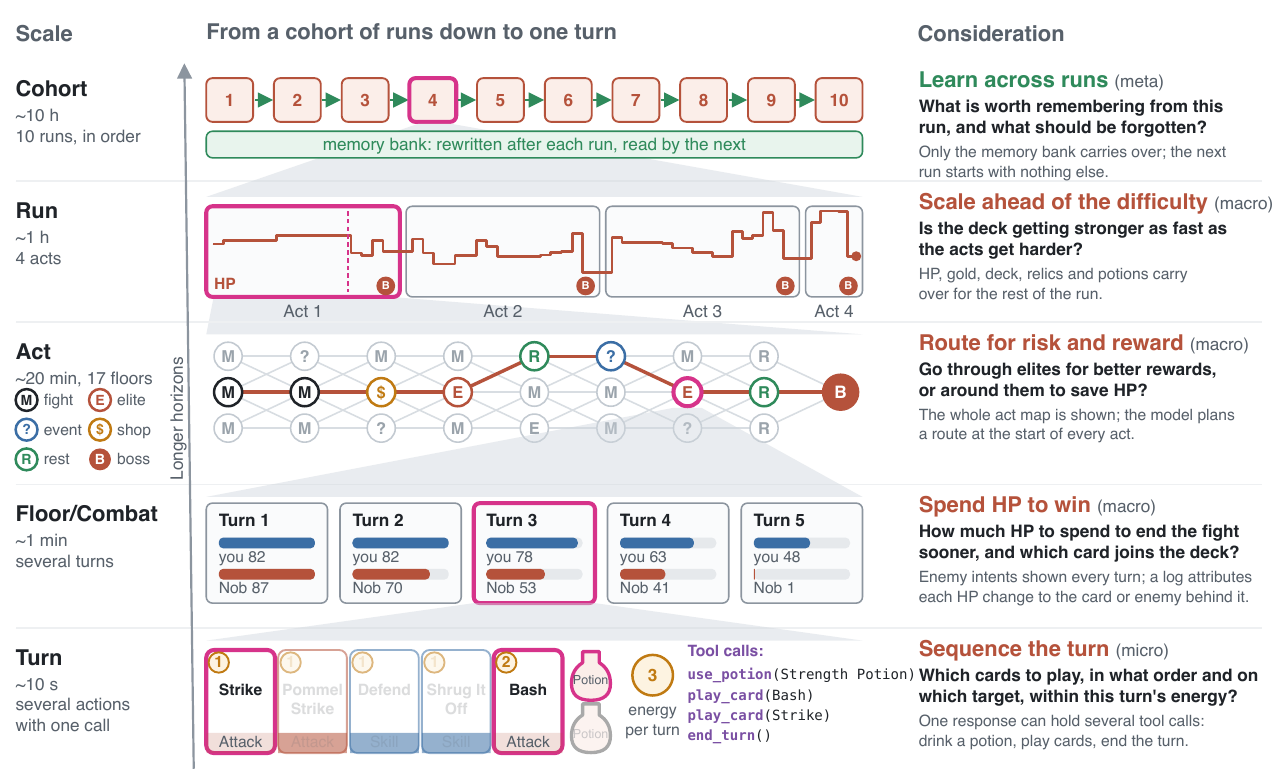}
\caption{\textbf{Multiple time horizons at play in \emph{Slay the Spire}.}
\textbf{Cohort:} the model plays a series of runs on different seeds, carrying memories it writes between them.
\textbf{Run:} four acts of escalating difficulty, each ending in a boss; the player must grow stronger fast enough to keep up.
\textbf{Act:} a map of 17 floors (Act~4: four floors) through which the player plans a route; some nodes are riskier, but offer larger rewards.
\textbf{Floor/Combat:} most floors are fights of several turns, each followed by a card reward that modifies the player's deck carried between floors. A generally strong card may not fit the current deck, which must be able to beat many types of encounter, so resolving its weaknesses against upcoming floors is important.
\textbf{Turn:} the player draws part of the deck; the model plans the turn and plays it in one response of several tool calls (purple). Sometimes the best play takes more damage now to end the fight sooner.
\definecolor{fighlight}{HTML}{D7388D}
\textbf{\color{fighlight}Magenta:} current point in the run.%
}
\label{fig:game-overview}
\vspace{-1em}
\end{figure}

StS combines tactical combat, deck construction, and route planning over 56 floors~(Fig.~\ref{fig:game-overview}). It is turn-based, fully describable in text, and its choices involve weighing short-term gains against long-term success at several time horizons, making it highly suitable for a long-horizon benchmark (\cref{app:why-sts}). 
Unlike previous StS benchmarks~\citep{bateni2024minists, orak}, SpireBench gives the model every aspect of the game, and separates knowledge from reasoning by reskinning the game. 

Our goal in this work is not to build the agent with the best performance at slaying the Spire, but to evaluate existing LLMs in a complex and challenging setting that demands long-horizon planning and reasoning with noisy signals. We extensively study this in one setting so we can hold the task fixed while varying the model, harness, memory mechanism, and the model's prior knowledge, and compare against human play on the same decisions.
Our contributions are as follows:
\begin{itemize}[nosep, leftmargin=*]

\item \textbf{A long-horizon reasoning benchmark calibrated against human data.}
We evaluate 20 LLMs at the game's highest difficulty, Ascension 20 (A20), and report their performance against cost and reasoning spent.
As human references we use telemetry released by the developers of \emph{Slay the Spire}, Mega Crit (462\,k runs), and the run histories of four experts who play professionally (Twitch streamers).
Frontier models beat rule-based and symbolic baselines and the average human player, yet remain far from expert play.
The benchmark tracks the Artificial Analysis Intelligence Index but is among the least predictable from other evaluations, suggesting it captures less measured aspects of intelligence.

\item \textbf{An efficient harness for the complete original \emph{Slay the Spire} game.} Agents play the unmodified commercial game through our harness, making every decision a human player makes (a first) and seeing the information a human player sees.
The model can return a whole turn's actions as multiple tool calls in one response, the harness reporting the effect of each action back to the model.
Compared with existing harnesses on the same models and seeds, it plays better and costs less (\cref{tab:harness-compare}). %

\item \textbf{Knowledge ablation with reskin.}
We rename every term in the game under one consistent theme, separating prior knowledge of the game from reasoning about its rules.
Hiding the real names costs performance, concentrated in a few models.

\item \textbf{Memory mechanisms do not reliably improve performance.}Models write notes within a run and carry memories between runs to try to improve future performance. Notes taken within a run raise acts-cleared by about half, but cross-run memories show mixed results. We investigate and find erroneous memories persist---rarely deleted even after being demonstrated to be incorrect (\cref{app:further-memory-banks}), and advice to be cautious outnumbers advice to take risks seven to one, reinforcing the poor long-term risk calibration of the models.

\item \textbf{Analysis of model understanding and failures.}
We ask the models to predict future game states to test their understanding of play: stronger models predict the near term better, but none predicts long-term outcomes.
Compared with human choices, even the best models are too conservative: they do not take the early risks needed to outpace the game's difficulty curve.
Puzzle-solving and agreement with decisions made by experts in their runs isolate these failures (\cref{app:puzzles,app:expert-decisions}).

\end{itemize}

\section{Background}

\emph{Slay the Spire} is a seminal roguelike deck-builder, in which players must balance decisions over several time horizons~(Fig.~\ref{fig:game-overview}). Each combat turn the player draws a hand from a deck of cards and
spends a limited pool of energy to attack, block, or build effects that scale over turns, while each enemy shows the action it will take next (its \emph{intent}); between fights the player adds cards to the deck,
sometimes removes them, and picks a route through a map whose nodes are fights, \emph{elite} fights (optional, harder fights that reward a \emph{relic}, a permanent bonus), events, shops, and rest sites (Fig.~\ref{fig:game-overview}, middle row). Decisions are taken under uncertainty and compound: a card once taken usually stays in the deck for the rest of the run, and threats met very late in the game must be prepared for early. Balancing risk
is central, since taking risks early yields the growth that the rising difficulty demands
(\cref{fig:game-overview}). Other long tasks that agents face share these properties (\cref{app:why-sts}).

\label{sec:related}

\mypara{\emph{Slay the Spire}.}
Analyses of the game measure its mix of luck and skill~\citep{andersson2023lucky,nam2023randomness} and the complexities of its map uncertainty~\citep{bazzaz2025paths}.
MiniStS~\citep{bateni2024minists} tests LLMs on 8 cards and 5 isolated encounters, so it covers single fights but not deck-building across a run.
The twelve-game benchmark \textsc{Orak}~\citep{orak} includes evaluation on \emph{Slay the Spire} (StS); its harness drives the real game, but its LLM plays only combat and card rewards, on the easiest difficulty (Ascension~0), on one seed, with a rule-based bot doing the rest.
Bottled AI~\citep{bottledai} is a hand-written symbolic player; it does not support every card or event, but reports high performance at the lowest difficulty.
Finally, the concurrent and most comparable work, AgenticSTS~\citep{cheng2026agenticsts}, plays complete runs of the sequel, \textit{Slay the Spire~2}. However, the sequel is still in early access with fortnightly balance patches, which makes it unsuitable as a benchmark, and AgenticSTS plays only at the lowest difficulty and was tested mainly with one model.
Overall, no existing evaluation has an LLM make every decision in the released game, tests it at the highest difficulty, studies more than a few models, compares against human play, or controls for the models' prior knowledge of the game. SpireBench does all five: 20 models make every decision at the highest difficulty (A20), and we compare them with extensive human telemetry and control for prior knowledge with a reskin. To our knowledge, ours is the first harness to give a model the complete original game.

\mypara{Games as benchmarks.} Games have a long history as benchmarks for AI. Deep reinforcement learning reached superhuman play in Atari, Go, chess,
and shogi~\citep{mnih2015dqn,silver2016alphago,silver2018alphazero} and hard-exploration
games~\citep{ecoffet2021goexplore} by training on the game. In contrast, LLM agents are usually evaluated untrained, such as in
NetHack, Crafter, and Minecraft~\citep{kuttler2020nethack,hafner2022crafter,wang2023voyager}, benchmark
suites~\citep{paglieri2024balrog,wu2024smartplay} and commercial games~\citep{zhang2025videogamebench}, and text adventures~\citep{cote2018textworld,hausknecht2020jericho,cui2025tales}. Overall, LLMs have mostly been tested on simplified states or partial games, and rarely with
learning across runs. BALROG~\citep{paglieri2024balrog} tests long-horizon play in a range of simplified games and in NetHack, a full game whose long-term decisions resemble those of StS, but which is so difficult that both humans and LLMs rarely win. Its authors report a \emph{knowing--doing gap}, in which models answer questions about the rules correctly but fail to act on them. SmartPlay~\citep{wu2024smartplay} isolates planning and reasoning about randomness, each in a separate short game, and TALES~\citep{cui2025tales} observes models recalling specific text adventures in their reasoning. Finally, similar deck-building games have been tested in BalatroBench~\citep{coder2026balatrobench} (shorter
horizons, and only the basic difficulty) and in HearthStone~\citep{garciasanchez2016hearthstone}, where whole decks are searched before play.
In contrast, \emph{Slay the Spire} tests planning, reasoning under randomness, and long-horizon play within one complete commercial game: each turn is an exact planning problem, rewards and draws are random, and choices compound over an hour-long run. A large human dataset at the same difficulty lets us calibrate the models against people, and demonstrates that expert players can reach high winrates (50+\%, \cref{tab:main-results}). Finally, the game's popularity lets us measure with a reskin, rather than only observe, how much play depends on knowledge of the game.

\mypara{Real-world and long-horizon benchmarks.} Tool-use, web, and agentic
benchmarks~\citep{jimenez2024swebench,mialon2024gaia,yao2024taubench,zhou2024webarena,koh2024visualwebarena,liu2024agentbench}
are mostly tens of steps long; others lengthen the horizon by task duration~\citep{kwa2025horizon},
execution length~\citep{sinha2026illusion}, or a simulated business over tens of millions of
tokens~\citep{backlund2025vendingbench}. SWE-bench~\citep{jimenez2024swebench} grounds agent evaluation in real GitHub issues, but each issue is an independent episode verified by unit tests, with nothing carried to the next. Finally, trading benchmarks~\citep{fintrade_li2025} span many decisions, but over a fixed historical price path with the same few actions each step. In contrast, a \emph{Slay the Spire} run couples hundreds of decisions through one pool of resources, and the agent builds its own later action space by building its deck.

\mypara{Memory across episodes.} Several earlier agents build persistent memories~\citep{park2023generative,packer2023memgpt}. Memory carried across episodes is generally reported to help~\citep{shinn2023reflexion,wang2023voyager,majumder2023clin,zhao2024expel,wang2024awm}, mostly where an external signal checks or adjusts what is stored. Where nothing checks it, models struggle to correct themselves~\citep{huang2024selfcorrect} and propagate the errors they store~\citep{xiong2025memory}. Closest to our work, Dynamic Cheatsheet~\citep{suzgun2026dynamic} shows that a self-curated memory can help without ground-truth labels, at times by a wide margin, on sequences of short problems whose answers the model can often verify itself (e.g.\ by storing a code solver for a simple arithmetic game). However, the problems tested were small and, as in that example, fully solvable. In SpireBench, a single delayed, noisy outcome follows hundreds of decisions, so no stored memory can be firmly verified within one run.

\mypara{Reskinning.} Reskinning and related interventions are common tools for dissecting model performance. Renaming entities while keeping the mechanics separates memorization from reasoning:
nonsense predicates~\citep{valmeekam2023planbench}, changed rules~\citep{wu2024counterfactual},
isomorphic worlds~\citep{gu2026synthworlds}, masked names~\citep{hu2025lmgame}, renamed security
tasks~\citep{honarvar2026ctf}, and re-textured game objects~\citep{dubey2018priors}; other benchmarks
generate environments a model cannot have seen~\citep{costarelli2024gamebench,li2025gvgaillm}. SpireBench improves on earlier reskins by covering the whole game under one consistent theme whose names evoke each
mechanic, so that each term keeps its meaning; random and permuted mappings serve as controls
(\cref{sec:methods-reskin}), paired by seed.

\section{Methods}
\label{sec:methods}

\providecommand{\spireicon}[1]{\raisebox{-0.25ex}{\includegraphics[height=1.7ex]{icons/#1.pdf}}}
\begin{table}[t!]
\centering
\caption{\textbf{Benchmark results.}
\emph{Slay the Spire}, Ascension~20 with the Ironclad character.
Each LLM is run at its highest reasoning setting (grey), played under our themed reskin, with note-taking across floors in a run and between runs in a cohort.
Act~$k$ win rate: share of runs that cleared Act~$k$.
Acts cleared: mean number of acts cleared per run.
Floor: final floor reached.
BottledAI: symbolic agent \citep{bottledai}.
Human players: public run histories of expert players (Twitch streamers who play \emph{Slay the Spire} professionally; see \cref{app:human-baselines} for details).
An author: non-professional player.
All humans: in-game telemetry (\cref{app:human-baselines}).
\best{Best} and \runnerup{second best} over LLMs.
$^{\dagger}$Self-hosted; priced at Artificial Analysis headline rates.
$^{\ddagger}$Peak-rate list price (upper bound).
}
\label{tab:main-results}
\begin{adjustbox}{max width=\linewidth}
\small
\begin{tabular}{@{}c@{\hspace{3pt}}lr rrrr r rrc r rr@{}}
\toprule
 & & & \multicolumn{4}{c}{Act win rate (\%)} & \multirow[b]{2}{*}{\makecell{Acts\\cleared}} & \multicolumn{3}{c}{Floor} & \multirow[b]{2}{*}{\makecell{Thinking\\(k/floor)}} & \multicolumn{2}{c}{Cost (\$)} \\
\cmidrule(lr){4-7} \cmidrule(lr){9-11} \cmidrule(lr){13-14}
 & Agent & Runs & 1 & 2 & 3 & 4 & & \multicolumn{1}{c}{mean} & \multicolumn{1}{c}{median} & \multicolumn{1}{c}{[min--max]} & & \multicolumn{1}{c}{/floor} & \multicolumn{1}{c}{/run} \\
\midrule
\multicolumn{14}{l}{\textit{Baseline policies}} \\
\spireicon{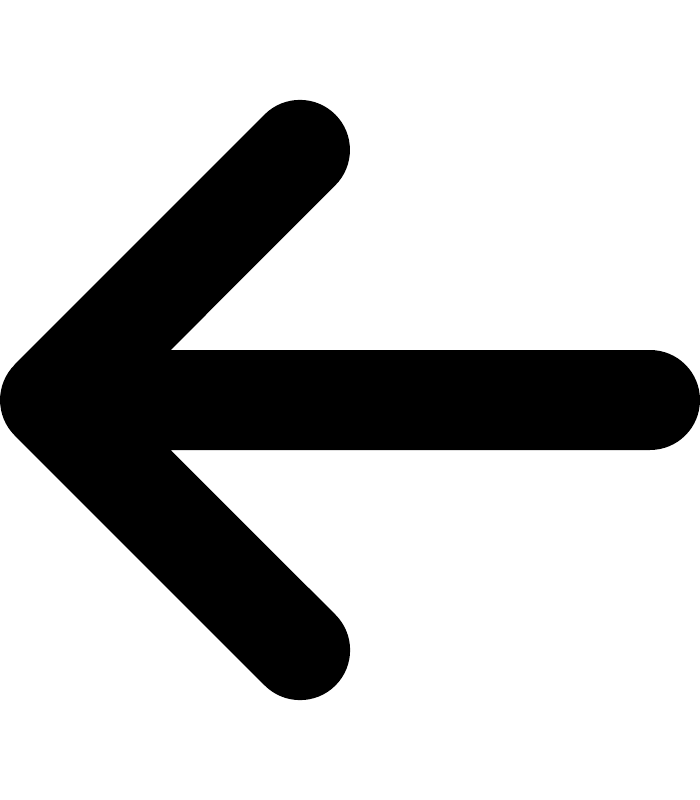} & Pick first option & 10 & 0 & 0 & 0 & 0 & 0.00 & 7.1 & 6.0 & [\phantom{0}1--16] & -- & -- & -- \\
\spireicon{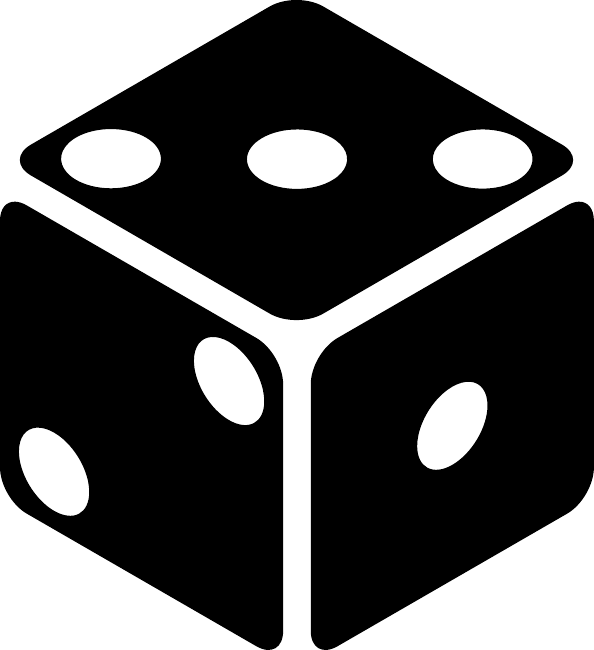} & Pick random option & 10 & 0 & 0 & 0 & 0 & 0.00 & 7.5 & 7.0 & [\phantom{0}4--14] & -- & -- & -- \\
\midrule
\multicolumn{14}{l}{\textit{Symbolic agent (Bottled AI)}} \\
\spireicon{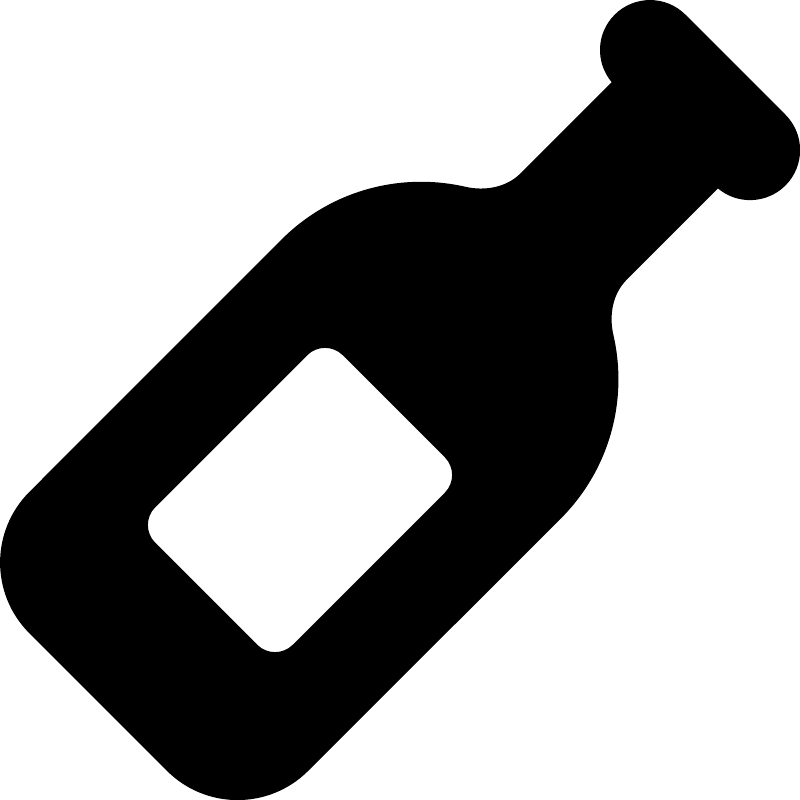} & BottledAI w/o Heart & 10 & 20 & 0 & 0 & 0 & 0.20 & 11.3 & 9.0 & [\phantom{0}5--25] & -- & -- & -- \\
\spireicon{bottled_ai} & BottledAI w/ Heart & 10 & 20 & 0 & 0 & 0 & 0.20 & 11.3 & 9.0 & [\phantom{0}5--25] & -- & -- & -- \\
\midrule
\multicolumn{14}{l}{\textit{All human players (Mega Crit telemetry, 2020; real game)}} \\
\spireicon{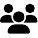} & All human players & 462{,}224 & 47 & 15 & 7 & 3 & 0.72 & 22.0 & 16.0 & [\phantom{0}1--57] & -- & -- & -- \\
\spireicon{crowd} & 1000+ hours played & 18{,}447 & 52 & 19 & 10 & 6 & 0.87 & 24.1 & 20.0 & [\phantom{0}1--57] & -- & -- & -- \\
\spireicon{crowd} & After prev.\ beat Act 4 & 7{,}490 & 58 & 29 & 20 & 16 & 1.23 & 27.9 & 23.0 & [\phantom{0}1--57] & -- & -- & -- \\
\midrule
\multicolumn{14}{l}{\textit{Human players: averages over a period (real game, own seeds)}} \\
\spireicon{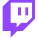} & Baalorlord {\scriptsize\textcolor{gray}{(2020--2026)}} & 687 & 89 & 77 & 70 & 63 & 2.98 & 47.7 & \best{57.0} & [\phantom{0}1--57] & -- & -- & -- \\
\spireicon{twitch} & panacea108 {\scriptsize\textcolor{gray}{(Sep--Nov 2022)}} & 51 & \best{92} & \best{78} & \best{71} & \best{65} & \best{3.06} & \best{48.3} & \best{57.0} & [\phantom{0}7--57] & -- & -- & -- \\
\spireicon{twitch} & vmService {\scriptsize\textcolor{gray}{(Sep--Dec 2022)}} & 50 & 88 & 74 & 64 & 58 & 2.84 & 46.8 & \best{57.0} & [\phantom{0}7--57] & -- & -- & -- \\
\spireicon{twitch} & cheg0 {\scriptsize\textcolor{gray}{(Feb 2024)}} & 46 & 85 & 65 & 56 & 46 & 2.52 & 43.1 & 56.0 & [\phantom{0}7--57] & -- & -- & -- \\
\spireicon{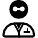} & An author {\scriptsize\textcolor{gray}{(all runs, 2020--2024)}} & 50 & 54 & 24 & 18 & 14 & 1.10 & 26.5 & 21.5 & [\phantom{0}6--57] & -- & -- & -- \\
\spireicon{scientist} & An author {\scriptsize\textcolor{gray}{(first 10 runs)}} & 10 & 50 & 20 & 20 & 10 & 1.00 & 24.4 & 20.0 & [\phantom{0}6--57] & -- & -- & -- \\
\midrule
\multicolumn{14}{l}{\textit{LLM agents (thematic reskin)}} \\
\spireicon{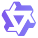} & Qwen3.8 27B {\scriptsize\textcolor{gray}{(xhigh)}} & 10 & 20 & 0 & 0 & 0 & 0.20 & 14.9 & 16.0 & [\phantom{0}5--29] & 18.1 & 0.091\rlap{$^{\dagger}$} & 1.45\rlap{$^{\dagger}$} \\
\spireicon{qwen} & Qwen3.6 35B-A3B & 10 & 0 & 0 & 0 & 0 & 0.00 & 10.2 & 8.0 & [\phantom{0}6--16] & 4.2 & 0.078\rlap{$^{\dagger}$} & 0.87\rlap{$^{\dagger}$} \\
\spireicon{qwen} & Qwen3.5 122B-A10B {\scriptsize\textcolor{gray}{(xhigh)}} & 10 & 0 & 0 & 0 & 0 & 0.00 & 9.8 & 9.0 & [\phantom{0}7--16] & \runnerup{3.7} & 0.072\rlap{$^{\dagger}$} & 0.77\rlap{$^{\dagger}$} \\
\spireicon{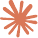} & Claude Fable 5.1 {\scriptsize\textcolor{gray}{(max)}} & 10 & \best{90} & \best{50} & \best{20} & 0 & \best{1.60} & \runnerup{36.2} & \runnerup{39.5} & [\phantom{0}7--52] & 17.7 & 1.664 & 61.91 \\
\spireicon{claude} & Claude Opus 5 {\scriptsize\textcolor{gray}{(max)}} & 10 & \best{90} & \runnerup{40} & 0 & 0 & 1.30 & 32.5 & 33.0 & [\phantom{0}7--51] & 21.3 & 1.743 & 60.62 \\
\spireicon{claude} & Claude Sonnet 5 {\scriptsize\textcolor{gray}{(max)}} & 10 & 40 & 10 & 0 & 0 & 0.50 & 19.3 & 16.0 & [\phantom{0}7--36] & 23.2 & 0.509 & 10.70 \\
\spireicon{claude} & Claude Haiku 4.5 {\scriptsize\textcolor{gray}{(12k)}} & 10 & 10 & 0 & 0 & 0 & 0.10 & 11.9 & 11.5 & [\phantom{0}4--22] & 7.6 & 0.094 & 1.23 \\
\spireicon{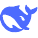} & DeepSeek V4.1 Flash {\scriptsize\textcolor{gray}{(max)}} & 10 & 30 & 0 & 0 & 0 & 0.30 & 16.9 & 16.0 & [\phantom{0}8--22] & 11.7 & 0.019\rlap{$^{\ddagger}$} & 0.35\rlap{$^{\ddagger}$} \\
\spireicon{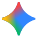} & Gemini 3.8 Flash {\scriptsize\textcolor{gray}{(high)}} & 10 & 50 & 10 & 0 & 0 & 0.60 & 22.9 & 18.0 & [\best{16}--47] & 13.7 & 0.117 & 2.89 \\
\spireicon{gemini} & Gemma 4 E4B & 10 & 0 & 0 & 0 & 0 & 0.00 & 7.7 & 6.5 & [\phantom{0}1--16] & 5.3 & \best{0.004}\rlap{$^{\dagger}$} & \best{0.03}\rlap{$^{\dagger}$} \\
\spireicon{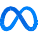} & Muse Spark 1.3 {\scriptsize\textcolor{gray}{(xhigh)}} & 10 & 60 & 10 & 0 & 0 & 0.70 & 21.9 & 22.5 & [\phantom{0}5--44] & 12.9 & 0.132 & 2.93 \\
\spireicon{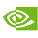} & Nemotron 3 Nano Omni & 10 & 0 & 0 & 0 & 0 & 0.00 & 6.8 & 6.0 & [\phantom{0}2--12] & 34.0 & 0.072\rlap{$^{\dagger}$} & 0.54\rlap{$^{\dagger}$} \\
\spireicon{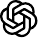} & GPT-6 Astra {\scriptsize\textcolor{gray}{(max)}} & 10 & \best{90} & \best{50} & \runnerup{10} & 0 & \runnerup{1.50} & \best{37.0} & \best{40.0} & [\phantom{0}5--\runnerup{55}] & \best{3.3} & 0.834 & 33.09 \\
\spireicon{openai} & GPT-5.6 Sol {\scriptsize\textcolor{gray}{(max)}} & 10 & \runnerup{70} & 20 & \runnerup{10} & \best{10} & 1.10 & 27.8 & 26.0 & [\phantom{0}7--\best{57}] & 4.0 & 0.272 & 8.39 \\
\spireicon{openai} & GPT-5.6 Terra {\scriptsize\textcolor{gray}{(max)}} & 10 & 50 & 0 & 0 & 0 & 0.50 & 18.2 & 18.0 & [\runnerup{10}--25] & 5.0 & 0.149 & 2.91 \\
\spireicon{openai} & GPT-5.6 Luna {\scriptsize\textcolor{gray}{(max)}} & 10 & 10 & 0 & 0 & 0 & 0.10 & 14.2 & 16.0 & [\phantom{0}7--24] & 5.0 & \runnerup{0.015} & \runnerup{0.24} \\
\spireicon{openai} & gpt-oss-120b {\scriptsize\textcolor{gray}{(high)}} & 10 & 10 & 0 & 0 & 0 & 0.10 & 10.4 & 8.0 & [\phantom{0}5--21] & 17.0 & 0.085\rlap{$^{\dagger}$} & 0.91\rlap{$^{\dagger}$} \\
\spireicon{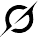} & Grok 4.6 {\scriptsize\textcolor{gray}{(xhigh)}} & 10 & 30 & 0 & 0 & 0 & 0.30 & 15.4 & 16.0 & [\phantom{0}5--28] & 11.7 & 0.346 & 5.68 \\
\spireicon{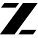} & GLM-5.3 {\scriptsize\textcolor{gray}{(max)}} & 10 & 60 & 0 & 0 & 0 & 0.60 & 22.1 & 22.0 & [\phantom{0}5--33] & 13.4 & 0.167 & 3.92 \\
\spireicon{zai} & GLM-5.3 Flash {\scriptsize\textcolor{gray}{(max)}} & 10 & 40 & 0 & 0 & 0 & 0.40 & 15.6 & 16.0 & [\phantom{0}5--27] & 14.4 & 0.024 & 0.40 \\
\bottomrule
\end{tabular}
\end{adjustbox}
\end{table}

\mypara{Agent harness.}%
\label{sec:methods-harness}
Agents play the released game through \texttt{CommunicationModPlus}, our fork of
CommunicationMod~\citep{communicationmod}, which exchanges JSON states and text commands and adds
snapshots around each enemy's turn (\cref{app:methods-game}). %
The harness holds one conversation per floor, shared by a combat, its rewards, and the post-floor
reflection. See \cref{fig:harness,app:harness-sequence} in the appendix for an overview and example floor.

A cached system prompt holds the task instructions, a compendium of the game's rules and items, and the memory bank
(\cref{sec:methods-memory}); each floor adds the act map, the run's notes, and the player's state. The
model acts only through tool calls, interleaving reasoning with action~\citep{yao2023react}, and may
return a whole turn's actions in one response (\emph{multi-action}); they run in order, and each game state change is returned as a tool response to the call that caused it.

Multi-action asks the model to plan the turn before acting and needs fewer calls: a turn costs one
response and one pass over the context, whereas one action per call pays for both again at every card.
It is also harder to use, because the actions available can change during a turn: the model must anticipate these changes to make valid tool calls, and the harness interrupts the queued actions when something unpredictable happens.
MiniStS makes one call per action~\citep{bateni2024minists}; Orak's agent may return
several actions, but it spends three calls on each decision (reflection, planning, action), shows the model only the
state after all of them, re-plans only once all have been played, and leaves every card selection that arises mid-fight to its rule bot~\citep{orak}.
\Cref{tab:harness-compare} compares the designs and
measures ours against corresponding variants and Orak on the same models and seeds.
We designed the prompts to show the model what a human player sees, and no more: the system prompt holds only the game's own tutorial, tips, and compendium, and the model is given no strategic advice (\cref{app:methods-harness}).
To estimate the model's calibration, at the end of each floor's conversation we prompt the model to forecast its chances of surviving the act and winning the run, and ask it to predict its upcoming hit points (HP) when it ends a turn.
Reasoning is kept across turns wherever the provider allows, and no response reached the output ceiling (\cref{app:methods-harness}).

\begin{table}[t]
\centering
\caption{\textbf{Comparison of \emph{Slay the Spire} harness designs.} %
\emph{Actions/call}: how many game actions one model response may issue; \emph{Effects}: whether each action's result is returned to the model;
\emph{Context}: what the model sees of earlier decisions; \emph{Choices}: what resolves card choices opened mid-fight (e.g. select a card to put in your hand, on top of your draw pile, or upgrade). Measured rows are paired with our harness on the same
models and seeds %
without the reskin or a cross-run memory bank for consistency with Orak which does not support either.
Calls are per floor on floors both arms played; cost is the run's API cost per floor it reached, priced as in \cref{tab:main-results}.
$^{\dagger}$Orak keeps no per-floor record: calls and cost per floor reached over whole runs, not floor-matched.
Further comparisons in \cref{tab:ablation-harness-by-model}.}
\label{tab:harness-compare}
\begin{adjustbox}{max width=\linewidth}
\small
\begin{tabular}{@{}l l l l l l rrrrr@{}}
\toprule
 & & & & & & \multicolumn{2}{c}{Per floor} & \multirow[b]{2}{*}{\makecell{Acts\\cleared}} & \multirow[b]{2}{*}{\makecell{Mean\\floor}} & \multirow[b]{2}{*}{\makecell{Tool errors\\/100 calls}} \\
\cmidrule(lr){7-8}
Harness & Model plays & Actions/call & Effects & Context & Choices & Calls & Cost (\$) & & & \\
\midrule
\textbf{Ours, multiaction} & whole game & a turn & per action & floor & model & 8.9 & 0.057 & 0.20 & 13.1 & 0.8 \\
\quad single action & whole game & one & per action & floor & model & \ddown{$+3.0$} & \ddown{$+23$\%} & \ddown{$-0.07$} & \ddown{$-1.5$} & \ddown{$+0.4$} \\
\quad single action, by index & whole game & one & per action & floor & model & \ddown{$+2.9$} & \ddown{$+24$\%} & \dzero{$\pm0.00$} & \ddown{$-0.3$} & \ddown{$+0.1$} \\
\quad continuous & whole game & a turn & per action & run & model & \dup{$-0.4$} & \ddown{$+22$\%} & \ddown{$-0.06$} & \ddown{$-1.2$} & \ddown{$+1.4$} \\
Orak~{\smaller\hypersetup{pdfborder={0 0 0}}\citep{orak}} & combat, card rewards & a turn, 3 calls & end state & reflection & rule bot & \ddown{$+3.3$}\rlap{$^{\dagger}$} & \dup{$-31$\%}\rlap{$^{\dagger}$} & \ddown{$-0.15$} & \ddown{$-3.3$} & -- \\ %
\bottomrule
\end{tabular}
\end{adjustbox}
\end{table}

\mypara{Notes and memory.}%
\label{sec:methods-memory}
After each floor the model may write, overwrite, or delete free-form \emph{run notes} under keys of its
choosing. After each run it consolidates them into its persistent \emph{memory bank} one operation at a time; the memory bank enters every later run's system prompt, so the runs of a cohort that writes to its bank form a chain,
played in seed order.
Ablations test what each part of this scheme contributes
(\cref{app:methods-memory}).

\mypara{Knowledge reskin.}%
\label{sec:methods-reskin}
Unless stated otherwise every run is reskinned: each game-specific name the model can see is replaced
by a name from a human-written, internally consistent mapping themed around tabletop RPGs.
This lets us evaluate the model's ability to reason about the game, removing as far as possible the rules of thumb for playing StS specifically that it may have seen in pretraining.
The model's own text is not renamed, so a real game name in its writing or reasoning is evidence that it recognizes the base game.
Ablations compare this themed renaming with the original names and with a randomized vocabulary (\cref{app:methods-reskin}).

\mypara{Models and reasoning settings.}%
\label{sec:methods-models}
We evaluate hosted models from Anthropic, DeepSeek, Google, Meta, OpenAI, xAI, and Z.ai, and
open-weight Qwen, Nemotron, and gpt-oss models on our own hardware. In \cref{tab:main-results}, each runs at its highest reasoning setting (grey, beside its name) (\cref{app:methods-models}).

\mypara{Evaluation protocol.}%
\label{sec:methods-protocol}
\label{sec:methods-metrics}
Every main result is on the Ironclad (the first of the game's four playable characters) at Ascension~20, the hardest difficulty, on the same ten seeds.
The seeds are new, both so that nothing could be tuned against the model/seed pairs of our preliminary experiments and so that model providers cannot have trained on our earlier results on them~\citep{sainz2023contamination}.
A \emph{cohort} is one model under one configuration on these seeds; each table cell is one cohort,
validated by the analysis code (\cref{app:methods-cohorts}).
We score a run by the number of acts it clears and also report the floor it reaches.
A run clears Acts~1 and~2 by beating their randomly selected bosses, Act~3 by beating two random bosses, and Act~4, which is reached only by collecting three keys during Acts~1--3, by killing the final boss, the Corrupt Heart.
The \emph{Act~$k$ win rate} is the share of runs that cleared Act~$k$; mean \emph{acts cleared} (0--4) is our headline summary.
\emph{Floor} is the final floor reached. %
A frontier cohort can cost hundreds of dollars (\cref{tab:budget}), so the choice of ten seeds is set by budget: too few to rank neighbouring models reliably, but enough for broader trends.

\mypara{Baselines and human references.}%
\label{sec:methods-humans}
Three non-LLM policies play the same seeds: \emph{pick first option} and
\emph{pick random option} are lower bounds for the models; \emph{BottledAI}~\citep{bottledai} is a
hand-written symbolic player (\cref{sec:methods-baselines}).
Human references at the same difficulty setting, scored alike, are taken from 462\,k finished runs of the game's telemetry, released by StS's developer Mega Crit, and 834 runs of four expert players (Twitch streamers; \cref{tab:main-results}).
The human runs are not seed-matched with the model's runs, providing an on-expectation reference point (\cref{app:human-baselines}).

\mypara{Diagnostic benchmarks.} Using the same harness, we also play through isolated decision points: a puzzle suite of 123 authored one-room puzzles with a known best answer (\cref{app:puzzles}), and decisions from the expert players' winning runs, scored by agreement with the expert's choice (\cref{app:expert-decisions}).

\section{Results}
\label{sec:results}

\begin{figure}[t]
\centering
\begin{subfigure}[t]{0.49\linewidth}
\centering
\includegraphics[width=\linewidth]{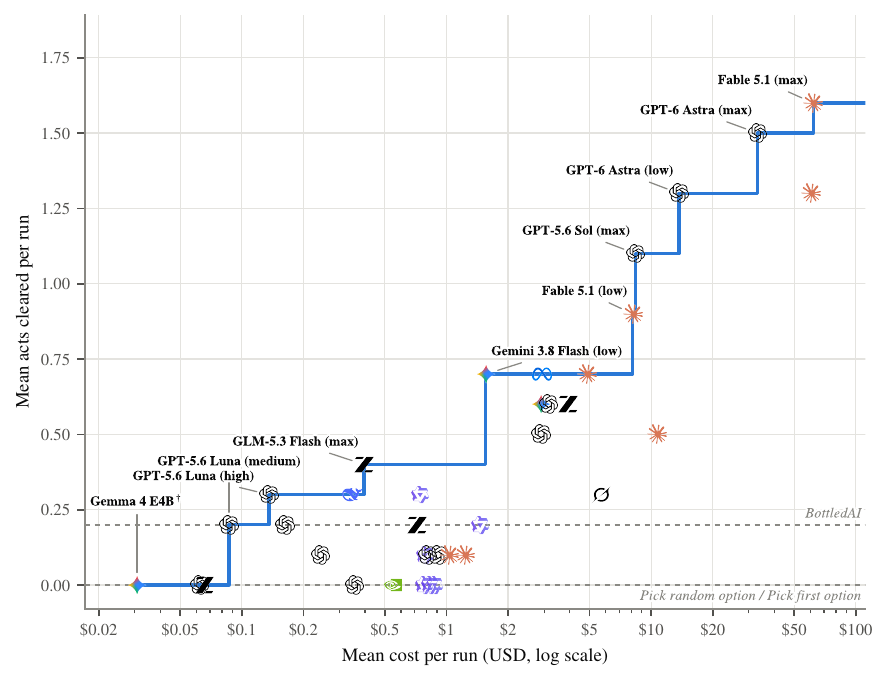}
\caption{Acts cleared against cost per run}
\label{fig:pareto}
\end{subfigure}\hfill
\begin{subfigure}[t]{0.49\linewidth}
\centering
\includegraphics[width=\linewidth]{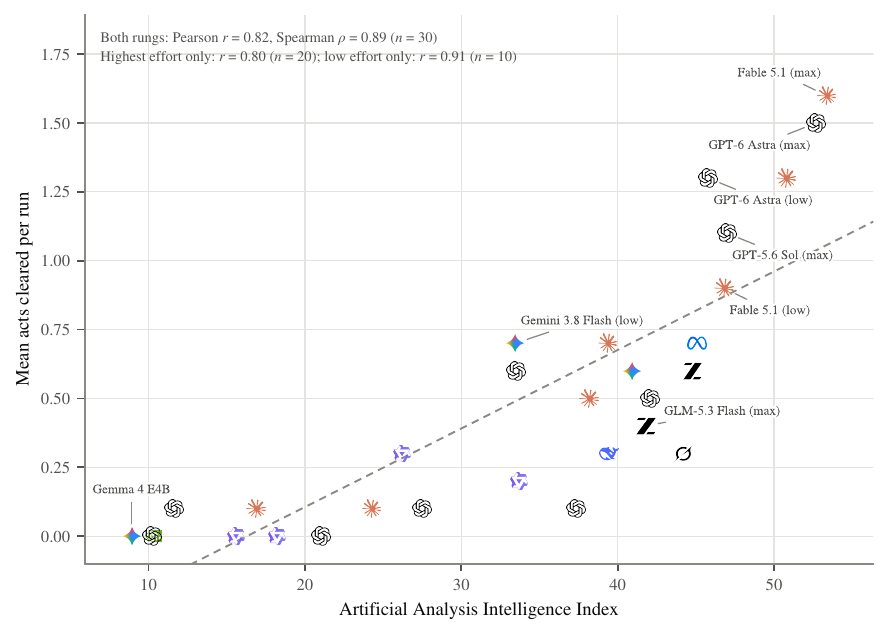}
\caption{Acts cleared against the Intelligence Index}
\label{fig:benchmarks}
\end{subfigure}
\caption{\textbf{Mean acts cleared per run} for the LLM agents of \cref{tab:main-results}.
(a)~Against mean dollar cost per run (log scale), across reasoning-effort levels.
Only points on the Pareto front (blue step line) are named; \cref{fig:pareto-full} shows all names.
Dashed lines: non-LLM baselines.
(b)~Against the Artificial Analysis Intelligence Index,
with each model at its highest effort and at low effort (labelled \emph{low}).
Dashed line: least-squares fit. Only models on the cost Pareto front of (a) are named.
For correlations with each Artificial Analysis evaluation, see \cref{tab:benchmarks}.}
\label{fig:pareto-benchmarks}
\end{figure}

\mypara{Performance and cost.} Of the 20 LLMs in \cref{tab:main-results}, the strongest
are Claude Fable~5.1, GPT-6 Astra, Claude Opus~5, and GPT-5.6 Sol. Only Sol completes the game, killing the final boss (the Corrupt Heart) in one run.
These four are also among the most expensive, per run and per floor (\cref{fig:pareto}). Mid-tier models outperform Bottled AI but not the
average human, and four models never clear an act, doing no better than the random-choice policy. Performance tracks
the Artificial Analysis Intelligence Index (\cref{fig:benchmarks}), but not the amount of reasoning:
GPT-6 Astra reasons least per floor (3.3k tokens) and clears 1.50 acts, while Nemotron~3 Nano Omni reasons most
(34.0k) and clears none.

\begin{figure}[t]
\centering
\includegraphics[width=\iftoggle{arxiv}{0.95}{0.85}\linewidth]{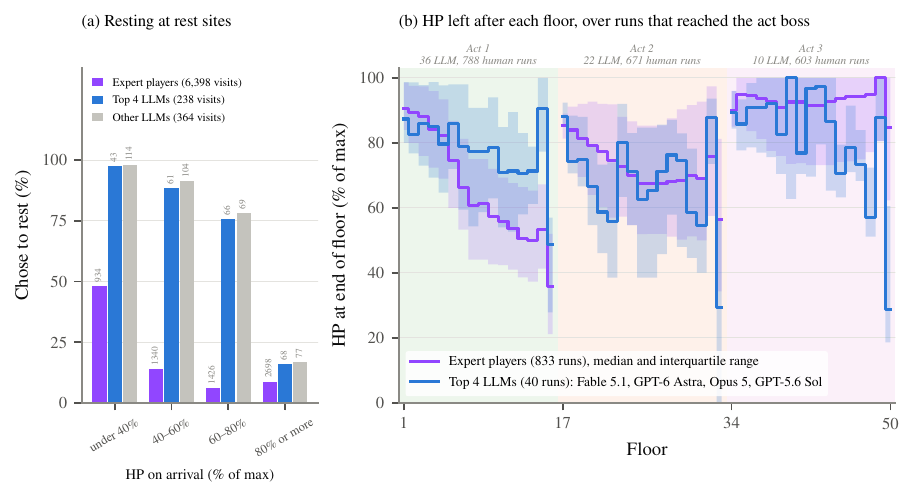}
\caption{\textbf{Do models use HP as a resource?} %
(a) At a rest site a player either rests, which heals, or upgrades a card. The top four LLMs (blue) and the other LLMs (grey) nearly always rest when they are missing any HP, whereas expert players (Twitch streamers; magenta) upgrade far more often.
Bars are labelled with their number of visits.
(b) Relative HP left at the end of each floor.
Each act is drawn over runs that reached the act boss (runs that die to the boss count as 0 HP for the boss floor); run counts are above each act. Lines: medians; bands: interquartile range.
Expert players spend HP in Act~1 to grow stronger faster than the game grows harder, and so take less damage on later floors.
}
\label{fig:hp-resource}
\end{figure}

\begin{figure}[t]
\centering
\includegraphics[width=\iftoggle{arxiv}{0.9}{0.78}\linewidth]{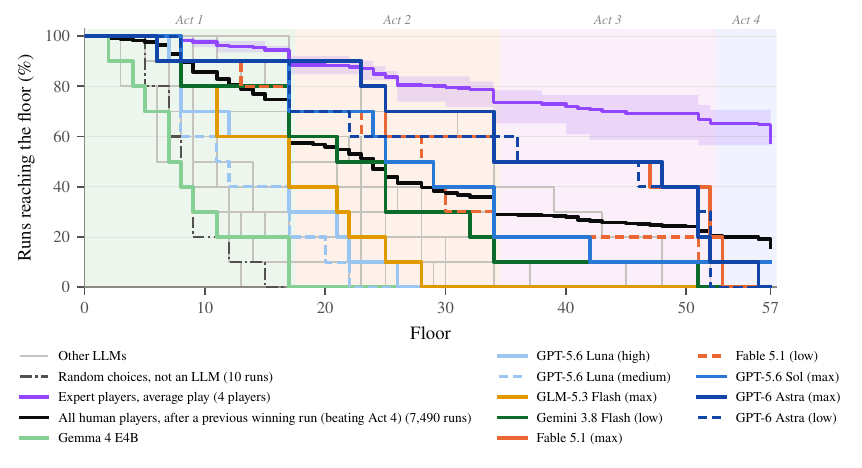}
\caption{\textbf{Survival over a run.} The share of each agent's runs that reached each floor. Every LLM agent in
\cref{tab:main-results} is shown; models on the cost Pareto front (\cref{fig:pareto}) are
coloured and named (solid: max effort; dashed: lower); the rest are grey.
Purple: mean of four expert players (band spans the four, weighted equally).
Black: the 7,490 A20 Ironclad telemetry runs that follow a previous winning A20 run (beating Act 4)
(\cref{app:human-baselines}). Shading: acts.}
\label{fig:survival}
\end{figure}

\mypara{Comparison with human players.} The strongest models clear more acts than the average
Ascension~20 Ironclad run in Mega Crit's telemetry (0.72) and than players with over 1000 hours
(0.87), but far fewer than the four experts (2.5--3.1 acts).
The gap opens as the run lengthens
(\cref{tab:main-results,fig:survival}): the three strongest models clear Act~1 in 90\% of their
runs, as the experts do (85--92\%), but Act~2 in 40--50\% against the experts' 65--78\%, and Act~3 in 0--20\%
against 56--71\%.
Their win rate for the full game (Act~4) is accordingly \emph{far} lower: 0\% against 46--65\%.

\mypara{How the models play.} The models play Act~1 cautiously. For elite enemies---harder floors which offer greater rewards---they path through 0.9--1.4 against an average of 2.0 for the telemetry population and 2.6 for the experts.
Rest sites offer a chance to \emph{either} heal for 30\% of max HP, or to upgrade a card---a boon which lasts the rest of the run---yet even when they arrive with at least 70\% HP, LLMs rest 31\% of the time (with some of the heal wasted).
The telemetry population over-heals at 2\% and the experts never.
An upgraded card can save HP in every fight throughout the rest of the run (blocking for more or killing enemies faster) so an upgrade is typically worth more HP in the long term than the heal would be.
Overall, the models heal at 57--75\% of rest sites, against 23\% and 9\%.
At the Act~1 boss the top four models hold a median 90\% of their maximum HP and the experts 53\% (\cref{fig:hp-resource}; \cref{app:deck-choices}).
But the transition to the next act restores 75\% of missing HP, so HP carried to the end of an act keeps only a quarter
of its value; the scarce resource is the \textit{floors} on which to outgrow enemies that scale as the game
progresses. Human runs that play Act~1 as the models do clear Act~2 in 15\% of cases, against 38\% for runs that
fight more elites and rest less (\cref{tab:act1-profile}), and in the models' own runs upgrading
rather than resting predicts clearing Act~2 ($r = 0.56$, 49 runs). Experience does not change this. Over a chain of
runs that share a memory bank, rest and elite rates change no more than in runs without memory; the banks advise caution seven times as often as
risk.
The four strongest models record the end-of-act heal correctly, but
mostly rest anyway (\cref{app:traces-rest}).
The Act~1 route explains less than a fifth of the gap to the experts (\cref{app:play-act1}).
Experts also lose less HP per fight, even from the first floors of the run, suggesting suboptimal decisions within a fight even for top models (\cref{fig:floor-fights}). %

\mypara{Strong models fight better.} Strong and weak models differ little in the route they take (\cref{app:deck-choices}), somewhat in the deck they build, and most in how they plan within a fight.
Stronger models build more as experts do: on the expert-decision benchmark the top four make the expert's pick on 45--70\% of card rewards, against 29--57\% for the seven bottom-tier models tested (\cref{tab:expert-decisions}), and in their own runs they take area-damage cards more often (46\% of Act~1 offers against 18\%; $\rho = 0.79$).
The top four
take 8.2 damage per ordinary Act~1 fight, the bottom tier 14.3 ($\rho = -0.89$; \cref{tab:combat-micro}).
The strong models turn energy
into damage (7.5 against 5.8 per energy spent on attacks, $\rho = 0.91$), rarely block on a turn when no enemy
attacks (6\% against 37\%), never miss a kill that the cards in hand allow (0\% against 11.7\%), and misstate rules far less often in their reasoning (6--11\% against up to 66\%; \cref{app:traces-facts}). The difference holds on the same encounter of the same seed, including the first six floors, before the decks diverge, and about three quarters of the top--bottom gap in Act~1 boss wins is not
explained by the state in which the models arrive.
With the hand and the enemy held fixed, the same holds: in 123 one-room puzzles with a known best answer (\cref{app:puzzles}), accuracy on those decided within one turn or one fight runs from 12--14\% for the weakest agent to 96--99\% for the strongest, and orders the agents as their runs do ($\rho = 0.95$ and $0.96$, with mean floor; \cref{app:puzzles-runs}).
The act boss, a long fight that neither benchmark contains, exposes an asymmetry (\cref{app:play-deck}): Gemini~3.8 Flash agrees with the experts' deck-building choices more than the Claude models do (\cref{tab:expert-decisions}), which it might have learned from its training data, but cannot pilot the deck it builds, beating the Act~1 boss in five of ten runs; Fable~5.1 and Opus~5 deviate from the experts' deck building but can use the decks they create, beating it in all nine they reach.

\mypara{Single decisions fail at the same horizon.} What separates the models is not what limits them all. In the same puzzles, accuracy falls with the time horizon of the decision for 17 of the 20 agents tested: 72\% on average where the answer is one turn, 62\% where it is a plan across a fight, and 48\% where the decision pays after the fight (\cref{tab:puzzle-horizons}).
The models under-invest even when the delayed payoff can be computed: in a puzzle where waiting three turns is certain to leave the player with more HP, 37 of 40 plays kill by the second turn.
They do so against human play too: shown the states of expert players' winning runs (\cref{app:expert-decisions}), they rest at 22--81\% of the rest sites where the expert instead upgraded a card.

\mypara{How the models reason.} In the reasoning of 16 models (\cref{app:traces}), what the top four still get wrong is knowledge built over many runs, such as what an enemy will do next (15\% of their statements about it).
Their caution is deliberate: Fable~5.1 and Opus~5 project their HP into a named upcoming fight at most rest sites and still rest; a model declining a risk usually cites its current HP or a remembered lesson, almost never the prompt's tips.
More reasoning is not better play: Fable~5.1 and Opus~5 make over half of all unprompted forecasts and go no further than Astra and Sol.
Nor do the models diagnose a loss well: they blame the final fight in 87 of 157 defeats, where an annotating model reading the same record does so in 27, and 67 of their accounts contradict the run's own history.

\mypara{Relation to public LLM benchmarks.} Across 30 model and effort points, mean acts cleared tracks the
Artificial Analysis Intelligence Index closely (Pearson $r{=}0.82$, Spearman $\rho{=}0.89$;
\cref{fig:benchmarks}),
yet SpireBench is one of the evaluations least
predictable from the others (\cref{tab:benchmarks}): for a held-out model they predict $50\%$ of
its variance, against $85\%$ of that of Humanity's Last Exam~\citep{phan2025hle}. %

\mypara{The harness moves scores as much as neighbouring models differ.} %
On the same models and seeds (\cref{tab:harness-compare}), Orak's
harness, whose rule bot plays every non-combat floor, reaches $3.3$ fewer floors than ours and clears $0.15$
fewer acts, a gap as large as that between many neighbouring models. %
The comparison uses eight mostly weaker models; the effect on frontier models is untested. %
Within our harness, allowing only one action per
response needs $3.0$ more calls per floor, costs $23\%$ more, and loses $0.07$ acts.
Keeping one conversation
for the whole run costs $22\%$ more per floor, loses $0.06$ acts, and raises tool-call errors from $0.8$ to $2.2$
per 100 calls~\citep{liu2024lost,sinha2026illusion}.

\mypara{Memory does not clearly compound.} %
Writing notes \emph{within} a run raises acts cleared by about half (0.19 to 0.28 per run), but carrying them \emph{across} runs does not: neither consolidating them into a memory bank nor rewriting them as one compact summary raises acts cleared at any point in the chain (\cref{fig:conditions,tab:ablation-memory,fig:memory-trend}), and carrying them forward verbatim raises them only slightly ($+0.03$ acts).
The banks are used, and more as they grow (\cref{app:traces-memory}), but nothing checks what is stored: with one delayed outcome per run and losses misdiagnosed (above), 17\% of the final memories are false, and even rules a model worked out itself are later misstated.
The errors stay (only 5 of the 20 models delete any memory, 30 in all) and cost play: models whose banks hold more
errors of their own clear fewer acts (Spearman $\rho = -0.67$, $p = 0.002$;
\cref{tab:memory-classification}).
Yet what a bank gets right pays once all of it is available: a finished bank, frozen and loaded on ten unseen seeds, \emph{does} help, adding about two
floors and up to nine points of Act~1 clears, on a par with removing the reskin, whoever wrote it ($p = 0.06$--$0.12$ on floors).
Finally, outside advice is not what was missing: an expert human-authored strategy guide \textit{costs} $1.9$ floors ($p = 0.016$; \cref{app:ablation-prompt}).

\providecommand{\spireicon}[1]{\raisebox{-0.25ex}{\includegraphics[height=1.7ex]{icons/#1.pdf}}}
\begin{table}[t]
\centering
\begin{minipage}[c]{0.25\linewidth}
\small
\caption{%
Real game names against our themed reskin. $\Delta$:~themed minus real. The other three reskin mappings are shown in \cref{tab:ablation-reskin}.}
\label{tab:reskin-compact}
\end{minipage}\hfill
\begin{minipage}[c]{0.72\linewidth}
\centering
\footnotesize
\begin{adjustbox}{max width=\linewidth}
\begin{tabular}{@{}c@{\hspace{3pt}}l rrr rrr@{}}
\toprule
 & & \multicolumn{3}{c}{Act~1 win rate (\%)} & \multicolumn{3}{c}{Mean floor reached} \\
\cmidrule(lr){3-5} \cmidrule(l){6-8}
 & Agent & Real & \!Reskin & $\Delta$ & Real & \!Reskin & $\Delta$ \\
\midrule
\spireicon{qwen} & Qwen3.8 27B {\scriptsize\textcolor{gray}{(xhigh)}} & 20 & 20 & \dzero{$\pm0$} & 14.0 & 14.9 & \dup{$+0.9$} \\
\spireicon{qwen} & Qwen3.6 35B-A3B & 0 & 0 & \dzero{$\pm0$} & 9.7 & 10.2 & \dup{$+0.5$} \\
\spireicon{qwen} & Qwen3.5 122B-A10B {\scriptsize\textcolor{gray}{(xhigh)}} & 10 & 0 & \ddown{$-10$} & 10.9 & 9.8 & \ddown{$-1.1$} \\
\spireicon{claude} & Claude Haiku 4.5 {\scriptsize\textcolor{gray}{(12k)}} & 10 & 10 & \dzero{$\pm0$} & 11.4 & 11.9 & \dup{$+0.5$} \\
\spireicon{deepseek} & DeepSeek V4.1 Flash {\scriptsize\textcolor{gray}{(max)}} & 70 & 30 & \ddown{$-40$} & 22.0 & 16.9 & \ddown{$-5.1$} \\
\spireicon{gemini} & Gemini 3.8 Flash {\scriptsize\textcolor{gray}{(low)}} & 40 & 60 & \dup{$+20$} & 20.2 & 22.6 & \dup{$+2.4$} \\
\spireicon{nvidia} & Nemotron 3 Nano Omni & 0 & 0 & \dzero{$\pm0$} & 5.3 & 6.8 & \dup{$+1.5$} \\
\spireicon{openai} & GPT-5.6 Luna {\scriptsize\textcolor{gray}{(max)}} & 60 & 10 & \ddown{$-50$} & 18.5 & 14.2 & \ddown{$-4.3$} \\
\spireicon{openai} & gpt-oss-120b {\scriptsize\textcolor{gray}{(high)}} & 0 & 10 & \dup{$+10$} & 10.4 & 10.4 & \dzero{$\pm0.0$} \\
\spireicon{zai} & GLM-5.3 {\scriptsize\textcolor{gray}{(max)}} & 60 & 60 & \dzero{$\pm0$} & 24.4 & 22.1 & \ddown{$-2.3$} \\
\spireicon{zai} & GLM-5.3 Flash {\scriptsize\textcolor{gray}{(max)}} & 70 & 40 & \ddown{$-30$} & 23.1 & 15.6 & \ddown{$-7.5$} \\
\midrule
 & Mean (11 models) & 31 & 22 & \ddown{$-9$} & 15.4 & 14.1 & \ddown{$-1.3$} \\
\bottomrule
\end{tabular}
\end{adjustbox}
\end{minipage}
\end{table}

\mypara{Prior knowledge is part of the score.} %
Hiding the real names costs $9$ points of Act~1 win rate and
$1.3$ floors, on average over the 11 models with all 10 seeds under both (\cref{tab:reskin-compact}).
The cost is concentrated in a few models: DeepSeek~V4.1 Flash falls from $0.80$ acts to $0.30$ and GPT-5.6 Luna from $0.60$ to
$0.10$ (\cref{tab:ablation-reskin-by-model}). On the seven models with every mapping (\cref{tab:ablation-reskin}), nonsense names cost least ($0.03$ acts), so unfamiliar words are
not the difficulty; the two permuted mappings, which give each entity another's name, cost
as much as the themed one or more ($0.11$ and $0.14$, against $0.09$). A familiar name pulls a model
towards its usual meaning even though the prompt states the rules~\citep{mccoy2024embers}. %
The reskin hides names, not the game:
in 125 of the 200 rethemed runs of \cref{tab:main-results}, from 18 of
the 20 models, the model recognizes \emph{Slay the Spire} or a
real entity the reskin replaced, yet recognition does not buy back the cost (Spearman $\rho = 0.18$,
permutation $p = 0.62$; \cref{tab:reskin-recognition}).
Withholding what each card and relic does costs about as much with the real names as with the reskinned ones (\cref{tab:ablation-names-blind}), so the models have not memorized what individual items do: what the real names recall is heuristic knowledge of the game.
In our puzzle benchmark, we find the reskin costs more the further ahead the decision looks (\cref{app:puzzles-reskin}), suggesting the models do better when they can utilize pretraining heuristics than when simply reasoning.

\mypara{Models forecast the next turn, not the rest of the run.} Models that forecast the
next turn's HP well clear more acts (Spearman $\rho = -0.84$ between error and acts;
\cref{tab:calibration-horizon}). This forecast measures the fight-reading skill above: on turns where the intents shown
determine the next HP exactly, the bottom tier still mispredicts it 31\% of the time and the top four 0.5\%. Forecasts of surviving the act are poorly calibrated: the four models that
never clear an act give themselves a 55--76\% chance, about the same as the four strongest (59--74\%), which go on to
survive the act 63--76\% of the time (\cref{fig:calibration-vs-play} in \cref{app:calibration}). The forecasts barely
sharpen as the act nears its end (\cref{fig:prediction-horizon}). Further analyses are in \cref{app:further-analyses}.

\section{Discussion}
\label{sec:discussion}

Our findings consistently show that agentic models misjudge risk over the long term.
Where the outcome is a turn or a fight away, the strongest models are competent: they win the first act as often as expert players do, win almost
every ordinary fight, and forecast the next turn well, though they still lose more HP in each fight than
humans do.
Where the outcome is an act or a run away, they fall behind: the gap to the expert players opens from Act~2, boss fights decide where each model stops,
forecasts of surviving an act barely sharpen, and what a model writes down for later runs does not
compound. This is the horizon gap of \citet{chen2026horizongap}, but within one task whose time scales share a
single pool of resources. Across time horizons \citep{kwa2025horizon} (turn, fight, act, or run), success does not decline gradually but drops sharply: every model succeeds on at least 87\% of its turns, yet falls below 50\% somewhere between a fight and a full run, and for most models already at the act (\cref{tab:time-horizon}).

The difficulty is not how long a
model can keep going: models perform well enough at micro-scale decisions to afford to take some risks which have positive expected value, but don't.
Nor is this only a result of the uncertainty of a run: in the puzzles the models take the immediate, safe outcome even when the delayed payoff is certain (\cref{app:puzzles-unsolved}).
No model we studied learns from its own runs that early floors should be spent on growth, with payoff coming later in the run.
Any agent planning over a long, stochastic task with one delayed outcome needs memories checked against
outcomes and discarded when wrong, forecasts that sharpen as the outcome approaches, decisions that follow the rules it can state, and a harness that plays the
whole task.

\label{sec:limitations}

Our findings have implications beyond the \emph{Slay the Spire} setting studied. \emph{Agents under-invest when the payoff is delayed}, even when they can state the rule that makes investing worthwhile, a choice any agent with a budget of compute, money, or time faces (the knowing--doing gap of BALROG~\citep{paglieri2024balrog}).
\emph{Unverified memory does not compound} wherever the outcome is delayed and noisy over many decisions, as
in most open-ended deployments: errors accumulate and existing biases are reinforced;
where a verifier exists, such as tests for code, the reported gains should hold.
\emph{Scores on a familiar task include recalled knowledge}, even when every rule is supplied,
so rankings on well-known tasks partly reflect training data.
\emph{Checks on single steps overstate competence on long tasks}, and so does the agent's own confidence.

\mypara{Limitations.} \emph{The ordering of models} is measured on one character in a
turn-based game played from parsed text; other characters, genres, real-time control, or perception would
weight the decisions differently. \emph{The size of the reskin cost} depends on how much of this game each
model saw in training, and the reskin hides names but not familiarity with deck-builders, so the measured cost is a lower
bound. \emph{The human comparison} over runs is not seed-matched (\cref{app:expert-decisions} compares single choices on the same states), and players in the telemetry may quit and reload the game to replay a fight, which our agents cannot.
Ten seeds per cohort cannot order neighbouring models, and providers can change the model behind an identifier without notice.
\emph{Reflection sees only a summary of the run}: the model's notes, the route, and one line per floor, not the turn-by-turn trajectory, which is often larger than its context window.
A cause of failure it did not note at the time cannot be recovered; whether a guided walk through the trajectory in hindsight would improve what it learns is untested.

\subsection*{Ethics statement}
SpireBench uses a commercial product, the video game \emph{Slay the Spire} by Mega Crit.
We do not redistribute the game; anyone wishing to run the benchmark themselves must use a legally purchased copy of the game such as from \href{https://store.steampowered.com/app/646570/Slay_the_Spire/}{Steam} or \href{https://www.gog.com/en/game/slay_the_spire}{GOG}.
Our harness modifies the game using the modding functionality built into \emph{Slay the Spire} and complies with Mega Crit's \href{https://www.megacrit.com/content-policy/}{content policy}.
This benchmark is not officially affiliated with Mega Crit.

We used anonymized human data as part of this work.
This anonymized data was collected as telemetry for \emph{Slay the Spire} by the game's developer, Mega Crit, under the \emph{Slay the Spire} terms of service.
It covers all human players from 2018 to 2020, and was \href{https://www.reddit.com/r/slaythespire/comments/jt5y1w/77_million_runs_an_sts_metrics_dump/}{released publicly by Mega Crit}.

Additionally, we used identifiable human data as part of our comparisons in this work.
We use publicly available run logs from four expert \emph{Slay the Spire} players (Twitch streamers) who openly shared their run logs for community analysis, and list their win rates under their Twitch pseudonyms.
The data used for this analysis predates our work and was not created as part of the study.
We do not anticipate any negative consequences to the individuals whose data was used as these aggregate statistics are already a matter of public record.
We do not rehost or redistribute any copyrighted content.

We also use personal data from one author of the paper who gave permission for their data to be included as a representative example of a proficient, but non-professional, player.

\subsection*{Acknowledgments}

Thanks to Sam Laing and Marvin da Silva for useful discussions and for providing feedback on the paper.

Resources used in preparing this research were provided, in part, by the Province of Ontario, the Government of Canada through CIFAR, and \href{https://vectorinstitute.ai/partnerships/current-partners/}{companies sponsoring} the Vector Institute.
This research was also enabled in part by support provided by the BC DRI Group and the \href{https://alliancecan.ca/}{Digital Research Alliance of Canada}.

Thanks also to Felix Dangel and Marvin da Silva for helping to support the work.

\FloatBarrier
\clearpage
\bibliography{main}
\bibliographystyle{iclr2027_conference}

\clearpage
\appendix
\crefalias{section}{appendix}\crefalias{subsection}{appendix}\crefalias{subsubsection}{appendix}
\section*{Appendices}
{
    \hypersetup{hidelinks}%
    \startcontents[sections]
    \printcontents[sections]{l}{1}{\setcounter{tocdepth}{2}}
}
\FloatBarrier
\clearpage

\section{Why \texorpdfstring{\emph{Slay the Spire}}{Slay the Spire}}
\label{app:why-sts}
\emph{Slay the Spire} is a seminal game that helped establish the roguelike deck-builder genre, and the properties that make it hard
are those that make any long task hard for an agent.

\textbf{Every choice recurs.} A card added to the deck
on the second floor is drawn again in every later fight, so it has to be useful not only against the next
enemy but against the act bosses an hour later, and a bad addition cannot simply be undone:
card removals are scarce, and the starting deck already holds cards (Strikes) that the player would
benefit from removing, so a bad addition costs a removal for the rest of the game.
The consequences of a
decision are therefore spread thinly over hundreds of later ones---this is the credit-assignment problem of
long-horizon execution, where small per-step errors compound over the length of a
task~\citep{sinha2026illusion}. The only verdict on the run as a whole, a win or a loss, comes at its end. %

\textbf{An archetype is not a deck.} Players learn named strategies, such as
building around Strength or around exhausting cards, but each run offers a random array of cards and relics based on its seed.
A plan must be devised as the run unfolds rather than determined in advance, and the agent has to judge when to commit to a plan and when to abandon it.
A card's worth also depends on the cards and relics already held: a card that scales with Strength is weak in a deck with no source of Strength and strong in one with several, so cards cannot be valued one at a time. %

\textbf{The agent builds its own action space.} In combat, the actions open to the player on a turn are the
cards in hand, and the hand is drawn from a deck the player assembled. Every card reward, purchase, removal, and
upgrade therefore decides which actions will exist in later fights: the agent chooses not only its moves but the
set of moves it will later choose among. This is a sharper form of what an agent does when it builds a library of
skills or tools for later use~\citep{wang2023voyager}, with the difference that here the library is also a
liability, since every card added is one more card that may be drawn in place of the one that was needed.
Even a single turn is a small planning problem, because the order of plays matters: an attack does half as much again
to an enemy made Vulnerable first, so Bash is played before the Strikes it sets up. %

\textbf{Each fight tests the deck differently.} A deck is not good or bad in general but against a particular
fight. Groups of enemies call for damage that hits all of them; long fights against enemies that grow stronger
call for cards that scale, while short ones call for damage up front; and some enemies punish a kind of play
outright: Gremlin Nob gains Strength whenever a Skill is played, the Awakened One whenever a Power is played,
and the Time Eater takes over the turn once a set number of cards has been played. A card bought to pass one of
these checks can weaken the deck against another, and every card added dilutes the draws for all of them, so
the player must decide which checks the deck has to pass, which it can afford to scrape through, and in what
order the run will set them.

\textbf{Risk is amortized over the run.} HP, gold, potions, and the deck form one irreversible pool that
carries from floor to floor. Taking an elite fight costs HP now for a relic that pays out over every later
fight; resting at a campfire recovers HP at the price of an upgrade. The right choice depends on what the
rest of the run will demand, so the agent must price present resources against a future it can only
estimate, and how well it does so shows in how well calibrated those estimates are.

\textbf{The challenges ahead are known in kind but not in detail.}
Each act draws its fights from a fixed pool, and the map shows
the types of room ahead, so a player who remembers what the act's elites and bosses do can
build the deck to meet them; one who does not meets each as a surprise. Doing well therefore needs memory
across runs as well as within one, which is the setting in which memory mechanisms for LLM agents are
proposed~\citep{park2023generative,packer2023memgpt,shinn2023reflexion} but seldom tested over long,
stochastic episodes.

\textbf{Some of the game is fixed and some is dealt.} Within a fight, each enemy shows its intended action for
the coming turn, and enemies follow move patterns that can be learned, so the threat on the current turn is known
exactly and a turn can be planned as a deterministic puzzle. What the player will draw next turn, which cards a
reward will offer, and what an unknown room holds are random; the player knows which cards are in the draw pile but
not their order. %
 The map is visible from the start of each act, but
not what will be on offer along a route. A good player therefore treats different parts of the same decision
differently, solving what is fixed and pricing what is dealt, and an agent has to tell the two apart, neither
gambling on a turn whose outcome it could have computed nor planning the run as though the draws were known.
The models we test divide along this line: they forecast the next turn's HP far better than their
chance of surviving the act (\cref{sec:results}). %

These properties are %
 the parts of the horizon gap~\citep{chen2026horizongap} that short benchmarks
cannot reach, and a single run of the game, about an hour of human play, sits at the task lengths on which
agent progress is now charted~\citep{kwa2025horizon}. The game also has three practical properties that
make it a good testbed. It is turn-based and fully described in text, so what is tested is decision-making
rather than perception or reaction time. Its outcomes mix skill with luck in a measured
proportion~\citep{trojanowski2021sts}, and its maps pose genuine route decisions under
uncertainty~\citep{bazzaz2025paths}, so a fixed set of seeds gives a paired, repeatable comparison. And it
has a large, well-documented population of human players, from general game telemetry to expert players, against which
models can be placed.

\FloatBarrier
\FloatBarrier
\section[\texorpdfstring{Sample game screens}{Sample game screens}]{Sample game screens}
\label{app:screens}

\definecolor{scrmark}{HTML}{FFD23F}
\tikzset{
  scrbox/.style={draw=scrmark, line width=1.1pt, rounded corners=2pt},
  scrbadge/.style={circle, fill=scrmark, draw=black, line width=0.4pt, text=black, font=\sffamily\bfseries\tiny,
    inner sep=0pt, minimum size=8.5pt},
  scrlab/.style={fill=black!82, text=scrmark, font=\sffamily\scriptsize, rounded corners=1pt, inner sep=1.6pt,
    align=center},
  scrtiny/.style={scrlab, font=\sffamily\tiny, inner sep=1pt},
}

This appendix exemplifies what a run looks like to a human, as frames of the replay videos of two runs in the main cohort (\cref{tab:main-results}): the turn of combat and the card reward are from Gemini 3.8 Flash's run of seed 1, and the other screens are from GPT-5.6 Sol's run of seed 5.
The replay drives the game with the commands the run sent.
The game shows its own names, so in the captions each name on screen is followed by the reskinned name the model saw in its place, in brackets (\cref{app:methods-reskin}).
The model itself never sees a screen: it receives the same state as text (\cref{app:prompt-combat-turn,app:prompt-floor-opening}).

\begin{figure}[!htbp]
\centering
\begin{tikzpicture}[x=\linewidth/1920, y=-\linewidth/1920]
\node[anchor=north west, inner sep=0] at (0,0) {\includegraphics[width=\linewidth]{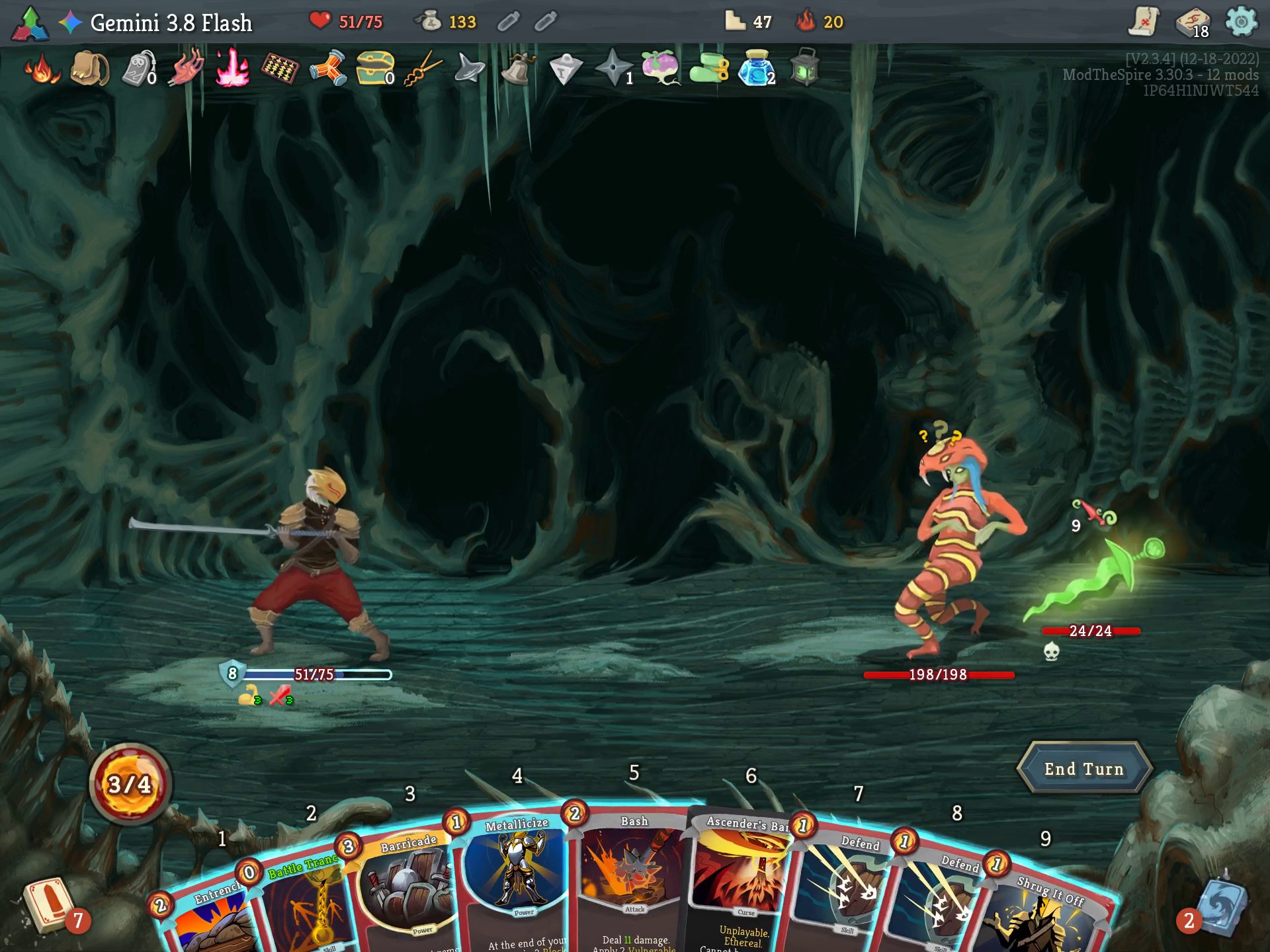}};
\foreach \n/\xa/\ya/\xb/\yb in {
    1/16/6/82/64, 2/458/6/850/58, 3/1088/6/1285/58, 4/1772/6/1846/62, 5/20/66/1252/140,
    6/1376/632/1466/684, 7/326/994/374/1040, 8/1298/1004/1542/1036, 10/130/1120/266/1256,
    12/24/1316/146/1428, 13/1774/1316/1904/1428} {
  \draw[scrbox] (\xa,\ya) rectangle (\xb,\yb);
  \node[scrbadge] at (\xa,\ya) {\n};
}
\foreach \n/\xa/\ya/\xb/\yb in {6/1606/748/1698/810, 8/1568/938/1732/1002} {
  \draw[scrbox] (\xa,\ya) rectangle (\xb,\yb);
  \node[scrbadge] at (\xa,\ya) {\n};
}
\draw[scrbox] (356,1040) rectangle (452,1072);
\node[scrbadge] at (452,1072) {9};
\draw[scrbox] (235,1210) rectangle (1700,1436);
\node[scrbadge] at (1700,1210) {11};
\end{tikzpicture}
\caption{\textbf{A turn of \emph{Slay the Spire} combat}. Gemini 3.8 Flash's run of seed 1, floor 47, the first turn against an elite, Reptomancer (reskinned: Husk Bomblord), and one of its Snake Daggers (Volatile Husk).
\textbf{(1)}~Keys: each third of the emblem lights up when its corresponding key is collected. All three keys need to be collected to progress to Act~4 upon completing Act~3.
\textbf{(2)}~Hit points (HP), gold, and the two potion slots, both empty here. These resources are carried between floors. If a player's character runs out of HP, they die. Potions are collected randomly at approximately every other fight and should be used to minimize HP loss, especially in harder fights or on turns when a poor hand is drawn. Potions should be preserved between floors to maximize their effectiveness, but not so much they are wasted by gaining a new potion when both slots are already full.
\textbf{(3)}~Floor (47 of 56) and difficulty level (Ascension 20 of 20).
\textbf{(4)}~Deck: 18 cards, carried floor to floor and added to over the course of the run.
\textbf{(5)}~Relics: items which offer passive buffs. Some have one-off out-of-combat effects (e.g. increase max HP). Some have once-per-combat effects (e.g. draw an extra two cards on turn 1). Some have triggered effects (e.g. \emph{Shuriken}: every third attack played in a turn, gain 1 strength). Some have triggered effects with counters carried across across combats (e.g. \emph{Ink Bottle}: every tenth card played, draw 1 card), meaning optimal play is to stall easier fights so the counters can be set up to trigger at the start of subsequent harder fights.
\textbf{(6)}~Enemies intents: what the enemies will do this turn, immediately after the player ends their turn. Reptomancer's is ``Unknown'', and the player learns from experience that this means they are summoning more minions to join the fight. The Snake Dagger minion's is an attack for 9 and a debuff (add a Wound to the player deck).
\textbf{(7)}~Player's Block, 8, which absorbs incoming attack damage before it reaches HP, and is removed at the start of the player's next turn.
\textbf{(8)}~Enemy HP. Non-minion enemies have to be reduced to 0 HP to win the fight.
\textbf{(9)}~Player statuses: Strength 3, which adds 3 to the damage of every Attack, and Strength Down 3 (Strength Drain), which takes that Strength away at the end of this turn.
\textbf{(10)}~Energy (3/4) for this turn, used to play cards. By default, players gain 3 energy per turn, but this can be increased by end-of-act Boss relics.
\textbf{(11)}~Player hand of cards: includes attacks (Bash), skills (Entrench, Battle Trance+, Defend, Shrug It Off), powers (Barricade, Metallicize) and a curse (Ascender's Bane). These are the real card names; our models primarily play with a reskin to ablate pretraining priors about the cards. At the start of each turn, 5 cards are drawn from the draw pile. More cards can be drawn during the turn by playing cards which draw more cards. Attacks damage enemies and may apply status effects on them. Skills often block the player but may have other effects such as drawing more cards, gaining more energy, or applying statuses to the player or the enemies. Powers have effects that apply each turn for the rest of the fight and are the main method of in-combat scaling; spending energy on them may mean skipping blocking this turn and taking damage.
\textbf{(12)}~Draw pile (7 cards), and \textbf{(13)}~the discard pile formed of played cards (2 cards). The discard pile is shuffled to create a new draw pile when drawing from a an empty draw pile.
\Cref{fig:screen-combat-prompt} shows the prompt the model received in this state.
}
\label{fig:screen-combat}
\end{figure}

\begin{figure}[!htbp]
\centering
\includegraphics[width=\linewidth]{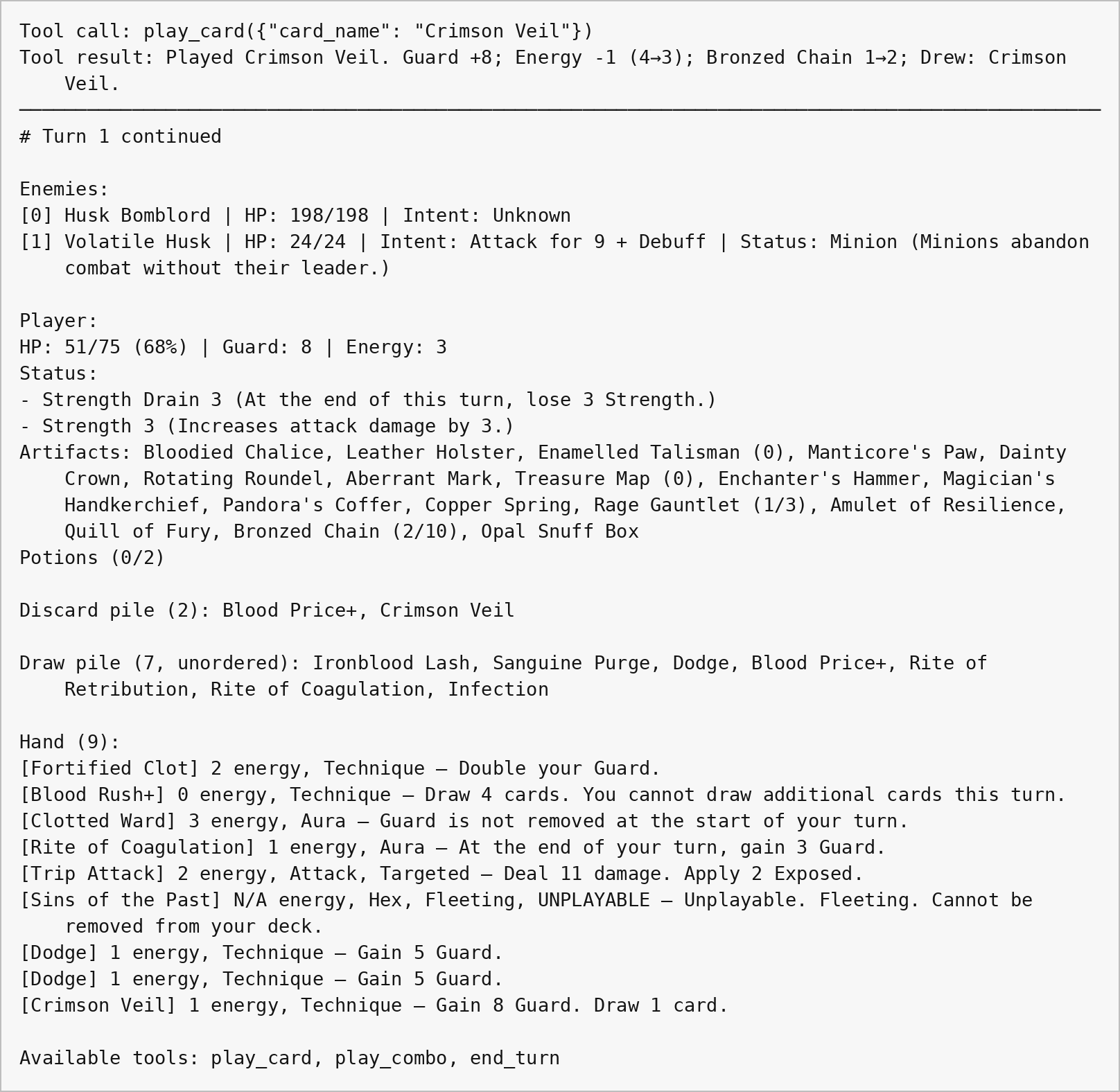}
\caption{\textbf{The prompt for the state of \cref{fig:screen-combat}}, verbatim.
Above the line are the model's previous tool call, playing Shrug It Off (Crimson Veil), and the result returned for it; below it is the message sent with that result, the last before the frame.
Every name shown is the reskinned one seen by the model, with Block reskinned as Guard.
The relic counters appear as Rage Gauntlet (1/3) and Bronzed Chain (2/10), and the tool result reports the Ink Bottle's count (Bronzed Chain) going from 1 to 2.
Relic effects are shown in the system prompt (shown in \cref{app:prompt-system}) and the first prompt of the floor (e.g. \cref{app:prompt-floor-opening}), but not repeated on subsequent turns within the same floor.
}
\label{fig:screen-combat-prompt}
\end{figure}

\begin{figure}[p]
\centering
\begin{subfigure}[t]{0.49\linewidth}
\centering
\begin{tikzpicture}[x=\linewidth/1140, y=-\linewidth/1140]
\node[anchor=north west, inner sep=0] at (0,0) {\includegraphics[width=\linewidth]{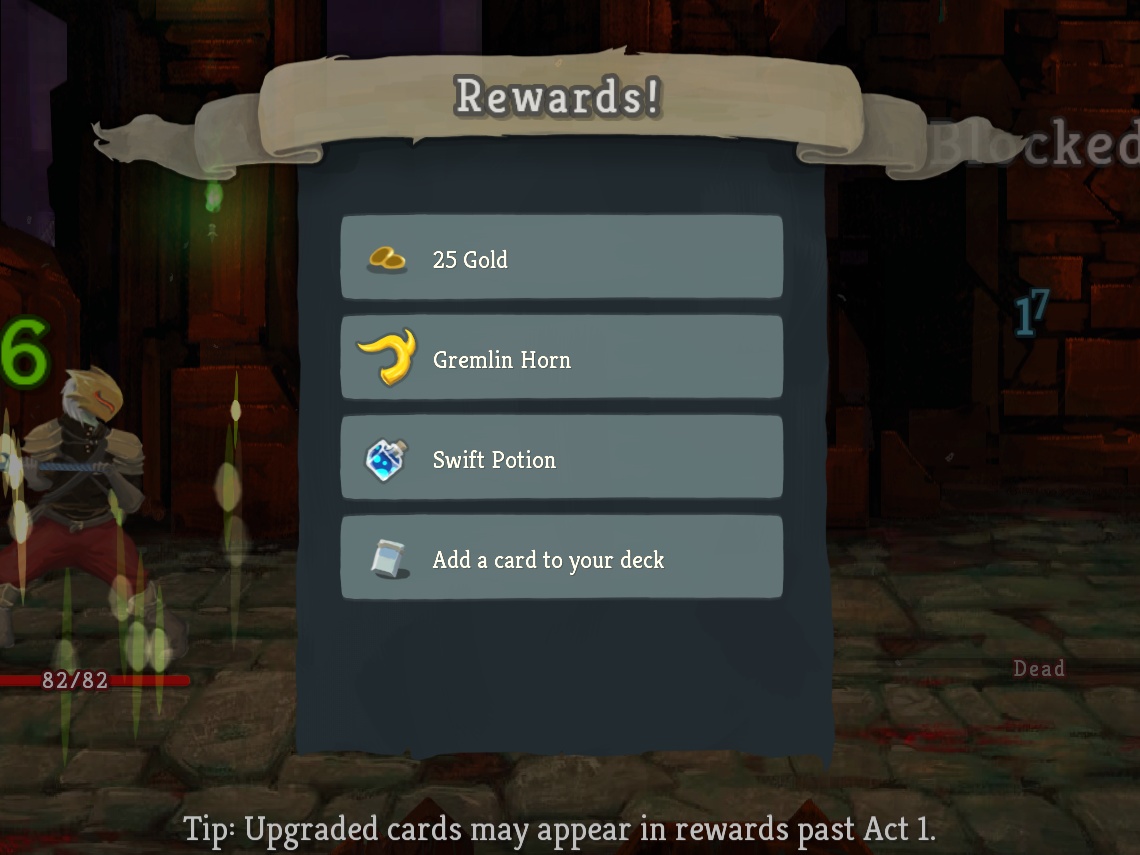}};
\foreach \ya/\lab in {214/gold, 314/relic, 414/potion, 514/card reward} {
  \draw[scrbox] (336,\ya) rectangle (788,\ya+88);
  \node[scrlab, anchor=west] at (802,\ya+44) {\lab};
}
\end{tikzpicture}
\caption{The rewards of an Elite fight: 25 gold (around twice as much as would be obtained from a regular fight), the relic \emph{Gremlin Horn}%
, the potion \emph{Swift Potion}%
, and a card reward, contents shown in panel \subref{fig:example-card-reward} (right).
The reward screen is accompanied with an in-game tip; we present all of these to the model in its system prompt (\cref{app:prompt-system}).
}
\end{subfigure}\hfill
\begin{subfigure}[t]{0.49\linewidth}
\centering
\begin{tikzpicture}[x=\linewidth/1140, y=-\linewidth/1140]
\node[anchor=north west, inner sep=0] at (0,0) {\includegraphics[width=\linewidth]{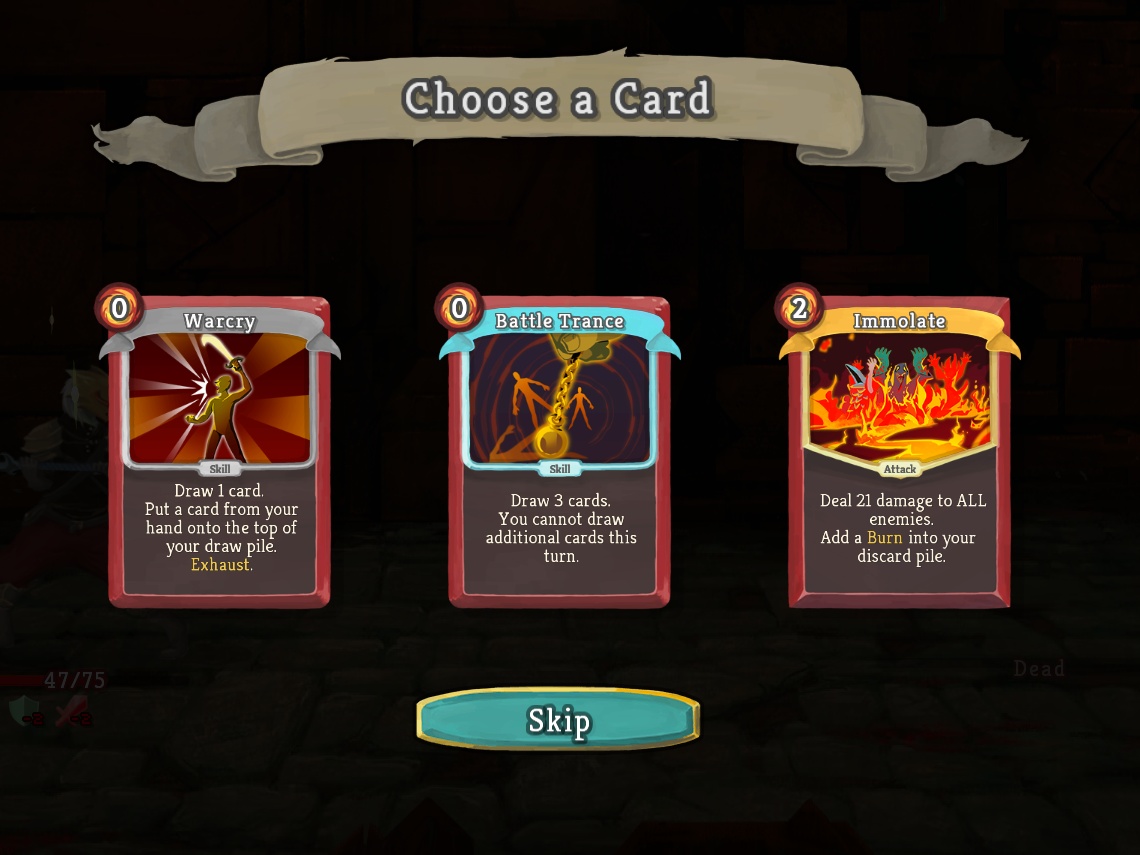}};
\draw[scrbox] (98,280) rectangle (1022,614);
\node[scrlab, anchor=north] at (560,622) {three cards};
\draw[scrbox] (414,686) rectangle (704,754);
\node[scrlab, anchor=west] at (716,720) {or none};
\end{tikzpicture}
\caption{Card reward options. Cards are seen at different frequencies depending on their rarity: Common (grey banner), Uncommon (blue), Rare (gold). Which card is best to take depends on the current deck composition, immediate threats from upcoming floors, and even potential threats from later acts. The card reward can be skipped if none are desirable.}
\label{fig:example-card-reward}
\end{subfigure}

\vspace{0.6em}
\begin{subfigure}[t]{0.49\linewidth}
\centering
\begin{tikzpicture}[x=\linewidth/1140, y=-\linewidth/1140]
\node[anchor=north west, inner sep=0] at (0,0) {\includegraphics[width=\linewidth]{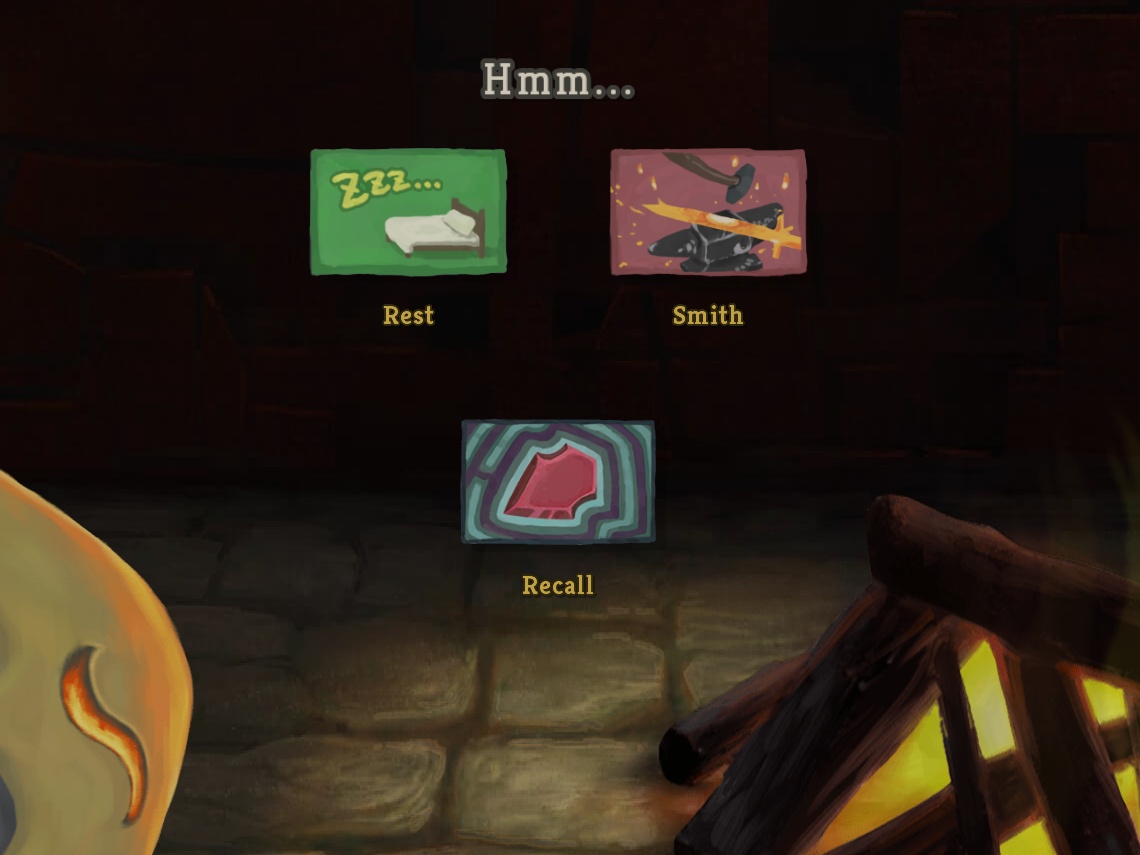}};
\end{tikzpicture}
\caption{Rest site: Three options are available, and only one can be selected. Rest: heal hit points (30\% of player's max HP); Smith: upgrade a card of the player's choice; Recall, grants the Ruby Key needed to reach Act~4.
We find models opt to Rest much more often than they should.
Upgrading a card is a permanent increase to the deck quality (e.g. increasing the block value of a card from 8 to 11) which can result in saving more hit points over the course of a run (blocking for 3 more every time the card is played) than the Rest option would heal (e.g. 23 hit points).
}
\end{subfigure}
\caption{\textbf{Example screens between fights.} Gold is collected by the harness without a model call; the model chooses among the other rewards, the cards, and the rest-site options (\cref{app:harness-features}).}
\label{fig:screen-rooms}
\end{figure}

\begin{figure}[p]
\centering
\begin{tikzpicture}[x=0.5\linewidth/1079, y=-0.5\linewidth/1079]
\node[anchor=north west, inner sep=0] at (0,0) {\includegraphics[width=0.5\linewidth]{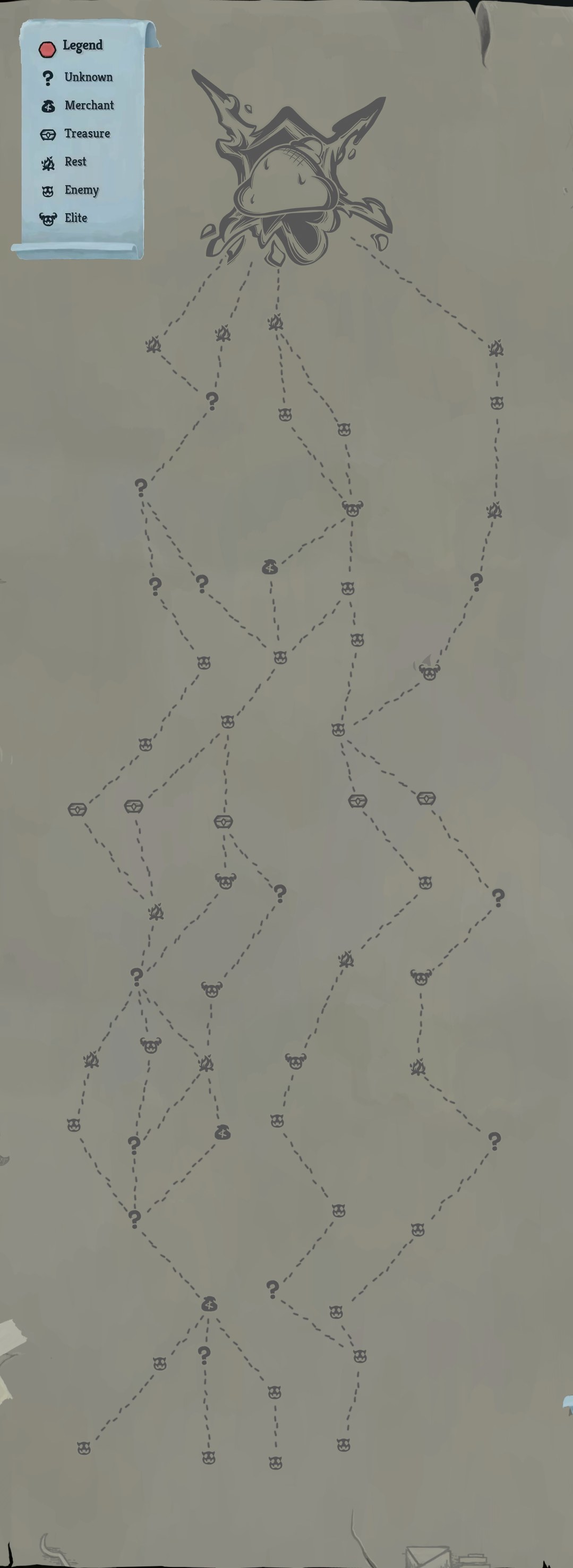}};
\draw[scrbox] (8,16) rectangle (324,504);
\node[scrtiny, anchor=north west] at (14,512) {the game's legend};
\foreach \f/\y in {1/2729, 2/2572, 3/2451, 4/2296, 5/2133, 6/1993, 7/1837, 8/1682, 9/1531, 10/1377, 11/1238,
    12/1095, 13/945, 14/776, 15/634} {
  \node[scrtiny] at (52,\y) {\f};
}
\foreach \c/\x in {0/157, 1/274, 2/403, 3/525, 4/654, 5/796, 6/924} {
  \node[scrtiny] at (\x,2846) {\c};
}
\node[scrtiny, anchor=west] at (960,2846) {column};
\node[scrtiny] at (52,584) {floor};
\end{tikzpicture}
\caption{\textbf{Act map.} The whole map of Act~1, from the bottom (floor 1) to the boss at the top.
The game's legend (top left) is pasted here once, from the map screen; the model never sees it.
The prompt's map names the rooms Enemy/Monster, Elite, Rest site, Merchant/Shop, Unknown, and Treasure, and each node by its floor and column, e.g.\ Monster(1,0); the numbers along the edges give both.
The act culminates in boss fight --- an especially hard encounter.
In the prompt, the map is a list of these nodes and the nodes each leads to, ending with the boss (Slime Boss) (\cref{app:prompt-floor-opening}).}
\label{fig:screen-map}
\end{figure}

\Cref{fig:screen-combat} is a turn of combat and \cref{fig:screen-combat-prompt} its prompt, \cref{fig:screen-rooms} the screens between fights, and \cref{fig:screen-map} the map of the first act, on which the model plans its route.
At the rest site of \cref{fig:screen-rooms}c, the model weighs healing now against a stronger deck later; its third option, Recall, takes the Ruby Key, one of the three keys a run needs to enter Act~4 (\cref{app:harness-coverage}).

\FloatBarrier

\section{Harness} %
\begin{figure}[h!]
\centering
\includegraphics[width=\linewidth]{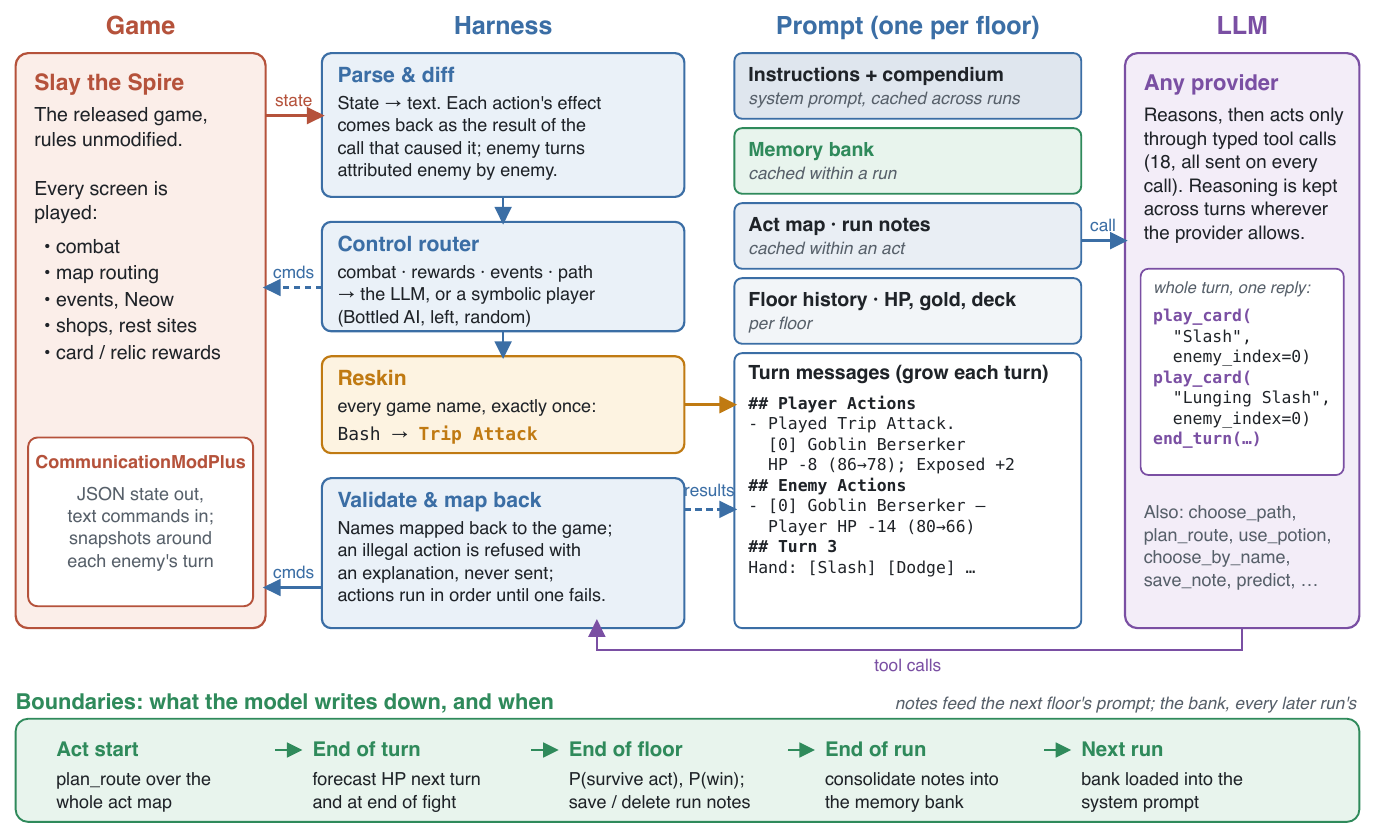}
\caption{\textbf{Our harness.} Game states become text, each action's effect attributed to its tool call; a
router sends each part of the game to the model or a symbolic player; in-game names are (optionally) reskinned;
tool calls are mapped back and validated. The prompt runs from cached layers to turn messages.
Bottom: when the model plans, forecasts, and writes notes.
}
\label{fig:harness}
\end{figure}

\begin{figure}[thb]
\centering
\includegraphics[width=\linewidth]{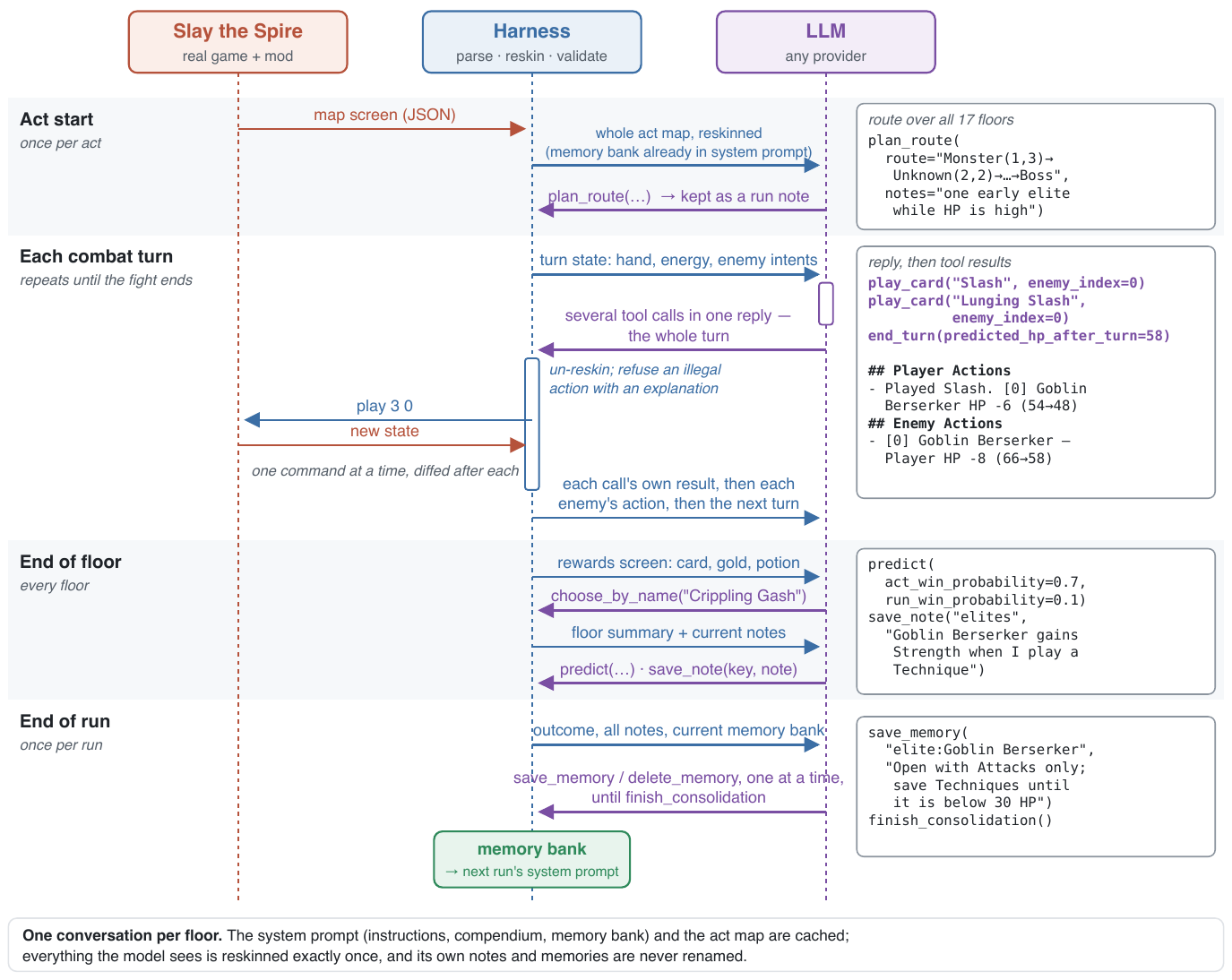}
\caption{\textbf{Harness details.} The harness as a sequence of messages, with time running downwards. In combat the model
returns a whole turn as several tool calls in one reply; the harness sends them to the game one
command at a time, diffs the state after each, and returns each call's own effect as that call's
result, followed by each enemy's action and the next turn's state. At the end of each floor the model
forecasts its chances and may update its run notes; at the end of the run it consolidates them into
the memory bank that the next run's system prompt carries. Code on the right is the shape of what the
model sends and receives, with names as it sees them under the reskin.}
\label{fig:harness-seq}
\end{figure}

\label{app:harness-sequence}
\Cref{fig:harness} gives an overview of the components of the harness.
\Cref{fig:harness-seq} follows the messages exchanged between the game, the harness, and the model
through the four points at which the harness calls the model: the start of an act, each combat turn,
the end of a floor, and the end of a run (\cref{sec:methods-harness,sec:methods-memory}).
The subsections then describe the game interface, the agent harness, notes and memory, and the
knowledge reskin, and \cref{app:harness-features} lists every feature of the harness that the model
can see. %

\subsection{Game interface}
\label{app:methods-game}
Our fork of CommunicationMod~\citep{communicationmod}, \texttt{CommunicationModPlus}, adds a file-based
channel that survives restarts of the game's Java process, tracking of the three keys that unlock
Act~4, and intermediate state snapshots around each enemy's turn, from which the harness attributes
damage, block, and status changes to the enemy that caused them. Every run loads the same mod list:
\texttt{BaseMod}, \texttt{StSLib}, \texttt{CommunicationModPlus}, \texttt{rngFix}, \texttt{permNeow}
(a Neow blessing on every run), \texttt{SuperFastMode} (animation speed only), \texttt{RelicStats},
and \texttt{fast\_multicast}. None changes the rules a human plays by. Each run executes in an
isolated game profile, and every act boss is marked as seen, so each run's seed, and not the
profile's history, picks the act bosses.
That profile has every card, relic, and character unlocked, so the pool of content is the same in
every run and does not change as more games are played. The model has no tool to save or reload, so it
cannot replay a floor to fix misplays or see what would have happened.
A run stopped by a failure of the harness or the model's API is resumed from an
autosave the game writes on entering the floor without the previous attempt in its context.

\subsection{Agent harness}
\label{app:methods-harness}
Each call carries a cached system prompt in two parts. The first is stable across runs: task
instructions and a compendium of the game's rules, keywords, status effects, cards, relics, and
potions with their effects. The second is stable within a run: the memory bank
(\cref{sec:methods-memory}). The first user message of a floor adds the act map, the run's notes, a
history of the floors played so far and the player's current state (HP, gold, deck, relics, potions,
and keys). In combat, each turn's message reports what each of the model's actions did, what each
enemy did, and the full state at the start of the new turn. Outside combat the message describes the
current screen and its legal choices.

The tools are: playing a card or a sequence of cards, ending the turn, using or discarding a potion,
choosing an option by index or by name, choosing a map node, planning a route, taking notes,
predicting the floor's outcome, and the end-of-run memory operations. %
The full tool set is sent on every call, so that the cached prefix never changes, and an action illegal
on the current screen is refused with an explanation. By default the model may return
several actions in one response (\emph{multi-action}), and the harness executes them in order,
stopping at the first that fails. At the start of each act the model is asked for a route plan over
the whole act map. At the end of each floor it is asked for its chance of surviving the current act
and of winning the run, and may reflect and update its notes; when it ends a turn it may also predict
its HP at the start of the next turn and at the end of the fight.

Two principles shaped the harness. The model is shown what a human player sees and nothing more: the
prompts describe the task on the current screen without steering towards a strategy (they never say,
for instance, to check first whether an enemy can be killed this turn), and the system prompt holds
only the documentation a player has in the game itself, namely its tutorial, tips, and compendium. And
the model is asked to do as much as possible per call, to keep cost down. Many combat actions have a
deterministic effect, so the model plans the whole turn and returns every action it wants in one
response, up to the first whose outcome is random. MiniStS makes one call per action and re-prompts
with the new state~\citep{bateni2024minists}. Orak's agent may also return several actions at once, but
spends three calls on each decision (reflection, planning, then the actions) and shows the model only
the state its actions leave~\citep{orak}. %
Multi-action play asks the model to hold the intervening states in mind, which is harder. Batching
also helped play in our preliminary runs. Because the model had to plan the whole turn before acting, it
committed to a plan, and it did not spend reasoning deriving the same plan again after each action when
that action had revealed nothing new. On the main seeds, one action per response needs 3.0 more calls
per floor, costs 23\% more, and clears 0.07 fewer acts (\cref{tab:harness-compare}). %
So that it can still see what each action did, the
harness diffs the game state between actions and returns each change as the result of the tool call
that caused it, with enemies' turns attributed enemy by enemy. A call that names a card imprecisely is
resolved where the intended card is unambiguous, and otherwise answered with a ``did you mean''
suggestion rather than executed.

\paragraph{Names, not indices.}The model refers to what it acts on by name: a card by the name
it is shown under (\texttt{play\_card(card\_name="Bash", enemy\_index=0)}), and an event option,
rest-site action, shop item, or card on a selection screen by its label
(\texttt{choose\_by\_name(name="Rest")}). Arguments are passed by name too, never by position. Only
enemies, card rewards, and the reward list are chosen by index, which is shown beside each entry.
There are two reasons for this.

The first is batching. A batch is written against the state at the start of the turn, but the hand
changes while the batch runs: every card played moves the positions of the others, and cards are
drawn, created, upgraded, and moved by the plays themselves. A position in the hand goes stale
after the first change, but a name still identifies the same card. A batch can therefore play a card that an earlier call in
it creates, and it can answer, in the same call, a choice that a card opens (the card that Armaments upgrades, the card that Headbutt
puts back), through \texttt{target\_card\_names}. This is what lets the model play a whole turn in
one response without giving up any of the game's mechanics. How Orak handles the same cases is in
\cref{app:harness-coverage}. %

The second is reliability. In preliminary runs, weaker models sometimes reasoned their way to one
option and then gave the index of another. A name states the intended option in the call itself,
so it cannot pick a different option by mistake. A name that matches nothing is refused with a
suggestion (\cref{app:harness-features}).
On the main seeds this concern did not show up: with one action per response, the models tested chose
no worse by index than by name (\cref{tab:harness-compare}). %

\subsection{Notes and memory}
\label{app:methods-memory}
Run notes are rendered back to the model as \texttt{<note key="\ldots">} tags. The memory ablation
(\cref{tab:ablation-memory}) varies the mechanism while holding everything else fixed.
\emph{XML Notes + Memo} is the default of \cref{sec:methods-memory}: at the end of each run the model rewrites its
notes into the memory bank (\emph{+ Memo}). \emph{XML Notes + Keep Notes} copies each
run's notes into the bank verbatim, with no consolidation call, so it keeps persistence but removes
the rewriting. \emph{XML Notes} discards the notes at the end of each run, and \emph{No notes or memory} writes
nothing within a run and keeps no bank, so that the notes themselves can be measured.
In the \emph{Compact Notes} arms the
model instead writes a single prose summary through a one-argument \texttt{compact} tool, overwriting
it at the end of every floor. \emph{Compact Notes} discards it at the end of each run; \emph{Compact Notes + Memo}
also keeps one summary of every run so far, which the model overwrites at the end of every run. Every prompt that asks
the model to write something down includes the same \emph{generic} guidance: distil insights rather
than log events, and record what went well, what went wrong, and what the model wishes it had known.
It names nothing specific to the game.

\Cref{tab:memory-arms} sets out what each arm keeps by default. An xml note stays until the model deletes or
overwrites it, whereas a compact summary is replaced at every boundary, so it keeps only what the model writes into it again.

\begin{table}[h]
\centering
\caption{\textbf{What each notes and memory arm keeps by default, between floors of a run and between runs.}
\keptmark{}:~information retained as written by default, unless the model deletes or overwrites it;
\partmark{}:~information can be retained, but only if the model actively rewrites it (not retained by default);
\lostmark{}:~information can not be retained.}
\label{tab:memory-arms}
\begin{adjustbox}{max width=\textwidth}
\small
\begin{tabular}{@{}l c >{\raggedright\arraybackslash}p{0.22\linewidth} c >{\raggedright\arraybackslash}p{0.36\linewidth}@{}}
\toprule
 & \multicolumn{2}{c}{Between floors} & \multicolumn{2}{c}{Between runs} \\
\cmidrule(lr){2-3} \cmidrule(l){4-5}
Arm & Retained & What happens & Retained & What happens \\
\midrule
No notes or memory & \lostmark & Nothing written & \lostmark & Nothing kept \\
XML Notes & \keptmark & Notes kept & \lostmark & Notes discarded \\
XML Notes + Keep Notes & \keptmark & Notes kept & \keptmark & Notes added to the bank verbatim; bank kept \\
XML Notes + Memo & \keptmark & Notes kept & \partmark & Model rewrites notes into memories; memories kept, notes discarded \\
Compact Notes & \partmark & Summary replaced & \lostmark & Summary discarded \\
Compact Notes + Memo & \partmark & Summary replaced & \partmark & Summary of all runs replaced; run's summary discarded \\
\bottomrule
\end{tabular}
\end{adjustbox}
\end{table}

A cohort with a read-write bank is a chain. Seed $k$ starts from the bank that seed $k-1$ left
behind, so the seeds are played in order and a cohort's bank is never shared with any other
configuration: each bank is partitioned by model, character, the full configuration tag, and the seed
list.

\subsection{Knowledge reskin}
\label{app:methods-reskin}
The names replaced are those of cards, relics, potions, enemies, events, powers, and keywords, the
character, and the map's node labels. Names are replaced through placeholders, so that one replacement can never trigger another, and each
replacement evokes the original's effect rather than its name. The harness's own text (the system
prompt, state descriptions, and tool results) is renamed; the model's own text is not, so a real game
name in a note is evidence of recognized prior knowledge rather than something to be laundered.

The five mappings of the reskin ablation (\cref{tab:ablation-reskin}) are all generated in the same
format, so that only the mapping file changes between arms. \emph{None} shows the real names.
\emph{Themed} is the default. \emph{Random} replaces every word with a pronounceable nonsense word,
consistently, so that multiword names keep their shared structure. \emph{Permuted} gives each entity
the real name of a different entity of the same kind (cards within their colour and type), which is
coherent but actively misleading. \emph{Permuted and themed} applies the same permutation in the
themed vocabulary. The permuted mappings keep every rule word that the themed mapping leaves
unrenamed, so the arms differ only in the names. \Cref{tab:reskin-examples} gives examples. %

\begin{table}[h]
\centering
\small
\caption{\textbf{Example entries of the five reskin mappings (\cref{tab:ablation-reskin}).} Under \emph{none} each
entity keeps the name in the second column.
Card names are shown verbatim.
Some in-game effects act on ``cards containing Strike''. Under a permutation the Strike family would no longer share a name the model can see, so in both permuted mappings its cards are marked with a dagger and the rule
refers to the mark. The themed reskin maps every Strike to Slash (and Random maps it to Drakae), so no marker is needed.
}
\label{tab:reskin-examples}
\begin{adjustbox}{max width=\textwidth}
\small
\begin{tabular}{llllll}
\toprule
Kind & None (real) & Themed & Random & Permuted & Permuted and themed \\
\midrule
Card & Bash & Trip Attack & Bek & Reckless Charge & Reckless Blow \\
Card & Strike & Slash & Drakae & \dag{}Defend & \dag{}Dodge \\
Mechanic & Block & Guard & Taedoast & Channel & Summon \\
Status & Vulnerable & Exposed & Baevimi & No Draw & Draw Lock \\
Relic & Burning Blood & Bloodied Chalice & Vriastild Tousorn & Fossilized Helix & Winged Statuette \\
Enemy & Gremlin Nob & Goblin Berserker & Slagloath Skial & Centurion & Guardian \\
Orb & Lightning Orb & Water Spirit & Sofoa Sem & Ruby Orb & Opal Spirit \\
\bottomrule
\end{tabular}
\end{adjustbox}
\end{table}

{%
\subsection[\texorpdfstring{Model-facing features}{Model-facing features}]{Model-facing features}
\label{app:harness-features}
This subsection lists everything the harness does that the model can see, beyond the outline in
\cref{app:methods-harness}. The harness never makes a strategic choice for the model.
When there is only one legal action (e.g. click Continue), or the action has no strategic content (e.g. gold rewards should always be picked up---there is only upside to doing so), the harness acts automatically without consulting the model and informs the model of what happened on the next call.
When the model makes an incorrect call, the harness corrects it if the intent is unambiguous, and otherwise refuses it with an explanation in the tool response.
Quoted messages are given as the model receives them.

\paragraph{What the model is shown.}
\begin{itemize}[leftmargin=*,itemsep=1pt,topsep=2pt]
\item \emph{Map.} The act map is drawn with labelled nodes (\texttt{Monster(5,2)}: floor~5,
column~2). Burning elites, which carry the Emerald Key, are marked, and the act boss is named. The
current floor is marked \texttt{>{}>} and the reachable next nodes \texttt{*}. On map screens the model
may revise its route plan before choosing where to navigate to next.
\item \emph{Player overview.} HP, gold, keys, the deck (identical copies compacted as e.g.
\texttt{3x Strike}), relics with their descriptions and counters (e.g.\ \texttt{(2/3)},
\texttt{(USED)}), and potions. This is omitted from a subsequent message when nothing in it has changed.
\item \emph{Combat turn.} Each enemy is shown with its index, HP, Block, intent (e.g.
\textit{Attack for 6\texttimes{}2}), and powers with their current descriptions. Dead enemies are removed and the
rest renumbered. The player's HP, Block, energy, powers, stance, and orbs are shown, and the hand lists
one card per line with cost, type, whether it targets, and why it cannot be played. The draw,
discard, and exhaust piles are listed in full when they are relevant (on a new turn, after a change
or a reshuffle, or when a card in hand refers to that pile) and as counts otherwise.
\item \emph{Card names.} Cards are referred to by name, not hand position. Copies that differ get a
qualifier, e.g.\ \texttt{[Strike (0 cost)]} when an effect (e.g. Snecko Eye, Mummified Hand) has changed a cost, and so does a card
whose value has grown past its base (Ritual Dagger, Genetic Algorithm).
\item \emph{Effects.} Each tool call yields its own response containing the change in game state that resulted from the tool call, e.g.
\textit{``Played Bash. Energy -2 (3$\to$1). Enemy [0] Jaw Worm HP -8 (40$\to$32); Vulnerable +2.''}
Cards drawn, created, and exhausted are named. Between turns, the next message reports the end of the
player's turn, then one line per acting enemy, then the end of the round (powers/statuses decaying or
expiring) and the start of the new turn (Block kept, start-of-turn relics, reshuffles).
These are split using game-state snapshots from the mod (\cref{app:methods-game}). Where several
effects land together (the end of a fight, a room, a selection screen), the harness reports the net
change without naming a cause, leaving the attribution to the model, which can see its relics.
\item \emph{Non-combat screens.} Each screen lists its legal choices with what the model needs to
choose: shop prices, with \textit{(can't afford)} on items above the current gold value;
the HP a rest would heal (without indicating if the heal would be clipped by reaching the max HP);
card, relic, and potion rewards as one numbered list;
event options with \textit{(DISABLED)} marked on non-applicable options as per the in-game interface.
Each choice is confirmed with what it
changed, e.g.\ \textit{``Chose to rest. HP +24 (\ldots)''}.
\item \emph{Reward screens.} 
When there are multiple card rewards as part of the same reward screen, the model gets to see all options for all card rewards before choosing how to resolve each.
\item \emph{Floor history.} One line per floor played so far (room type, encounter, turns taken, HP and
gold change, cards added, upgraded, and removed, relics gained), with the act's route plan and the
run notes.
\end{itemize}

\paragraph{Reading the model's calls.}
The harness interprets the model's tool calls and converts them into in-game actions as follows.
\begin{itemize}[leftmargin=*,itemsep=1pt,topsep=2pt]
\item \emph{Lenient matching.} Card, potion, relic, and option names are matched case-insensitively,
with brackets, stray quotes, parentheses, and non-breaking spaces removed. A qualified name (e.g.\ \texttt{[Strike (0 cost)]})
matches if its qualifiers match the card's current values (e.g.\ only matches a 0 cost Strike and not a 1 or 2 cost Strike).
A bare name resolves to the upgraded copy when only that copy is in hand (\texttt{Bash} $\to$ \texttt{Bash+}), but \textit{not} the reverse.
Numbers given as strings are converted.
Under the reskin, names are mapped back to the game's names when the model is prompted, and then the tool call names are inverted and checked against what is on screen to create the in-game command.
\item \emph{Deferred card resolution.} The player's hand may have multiple copies of a card bearing the same name.
Earlier tool calls in the same message may play or discard some copies, or more copies may enter the player's hand.
The harness plays any copy of the card which is in hand when resolving a given action in the tool call queue.
\item \emph{Targets.} An attack or potion that needs a target should have that target specified by the model.
If omitted, it is automatically resolved to the only living enemy when there is one, and is refused otherwise.
Enemy indices refer to the list the model was shown, not the game's internal one, which may differ in the event that an enemy dies mid-turn.
If a targeted enemy has died, the refusal says that the remaining enemies have been renumbered.
\item \emph{Pre-targeted selections.} A card that makes the player choose another card (Armaments,
Headbutt, a discard) can name its target in the same call (\texttt{target\_card\_names}).
In-game, these card plays trigger a card selection screen, which the harness resolves using the specified target.
This enables the batch of tool calls to continue without interrupting to reprompt the model for the selection screen.
Similarly, the \texttt{choose\_by\_name(name, target\_name)} tool specifies to buy a card removal, or picks Smith or Toke at a rest site, while naming the target card in a single tool call.
\item \emph{Aliases.} \texttt{choose\_by\_name("skip")} or \texttt{("leave")} means skip, and \texttt{("proceed")} means proceed.
A number given as a name to \texttt{choose\_by\_name} is treated as an index to \texttt{choose\_by\_index}.
Phrases like \textit{``remove''} map to the shop's card removal.
An event option can also be named by its effect text when exactly one option matches.
\item \emph{Calls written as text.} When a response contains no tool call, the harness reads compliant calls written in its output text instead: function-call notation (\texttt{play\_card(card\_name="Strike")}), JSON objects in the shapes models commonly use, a bare argument object that fits exactly one expected tool, or \texttt{key: value} lines when one tool is required.
It reads the reasoning only when there is no visible text, and then only the final draft of the turn.
It does not read a response that was cut off.
A recovered call is executed like any other and counted as an error in \cref{tab:tool-errors}.
\end{itemize}

\paragraph{Corrections and refusals.} Tool calls can be rejected by the harness if they do not correspond to the game's state.
A refused call costs nothing in the game, but costs us an additional API call.
The rejected call's result says why it was rejected, and usually a hint to indicate what would be accepted on the current screen as follows:
\begin{itemize}[leftmargin=*,itemsep=1pt,topsep=2pt]
\item \emph{Did you mean.} A requested card not in hand is refused, along with the cards that are in hand, plus an example tool call for the closest
name in hand if one is close enough (fuzzy match, similarity at least 0.6). For example, the model tries to play Strike (0 cost) when it does not have one, resulting in
\textit{``Did you mean `Strike (1 cost)'? Example: play\_card(card\_name="Strike (1 cost)",
enemy\_index=0)''}.
Names that match more than one card are refused with the candidates shown in response.
In preliminary experiments, these suggestions stopped models from simply repeating the same refused call again without modification. %
\item \emph{Wrong tool or screen.} A tool that doesn't match the current screen is refused with the tools that do:
\textit{``not available on \{SCREEN\}. Valid tools here: [\ldots]''}.
The same applies to an action the game reports as unavailable (e.g.\ too little energy), to a disabled event option, and
to an unaffordable shop item.
\item \emph{Energy check.} \texttt{end\_turn} takes, as a required argument, the energy the model expects to have left at the end of its turn.
If the model ends its turn with more energy than it declared while a card in hand is still playable, the end of turn is refused: \textit{``Rejected end\_turn: declared expected\_energy\_remaining=$E$, but you actually have $A$ energy. Your turn plan didn't match what
happened\ldots''}.
The model can then play more cards or end the turn again.
This catches a model that has lost track of its own turn.
For example, the relic Nunchaku gives an energy for every tenth Attack played; a model that does not notice the tenth Attack coming would end its turn with energy it could have spent.
These miscounts are reported as tool errors (\cref{tab:tool-errors}). %
\item \emph{Actions the game would ignore.} The game silently ignores some actions, which would leave the harness waiting for a state change that never comes.
These are refused before they are sent: buying or picking up a potion with every potion slot full, drinking a potion the game marks as unusable here, taking a potion under Sozu, and buying a second card removal in one shop visit.
The refusal names what is possible instead.
\item \emph{Event fallback.} If a named choice matches no option on an event, the event is shown
again with both indices and labels, and \texttt{choose\_by\_index} is offered. Events whose options
share a label (Memory Stone) are numbered from the start.
\item \emph{Malformed arguments.} A required argument left as a \texttt{\ldots} placeholder, a
missing argument, or a text argument under the wrong name is refused with an example of the correct
call. A note given under a sibling field name is accepted, and the confirmation says so.
\item \emph{No tool call.} A response with no executable call gets one retry:
\textit{``Your reply contained no tool call. Text describing a call, whether JSON or
play\_card(\ldots) notation, is not a call. Call one or more of these tools now: \ldots''}. The
model's own reasoning is quoted back above this message. If the response was cut off at the length
limit, the message says that nothing in it was executed.
\item \emph{Game errors.} As a fallback, an error from the mod is passed on as \textit{``Action failed:
\{error\}. Choose a different action.''}, with the player's remaining energy added for an unplayable
card. The harness does not retry the action itself.
\item \emph{Repeated failure.} After two batches in a row on the same screen with every call
refused, the full screen state is sent again with the available tools.
\end{itemize}

\paragraph{Batches and interruptions.} The model can send multiple tool calls in response to the same prompt.
The model's calls are executed one at a time, and the game state is read after each.
The rest of the batch is dropped, with a result for every dropped call, when:
\begin{itemize}[leftmargin=*,itemsep=1pt,topsep=2pt]
\item a call is refused (\textit{``Skipped: an earlier action in your batch was rejected\ldots\ Your
planned sequence is stale\ldots''});
\item cards are drawn or created mid-turn that the plan could not have foreseen (cards that fetch a named card do not break the action chain, but ones which draw a random card do);
\item an enemy's intent changes to a different move;
\item a Lightning orb hits a random enemy while more than one is alive;
\item a selection screen opens that the plan did not account for;
\item the fight or the floor ends.
\end{itemize}
Calls after \texttt{end\_turn} are dropped as stale. Note writes always run. A batch with a single
card play gets a reminder to send the whole turn in one response, unless that card draws cards or
creates random ones, which the model cannot plan past. A card that makes the player choose another card
does not excuse the batch, since the choice can be named in the same call.

\paragraph{Actions the harness takes by itself.}
\begin{itemize}[leftmargin=*,itemsep=1pt,topsep=2pt]
\item \emph{Ending the turn} when no card is playable and no potion usable
(\textit{``No more playable options. Turn ended automatically.''}).
\item \emph{Single legal choices.} The harness takes an event window with one option, a single
reachable map node, and a selection screen where the number of eligible cards equals the number to
pick. The skipped event windows are shown to the model at the start of its next call, with what
they changed.
\item \emph{Screens with no choice.} The harness proceeds when proceeding is the only action,
confirms a selection once the model has picked, opens chests (except with Cursed Key), and enters a
shop it has walked into. It leaves a shop when nothing is affordable, or after 30 actions in it.
\item \emph{Rewards with no trade-off.} Gold, stolen gold and the Emerald Key are collected
automatically, as are potions when there is capacity to pick them all up (and the player does not have Entropic Brew).
\item \emph{Waiting for the game} to deal the opening hand, finish a transform, or show a purchase
before reporting it, so a reported effect is never one the game has not yet applied.
\end{itemize}

\paragraph{Guards.} Ten consecutive failed model calls, or ten consecutive responses with no
executable game action, end the run as a harness failure, and it is resumed from its autosave.
Over the main runs (200 runs and 37{,}306 model calls) the first fired three times, each when an API
account ran out of credit, and the second 16 times, all for Grok~4.6 on five of its seeds, each after ten
responses that chose an option with an empty name. Transient API errors are retried with
exponential backoff (five retries by default); 48 calls (0.13\%) still failed after their retries.
\texttt{play\_combo} repeats a card sequence for
an infinite combo, up to 1{,}000 times or until a pass changes nothing; it was called 149 times, and neither limit
was reached. Reflection is capped at 20
turns and consolidation at 200; the longest took 13 and 47. When a conversation outgrows the model's context window, the
oldest turns of the floor are dropped (always keeping the floor's first message and the latest turn)
and a note tells the model how many were removed. This happened on 475 calls (1.3\%), all from three models: Gemma~4 E4B, run with a 32k-token window (321 of its
820 calls, in every run), and two models with 128k-token windows, Nemotron~3 Nano Omni (139 of its 708 calls, in
every run) and gpt-oss-120b (15 calls, in three runs).

\subsection[\texorpdfstring{What the model receives}{What the model receives}]{What the model receives}
\label{app:harness-prompts}
\label{app:harness-prompts-full}
The samples below are quoted as sent, in the order the model meets them: the system prompt, tool
definitions, a combat turn, the end of that floor, the opening of a later floor, the end of the run, and the
memories carried into the next run. All come from one model's first run with the
game's real names, so that they
can be read without the reskin's vocabulary. In the main cohort the text is the same with every
name replaced, and each list of the compendium sorted by its new names. Lines longer than the page are wrapped
with an indent.

\subsubsection{System prompt}\label{app:prompt-system}
A short preamble states the task, followed by the compendium: the game's
rules and tips (40 points adapted from the game's own tutorial and loading-screen tips), its keywords
and status effects, and the character's cards, the colourless, status, and curse cards, and every relic
and potion with its effect.
The descriptions of keywords, status effects, cards, relics, and potions are taken from the community
Slay the Spire Wiki (\url{https://slaythespire.wiki.gg}), whose page content is licensed under
CC~BY-SA~4.0. On the first seed, before any memory exists, this is 29{,}513 characters: the
preamble is 591, the rules and tips 2{,}763, the keywords and status effects 1{,}393, the cards 13{,}118, the
relics 9{,}691, and the potions 1{,}951. With the tool definitions and the first message, the first call was
12{,}291 input tokens. Beyond the preamble the prompt holds no instruction of ours. What an upgrade does to a
card is given on that card's line (\texttt{Deal 6($\to$9) damage});
the notation used for upgrades is shown at the start of the cards section.
The runs on the evaluation seeds (\cref{tab:ablation-prompt,tab:ablation-prompt-by-model}) write no memory bank,
so their system prompt differs from the one shown here in two places: the preamble leaves out the three sentences
telling the model that it will play the game several times and should learn on each run to win more
consistently, and no memory section ever
follows the compendium. In the Bestiary and Reddit-guide arms, the added material is also part of the
system prompt (\cref{app:prompt-material}).

{\parskip=0pt\parindent=0pt\offinterlineskip
\includegraphics[width=\linewidth,viewport=0 4776.24 387.84 4837.92,clip]{figures/prompt_system.png}\par\vskip-0.24bp
\includegraphics[width=\linewidth,viewport=0 4721.52 387.84 4776.48,clip]{figures/prompt_system.png}\par\vskip-0.24bp
\includegraphics[width=\linewidth,viewport=0 4666.80 387.84 4721.76,clip]{figures/prompt_system.png}\par\vskip-0.24bp
\includegraphics[width=\linewidth,viewport=0 4612.08 387.84 4667.04,clip]{figures/prompt_system.png}\par\vskip-0.24bp
\includegraphics[width=\linewidth,viewport=0 4557.36 387.84 4612.32,clip]{figures/prompt_system.png}\par\vskip-0.24bp
\includegraphics[width=\linewidth,viewport=0 4502.64 387.84 4557.60,clip]{figures/prompt_system.png}\par\vskip-0.24bp
\includegraphics[width=\linewidth,viewport=0 4447.92 387.84 4502.88,clip]{figures/prompt_system.png}\par\vskip-0.24bp
\includegraphics[width=\linewidth,viewport=0 4393.20 387.84 4448.16,clip]{figures/prompt_system.png}\par\vskip-0.24bp
\includegraphics[width=\linewidth,viewport=0 4338.48 387.84 4393.44,clip]{figures/prompt_system.png}\par\vskip-0.24bp
\includegraphics[width=\linewidth,viewport=0 4283.76 387.84 4338.72,clip]{figures/prompt_system.png}\par\vskip-0.24bp
\includegraphics[width=\linewidth,viewport=0 4229.04 387.84 4284.00,clip]{figures/prompt_system.png}\par\vskip-0.24bp
\includegraphics[width=\linewidth,viewport=0 4174.32 387.84 4229.28,clip]{figures/prompt_system.png}\par\vskip-0.24bp
\includegraphics[width=\linewidth,viewport=0 4119.60 387.84 4174.56,clip]{figures/prompt_system.png}\par\vskip-0.24bp
\includegraphics[width=\linewidth,viewport=0 4064.88 387.84 4119.84,clip]{figures/prompt_system.png}\par\vskip-0.24bp
\includegraphics[width=\linewidth,viewport=0 4010.16 387.84 4065.12,clip]{figures/prompt_system.png}\par\vskip-0.24bp
\includegraphics[width=\linewidth,viewport=0 3955.44 387.84 4010.40,clip]{figures/prompt_system.png}\par\vskip-0.24bp
\includegraphics[width=\linewidth,viewport=0 3900.72 387.84 3955.68,clip]{figures/prompt_system.png}\par\vskip-0.24bp
\includegraphics[width=\linewidth,viewport=0 3846.00 387.84 3900.96,clip]{figures/prompt_system.png}\par\vskip-0.24bp
\includegraphics[width=\linewidth,viewport=0 3791.28 387.84 3846.24,clip]{figures/prompt_system.png}\par\vskip-0.24bp
\includegraphics[width=\linewidth,viewport=0 3736.56 387.84 3791.52,clip]{figures/prompt_system.png}\par\vskip-0.24bp
\includegraphics[width=\linewidth,viewport=0 3681.84 387.84 3736.80,clip]{figures/prompt_system.png}\par\vskip-0.24bp
\includegraphics[width=\linewidth,viewport=0 3627.12 387.84 3682.08,clip]{figures/prompt_system.png}\par\vskip-0.24bp
\includegraphics[width=\linewidth,viewport=0 3572.40 387.84 3627.36,clip]{figures/prompt_system.png}\par\vskip-0.24bp
\includegraphics[width=\linewidth,viewport=0 3517.68 387.84 3572.64,clip]{figures/prompt_system.png}\par\vskip-0.24bp
\includegraphics[width=\linewidth,viewport=0 3462.96 387.84 3517.92,clip]{figures/prompt_system.png}\par\vskip-0.24bp
\includegraphics[width=\linewidth,viewport=0 3408.24 387.84 3463.20,clip]{figures/prompt_system.png}\par\vskip-0.24bp
\includegraphics[width=\linewidth,viewport=0 3353.52 387.84 3408.48,clip]{figures/prompt_system.png}\par\vskip-0.24bp
\includegraphics[width=\linewidth,viewport=0 3298.80 387.84 3353.76,clip]{figures/prompt_system.png}\par\vskip-0.24bp
\includegraphics[width=\linewidth,viewport=0 3244.08 387.84 3299.04,clip]{figures/prompt_system.png}\par\vskip-0.24bp
\includegraphics[width=\linewidth,viewport=0 3189.36 387.84 3244.32,clip]{figures/prompt_system.png}\par\vskip-0.24bp
\includegraphics[width=\linewidth,viewport=0 3134.64 387.84 3189.60,clip]{figures/prompt_system.png}\par\vskip-0.24bp
\includegraphics[width=\linewidth,viewport=0 3079.92 387.84 3134.88,clip]{figures/prompt_system.png}\par\vskip-0.24bp
\includegraphics[width=\linewidth,viewport=0 3025.20 387.84 3080.16,clip]{figures/prompt_system.png}\par\vskip-0.24bp
\includegraphics[width=\linewidth,viewport=0 2970.48 387.84 3025.44,clip]{figures/prompt_system.png}\par\vskip-0.24bp
\includegraphics[width=\linewidth,viewport=0 2915.76 387.84 2970.72,clip]{figures/prompt_system.png}\par\vskip-0.24bp
\includegraphics[width=\linewidth,viewport=0 2861.04 387.84 2916.00,clip]{figures/prompt_system.png}\par\vskip-0.24bp
\includegraphics[width=\linewidth,viewport=0 2806.32 387.84 2861.28,clip]{figures/prompt_system.png}\par\vskip-0.24bp
\includegraphics[width=\linewidth,viewport=0 2751.60 387.84 2806.56,clip]{figures/prompt_system.png}\par\vskip-0.24bp
\includegraphics[width=\linewidth,viewport=0 2696.88 387.84 2751.84,clip]{figures/prompt_system.png}\par\vskip-0.24bp
\includegraphics[width=\linewidth,viewport=0 2642.16 387.84 2697.12,clip]{figures/prompt_system.png}\par\vskip-0.24bp
\includegraphics[width=\linewidth,viewport=0 2587.44 387.84 2642.40,clip]{figures/prompt_system.png}\par\vskip-0.24bp
\includegraphics[width=\linewidth,viewport=0 2532.72 387.84 2587.68,clip]{figures/prompt_system.png}\par\vskip-0.24bp
\includegraphics[width=\linewidth,viewport=0 2478.00 387.84 2532.96,clip]{figures/prompt_system.png}\par\vskip-0.24bp
\includegraphics[width=\linewidth,viewport=0 2423.28 387.84 2478.24,clip]{figures/prompt_system.png}\par\vskip-0.24bp
\includegraphics[width=\linewidth,viewport=0 2368.56 387.84 2423.52,clip]{figures/prompt_system.png}\par\vskip-0.24bp
\includegraphics[width=\linewidth,viewport=0 2313.84 387.84 2368.80,clip]{figures/prompt_system.png}\par\vskip-0.24bp
\includegraphics[width=\linewidth,viewport=0 2259.12 387.84 2314.08,clip]{figures/prompt_system.png}\par\vskip-0.24bp
\includegraphics[width=\linewidth,viewport=0 2204.40 387.84 2259.36,clip]{figures/prompt_system.png}\par\vskip-0.24bp
\includegraphics[width=\linewidth,viewport=0 2149.68 387.84 2204.64,clip]{figures/prompt_system.png}\par\vskip-0.24bp
\includegraphics[width=\linewidth,viewport=0 2094.96 387.84 2149.92,clip]{figures/prompt_system.png}\par\vskip-0.24bp
\includegraphics[width=\linewidth,viewport=0 2040.24 387.84 2095.20,clip]{figures/prompt_system.png}\par\vskip-0.24bp
\includegraphics[width=\linewidth,viewport=0 1985.52 387.84 2040.48,clip]{figures/prompt_system.png}\par\vskip-0.24bp
\includegraphics[width=\linewidth,viewport=0 1930.80 387.84 1985.76,clip]{figures/prompt_system.png}\par\vskip-0.24bp
\includegraphics[width=\linewidth,viewport=0 1876.08 387.84 1931.04,clip]{figures/prompt_system.png}\par\vskip-0.24bp
\includegraphics[width=\linewidth,viewport=0 1821.36 387.84 1876.32,clip]{figures/prompt_system.png}\par\vskip-0.24bp
\includegraphics[width=\linewidth,viewport=0 1766.64 387.84 1821.60,clip]{figures/prompt_system.png}\par\vskip-0.24bp
\includegraphics[width=\linewidth,viewport=0 1711.92 387.84 1766.88,clip]{figures/prompt_system.png}\par\vskip-0.24bp
\includegraphics[width=\linewidth,viewport=0 1657.20 387.84 1712.16,clip]{figures/prompt_system.png}\par\vskip-0.24bp
\includegraphics[width=\linewidth,viewport=0 1602.48 387.84 1657.44,clip]{figures/prompt_system.png}\par\vskip-0.24bp
\includegraphics[width=\linewidth,viewport=0 1547.76 387.84 1602.72,clip]{figures/prompt_system.png}\par\vskip-0.24bp
\includegraphics[width=\linewidth,viewport=0 1493.04 387.84 1548.00,clip]{figures/prompt_system.png}\par\vskip-0.24bp
\includegraphics[width=\linewidth,viewport=0 1438.32 387.84 1493.28,clip]{figures/prompt_system.png}\par\vskip-0.24bp
\includegraphics[width=\linewidth,viewport=0 1383.60 387.84 1438.56,clip]{figures/prompt_system.png}\par\vskip-0.24bp
\includegraphics[width=\linewidth,viewport=0 1328.88 387.84 1383.84,clip]{figures/prompt_system.png}\par\vskip-0.24bp
\includegraphics[width=\linewidth,viewport=0 1274.16 387.84 1329.12,clip]{figures/prompt_system.png}\par\vskip-0.24bp
\includegraphics[width=\linewidth,viewport=0 1219.44 387.84 1274.40,clip]{figures/prompt_system.png}\par\vskip-0.24bp
\includegraphics[width=\linewidth,viewport=0 1164.72 387.84 1219.68,clip]{figures/prompt_system.png}\par\vskip-0.24bp
\includegraphics[width=\linewidth,viewport=0 1110.00 387.84 1164.96,clip]{figures/prompt_system.png}\par\vskip-0.24bp
\includegraphics[width=\linewidth,viewport=0 1055.28 387.84 1110.24,clip]{figures/prompt_system.png}\par\vskip-0.24bp
\includegraphics[width=\linewidth,viewport=0 1000.56 387.84 1055.52,clip]{figures/prompt_system.png}\par\vskip-0.24bp
\includegraphics[width=\linewidth,viewport=0 945.84 387.84 1000.80,clip]{figures/prompt_system.png}\par\vskip-0.24bp
\includegraphics[width=\linewidth,viewport=0 891.12 387.84 946.08,clip]{figures/prompt_system.png}\par\vskip-0.24bp
\includegraphics[width=\linewidth,viewport=0 836.40 387.84 891.36,clip]{figures/prompt_system.png}\par\vskip-0.24bp
\includegraphics[width=\linewidth,viewport=0 781.68 387.84 836.64,clip]{figures/prompt_system.png}\par\vskip-0.24bp
\includegraphics[width=\linewidth,viewport=0 726.96 387.84 781.92,clip]{figures/prompt_system.png}\par\vskip-0.24bp
\includegraphics[width=\linewidth,viewport=0 672.24 387.84 727.20,clip]{figures/prompt_system.png}\par\vskip-0.24bp
\includegraphics[width=\linewidth,viewport=0 617.52 387.84 672.48,clip]{figures/prompt_system.png}\par\vskip-0.24bp
\includegraphics[width=\linewidth,viewport=0 562.80 387.84 617.76,clip]{figures/prompt_system.png}\par\vskip-0.24bp
\includegraphics[width=\linewidth,viewport=0 508.08 387.84 563.04,clip]{figures/prompt_system.png}\par\vskip-0.24bp
\includegraphics[width=\linewidth,viewport=0 453.36 387.84 508.32,clip]{figures/prompt_system.png}\par\vskip-0.24bp
\includegraphics[width=\linewidth,viewport=0 398.64 387.84 453.60,clip]{figures/prompt_system.png}\par\vskip-0.24bp
\includegraphics[width=\linewidth,viewport=0 343.92 387.84 398.88,clip]{figures/prompt_system.png}\par\vskip-0.24bp
\includegraphics[width=\linewidth,viewport=0 289.20 387.84 344.16,clip]{figures/prompt_system.png}\par\vskip-0.24bp
\includegraphics[width=\linewidth,viewport=0 234.48 387.84 289.44,clip]{figures/prompt_system.png}\par\vskip-0.24bp
\includegraphics[width=\linewidth,viewport=0 179.76 387.84 234.72,clip]{figures/prompt_system.png}\par\vskip-0.24bp
\includegraphics[width=\linewidth,viewport=0 125.04 387.84 180.00,clip]{figures/prompt_system.png}\par\vskip-0.24bp
\includegraphics[width=\linewidth,viewport=0 70.32 387.84 125.28,clip]{figures/prompt_system.png}\par\vskip-0.24bp
\includegraphics[width=\linewidth,viewport=0 0.00 387.84 70.56,clip]{figures/prompt_system.png}\par\vskip-0.24bp
}

\subsubsection{Tools}\label{app:prompt-tools}
{\raggedright Seventeen tools are sent on every call (required arguments marked *):
\texttt{play\_card(card\_name*, enemy\_index, target\_card\_names)},
\texttt{play\_combo(card\_names*, enemy\_index)},
\texttt{end\_turn(expected\_energy\_remaining*, predicted\_hp\_after\_turn, predicted\_hp\_after\_fight)},
\texttt{use\_potion(potion\_name*, enemy\_index)},
\texttt{discard\_potion(potion\_name*)},
\texttt{choose\_by\_index(choice\_index*)},
\texttt{choose\_by\_name(name*, target\_name)},
\texttt{choose\_path(choice\_index*)},
\texttt{skip()},
\texttt{plan\_route(route*, notes*)},
\texttt{predict(act\_win\_probability*, run\_win\_probability*)},
\texttt{save\_note(key*, note*)},
\texttt{delete\_note(key*)},
\texttt{finish\_reflection()},
\texttt{save\_memory(key*, memory*)},
\texttt{delete\_memory(key*)},
\texttt{finish\_consolidation()}.\par}
The definitions follow in full. Each tool's description is as sent; its arguments are listed
one per line with their type and description. The description of
\texttt{save\_note} informs the model that notes will outlast a floor and need to include information useful for future runs too:
notes ``persist for the rest of this run, and will be used to update your permanent memories at the end of the run''.

{\parskip=0pt\parindent=0pt\offinterlineskip
\includegraphics[width=\linewidth,viewport=0 1265.04 387.84 1326.72,clip]{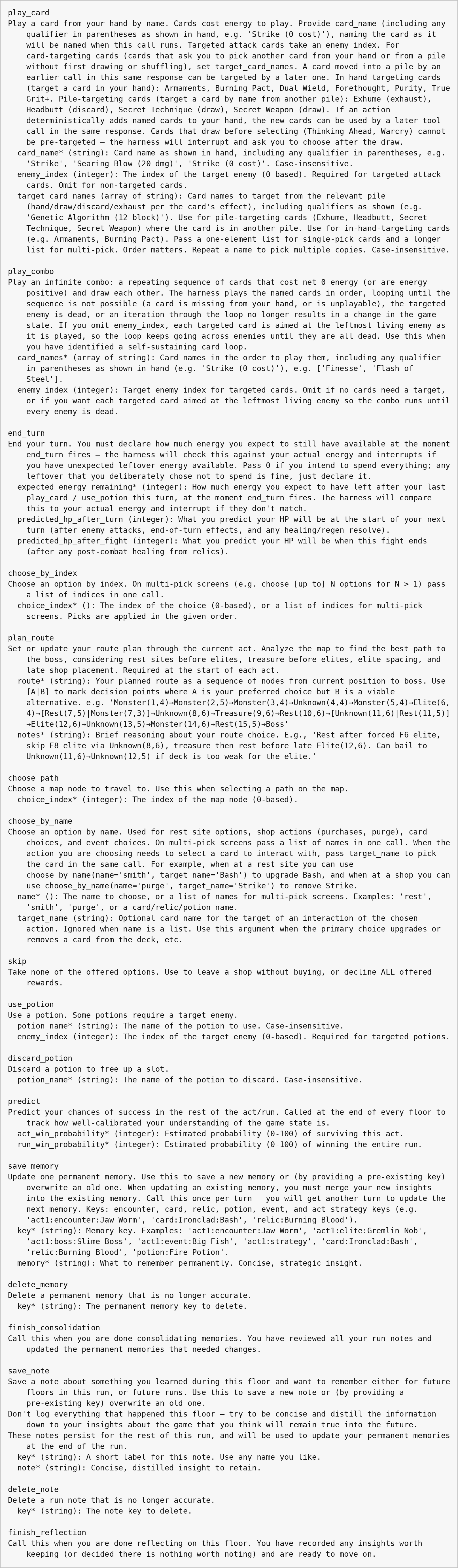}\par\vskip-0.24bp
\includegraphics[width=\linewidth,viewport=0 1210.32 387.84 1265.28,clip]{figures/prompt_tools.png}\par\vskip-0.24bp
\includegraphics[width=\linewidth,viewport=0 1155.60 387.84 1210.56,clip]{figures/prompt_tools.png}\par\vskip-0.24bp
\includegraphics[width=\linewidth,viewport=0 1100.88 387.84 1155.84,clip]{figures/prompt_tools.png}\par\vskip-0.24bp
\includegraphics[width=\linewidth,viewport=0 1046.16 387.84 1101.12,clip]{figures/prompt_tools.png}\par\vskip-0.24bp
\includegraphics[width=\linewidth,viewport=0 991.44 387.84 1046.40,clip]{figures/prompt_tools.png}\par\vskip-0.24bp
\includegraphics[width=\linewidth,viewport=0 936.72 387.84 991.68,clip]{figures/prompt_tools.png}\par\vskip-0.24bp
\includegraphics[width=\linewidth,viewport=0 882.00 387.84 936.96,clip]{figures/prompt_tools.png}\par\vskip-0.24bp
\includegraphics[width=\linewidth,viewport=0 827.28 387.84 882.24,clip]{figures/prompt_tools.png}\par\vskip-0.24bp
\includegraphics[width=\linewidth,viewport=0 772.56 387.84 827.52,clip]{figures/prompt_tools.png}\par\vskip-0.24bp
\includegraphics[width=\linewidth,viewport=0 717.84 387.84 772.80,clip]{figures/prompt_tools.png}\par\vskip-0.24bp
\includegraphics[width=\linewidth,viewport=0 663.12 387.84 718.08,clip]{figures/prompt_tools.png}\par\vskip-0.24bp
\includegraphics[width=\linewidth,viewport=0 608.40 387.84 663.36,clip]{figures/prompt_tools.png}\par\vskip-0.24bp
\includegraphics[width=\linewidth,viewport=0 553.68 387.84 608.64,clip]{figures/prompt_tools.png}\par\vskip-0.24bp
\includegraphics[width=\linewidth,viewport=0 498.96 387.84 553.92,clip]{figures/prompt_tools.png}\par\vskip-0.24bp
\includegraphics[width=\linewidth,viewport=0 444.24 387.84 499.20,clip]{figures/prompt_tools.png}\par\vskip-0.24bp
\includegraphics[width=\linewidth,viewport=0 389.52 387.84 444.48,clip]{figures/prompt_tools.png}\par\vskip-0.24bp
\includegraphics[width=\linewidth,viewport=0 334.80 387.84 389.76,clip]{figures/prompt_tools.png}\par\vskip-0.24bp
\includegraphics[width=\linewidth,viewport=0 280.08 387.84 335.04,clip]{figures/prompt_tools.png}\par\vskip-0.24bp
\includegraphics[width=\linewidth,viewport=0 225.36 387.84 280.32,clip]{figures/prompt_tools.png}\par\vskip-0.24bp
\includegraphics[width=\linewidth,viewport=0 170.64 387.84 225.60,clip]{figures/prompt_tools.png}\par\vskip-0.24bp
\includegraphics[width=\linewidth,viewport=0 115.92 387.84 170.88,clip]{figures/prompt_tools.png}\par\vskip-0.24bp
\includegraphics[width=\linewidth,viewport=0 61.20 387.84 116.16,clip]{figures/prompt_tools.png}\par\vskip-0.24bp
\includegraphics[width=\linewidth,viewport=0 0.00 387.84 61.44,clip]{figures/prompt_tools.png}\par\vskip-0.24bp
}

\subsubsection{A floor's opening message}\label{app:prompt-floor-opening}
Each floor begins a new conversation, which opens with
the history of the run so far (one line per floor), the rest of the act's map (pruned to the nodes still accessible),
the route plan, the run notes, the player's state, and the screen to act on. The example is the rest site on
floor~8, where the choice is between healing and upgrading a card.
The model is not told in so many words that the conversation ends with the floor; it is told what
persists. The reflection that closes each floor asks for anything ``worth remembering for later floors or
future runs'', the tool that saves a note says that notes ``persist for the rest of this run'', and every
floor opens with them under ``Your notes from this run. Updated after each floor.''
For brevity, the message below is cut after the first two of the model's twelve notes; the model was shown all twelve.

{\parskip=0pt\parindent=0pt\offinterlineskip
\includegraphics[width=\linewidth,viewport=0 690.48 387.84 752.16,clip]{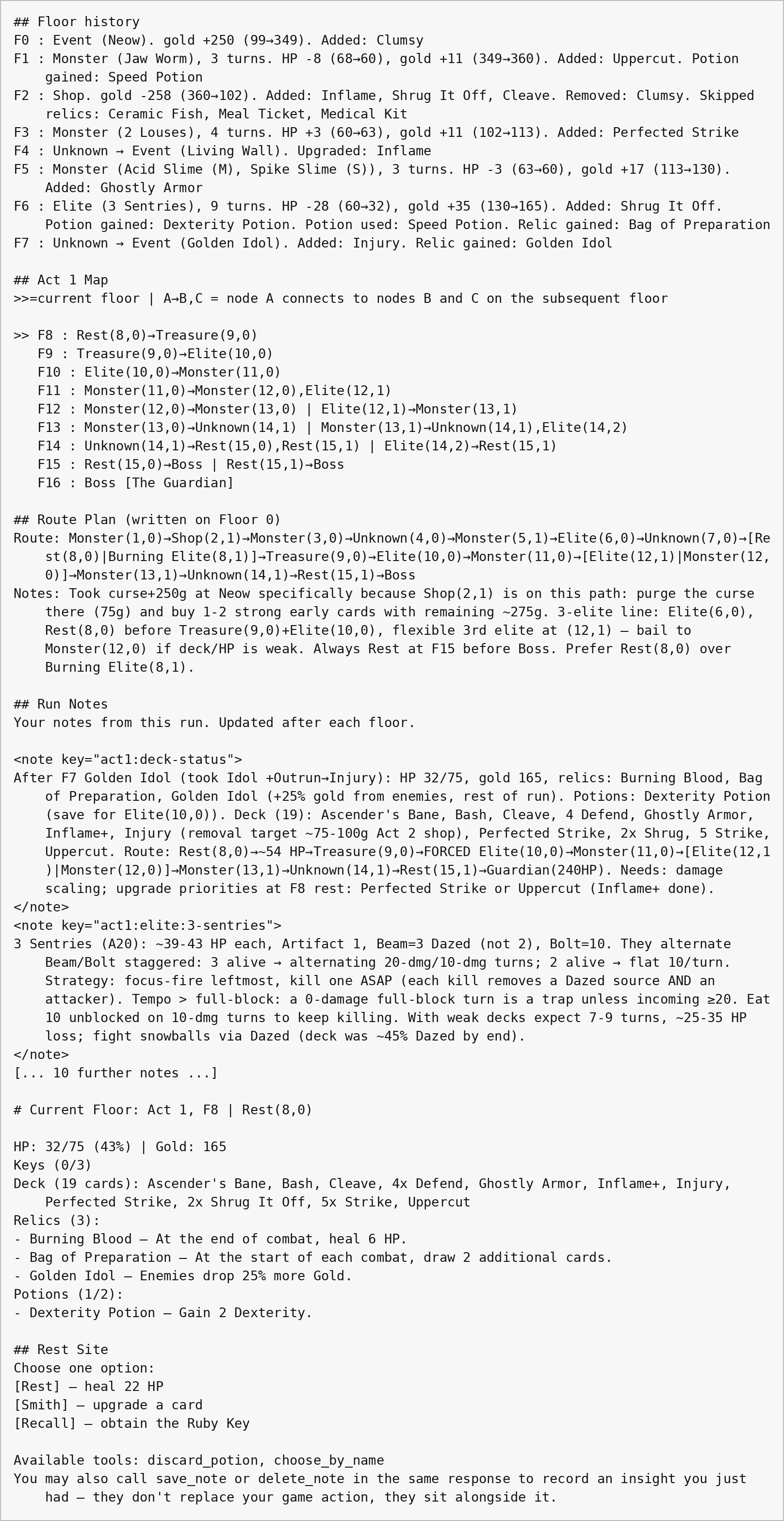}\par\vskip-0.24bp
\includegraphics[width=\linewidth,viewport=0 635.76 387.84 690.72,clip]{figures/prompt_floor_opening.png}\par\vskip-0.24bp
\includegraphics[width=\linewidth,viewport=0 581.04 387.84 636.00,clip]{figures/prompt_floor_opening.png}\par\vskip-0.24bp
\includegraphics[width=\linewidth,viewport=0 526.32 387.84 581.28,clip]{figures/prompt_floor_opening.png}\par\vskip-0.24bp
\includegraphics[width=\linewidth,viewport=0 471.60 387.84 526.56,clip]{figures/prompt_floor_opening.png}\par\vskip-0.24bp
\includegraphics[width=\linewidth,viewport=0 416.88 387.84 471.84,clip]{figures/prompt_floor_opening.png}\par\vskip-0.24bp
\includegraphics[width=\linewidth,viewport=0 362.16 387.84 417.12,clip]{figures/prompt_floor_opening.png}\par\vskip-0.24bp
\includegraphics[width=\linewidth,viewport=0 307.44 387.84 362.40,clip]{figures/prompt_floor_opening.png}\par\vskip-0.24bp
\includegraphics[width=\linewidth,viewport=0 252.72 387.84 307.68,clip]{figures/prompt_floor_opening.png}\par\vskip-0.24bp
\includegraphics[width=\linewidth,viewport=0 198.00 387.84 252.96,clip]{figures/prompt_floor_opening.png}\par\vskip-0.24bp
\includegraphics[width=\linewidth,viewport=0 143.28 387.84 198.24,clip]{figures/prompt_floor_opening.png}\par\vskip-0.24bp
\includegraphics[width=\linewidth,viewport=0 88.56 387.84 143.52,clip]{figures/prompt_floor_opening.png}\par\vskip-0.24bp
\includegraphics[width=\linewidth,viewport=0 33.84 387.84 88.80,clip]{figures/prompt_floor_opening.png}\par\vskip-0.24bp
\includegraphics[width=\linewidth,viewport=0 0.00 387.84 34.08,clip]{figures/prompt_floor_opening.png}\par\vskip-0.24bp
}

\subsubsection{A combat turn}\label{app:prompt-combat-turn}
The message that opens a turn, here turn~3 of the elite fight on floor~6. The
effect of each card played is returned as the result of its own tool call (e.g.\ \textit{``Played Perfected
Strike. Energy -2 (3$\to$1). Enemy [0] Sentry HP -18 (35$\to$17).''}), so the message begins with what the
enemies then did.
\begin{center}
\includegraphics[width=\linewidth]{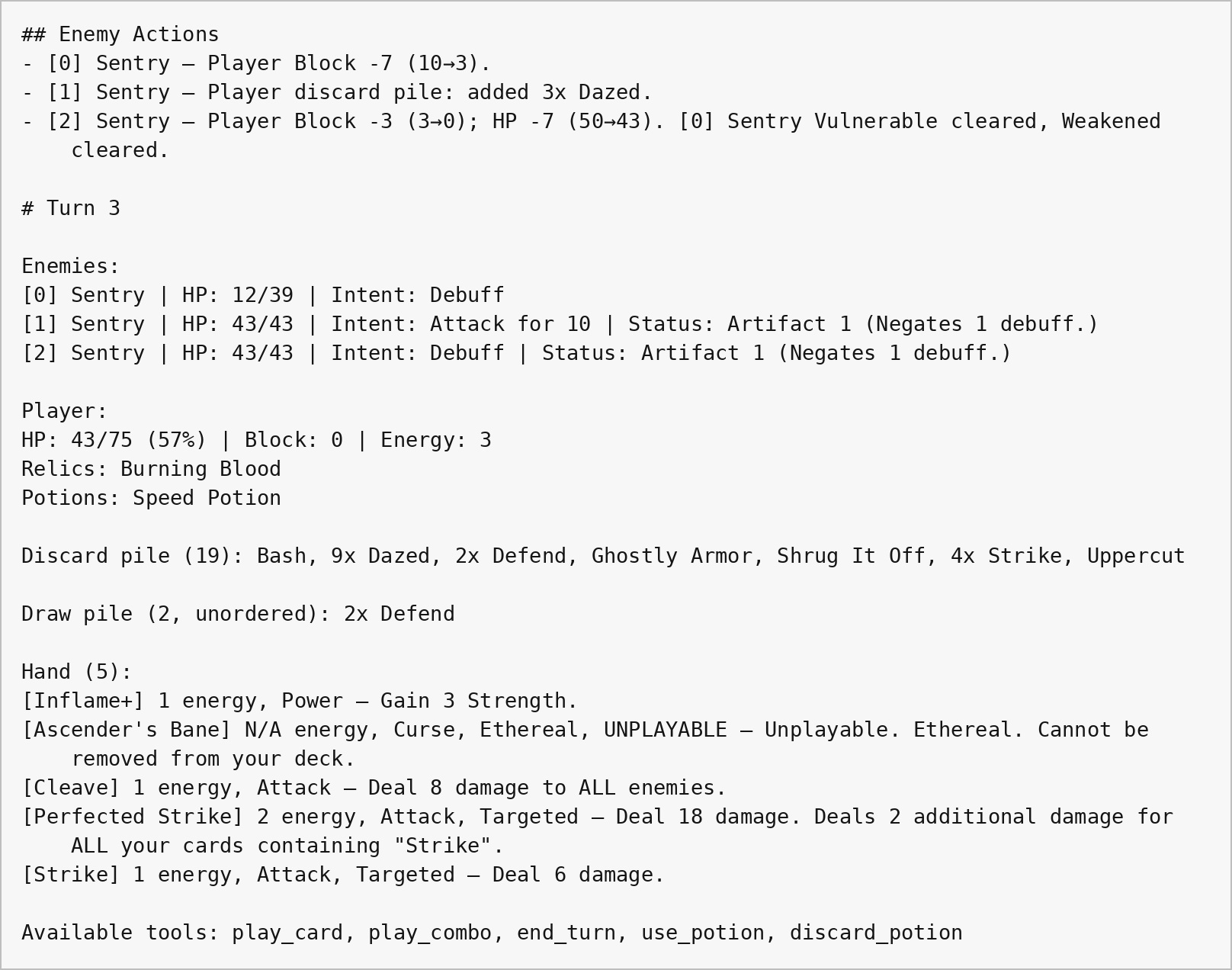}
\end{center}

\subsubsection{End of a floor}\label{app:prompt-floor-end}
After the floor's last screen the model is asked to reflect and update its notes (its memory within the run),
then to forecast its chances of success (used by us for calibration only). These are the two messages that closed floor~6:
\begin{center}
\includegraphics[width=\linewidth]{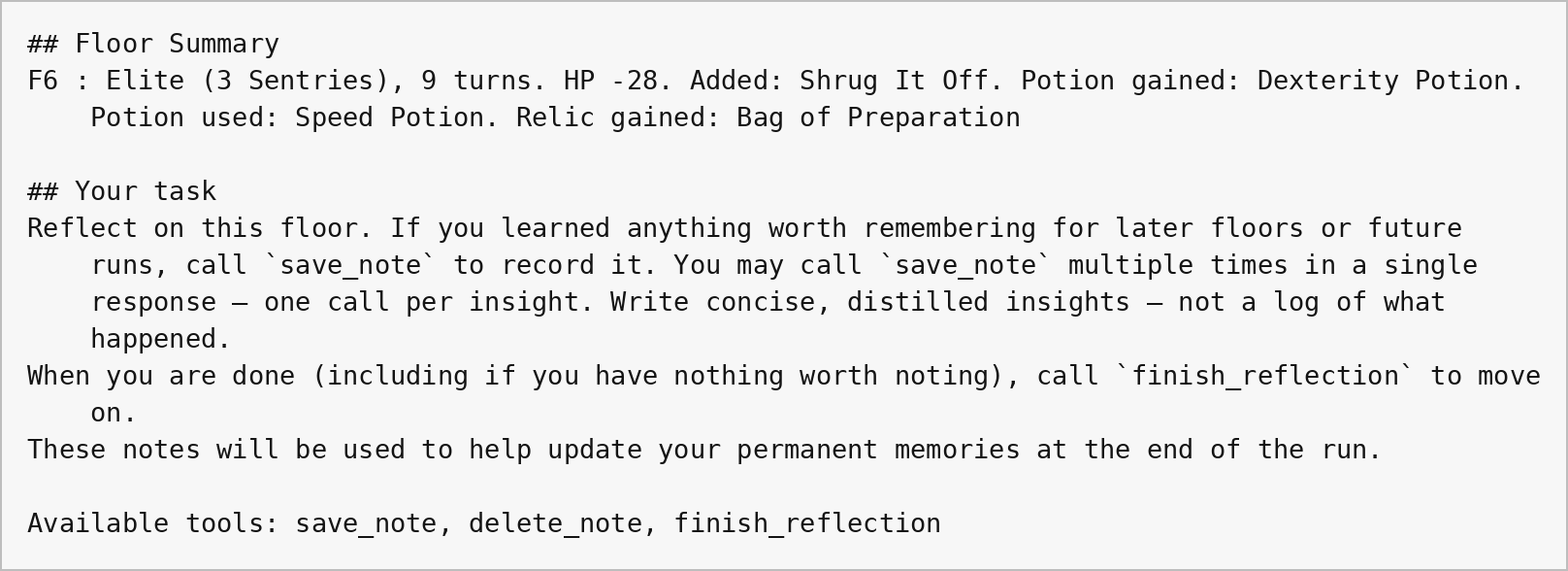}
\end{center}
\begin{center}
\includegraphics[width=\linewidth]{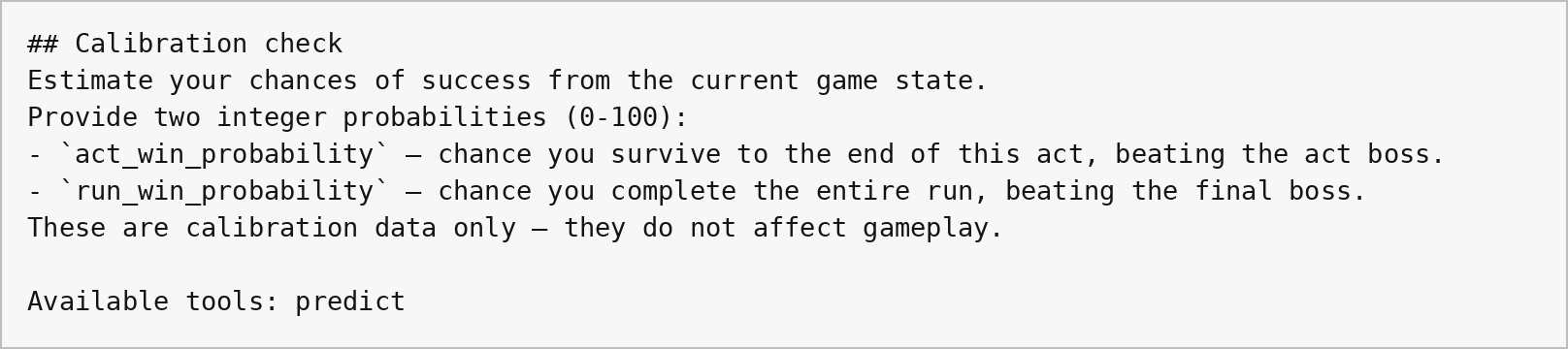}
\end{center}

\subsubsection{End of the run}\label{app:prompt-run-end}
The run ended on floor~46. The consolidation message consists of
the run's result and final state, each act's full map, route plan, and path taken,
the history of every floor, and the run notes, and ends:
\begin{center}
\includegraphics[width=\linewidth]{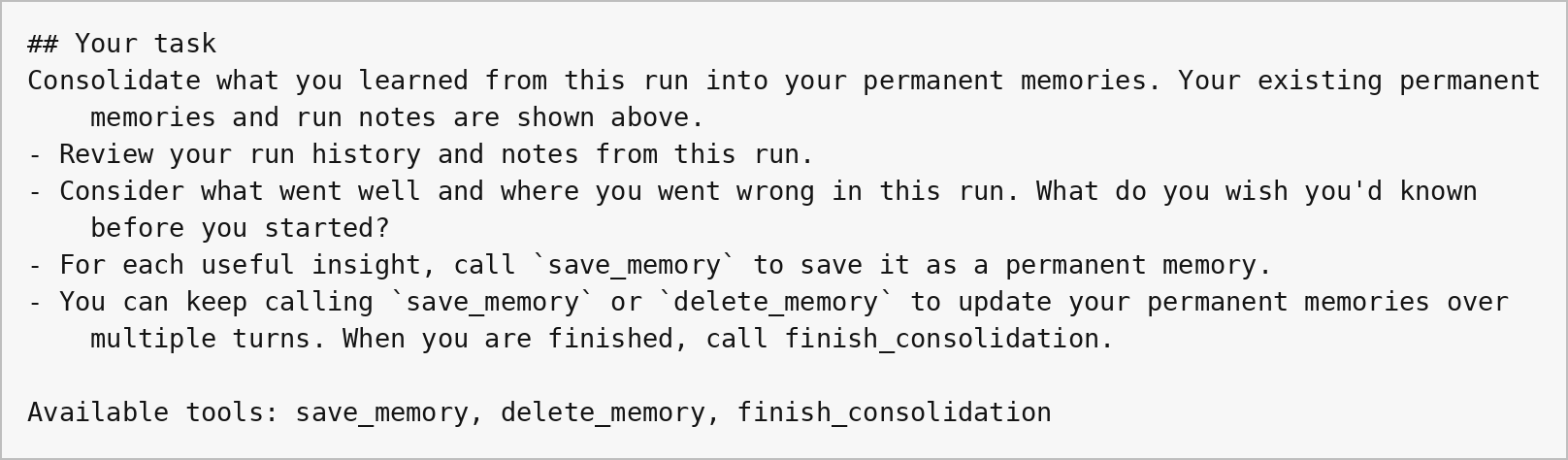}
\end{center}

\subsubsection{The next run}\label{app:prompt-next-run}
From the second seed on, the memory bank follows the compendium as the last section
of the system prompt. In this example it held 15 memories at the start of the second seed, and the prompt had grown to
41{,}685 characters; by the tenth seed this model's bank held 64 memories (91{,}090
characters).

{\parskip=0pt\parindent=0pt\offinterlineskip
\includegraphics[width=\linewidth,viewport=0 188.88 387.84 250.56,clip]{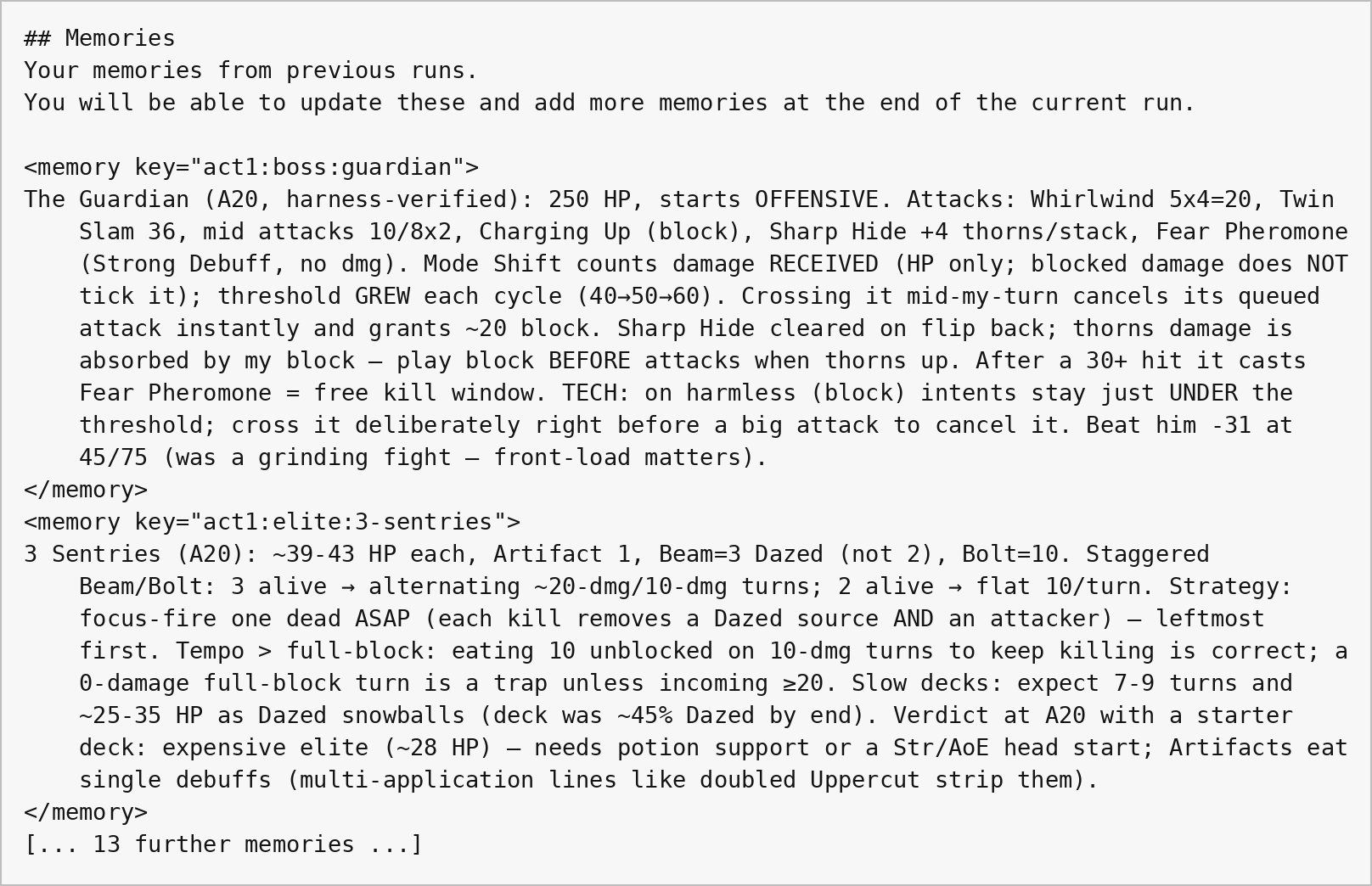}\par\vskip-0.24bp
\includegraphics[width=\linewidth,viewport=0 134.16 387.84 189.12,clip]{figures/prompt_memories.png}\par\vskip-0.24bp
\includegraphics[width=\linewidth,viewport=0 79.44 387.84 134.40,clip]{figures/prompt_memories.png}\par\vskip-0.24bp
\includegraphics[width=\linewidth,viewport=0 0.00 387.84 79.68,clip]{figures/prompt_memories.png}\par\vskip-0.24bp
}

\subsection{Coverage compared with Orak and Bottled AI}
\label{app:harness-coverage}
Orak's own rule bot plays the map, events, shops, rest sites, chests, boss relics, Neow, and
potions, so its model decides combat and card rewards only. Even within a fight its model decides less than ours.
Orak asks the model only when a card can be played,
so every card choice that a card or potion opens mid-fight (Armaments, True Grit+, Dual Wield, Headbutt,
Exhume, an Attack Potion) goes to the rule bot, which picks from a fixed card ranking rather than the fight
at hand: the best-ranked card to upgrade or copy, the worst-ranked to exhaust, and, since the flag that
would select the best is never set, the worst-ranked card for Headbutt and Exhume. The model is not told
which card was chosen, and the rest of its planned turn is dropped. Orak's agent returns
several actions at once, as ours does, but plays each card by its position in the hand as shown when the
model decided, so a response cannot play a card that arrives mid-turn. Queued actions keep executing through
draws and random effects, re-resolved by card identity so a shifted hand still plays the intended card, but
the model sees what was drawn only once its queue is spent; Orak's paper describes the state
as returned after each action, where its code returns it after the queue. Ours puts every such choice to the
model, lets a card-targeting card name its target in the same call, and interrupts the queue when cards are
drawn or created, orbs evoke at random, or an enemy's intent changes, naming what
changed~\citep{orak}.

\paragraph{Game coverage.} \Cref{tab:feature-coverage} goes through the game by feature. It refines
the \emph{Model plays} and \emph{Choices} columns of \cref{tab:harness-compare}, and adds Bottled AI.
Orak's model sees no orbs, stance, potions, or card piles, so a Defect under Orak plays its orb cards
without seeing the orbs they make. Orak builds on spirecomm, a Python client for CommunicationMod
whose last commit predates the Watcher, and so cannot play the Watcher at all; our harness adds the
Watcher to spirecomm itself. Orak's rule bot never chooses Recall at a rest site, so it never
collects the Ruby Key and can never reach Act~4.

\begin{table}[h]
\centering
\small
\caption{\textbf{Which parts of the game each harness covers.}
\correctmark{}: covered (for our harness and Orak, the LLM sees or decides it).
\emph{rule bot}: Orak's rule-based bot decides it, without the LLM.
\incorrectmark{}: not covered.
For Bottled AI, a symbolic player with no model, \correctmark{} means it handles it, and a percentage is the share it handles.
}
\label{tab:feature-coverage}
\begin{adjustbox}{max width=\linewidth}
\small
\begin{tabular}{lccc}
\toprule
 & Ours & Orak & Bottled AI \\
\midrule
\multicolumn{4}{@{}l}{\emph{Characters}} \\
\quad Ironclad, Silent, Defect & \correctmark & \correctmark & \correctmark \\
\quad Watcher & \correctmark & \incorrectmark & \correctmark \\
\multicolumn{4}{@{}l}{\emph{Combat: what is seen}} \\
\quad Hand, energy, HP, Block, relics & \correctmark & \correctmark & \correctmark \\
\quad Enemy intents and powers & \correctmark & \correctmark & \correctmark \\
\quad Draw, discard, and exhaust piles & \correctmark & \incorrectmark & \correctmark \\
\quad Orbs (Defect) & \correctmark & \incorrectmark & \correctmark \\
\quad Stance (Watcher) & \correctmark & \incorrectmark & \correctmark \\
\quad Potions & \correctmark & \incorrectmark & \correctmark \\
\multicolumn{4}{@{}l}{\emph{Combat: what is decided}} \\
\quad Cards to play, and their targets & \correctmark & 91\%$^{a}$ & 85\%$^{b}$ \\ %
\quad Choices a card opens & \correctmark & rule bot & 42\%$^{b}$ \\ %
\quad Potions to drink & \correctmark & rule bot$^{c}$ & 88\%$^{d}$ \\
\multicolumn{4}{@{}l}{\emph{Outside combat}} \\
\quad Card rewards after a fight & \correctmark & \correctmark & \correctmark \\
\quad Relic, potion, and other rewards & \correctmark & rule bot & \correctmark \\
\quad Route on the map & \correctmark & rule bot & \correctmark \\
\quad Neow and events & \correctmark & rule bot$^{e}$ & \correctmark \\
\quad Shops & \correctmark & rule bot & \correctmark \\
\quad Rest sites & \correctmark & rule bot & \correctmark \\
\quad Boss relics and chests & \correctmark & rule bot & \correctmark \\
\multicolumn{4}{@{}l}{\emph{Act 4}} \\
\quad Keys (needed to reach Act 4) & \correctmark & \incorrectmark & w/ Heart$^{f}$ \\
\bottomrule
\end{tabular}
\end{adjustbox}

\smallskip
\begin{minipage}{0.9\linewidth}\footnotesize
$^{a}$The model plays every card, but 33 of 351 target another card and that selection goes to the rule bot (\cref{tab:card-coverage}).
$^{b}$Share of the 351 cards that its combat simulator models (\cref{tab:card-coverage}).
$^{c}$Only deployed at the start of boss fights.
$^{d}$Share of the Ironclad's 32 drinkable potions that it can drink (it skips Smoke Bomb, Elixir, Liquid Memories, and Snecko Oil). Only deployed in elite, boss, and event-triggered fights.
$^{e}$Context-free policy: always the first or last option, depending on the event.
$^{f}$Only with the Heart strategy (\cref{sec:methods-baselines}).
\end{minipage}
\end{table}

\paragraph{Card coverage.} \Cref{tab:card-coverage} estimates how much of the game each harness
lets its player use in full. For Orak, a card counts as unsupported if playing it opens a choice
(either version, so Armaments counts although Armaments+ upgrades every card without asking). Orak's
rule bot makes those choices, not the model. 33 cards open a choice when played, among them Headbutt,
Warcry, Dual Wield, and Exhume, and every card that has the player pick a card to discard, scry, or
keep, or pick one of three. Bottled AI plays only cards its combat simulator models, and treats any other card as one
it can never play. Its gaps are mostly cards that open a choice or have a random effect (Armaments,
Headbutt, Discovery, Infernal Blade), plus a few powers (Brutality, Combust, Rupture). Our harness passes every
selection a card opens to the model, so every card is supported. The count covers playing a card, not
how well a player uses it.

\begin{table}[h]
\centering
\small
\caption{\textbf{Cards each harness supports in full, by card pool.}
A card counts if a player of that character could use it in full; which
characters each harness can play, and which parts of the game, are in \cref{tab:feature-coverage}. %
} %
\label{tab:card-coverage}
\setlength{\tabcolsep}{5pt}
\begin{adjustbox}{max width=\linewidth}
\small
\begin{tabular}{l*{6}{r@{\hspace{3pt}}l}}
\toprule
 & \multicolumn{2}{c}{Ironclad} & \multicolumn{2}{c}{Silent} & \multicolumn{2}{c}{Defect} & \multicolumn{2}{c}{Watcher} & \multicolumn{2}{c}{Colourless} & \multicolumn{2}{c}{All cards} \\
 & \multicolumn{2}{c}{(75)} & \multicolumn{2}{c}{(75)} & \multicolumn{2}{c}{(75)} & \multicolumn{2}{c}{(75)} & \multicolumn{2}{c}{(51)} & \multicolumn{2}{c}{(351)} \\
\midrule
Ours & 75 & (100\%) & 75 & (100\%) & 75 & (100\%) & 75 & (100\%) & 51 & (100\%) & 351 & (100\%) \\
Bottled AI & 63 & \phantom{0}(84\%) & 66 & \phantom{0}(88\%) & 68 & \phantom{0}(91\%) & 70 & \phantom{0}(93\%) & 33 & \phantom{0}(65\%) & 300 & \phantom{0}(85\%) \\
Orak & 68 & \phantom{0}(91\%) & 66 & \phantom{0}(88\%) & 72 & \phantom{0}(96\%) & 67 & \phantom{0}(89\%) & 45 & \phantom{0}(88\%) & 318 & \phantom{0}(91\%) \\
\bottomrule
\end{tabular}
\end{adjustbox}
\end{table}

\FloatBarrier
\section{Experimental details}
\label{app:methods-details}
This appendix gives the details of the experimental setup summarized in \cref{sec:methods}. The
harness itself is described in \cref{app:harness-sequence}. %

\subsection{Models and reasoning settings}
\label{app:methods-models}
Reasoning intensity is set on one scale (\texttt{low}, \texttt{medium}, \texttt{high},
\texttt{xhigh}, \texttt{max}) and resolved for each provider to the nearest level it supports at or
above the one asked for, so a model is never run below its label. Claude Haiku~4.5 takes a thinking
budget instead of a level and is run at 12{,}288 tokens; Nemotron and Gemma~4 E4B take no reasoning
setting and run at their server's default. %
Qwen3.6 35B-A3B is sent a level, but its deployment ignores it, so it too runs at its default under any level. %
 Test-time compute can matter as much as model size~\citep{snell2025scaling}, so the effort
ablation (\cref{tab:ablation-effort}) reruns the same configuration at every other level each model
was run at. %
A model name that the provider points at successive versions is always recorded as the version
that answered. The models, how each was reached, and the setting each ran at are listed in
\cref{tab:models}. %

\begin{table}[t]
\centering
\small
\caption{%
\textbf{The models of \cref{tab:main-results}.} Developer, how each was reached, the model ID sent, and the
reasoning setting it ran at in the main table (an effort level, or a thinking budget for Claude Haiku~4.5).
Self-hosted models were served through an OpenAI-compatible endpoint under the deployment name given.
Every main-table run was played between 19 and 25 September 2026.
\na{}: No reasoning control available; the model runs at its default.
}
\label{tab:models}
\begin{adjustbox}{max width=\textwidth}
\small
\begin{tabular}{@{}c@{\hspace{3pt}}lllll@{}}
\toprule
 & Model & Developer & Reached through & Model ID & Reasoning \\
\midrule
\spireicon{qwen} & Qwen3.8 27B & Alibaba & self-hosted & \texttt{Qwen3\_8-27B} & xhigh \\
\spireicon{qwen} & Qwen3.6 35B-A3B & Alibaba & self-hosted & \texttt{Qwen3\_6-35B-A3B} & \na{} \\
\spireicon{qwen} & Qwen3.5 122B-A10B & Alibaba & self-hosted & \texttt{Qwen3\_5-122B-A10B} & xhigh \\
\spireicon{claude} & Claude Fable~5.1 & Anthropic & Anthropic API & \texttt{claude-fable-5-1} & max \\
\spireicon{claude} & Claude Opus~5 & Anthropic & Anthropic API & \texttt{claude-opus-5} & max \\
\spireicon{claude} & Claude Sonnet~5 & Anthropic & Anthropic API & \texttt{claude-sonnet-5} & max \\
\spireicon{claude} & Claude Haiku~4.5 & Anthropic & Anthropic API & \texttt{claude-haiku-4-5-20251001} & 12{,}288 tokens \\
\spireicon{deepseek} & DeepSeek V4.1 Flash & DeepSeek & DeepSeek API & \texttt{deepseek-flash} & max \\
\spireicon{gemini} & Gemini~3.8 Flash & Google & Gemini API & \texttt{gemini-3.8-flash} & high \\
\spireicon{gemini} & Gemma~4 E4B & Google & self-hosted & \texttt{gemma4:e4b-highcontext} & \na{} \\
\spireicon{meta} & Muse Spark 1.3 & Meta & Meta Model API & \texttt{muse-spark-1.3} & xhigh \\
\spireicon{nvidia} & Nemotron~3 Nano Omni & NVIDIA & self-hosted & \texttt{Nemotron-3-Nano-Omni-30B-A3B} & \na{} \\
\spireicon{openai} & GPT-6 Astra & OpenAI & Responses API & \texttt{gpt-6-astra} & max \\
\spireicon{openai} & GPT-5.6 Sol & OpenAI & Responses API & \texttt{gpt-5.6-sol} & max \\
\spireicon{openai} & GPT-5.6 Terra & OpenAI & Responses API & \texttt{gpt-5.6-terra} & max \\
\spireicon{openai} & GPT-5.6 Luna & OpenAI & Responses API & \texttt{gpt-5.6-luna} & max \\
\spireicon{openai} & gpt-oss-120b & OpenAI & self-hosted & \texttt{gpt-oss-120b} & high \\
\spireicon{grok} & Grok~4.6 & xAI & xAI API & \texttt{grok-4.6} & xhigh \\
\spireicon{zai} & GLM-5.3 & Z.ai & Z.ai API & \texttt{glm-5.3} & max \\
\spireicon{zai} & GLM-5.3 Flash & Z.ai & Z.ai API & \texttt{glm-5.3-flash} & max \\
\bottomrule
\end{tabular}
\end{adjustbox}
\end{table}

\paragraph{Reasoning, temperature, and response length.} Reasoning is kept wherever the provider allows.
Claude's signed thinking blocks are returned unchanged; OpenAI's reasoning models
keep their reasoning server side through the Responses API; Gemini's thought signatures are
round-tripped; and models that return their reasoning as text (DeepSeek, Z.ai GLM, gpt-oss, Nemotron) have it
sent back in later turns. Qwen's reasoning is sent back within a chain of tool calls and dropped
at each new user message, as its own chat template would drop it. Temperature is left at each
provider's default, and a response may use up to 65{,}536
tokens, reduced to the model's own output limit or a quarter of its context window where either is
smaller. No response reached this ceiling: none of the 274{,}306 model calls in the 1{,}908
finished runs behind this paper was truncated.

\subsection{Seeds and cohort selection}
\label{app:methods-cohorts}
Every main result, ablation, and difficulty level is played on the same ten seeds, minted for this paper.
We replaced the seed list the harness had
used during preliminary development since the opening states of a seed (the Neow choice, the first map, and the
first card reward) are identical in every run of it, and some providers' terms permit training on
API traffic, which is the route by which static benchmarks become
contaminated~\citep{sainz2023contamination,golchin2024timetravel}. New seeds close that route to
memorization, though not the possibility that a model has seen the harness's prompt format and learned about the domain.

Runs are recorded in Weights \& Biases, and each table cell is built from exactly one cohort,
selected and then checked by the analysis code:
(i)~every run shares one run-name stem, which encodes the model, character, full configuration tag,
and seed list, so that runs from other configurations or seed lists cannot pool;
(ii)~only runs that finished and recorded a final floor are used;
(iii)~a seed played more than once contributes its latest run;
(iv)~every recorded configuration value other than the seed is identical across the cohort; and
(v)~for a read-write memory cohort, the runs form one unbroken chain: the first seed starts from an
empty bank and each seed starts from the bank the previous one left.
When a restart left a stray run newer than the chain it would be spliced into, the chain is rebuilt
backwards from the last seed by matching each run's starting memory count to the previous run's
final one; this changed one cohort.

\subsection{Metrics and statistics}
\label{app:methods-metrics}
Act~1 ends at floor 16, Act~2 at floor 33, and Ascension~20's two Act~3 bosses are at floors 50 and 51.
Our main metric is mean acts cleared, which rewards reaching the next act and is not moved by how far into an act a run died.
Floor is reported as mean, median, and range. Per-floor figures divide by the final floor plus one,
since Neow (``floor 0'') counts as a floor for this.
We prefer mean acts cleared to mean floor reached because a model can reach more floors within an act by taking an easier path (e.g.\ avoiding elites), and such conservative pathing works against the objective of winning the run.
The player must grow the deck's strength during the run, and harder paths yield larger rewards: a higher risk of death now for better odds of eventual success.
Each act ends in a hard boss fight that cannot be avoided, a step up in difficulty.
An act is thus a smaller version of the whole game, since the agent must grow strong enough within it to beat the boss at its end.
The metric is still somewhat misaligned with winning the game: models play the first act more cautiously than average humans (\cref{app:play}) and the top models clear it more often, yet all models win runs less often than humans. However, the models do not win the game often enough for us to rank them on Act~4 win rate alone.

\emph{Thinking tokens} are the reasoning tokens a provider reports;
where a self-hosted server does not separate them, the count is apportioned by the reasoning text's
share of the response's characters. Paired comparisons across models, such as an ablation against its
baseline, use exact sign-flip permutation tests~\citep{fisher1935design}; correlations across models
are tested by permutation (Spearman unless stated otherwise), and within-run contrasts by exact sign tests.

Several analyses group the 20 models of \cref{tab:main-results} into three tiers by acts cleared: the
\emph{top four} (Claude Fable~5.1, GPT-6 Astra, Claude Opus~5, and GPT-5.6 Sol), the \emph{next eight} and the
\emph{bottom eight}. %

\subsection{Cost}
\label{sec:methods-cost}
Hosted models are priced at each provider's list price for input, cached input, cache writes, and
output, with reasoning billed as output. Two conventions apply. OpenAI reports cache reads but not
cache writes, so the written share is inferred from the growth of the cached prefix between calls,
and can only be underestimated. DeepSeek's prices vary with the time of day; we charge the peak rate,
so its figures are upper bounds. Z.ai's GLM models were run on a flat-rate subscription and are
priced at the list price of the tokens they used.

Self-hosted models have no bill, so we price their recorded tokens at the headline prices published
by Artificial Analysis (retrieved 22 September 2026). These are the model's first-party API price
where one exists, otherwise the median across providers. Input, cached input, and output are each
charged at their listed rate; where no cache-hit price is listed, cached input is charged in full.
The per-million-token prices are: gpt-oss-120b \$0.15 / \$0.125 / \$0.59; Qwen3.8~27B \$0.50 / \$0.05
/ \$3.00; Qwen3.6~35B-A3B \$0.38 / \$0.38 / \$2.25; Qwen3.5~122B-A10B \$0.40 / \$0.40 / \$3.20;
Nemotron~3 Nano Omni \$0.20 / \$0.195 / \$1.09; Gemma~4 E4B \$0.02 / \$0.02 / \$0.10 (input / cache hit / output). Each reproduces the
blended price that Artificial Analysis publishes, except Gemma~4 E4B's, for which our
snapshot lists no blended price.

\subsection{Non-LLM baselines}
\label{sec:methods-baselines}
Two policies are lower bounds for the models. \emph{Pick first option} always takes the leftmost valid
option, including the leftmost enemy as its target, and is deterministic. \emph{Pick random option}
draws uniformly among the valid options, from a random stream seeded by the game seed. Both drink
every usable potion at the first opportunity, since hoarding potions is a known failure of the
models. In shops, the first policy buys only relics (the leftmost it can afford), so the stock of
slot zero does not decide the run, while the random policy buys random affordable items until it can
afford none. Above them, \emph{BottledAI}~\citep{bottledai} is a hand-written
symbolic Ironclad player, run with and without its Heart strategy. All three play the same seeds,
ascension, and game build as the models, but see no names at all, so the reskin does not apply to
them.

\paragraph{Bottled AI with the Heart.}
The latest released Bottled AI implementation stops at Act~3.
It does not collect the three keys that unlock Act~4, so it can never meet the final boss (the Corrupt Heart of the Spire).
The Heart variant comes from an open, unmerged pull request to Bottled AI, \emph{Slay the heart}, by the GitHub user Typhi~\citep{bottledaiheart}.
It adds a \texttt{slay\_heart} switch to any strategy. With the switch on, the bot:
\begin{itemize}[leftmargin=*,itemsep=1pt,topsep=2pt]
\item takes the Emerald Key from a burning elite's rewards;
\item takes the Sapphire Key instead of the relic from the Act~3 chest on floor~43;
\item chooses Recall to obtain the Ruby Key at the last rest site before the Act~3 boss;
\item rewards map paths through the burning elite, which provides the Emerald Key.
\end{itemize}
The same pull request also adds combat knowledge that applies with the switch on or off: Act~4's
enemy powers (the Heart's Beat of Death and Invincible), a preference for killing the Spire Spear
before the Spire Shield, and a guard against playing a pointless infinite loop of cards.

We run Bottled AI's Ironclad strategy, \emph{Requested Strike}, with the switch off (\emph{w/o
Heart}) and on (\emph{w/ Heart}). The code is the pull request's final commit, with the commits
added to Bottled AI's main branch since then applied on top, plus three fixes of our own:
\begin{itemize}[leftmargin=*,itemsep=1pt,topsep=2pt]
\item two handlers added on main after the pull request still used the handler signature the pull
request had changed, and failed when called;
\item the Match and Keep event handler always chose the first card. On the second flip of a pair,
that card is already face up, so the game ignored the choice and the bot waited forever;
\item the Match and Keep event's name was misspelled (it lacks the final ``!''), so the fix above was never reached.
\end{itemize}

Since neither variant reaches Act~3 on the ten main seeds at A20 difficulty (\cref{tab:main-results}), in practice they play identical games on all ten. At lower ascension levels they differ (\cref{fig:difficulty}): the Heart-seeking variant does worse, because it must forgo a relic and a card upgrade to pick up two of the three keys, which leaves it weaker at the Act~3 boss.

\subsection{Ablations and analyses}
\label{sec:methods-ablations}
Each ablation changes a single factor from the main table's configuration and keeps each model at
its main-table reasoning setting, unless stated otherwise. Gemini~3.8 Flash is also run at low effort,
where it lies on the cost Pareto front (\cref{fig:pareto}), and is pooled as a separate entry.

\paragraph{Harness.} \Cref{tab:harness-compare} compares our default harness with four
alternatives. \emph{Single action} allows one action per response. \emph{Single action, by index} does
the same and also makes every choice by its number: the tool that chooses an option by name is withdrawn,
and every choice screen lists its options as numbered entries, so the two single-action arms differ only
in how a choice is named.
How a problem is represented can bound what a reasoning system can do with it~\citep{Yihe2019_ProgramSynthesis},
and this arm isolates that factor in the interface.
\emph{Continuous} keeps one
conversation for the whole run instead of one per floor.
Its history is kept as sent, with no summarization; a conversation that outgrows the model's context window loses its oldest turns until it fits
(\cref{app:harness-features}). This happened on 12\% of its calls (809 of 6{,}655), from three models: gpt-oss-120b (494 of its 1{,}398 calls) and Nemotron~3 Nano Omni (278 of 736), both with 128k-token windows and in every run, and Qwen3.5 122B-A10B with a 64k-token window (37 of 654, in five runs).
\emph{Orak} hands the whole run to the
reflection-and-planning agent of the Orak benchmark~\citep{orak}, run from its pinned upstream
release (commit \texttt{c14e5a0}) with its own prompts and a budget of 200 decisions, and passed
the reasoning level each model runs at in our harness, so that the two harnesses are compared at the same
setting.
Orak's own rule
bot plays the map, events, shops, rest sites, chests, boss relics, Neow, and potions, so the comparison
covers combat and card rewards only; what its model decides within a fight is set out in
\cref{app:harness-coverage}. %
Orak predates the current effort-based reasoning models, so we made minor changes to its
provider code, none of which touch its prompts, agent, or decisions. It sends each model the
reasoning level, in the form that model's API takes (for OpenAI models, through the Responses API,
whose scale reaches the highest levels). It lets each model use the context window and response
ceiling our harness gives it, where Orak's defaults would cut a reasoning model's reply short or
truncate its prompt. It reaches Gemini with an API key rather than Orak's authors' own cloud project.
And it plays through Orak's own build of CommunicationMod rather than ours, whose choice lists
differ on some event screens. These variants were run without the reskin or a memory
bank, and so the default arm uses that configuration too.
Because each harness was run on a different set of models, the table pools over models and reports
each harness's difference from the default, which is what the comparison is about, on matched pairs
of runs (the same model on the same seed). Besides performance it compares efficiency: model calls
and tokens per floor. Those are compared on like-for-like floors only, meaning the same model, seed,
act, room type, and occurrence of that room type within the act (the second elite of Act~1, say), and
only where every harness run on that model and seed completed the floor. A harness that dies early
would otherwise be credited with the cheap early floors, and one that goes deeper charged for the
expensive late ones. Orak keeps no per-floor record, so its usage can only be compared per floor
reached over whole runs. \Cref{app:further-harness-ended} reports the runs each harness ended itself,
through a crash or a loop, since a run stopped by
the harness rather than by a death counts against the model in every per-run figure.
The memory ablation (\cref{tab:ablation-memory}) is reported the same way,
against the arm without memory for the same kind of notes. Per-model results are in
\cref{app:ablations-by-model}.

\paragraph{Knowledge given to the model.}\label{app:prompt-material}\Cref{tab:ablation-prompt,tab:ablation-prompt-by-model}
change the knowledge base the model is given, on top of the themed reskin (it reaches the model through the
cached system prompt).
\emph{No compendium} removes the card, potion, and relic compendium (\cref{app:prompt-system}).
\emph{Bestiary}
extends the compendium with the normal enemies, elites, and bosses of each act: their HP, intents, and
starting powers, together with each enemy's attack pattern, all taken from the Slay the Spire Wiki
(CC~BY-SA~4.0) like the rest of the compendium (\cref{app:prompt-system}). %
The \emph{Reddit guide} is ``Slay the Spire: High Ascension Guide'', posted to
r/slaythespire by the user screwyioux in January 2020 (about 7{,}000 words), included verbatim apart from
putting game names in their canonical spelling so that the reskin renames them like the rest of the prompt.

\paragraph{Reskin.} \Cref{tab:ablation-reskin} compares each reskin with the real game names on the
same model and seed, pooled over the seven models with all ten seeds in every mapping, in the same way as the harness and memory comparisons above. %
The permuted and nonsense mappings separate the two things the themed reskin does at once, hiding each
real name and supplying a coherent replacement for it (\cref{sec:methods-reskin}). Per-model results
are in \cref{tab:ablation-reskin-by-model}.

\paragraph{Difficulty.} The difficulty sweep is described with its results in \cref{app:difficulty-outcomes}. %

\paragraph{Cost and reasoning frontiers.} \Cref{fig:pareto,fig:pareto-cost-floor,fig:pareto-thinking} plot mean acts
cleared against mean cost per run, cost per floor, and thinking tokens per floor, with one point for
each model at each reasoning level it was run at, since an agent's accuracy means little without its
cost~\citep{kapoor2025agents}.
A point lies on the Pareto front when no other
point clears at least as many acts for no more spend; the front is computed over cohorts with all
ten seeds.

\paragraph{Public benchmarks.} \Cref{fig:benchmarks,tab:benchmarks} compare mean acts cleared with
the scores Artificial Analysis publishes for the same models, taken from a snapshot of its leaderboard
(the retrieval date is in the table caption), for the LLM agents of \cref{tab:main-results} with all ten
seeds at their highest reasoning effort, and again at low effort where that cohort has all ten seeds and
Artificial Analysis lists the setting (it does not for Qwen3.5 122B-A10B, Qwen3.6 35B-A3B, or either GLM-5.3
model). %
Each point is matched to the Artificial Analysis entry for the reasoning setting it ran at here;
two matches are approximate, since Artificial Analysis lists Claude Haiku~4.5 without its thinking budget
and GLM-5.3 Flash under a single setting. We use the evaluations reported for nearly every one of these
models, among them Humanity's Last Exam~\citep{phan2025hle}, GPQA Diamond~\citep{rein2024gpqa},
and SciCode~\citep{tian2024scicode}.
A correlation with the public suite says little on its own, because the suite's
evaluations are correlated with each other~\citep{burnell2023revealing,ruan2024observational}; so we also add SpireBench to the suite as one more row and
ask, for every row, how well the rest stand in for it: its strongest correlation with any other single
evaluation, and how much of it the first principal component of the others predicts for a model left out
of the fit. The Intelligence Index is a composite of evaluations like these and is left out of that part.
The two effort levels of one model are not independent points, so a model is held out at both levels together, and
we also report the correlations within each level. %

\paragraph{Learning from memory.} Described with its results in \cref{app:play-learning}. %

\paragraph{Memory growth.} Every model call logs the length of its system prompt, and within a cohort
the memory bank is the only part of the system prompt that changes between runs. The first call of
run~$k+1$ therefore measures the bank left by run~$k$, relative to run~1's empty bank
(\cref{fig:memory-growth}). This is measured in characters so that sizes are comparable across
tokenizers; the bank left after the last run is never read, so the series ends at run~9.

\paragraph{Tool-call errors.} The harness counts four kinds of tool-use failure, reported per 100 model
calls in \cref{tab:tool-errors} so that models making more calls are not charged for it: a response in
which no call could be executed, because every call was malformed or refused or none could be parsed;
a single call refused before it reached the game, as illegal on the current screen or with invalid
arguments; a turn ended with a different amount of energy left than the model declared, which marks a
model that has lost track of its own turn; and a call written as prose rather than through the tool
interface, which the harness recovers. The model is told about the first three and the harness asks
again or carries on, so none of them ends a run; they measure how precisely a model acts, separately from
how well it plays, as function-calling benchmarks measure tool use on its own~\citep{patil2025bfcl}.
The harness comparison (\cref{tab:harness-compare}) reports the same rate.

\paragraph{Calibration.} How the forecasts are scored is described with the results in \cref{app:calibration}.

\FloatBarrier
\section{Human baselines}
\label{app:human-baselines}
We compare the models with three kinds of human player, described in turn below: four expert players (Twitch streamers) %
(\cref{tab:human-collection}), the whole population of players at Ascension~20, from the
telemetry the game uploads (\cref{tab:crowd-baseline,tab:crowd-by-hours}), and one of the authors, a
non-professional player, over all 50 of their A20 Ironclad runs (2020--2024) and over the first ten.
The \emph{four Twitch streamers}, Baalorlord, panacea108, vmService, and cheg0, have per-run files with a record of every floor; these are compared floor by floor with the models in \cref{app:play}.

\paragraph{Human baselines.}
We compare against published A20 Heart Ironclad results from four top players, with the
same act-clearing definitions as our harness. They differ in how the data were collected,
and this determines what each figure means. Only Baalorlord's is an automatically logged
long-run average: the Spireblight client uploads every run file from his game directory,
abandoned runs included, and we use all 687 Ironclad A20 runs on his main profiles from
2020 to 2026 (\url{https://baalorlord.tv/runs})\footnote{This includes Baalorlord's ``mastery'' runs, which are abnormal as he tries to win a run specifically with two copies of a card, but we don't find a significant difference in his overall performance when including these.}.
Three further samples of per-run files
(panacea108, vmService, cheg0) come from a public archive
(\url{https://github.com/MaT1g3R/Slay-the-Spire-data}), \href{https://github.com/MaT1g3R/Slay-the-Spire-data/blob/097aaf3564c2247835162d267cbc7c55d2c9039e/LICENSE}{released under the Apache~2.0 license}.
They are contiguous blocks of one to
three months, and the two rotating samples cycle through the characters without a break,
consistent with no runs missing within them.

\begin{table}[t]
\centering
\caption{\textbf{How each human baseline was collected.} \emph{Complete block}: every run over the
period, logged automatically. \emph{Player-selected}: a contiguous block shared by or for
the player, chosen by unstated means. \emph{Record set}: a player-kept set known because
it set a record. \emph{Summary}: a figure copied from the player's own summary, with no
per-run data.}
\label{tab:human-collection}
\adjustbox{max width=\textwidth}{
\small
\setlength{\tabcolsep}{4pt}
\begin{tabular}{@{}llrcll@{}}
\toprule
Player & Source & Runs & Period & Coverage & Notes \\
\midrule
Baalorlord & site log of \texttt{.run} files & 687 & 2020-04 to 2026-02 & complete block & abandons kept; older modded profile excluded \\
panacea108 & archive, \texttt{.run} files & 51 & 2022-09 to 2022-11 & player-selected & Ironclad-only sample \\
vmService & archive, \texttt{.run} files & 50 & 2022-09 to 2022-12 & player-selected & Ironclad runs of a 200-run rotation \\
cheg0 & archive, \texttt{.run} files & 46 & 2024-02 & player-selected & Ironclad runs of a 184-run rotation \\
\bottomrule
\end{tabular}
}
\end{table}

\paragraph{Population baseline.}
For the general player population we use Mega Crit's own run telemetry, which the game uploads
automatically at the end of every run. We use all 20.25M runs the developers published from
July to mid-November 2020, which include every run in the dump on the final content build (\texttt{2020-07-30}), at
\url{https://drive.google.com/drive/folders/1c7MwTdLxnPgvmPbBEfNWa45YAUU53H0l}. Unlike
self-uploaded run archives it is not selected by the player, so it reflects everyone playing
Ironclad at Ascension~20, a level that is itself only unlocked by winning at Ascension~19 with
the Ironclad. We keep Ironclad A20 runs (773,205), drop daily, endless, seeded, trial, and
modded runs and any on another build or the beta branch (560,064 left), and remove runs
abandoned before a death or victory, nearly all quit at Neow, leaving 462,224. Act clears use the
harness's definitions, with floors checked against where each boss's victims died. The record has no
player identifier, but each run carries its player's cumulative \emph{Slay the Spire} play time
(the in-game statistic, not a cross-game or Steam figure), so we also stratify by hours played. By
requiring each run's cumulative play time to equal the previous run's plus that run's length, with
the run starting within an hour of the previous one ending, we reconstruct chains of consecutive
runs (61\% of runs link, against
0.9\% with a deliberately wrong offset). Chains break at longer
pauses, so they are play sessions rather than players, but they let us count only runs made after a
previous winning A20 run (beating Act~4) in the same chain, avoiding the bias of selecting on the outcome being
measured. The record does not show whether a player quit and reloaded to replay a fight, which the
game allows and our agents cannot do (\cref{app:methods-game}); where players did, the telemetry overstates
play without it, more so among the players with the highest win rates. The population clears Act~1 in
47\% of runs and
kills the Heart in 3.3\%.

\begin{table}[t]
\centering
\caption{\textbf{Population baseline for A20 Heart Ironclad.} From Mega Crit's run telemetry (build
\texttt{2020-07-30}, 2020-07-30 to 2020-11-12). Hours is the median cumulative Slay the
Spire play time of the player at the run.
Acts is the mean number of acts cleared per run.}
\label{tab:crowd-baseline}
\begin{adjustbox}{max width=\linewidth}
\small
\begin{tabular}{@{}lrrrrrrrrr@{}}
\toprule
 & & & \multicolumn{2}{c}{Floor} & \multicolumn{4}{c}{Act cleared (\%)} & \\
\cmidrule(lr){4-5} \cmidrule(lr){6-9}
Population & Runs & Hours & mean & median & 1 & 2 & 3 & 4 & Acts \\
\midrule
All A20 Ironclad runs & 462,224 & 295 & 22.0 & 16 & 47.0 & 15.1 & 6.8 & 3.3 & 0.72 \\
All A20 Ironclad runs, incl. abandoned & 560,064 & 292 & 18.5 & 16 & 39.5 & 12.6 & 5.6 & 2.8 & 0.60 \\
StS hours < 100 h & 32,550 & 78 & 21.2 & 16 & 44.0 & 15.0 & 7.0 & 2.4 & 0.68 \\
StS hours 100--300 h & 202,802 & 198 & 21.5 & 16 & 45.3 & 14.3 & 6.2 & 2.7 & 0.69 \\
StS hours 300--1000 h & 208,425 & 460 & 22.5 & 16 & 48.6 & 15.6 & 7.1 & 3.9 & 0.75 \\
StS hours $\geq$ 1000 h & 18,447 & 1223 & 24.1 & 20 & 52.5 & 18.8 & 9.6 & 6.1 & 0.87 \\
Session chains with a winning A20 run (biased) & 35,279 & 368 & 36.7 & 40 & 69.3 & 50.8 & 46.0 & 43.8 & 2.10 \\
Runs after a previous winning A20 run in the chain & 7,490 & 433 & 27.9 & 23 & 57.5 & 29.0 & 20.5 & 15.8 & 1.23 \\
\bottomrule
\end{tabular}
\end{adjustbox}
\end{table}

\paragraph{Experience and performance.}
Each telemetry record carries the player's cumulative \emph{Slay the Spire} play time when the run
ended, so we can stratify the 462,224 A20 Ironclad runs of \cref{tab:crowd-baseline} by hours
played (\cref{tab:crowd-by-hours}). Performance rises with experience, but slowly: runs
by players with 50--100\,h clear 0.68 acts on average and
kill the Heart in 2.3\% of runs, against 0.98 acts and
7.1\% at 1500--2000\,h, the best bin. Play time is a lower
bound on experience: it lives in the save profile and resets with a new one, so the lowest bin
(774 runs under 25\,h at A20, which takes far longer than that to
unlock) likely holds experienced players on a fresh profile. It also measures time, not skill.
Every bin holds at least 200 runs, so none was merged.

\begin{table}[t]
\centering
\caption{\textbf{Human play by hours played.} A20 Heart Ironclad by hours of \emph{Slay the Spire} played, from Mega Crit's run telemetry (build
\texttt{2020-07-30}). Hours is the player's cumulative play time when the run ended (bin and bin median).
Acts is the mean number of acts cleared per run.
}
\label{tab:crowd-by-hours}
\begin{adjustbox}{max width=\linewidth}
\small
\begin{tabular}{crrrrrrrrr}
\toprule
 & & & \multicolumn{2}{c}{Floor} & \multicolumn{4}{c}{Act cleared (\%)} & \\
\cmidrule(lr){4-5} \cmidrule(lr){6-9}
Hours & Runs & Median h & mean & median & 1 & 2 & 3 & 4 & Acts \\
\midrule
0--25 & 774 & 16 & 23.8 & 16 & 47.7 & 21.6 & 13.4 & 3.9 & 0.87 \\
25--50 & 3,791 & 42 & 21.3 & 16 & 43.4 & 16.7 & 8.2 & 2.6 & 0.71 \\
50--100 & 27,985 & 81 & 21.1 & 16 & 44.0 & 14.5 & 6.7 & 2.3 & 0.68 \\
100--150 & 46,501 & 128 & 21.0 & 16 & 43.7 & 13.7 & 5.8 & 2.1 & 0.65 \\
150--200 & 56,982 & 175 & 21.4 & 16 & 45.2 & 14.3 & 6.2 & 2.8 & 0.68 \\
200--300 & 99,319 & 246 & 21.8 & 16 & 46.2 & 14.5 & 6.4 & 3.0 & 0.70 \\
300--400 & 72,033 & 345 & 22.1 & 16 & 47.4 & 15.2 & 6.7 & 3.6 & 0.73 \\
400--500 & 48,734 & 444 & 22.5 & 16 & 48.6 & 15.7 & 7.0 & 3.7 & 0.75 \\
500--750 & 65,828 & 600 & 22.6 & 16 & 49.1 & 15.6 & 7.3 & 4.0 & 0.76 \\
750--1000 & 21,830 & 845 & 23.3 & 19 & 50.9 & 16.7 & 8.0 & 4.6 & 0.80 \\
1000--1500 & 13,613 & 1155 & 23.8 & 19 & 51.4 & 18.5 & 9.7 & 6.2 & 0.86 \\
1500--2000 & 3,018 & 1707 & 25.8 & 22 & 57.9 & 22.0 & 11.4 & 7.1 & 0.98 \\
2000--3000 & 1,564 & 2228 & 23.6 & 19 & 51.9 & 17.0 & 6.6 & 3.8 & 0.79 \\
$\geq$3000 & 252 & 3498 & 22.1 & 19 & 52.8 & 10.3 & 3.6 & 2.0 & 0.69 \\
\bottomrule
\end{tabular}
\end{adjustbox}
\end{table}

\Cref{fig:human-hours} plots \cref{tab:crowd-by-hours} against hours played. Between 25 and 1{,}000 hours
performance barely moves (0.65--0.80 acts per run, 2.1--4.6\% of runs killing the Heart); it rises between
1{,}000 and 2{,}000 hours (0.86--0.98 acts, 6.2--7.1\%) and falls again in the 1{,}816 runs above 2{,}000 hours
(0.69--0.79 acts, 2.0--3.8\%). The lowest bin is high because play time resets with a new save
profile, so it likely holds experienced players on fresh profiles.

\begin{figure}[t]
\centering
\includegraphics[width=\linewidth]{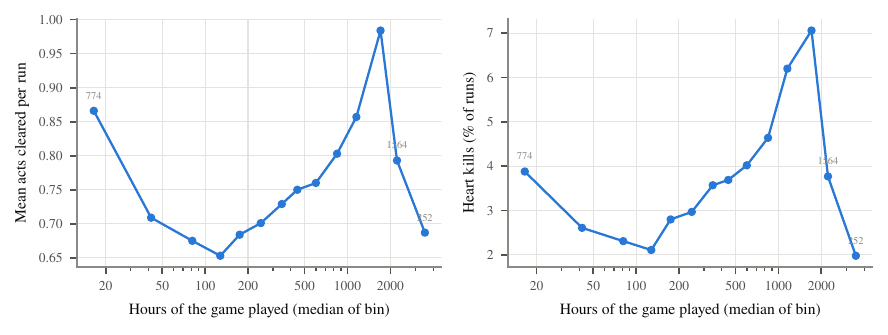}
\caption{\textbf{Human performance by experience.} Human performance at A20 with the Ironclad against hours of the game played, from the
telemetry in \cref{tab:crowd-by-hours}: mean acts cleared per run (left) and the share of runs that
kill the Heart (right). Each point is a bin of \cref{tab:crowd-by-hours} at its median hours; bins
with fewer than 2{,}000 runs are labelled with their count.
Play time resets with a new save profile, so the lowest bin likely holds experienced players on
fresh profiles.}
\label{fig:human-hours}
\end{figure}

\FloatBarrier
\section{Strategy and play quality}
\label{app:play}

This appendix asks two questions of the runs of \cref{tab:main-results}: why the models fall behind human experts,
and why the strong models beat the weak ones. Tiers and the tests are as in
\cref{app:methods-metrics}. %

\subsection{Where agents die: agents against humans}
\label{app:deck-choices}
We compare the decks and in-game choices of our agents with those of human players at the floors
where the agents die. The agents are the 200 main-table runs (20 LLMs, A20 Ironclad on the
held-out seeds, themed reskin, xml notes and memory, generic guidance), with per-floor state from
each run's logged history. The humans are the 462,224 finished A20 Ironclad runs of the game's final
content build (2020-07-30) in Mega Crit's \hreffoot{https://www.reddit.com/r/slaythespire/comments/jt5y1w/77_million_runs_an_sts_metrics_dump/}{released run telemetry} for July to
mid-November 2020, with beta, daily, seeded, trial, modded, anomalous,
and abandoned runs removed. Of the agent runs that die, 27\% die at the Act~1 boss (floor~16),
11\% on floor~7, 7\% on floor~22, and 6\% at the Act~2 boss (floor~33). On
reaching these floors the agents die more often than humans: 41\% against 32\% at the
Act~1 boss, 12\% against 4\% on floor~7, and 20\% against 6\% on
floor~22 (\cref{tab:death-floors}).

The agents reach the Act~1 boss healthier than humans but built differently
(\cref{tab:deck-choices}). They arrive with 82\% of their maximum HP against
62\%, having spent 68\% of their campfires resting where humans rest at
23\% and upgrade a card at the rest. They fight fewer elites (1.25 against
1.95), and they take a card after almost every fight (none after 8\% of
fights, against 23\%), so their decks are larger (19.3 cards against
16.8). They also drink more of their potions before the boss (0.50
of those obtained, against 0.27); among humans, those who then die at the boss
had drunk 0.40 of theirs and those who pass it
0.21. At death, against humans who died on the same floor, the
agents' decks hold 1.4 fewer upgraded cards and 2.5 more cards, and they carry 1.9 fewer
relics (1.8 fewer upgrades and 2.2 fewer relics at the Act~1 boss).

These are descriptive differences, not matched comparisons: the agents play a reskinned game on at most ten
held-out seeds, while the humans play the real game on many different seeds, and every human at A20 has,
by the game's unlock rule, won at A19 with the Ironclad. Four measurements are approximate. Human deck
size on entering floor~16 is reconstructed backwards from the final deck, since the record lists only
the final deck and does not itemize Neow's card effects; it agrees with an independent forward count
for 79\% of runs. Agent potion use counts \texttt{use\_potion} calls, and the potions obtained are
inferred from the potion count, which misses discards, so the agent rate is an upper bound. The agent
skip rate counts fights after which no card entered the deck. Some human records carry implausible gold
values (0.28\% reach 5,000 or more), so gold is reported as a median.

\begin{table}[t]
\centering
\small
\setlength{\tabcolsep}{4pt}
\caption{\textbf{Runs that died at the Act~1 boss against those that passed it.} State and Act~1 choices on
entering the Act~1 boss (floor~16), for the runs that reached
it, split by whether the run then died at the boss. Humans are Mega Crit's telemetry; expert humans are the
four expert players of \cref{tab:main-results}, from their own run files. Campfire, fight, and potion rates
are pooled over campfires, fights, and potions on floors 1--15 (potions 0--15); HP, elites, deck size, and
potions used at the boss are means over runs.}
\label{tab:deck-choices}
\begin{adjustbox}{max width=\linewidth}
\small
\begin{tabular}{lrrrrrrrrr}
\toprule
 & \multicolumn{3}{c}{Agents} & \multicolumn{3}{c}{Humans} & \multicolumn{3}{c}{Expert humans} \\
\cmidrule(lr){2-4}\cmidrule(lr){5-7}\cmidrule(lr){8-10}
 & All & Died & Passed & All & Died & Passed & All & Died & Passed \\
\midrule
HP (\% of max) & 82 & 75 & 86 & 62 & 56 & 65 & 55 & 46 & 56 \\
Campfires spent resting (\%) & 68 & 75 & 63 & 23 & 26 & 22 & 9 & 10 & 9 \\
\quad excluding Recall (\%) & 72 & 81 & 67 & 23 & 27 & 22 & 10 & 10 & 9 \\
Elites fought & 1.25 & 1.21 & 1.28 & 1.95 & 1.84 & 2.00 & 2.61 & 2.53 & 2.61 \\
Fights with no card taken (\%) & 8 & 12 & 6 & 23 & 26 & 21 & 15 & 17 & 15 \\
Deck size & 19.3 & 18.4 & 19.9 & 16.8 & 16.5 & 17.0 & 18.5 & 18.8 & 18.4 \\
Potions used / obtained & 0.50 & 0.55 & 0.46 & 0.27 & 0.40 & 0.21 & 0.46 & 0.53 & 0.46 \\
Potions used at the boss & 1.08 & 1.19 & 1.00 & 0.56 & 0.81 & 0.44 & 0.57 & 1.09 & 0.54 \\
Gold (median) & 102 & 112 & 80 & 153 & 160 & 149 & 178 & 182 & 178 \\
\midrule
Runs & 127 & 52 & 75 & 320,719 & 103,465 & 217,254 & 788 & 47 & 741 \\
\bottomrule
\end{tabular}
\end{adjustbox}
\end{table}

\begin{table}[t]
\centering
\small
\caption{\textbf{Where runs end.} The share of all deaths that fall on a floor, and the death rate among the
runs that reached it. Agents: 196 deaths in 200 runs (the rest cleared
Act~3, one of them going on to kill the Heart). Humans: 438,944
deaths in 462,224 runs. Expert humans (the four expert players of \cref{tab:main-results}):
319 deaths in 834 runs.}
\label{tab:death-floors}
\begin{adjustbox}{max width=\linewidth}
\small
\begin{tabular}{lrrrrrr}
\toprule
 & \multicolumn{3}{c}{Share of deaths (\%)} & \multicolumn{3}{c}{Death rate on reaching (\%)} \\
\cmidrule(lr){2-4}\cmidrule(lr){5-7}
Floor & Agents & Humans & Experts & Agents & Humans & Experts \\
\midrule
7 & 11 & 4 & 2 & 12 & 4 & 1 \\
16 (Act~1 boss) & 27 & 24 & 15 & 41 & 32 & 6 \\
22 & 7 & 3 & 1 & 20 & 6 & 1 \\
33 (Act~2 boss) & 6 & 8 & 12 & 37 & 34 & 6 \\
\bottomrule
\end{tabular}
\end{adjustbox}
\end{table}

\paragraph{Health as a resource.} The telemetry above shows the agents reaching the Act~1 boss with more HP
than humans because they rest where humans upgrade. \Cref{fig:hp-resource} follows HP over the whole run
against the four expert players of \cref{app:human-baselines} (833 runs with per-floor records). The
experts spend HP through Act~1: after the floor-15 rest site, the last before the boss, their median is
53\% of maximum HP, against 90\% for the four models that clear the most acts. At a rest site the models
rest unless they are nearly full, 76--98\% of the time when they arrive below 80\% of maximum HP, where the
experts rest 6--48\% of the time and upgrade a card otherwise; over all visits the experts rest at
15\% (6{,}398 visits), the top four models at 66\% (238) and the other models at 75\% (364). The strongest
models manage HP no differently from the weakest: none treats it as a resource to trade for a stronger
deck. The experts play their own seeds, so this is a reference rather than a matched comparison.

\paragraph{Routes, relics, and deck.} \Cref{fig:macro-routes,fig:macro-trajectories} compare the run-level
choices of the LLM agents of \cref{tab:main-results}, in three groups by acts cleared, with the telemetry population
and the four expert players. In Act~1 the models fight fewer elites (1.4 for the top four models, 1.3 and 0.9 for
the next eight and the bottom eight, against 2.0 for the population and 2.6 for the experts; 56--77\% of model runs
reaching the boss fought at most one, against 29\% and 4\%), enter more shops (1.3--1.6 against 1.0 and 0.8),
rest at 57--75\% of rest sites against 23\% and 9\%, and take a card after almost every fight (3--13\% of fights
with no card taken, against 23\% and 15\%). They therefore fall behind in relics from the first elites and never
catch up (9 against the experts' 12 at floor 32), carry fewer upgraded cards, and hold a larger deck.
In Act~1 the top four and the next eight gain about as much gold on the map floors as humans, and the bottom eight less (\cref{fig:macro-gold}; median 226, 203, and 175, against 199 for the population and 215 for the experts); the top four enter shops with about as much as the experts (208 against 219; the next eight 157, the bottom eight 127, the population 141).
Yet all three groups reach the Act~1 boss with less (a median of 104, 113, and 64, against 153 and 178), so the top four and the next eight spend or lose more of what they gain.
The top four also reach the Act~2 and Act~3 bosses with less (111 and 92, against 139 and 164 for the population and 164 and 235 for the experts).

These choices do not separate the stronger models from the weaker ones (\cref{fig:macro-models}). Across the 16
models with at least three runs reaching the Act~1 boss, Spearman's $\rho$ with acts cleared is
$+0.37$ for elites fought ($p=0.16$), $-0.25$ for shops, $-0.20$ for resting, and $-0.21$ for skipped card rewards
(permutation tests), whereas the net HP change over a hallway fight has $\rho=+0.62$ ($p=0.01$), from $-2.4$~HP for
the top four to $-6.5$ for the bottom eight ($-0.4$ for the population and $-0.2$ for the experts). A logistic model of beating the Act~1
boss from the state on reaching it (HP, maximum HP, relics, deck size, and upgrades), fitted on the population, rates
the top four's positions at 78\% and the next eight's at 70\%; they actually won
94\% and 55\% of those fights. The experts arrive in positions the model rates at 68\%, having spent HP for relics
and upgrades, and win 94\%. The stronger models are stronger mainly in the fights, not in the route they take to
them. The state model explains little of the outcome on its own (McFadden $R^2$ 0.05), and the human deck, relic, and
upgrade counts are reconstructed backward from each run's final state, since the telemetry records
only that.

The gap to the experts in the fights grows along the run (\cref{fig:floor-fights}). In the first three fights,
when every deck is close to the starter deck, the top four lose 2.0~HP per hallway fight and the experts 1.0;
the next eight lose 3.7 and the bottom eight 6.2, so for the weaker models the gap is already in how they play the
turn. Over the rest of Act~1 the top four lose about 3.5~HP per fight more than the experts, in Act~2 about 6
more, and in Act~3 7--11 more, as the experts' decks and relics pull ahead. The top four
already lose more than the average human on floors 5--15 (2.8--5.7~HP per fight against 1.2--2.3), before the decks
have diverged much, and from Act~2 they lose about as much as the average human.

\begin{figure}[t]
\centering
\includegraphics[width=\linewidth]{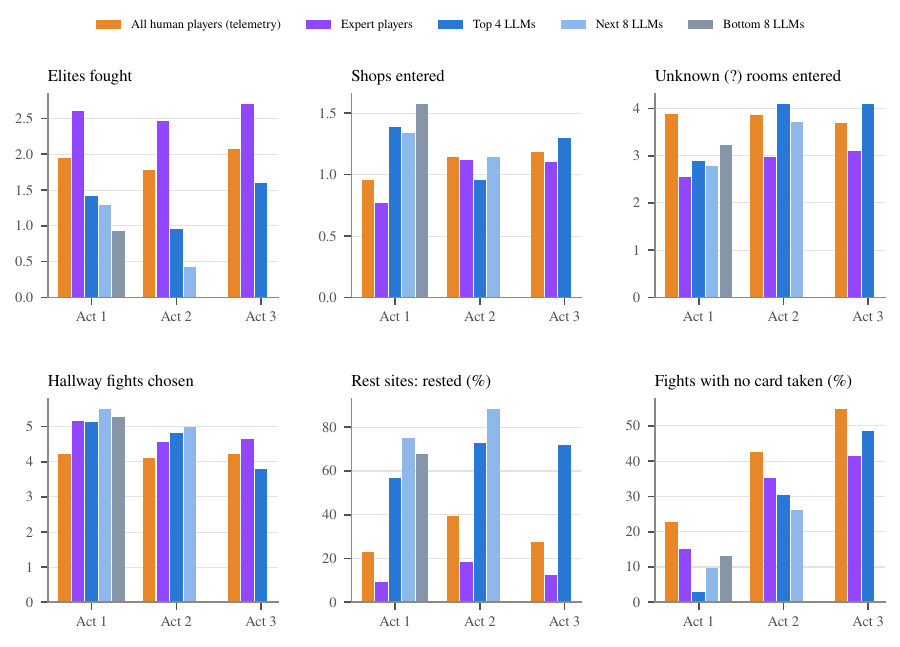}
\caption{\textbf{Route choices by act.} What runs did on each act's 15 map floors, over the runs that reached that act's
boss (floors 16, 33, and 50): the telemetry population (320{,}719, 105{,}907 and 50{,}405 runs), the four expert players
(788, 671, and 603) and the LLM agents of \cref{tab:main-results} in three groups by mean acts cleared: the top 4
(36, 23, and 10 runs), the next 8 (65 and 7) and the bottom 8 (26). A group is drawn for an act only when at least
five of its runs reached that boss. Shops and unknown rooms count the map nodes entered; elites fought and hallway
fights the rooms entered. Rest-site and card-reward shares pool the visits and fights of all runs in the group.}
\label{fig:macro-routes}
\end{figure}

\begin{figure}[t]
\centering
\includegraphics[width=0.8\linewidth]{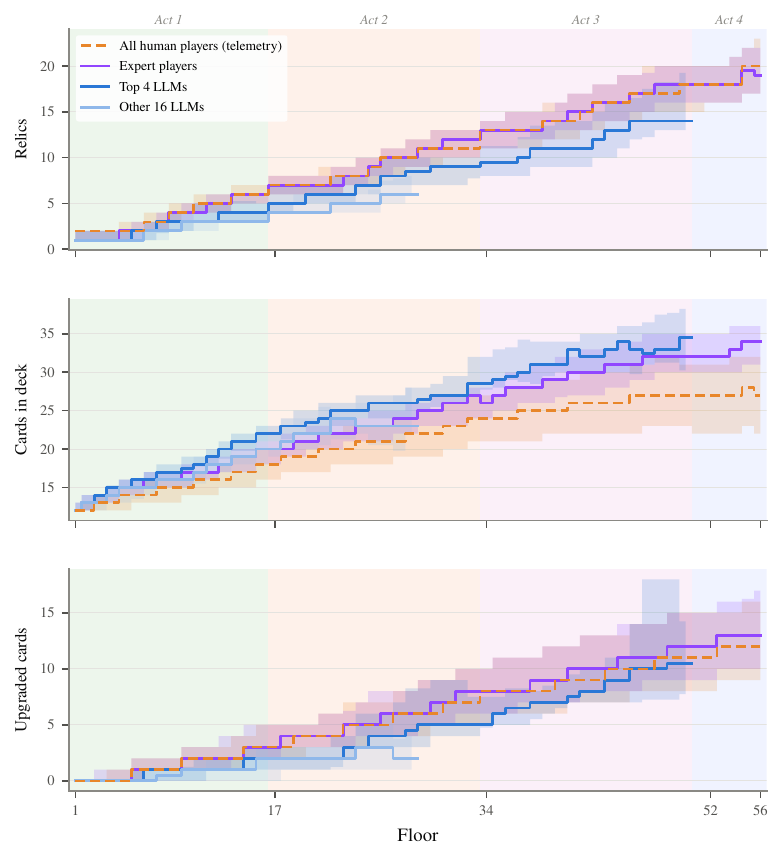}
\caption{\textbf{How relics and decks grow over a run.} Relics, cards in the deck, and upgraded cards at the end of each floor survived: median (lines) and
interquartile range (bands), for the telemetry population (dashed), the four expert players, the top 4 LLM agents
of \cref{tab:main-results} by acts cleared and the other 16. A floor is drawn only while at least 20 human or 8 LLM
runs are alive on it. Human counts are reconstructed backward from each run's final deck and relics,
removing every change the record itemizes after the floor.}
\label{fig:macro-trajectories}
\end{figure}

\begin{figure}[t]
\centering
\includegraphics[width=\linewidth]{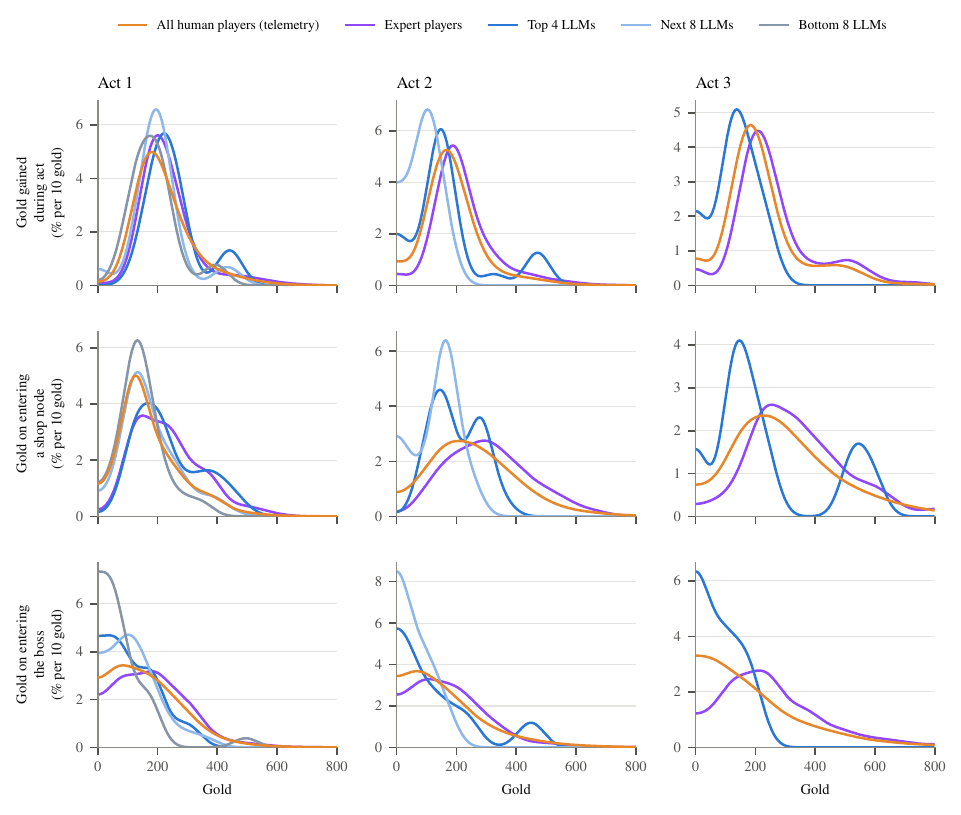}
\caption{\textbf{Gold gained and held, by act.} Gold in each act (columns), over the runs that reached that act's boss. Rows: gold gained on the
act's 15 map floors (the sum of each floor's net change, over the floors where it was positive); gold held on
entering each shop node (pooled over visits; an unknown room that turned out to be a shop is not counted); and gold
held on entering the boss.
Each curve is a smoothed distribution of a group's visits or runs (a Gaussian kernel density with a standard
deviation of 40 gold, reflected at 0 gold), in percent per 10 gold; gold above 800 is off the axis.
The groups, and the rule of drawing a group for an act only when at least five of its runs reached that boss, are
those of \cref{fig:macro-routes} (top 4 LLMs: 36, 23, and 10 runs; next 8: 65 and 7; bottom 8: 26).}
\label{fig:macro-gold}
\end{figure}

\begin{figure}[t]
\centering
\includegraphics[width=\linewidth]{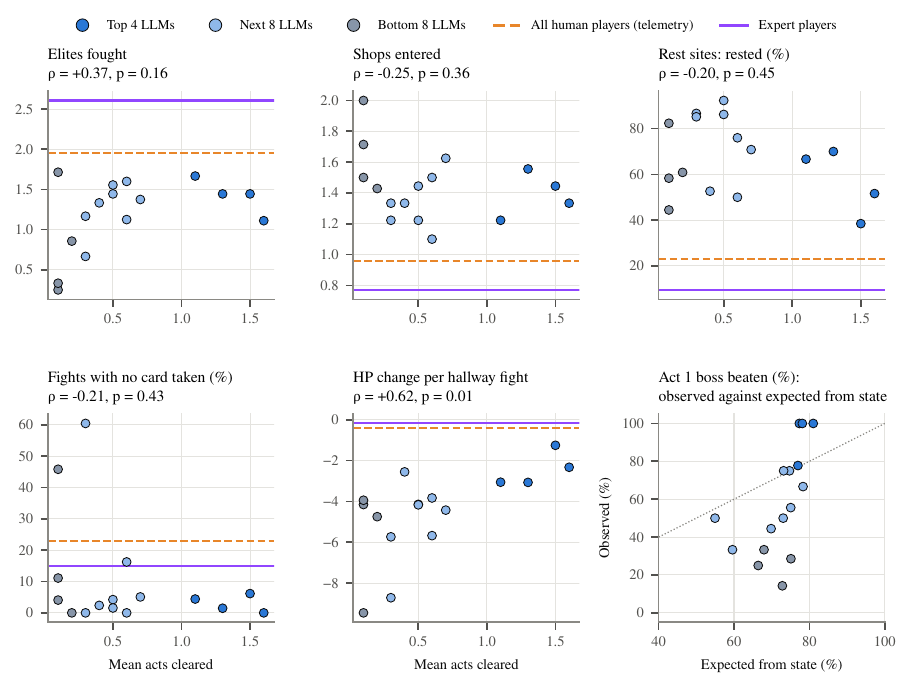}
\caption{\textbf{Act~1 play against performance, by model.} Act~1 choices and combat against mean acts cleared, one point per LLM agent of \cref{tab:main-results}
with at least three runs reaching the Act~1 boss (16 models), over those runs; lines are the telemetry population
(dashed) and the four expert players. $\rho$ is Spearman's correlation across the models and $p$ a
two-sided permutation test. HP change per hallway fight is the net HP change over a hallway-fight floor, including
the relic heal after it, computed the same way for humans. Last panel: each model's rate of beating the
Act~1 boss against the rate expected from its runs' states on reaching it, by a logistic model fitted on
the 320{,}562 telemetry runs that reached it with a complete state record; the dotted line is equality.}
\label{fig:macro-models}
\end{figure}

\begin{figure}[t]
\centering
\includegraphics[width=0.85\linewidth]{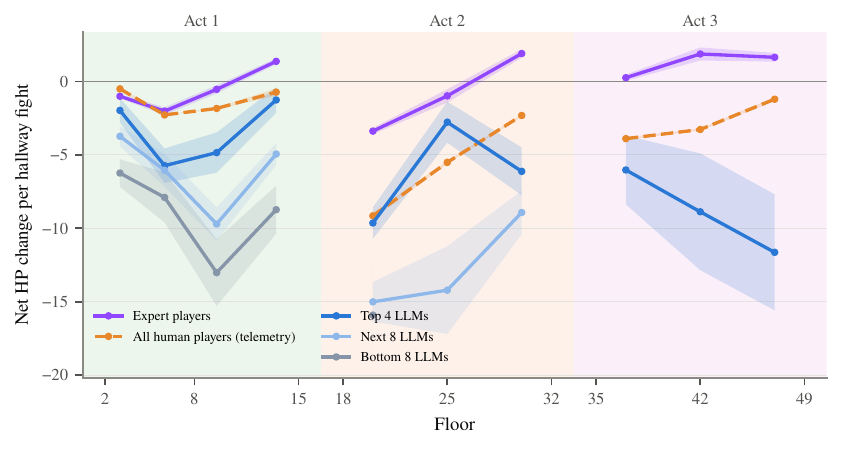}
\caption{\textbf{HP change over hallway fights, by floor.} Net HP change over a hallway fight, by floor: the fight-weighted mean over the floors of each bin (floors
2--4, 5--7, 8--11, and 12--15 in Act~1, five floors in Acts~2 and~3), with $\pm$1 standard error, for the telemetry
population (dashed), the four expert players, and the LLM agents of \cref{tab:main-results} in three groups by
mean acts cleared. A point needs at least 10 fights. The change includes the relic heal after the fight, which every
group gets; a run's first floor is not scored, and later floors hold only the runs still alive.
Humans play their own seeds.}
\label{fig:floor-fights}
\end{figure}

\subsection{Conservative Act~1 play among humans}
\label{app:play-act1}
In Mega Crit's A20 Ironclad telemetry we took the
human runs that beat the Act~1 boss and compared those whose Act~1 looks like the models' (at most one
elite, resting at half or more of the rest sites, at least 70\% HP at the boss) with those that fought two or more
elites and rested at fewer than half (\cref{tab:act1-profile}). The model-like profile covers 3.2\% of human runs and
36\% of the models'. It clears Act~2 in 16.2\% of cases against 37.5\%, and Act~3 in 3.2\% against
18.4\%, and the gap holds in every band of hours played (12\% against 43\% for Act~2 among players with over 1000
hours). Clearing Act~2 rises steadily with Act~1 elites (22, 25, 32, 39, and 49\% for 0 to 4 or more) and falls with
rests (36, 29, 24, and 21\% for 0 to 3 or more). The telemetry has no player identifier, so the strongest design
compares runs within one play session: there each Act~1 elite multiplies the odds of clearing Act~2 by 1.37
(1.34--1.40) and each rest by 0.87 (0.84--0.89), over 19{,}339 sessions, and model-like runs clear Act~2 5.9 points
less often than the same session's other runs (permutation $p = 10^{-4}$). Within each of the four
experts the odds ratios are 1.68 per elite and 0.64 per rest. Resting at a rest site entered with at least 70\% HP,
which the models do at 31\% of such sites, the population at 2\%, and the experts never, predicts worse outcomes
across the population but not within a session, so its association is likely selection on weaker players.

How much of the gap does the route explain? Shifting the experts' odds by the effect of the
models' Act~1 profile accounts for 15\% of their Act~2 advantage and 19\% of their Act~3 advantage
under the within-session model, and 32--64\% under cross-sectional models, which also absorb the effects of deck and
relics and so overstate it. At their own profile the models are not worse than humans: humans with that profile
clear Act~2 in about 20\% of cases, the models in 25\% (the top four in 47\%), while the experts clear 85\%, 46
points above what their profile predicts. These are associations: weak runs skip elites and rest partly because they
are weak, and the telemetry does not record which elites the map offered.

\begin{table}[t]
\centering\small
\caption{\textbf{Act~1 play and what follows.} Over runs that beat the Act~1 boss. Humans are Mega
Crit's A20 Ironclad telemetry and the four expert players (Twitch streamers) of \cref{tab:main-results}; agents are
the LLM runs of \cref{tab:main-results}. Elites, rest share (rest-site visits spent resting rather than upgrading),
HP (\% of maximum), relics, and upgrades are on entering the Act~1 boss. The lower block splits the telemetry by
profile: \emph{model-like} is at most one elite, a rest share of at least 0.5 and at least 70\% HP at the boss;
\emph{aggressive} is two or more elites and a rest share below 0.5. Human runs pass a plausibility filter first,
so their counts differ slightly from \cref{tab:deck-choices}.}
\label{tab:act1-profile}
\begin{adjustbox}{max width=\linewidth}
\small
\begin{tabular}{@{}l r rrrrr rr@{}}
\toprule
 & Runs & Elites & Rest share & HP & Relics & Upgrades & Clear Act 2 (\%) & Clear Act 3 (\%) \\
\midrule
Telemetry population & 217{,}133 & 2.00 & 0.22 & 64 & 5.7 & 3.3 & 32 & 14 \\
Experts (streamers) & 741 & 2.61 & 0.09 & 56 & 5.7 & 3.4 & 85 & 77 \\
LLM agents, all & 75 & 1.28 & 0.63 & 86 & 3.9 & 2.0 & 25 & 5 \\
LLM agents, top four & 34 & -- & -- & -- & -- & -- & 47 & 12 \\
\midrule
Telemetry, model-like & 6{,}924 & & & & & & 16.2 & 3.2 \\
Telemetry, aggressive & 120{,}417 & & & & & & 37.5 & 18.4 \\
\bottomrule
\end{tabular}
\end{adjustbox}
\end{table}

\subsection{Risk in the models' own runs}
\label{app:play-risk}
Elites are where the models die before the Act~1 boss: 40
of the 73 such deaths are in elite rooms, a death rate of 18.8\% per elite fight against 3.0\% per ordinary fight,
higher for 16 of 19 models (sign test $p = 0.004$). The deaths are concentrated in the weak models (Qwen3.5 122B
6 of 10 elite fights, gpt-oss-120b 5 of 7), while the top four die in 0--9\% of theirs, so for a model that fights
badly, avoiding elites is partly the right call. Among runs that beat the Act~1 boss (49 runs, 7 models), upgrades
on reaching the boss predict clearing Act~2 ($r = 0.56$, Holm-adjusted $p = 0.006$) and rests predict against it
($r = -0.44$), while the number of elites fought does not ($r = -0.12$). Part of the upgrade effect is that a run
with HP to spare can afford to upgrade: controlling for HP on arrival weakens it to $r = 0.27$ ($p = 0.11$). HP on
reaching the boss itself predicts clearing Act~2 ($r = 0.32$), which says that a deck able to clear Act~1 unhurt is
strong, not that HP should be hoarded. In 72 of the 75 Act~1 clears the transition heal was exactly 75\% of missing
HP.

\subsection{Learning over the chain}
\label{app:play-learning}
Whether acts cleared improve along the chain is in \cref{app:ablation-memory} (\cref{fig:learning,fig:memory-trend});
here we ask whether play itself changes.
Comparing runs 1--5 with runs 6--10 of each memory chain, with no-memory chains
on the same seeds as the control for how hard each half is (9 models), memory changes nothing measurable: its added
effect is $-0.04$ on the share of rest sites spent resting ($p = 0.51$), $-0.25$ Act~1 elites ($p = 0.69$, 6 models
reaching the boss in both halves) and null on upgrades, HP at the boss, floor, and acts ($p \ge 0.22$). Power is low,
but the direction is towards more caution, not less. The final memory banks lean the same way: 119 of 1{,}374
memories advise caution and 17 advise risk (keyword-coded, 20 of 25 sampled caution codes confirmed by hand), and
18 models hold more cautious than risky memories, none the reverse. The main exception is Claude Opus~5
(``DODGING FIGHTS IS AN ERROR, BECAUSE UNSPENT SLACK EXPIRES''). Only the four strongest models state the act heal
correctly; Gemini~3.8 Flash writes that it ``restores player HP up to 80\% of Max HP'', GLM-5.3 that it is ``likely
flat \textasciitilde50'', and ten models never record it. Knowing the rule does not settle the decision: 32 of 212
late-act rest-site decisions (floors 13--15, 30--32, and 47--49) reason about the heal, mostly by Claude Fable~5.1
and Opus~5, and still rest in 21 (66\%, against 77\% for the others). Opus's own bank records the
conflict: ``PRE-BOSS CAMP \ldots\ the Sleep LOSES ON THE LEDGER \ldots\ AND WINS ANYWAY''.

\subsection{Deck building and boss preparation}
\label{app:play-deck}
We scored 1{,}659 card rewards against a choice model fitted on
about two million reward screens from the telemetry, as the probability that an average human makes the same pick
on the same screen. Agreement barely rises with strength (0.324, 0.313, and 0.296 by tier; $\rho = 0.10$, $p = 0.66$);
on the 92 identical screens seen by both a top and a bottom model the top model agrees more by 0.038 ($p = 0.02$).
The models skip 10--13\% of rewards where the human model expects 23--30\% on the same screens, unrelated to strength.
The average player is not a standard of quality, and agreement with expert players' picks on the experts' own reward screens does rise with strength (\cref{app:expert-decisions-results}).
In the models' own picks, the one clear signal is area damage: on Act~1 screens offering such a card, the top four take one 46\% of the time,
the next eight 30\%, and the bottom eight 18\%, against about 34\% for humans on the same screens ($\rho = 0.79$,
$p < 0.001$; 30 points more on 28 identical screens, $p = 0.002$), and their reasoning cites multi-enemy fights
rather than the boss. Scaling, block, and draw cards show nothing. Although 85--98\% of runs in every tier name the
upcoming boss before reaching it, no tier adapts its deck to it: top decks facing the Slime Boss, which splits into
several enemies, hold 1.94 area-damage cards against 1.50 for other bosses ($p = 0.37$), and the models'
picks follow none of the shifts humans make by boss. Given the state in which each run reaches the Act~1 boss (HP,
maximum HP, relics, deck size, and upgrades), 13\% of the 75-point top--bottom gap in boss wins is explained by a model
fitted on humans and 24--30\% by one fitted on the models; the residual, 52--57 points ($p \le 0.01$), is in the fight
or in what the state omits.

\paragraph{What the diagnostic benchmarks do not see: the act boss.}
Claude Fable~5.1 and Claude Opus~5 clear more acts than Gemini~3.8 Flash (1.60 and 1.30 against 0.60), although Gemini answers as many of the puzzles under the reskin, the naming the runs are played under (83\%, against 83\% and 81\%; \cref{tab:puzzle-reskin}), and makes the expert's choice more often (\cref{tab:expert-decisions}).
The ordinary fights of Act~1 do not separate the three: they take 9.1, 8.6, and 8.7 damage per fight.
The act boss does.
Gemini reaches the Act~1 boss in all ten of its runs and beats it in five; Fable, Opus, and GPT-6 Astra each reach it in nine runs and beat it in all nine.
The state on arrival does not account for this: the model fitted on humans rates the positions alike, at 73\% for Gemini and 77\% to 81\% for the other three.
Gemini builds its deck more as human players do.
It skips 25\% of its card rewards, where the choice model expects 27\% on the same screens and the two Claude models skip 2\% and 5\%, and it reaches the boss with 17.7 cards, against their 20.4 and 20.7.
The Claude models take nearly every card they are offered, arrive with larger decks, and win the fight.
This might have come from an asymmetry in what the models have learnt.
Gemini appears to have learnt from its training data how experts build a deck, but not how to pilot the deck it builds; the Claude models deviate from the experts' deck building, but can make better use the decks they create.
The evidence that Gemini builds as experts do is the expert-decision benchmark (\cref{tab:expert-decisions}): it makes the expert's choice more often than either Claude model in all five deck-building categories, from card rewards (67\% against 55\% and 45\%) to shop visits (27\% against 9\% and 11\%).
That these choices are learnt in pretraining is an inference, supported by the puzzles, where the reskin costs Gemini 10 points of accuracy against 2 and 5 for the two Claude models (\cref{tab:puzzle-reskin}).
The reading rests on eight to ten boss fights per model, and expert-like deck building does not itself prevent good piloting: GPT-6 Astra both follows the experts and wins these fights.
Neither diagnostic benchmark contains such a fight.
The puzzle fights are short and their enemies repeat one attack (\cref{app:puzzles-implementation}), and the expert decisions contain no combat, whereas an act boss is a long fight against an enemy whose behaviour changes as the fight goes on.

\subsection{Combat}
\label{app:play-combat}
From the full logs we measured each combat turn against the state the model was shown
(\cref{tab:combat-micro}). The measures were validated first: a card's displayed damage matches the HP
and block it removed on 99.4\% of 3{,}082 checked plays, and the shown intents less block match the damage taken on
92\% of turns; %
intent-based measures use
only the consistent turns. Measures that do not separate the tiers are as informative as those that do: unblocked
damage the hand could have prevented is about 0.1 HP per turn in every tier, unspent energy 5--6\%, energy spent
2.6--2.7 per turn, and the share of incoming damage blocked 48--52\%. The strong models do not defend better; they
attack better. Paired on the same hallway encounter of the same seed (74 encounters), the top four take 6.9 less
damage than the bottom tier, in 1.10 fewer turns, dealing 5.4 more damage per turn (sign-flip $p < 0.001$ each); on
floors 1--6, where decks are still close to the starter deck, the gaps are 3.2 HP ($p = 0.001$) and 1.03 turns. About
half of the difference in damage taken comes from the extra turns and half from damage per turn, mostly from
attackers left alive. The next-turn HP forecast adds nothing in Act~1 beyond these measures (partial $\rho = -0.05$
after controlling for hallway damage, blocking into non-attacks, and missed lethal). Two examples, with the names the
model saw: on seed 6, floor 8, facing a Zealot [Cultist] whose intent is a buff, Qwen3.8 27B plays three block cards
(``The 20G fully covers the T2 hit \ldots\ at zero HP cost'') and deals 4 damage, while GPT-6 Astra on the same turn
deals 30; and Qwen3.5 122B, with a 9-HP Corrosive Ooze [Acid Slime] and 6 + 4 damage in hand, computes ``Trip Attack
(2 energy): 4 damage \ldots\ not enough to kill'' by applying its own Fatigue [Weak] a second time, and leaves it at
3 HP.

\begin{table}[t]
\centering\small
\caption{\textbf{Combat play in Act~1, by tier.} Tiers of \cref{tab:main-results} (top four, next eight, bottom eight),
with Spearman's $\rho$ against mean acts cleared over the 20 models and its permutation $p$.
Hallway fights are ordinary combats. Block into no attack: turns on which a block card is played although no
enemy intends to attack. Over-block: a block-only card played beyond the incoming damage while an attack was
playable. Missed lethal: turns with one enemy left and enough damage in hand to kill it. The lower block shows
measures that do not separate the tiers ($p > 0.2$).}
\label{tab:combat-micro}
\begin{adjustbox}{max width=\linewidth}
\small
\begin{tabular}{l rrr rr}
\toprule
 & Top 4 & Next 8 & Bottom 8 & $\rho$ & $p$ \\
\midrule
Damage taken per ordinary fight & 8.2 & 9.6 & 14.3 & $-0.89$ & $<0.001$ \\
Turns per ordinary fight & 3.09 & 3.58 & 4.20 & $-0.70$ & $<0.001$ \\
Damage taken per ordinary-fight turn & 2.65 & 2.77 & 3.39 & $-0.57$ & 0.010 \\
Damage dealt per energy on attacks & 7.5 & 6.7 & 5.8 & $+0.91$ & $<0.001$ \\
Damage dealt per energy, all cards & 5.7 & 4.7 & 3.7 & $+0.78$ & $<0.001$ \\
Damage dealt / attack damage in hand & 0.95 & 0.86 & 0.71 & $+0.69$ & 0.001 \\
Block into no attack (\% of turns) & 6 & 15 & 37 & $-0.87$ & $<0.001$ \\
Over-block (\%) & 0.8 & 3.0 & 8.0 & $-0.72$ & $<0.001$ \\
Missed lethal (\%) & 0 & 0.8 & 11.7 & $-0.81$ & $<0.001$ \\
Plays refused as not in hand (per 100 turns) & 1.1 & 2.0 & 10.4 & $-0.61$ & 0.006 \\
Next-turn HP forecast error (HP) & 0.06 & 0.44 & 2.42 & $-0.85$ & $<0.001$ \\
Potions drunk in ordinary fights (\%) & 27 & 27 & 61 & $-0.79$ & $<0.001$ \\
\midrule
Preventable unblocked damage (HP per turn) & 0.11 & 0.06 & 0.06 & & \\
Energy left unspent (\%) & 6 & 5 & 5 & & \\
Energy spent per turn & 2.57 & 2.68 & 2.75 & & \\
Share of incoming damage blocked & 0.48 & 0.52 & 0.51 & & \\
\bottomrule
\end{tabular}
\end{adjustbox}
\end{table}

\clearpage
\section{Difficulty and where runs end}
\label{app:difficulty-outcomes}
\Cref{fig:difficulty} repeats the main configuration at a range of pinned ascension levels: every level
from 0 to 20 for the non-LLM baselines, and levels 0, 1, 5, 10, 15, and 20 for the models run at more than
one. Each level is a separate cohort with its own memory chain. \Cref{fig:outcomes} breaks down in which
act the runs of \cref{tab:main-results} ended.
\Cref{fig:outcomes-series} shows the same runs one by one, in the order each agent played them.
Run order makes little difference: 10 of the 20 models end deeper on average over runs 6--10 than
over runs 1--5, and the other 10 end shallower (mean change $-0.5$ floors). What varies along a row is mostly
the seed. The models agree on which seeds go deep (Kendall's $W = 0.34$ over the 20 models, $p < 10^{-9}$):
the mean final floor ranges from 9.2 on seed~4, where each of the four strongest models dies by floor~7,
to 25.6 on seed~8.

\begin{figure}[htbp]
\centering
\includegraphics[width=\linewidth]{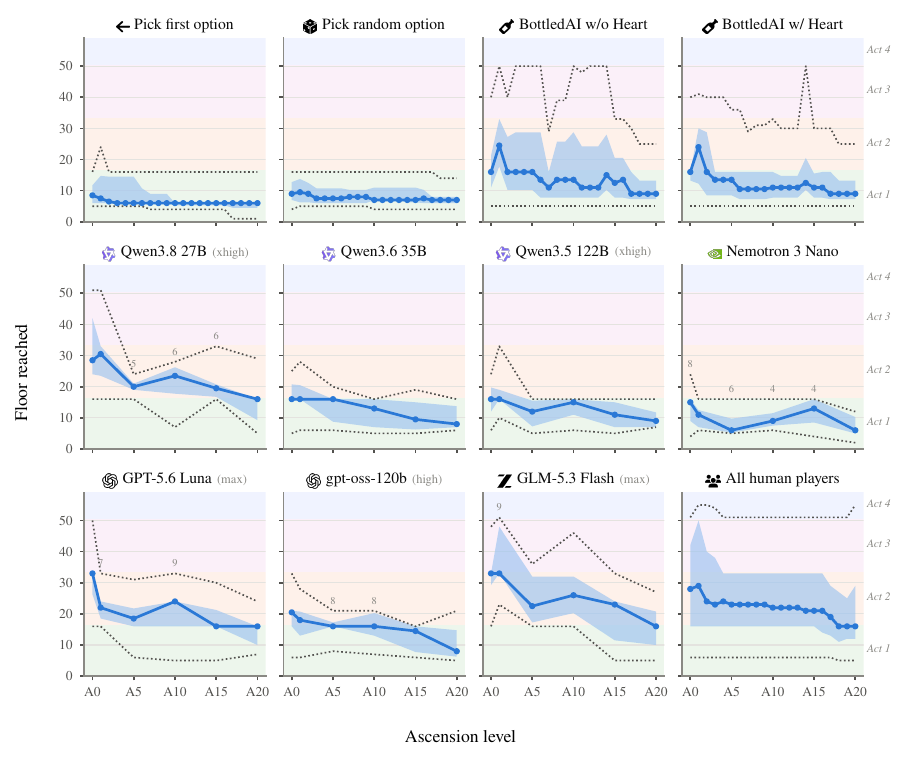}
\caption{\textbf{Effect of difficulty on performance.} Final floor reached against the pinned ascension level,
for every agent in \cref{tab:main-results} with runs at more than one level, on the held-out seeds
under the same configuration as the table. Each level is a separate cohort (its own memory chain). The
solid line is the median floor at each level, the blue band the interquartile range, and the dotted lines
the lowest and highest floor reached. Background shading marks the floors of each act. Levels with fewer than 10 completed seeds are annotated with their seed count.
The last panel is every Ironclad run in Mega Crit's released telemetry (July--November 2020,
final content build; abandoned runs excluded), with the dotted lines at the 5th and 95th percentiles
instead, since over tens of thousands of runs per level the extremes are always the first and last floors reachable. Each
level there is a different population: a level is unlocked by winning at the one below it, so higher
levels hold more experienced players, and only Ascension~20 includes runs that continue to the
Heart.} %
\label{fig:difficulty}
\end{figure}

\begin{figure}[t]
\centering
\includegraphics[width=\linewidth]{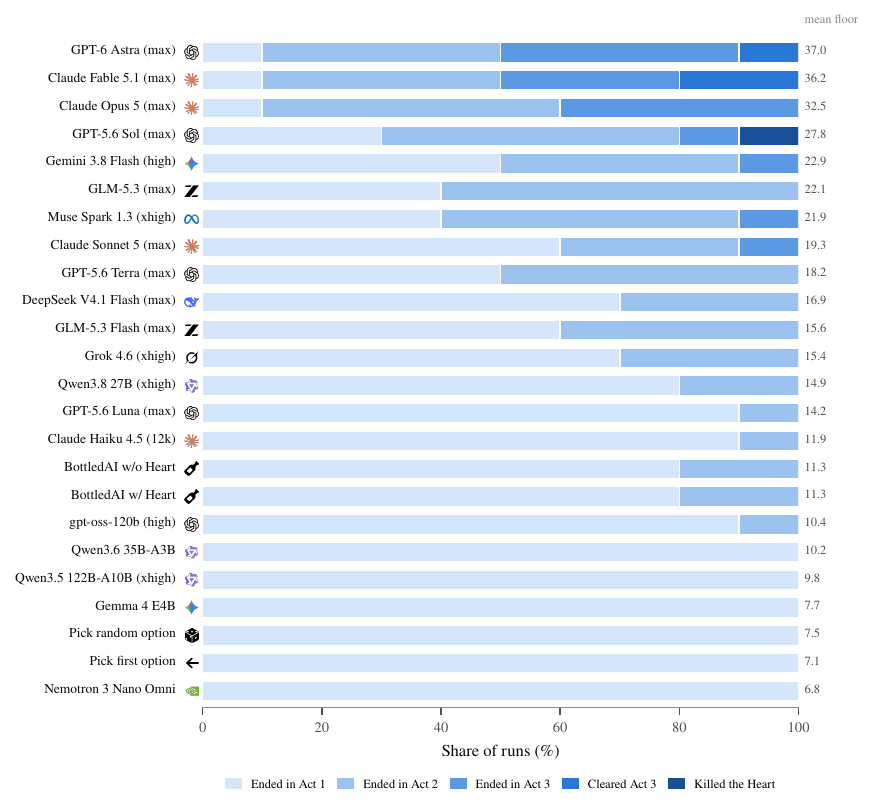}
\caption{\textbf{Where runs end.} For each agent in \cref{tab:main-results}, the share of its runs that ended in each
act, sorted by mean floor (right). Agents with a similar mean floor can differ in whether they reliably reach
the next act or split between early deaths and deep runs.}
\label{fig:outcomes}
\end{figure}

\begin{figure}[htbp]
\centering
\includegraphics[width=0.72\linewidth]{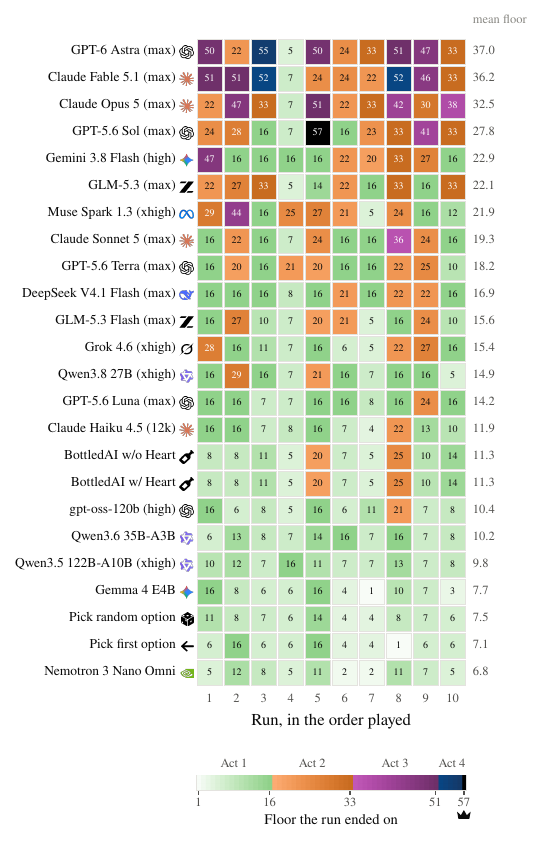}
\caption{%
\textbf{Where each run ended, in the order played.} One square per run of each agent in
\cref{tab:main-results}, with agents sorted by mean floor (right). The hue is the number of acts the run
beat. The lightness is the floor it ended on (the number in the square), on one scale over the whole
game: the deeper the run, the darker the square.
A run that beat Act 4 by killing the Heart (indicated as floor 57) is black.
Every agent plays the same ten seeds in the same order, so each column is one seed: the columns that are pale for every agent (seeds 4, 6, and 7) are presumably harder seeds.
For the LLMs the memory bank written in one run is read by the next, so a row read left to right is the agent's sequence of experience.
}
\label{fig:outcomes-series}
\end{figure}

\clearpage
\section{Calibration}
\label{app:calibration}
\paragraph{Forecasting.} At the end of every floor the model is asked, through the \texttt{predict} tool, for
two probabilities: that it survives the current act, and that it wins the run. When it ends a turn it may
also give, as optional arguments of \texttt{end\_turn}, the HP it expects at the start of its next turn and at
the end of the fight (\cref{app:methods-harness}). %

\paragraph{Scoring.} \Cref{tab:calibration} scores these predictions against what happened, over
the runs of \cref{tab:main-results}. A floor's act-survival prediction is scored as correct if the run
went on to clear that act, and a run-win prediction if it killed the Heart. We report the mean predicted
and observed rates, the Brier score~\citep{brier1950verification}, whose reliability term~\citep{murphy1973vector}
is the calibration error of \cref{fig:calibration-vs-play}, and the AUC,
which measures whether the model gives higher
predictions in the acts it goes on to survive regardless of their overall level. HP predictions are
matched to the next recorded HP on the same floor, and a prediction made in the fight that killed the
player is scored against 0~HP.
A prediction is easier the closer it is made to the moment it resolves: the end of the fight, the end of
the act (or the death that ends it early), the end of the run. Stronger models survive longer and so make
more of their predictions close to an act's end, which would flatter their raw scores. We
therefore bin every prediction by its distance from resolution (\cref{fig:prediction-horizon}) and
report scores standardized to a common mix of distances: each error is fitted as a model effect plus a
distance-bin effect, and each model's fit is averaged over the distance mix of all
models' predictions pooled. The AUC is computed within each distance bin and averaged
with the same weights. Next-turn HP is always one turn ahead and needs no adjustment.

\paragraph{Results.} \Cref{tab:calibration} scores the models' predictions of their own game; how the error depends on the
distance to resolution is in \cref{fig:prediction-horizon}.
\Cref{fig:calibration-vs-play} relates each model's end-of-floor calibration error to the acts
it clears, and \cref{fig:calibration-vs-play-within-floor} does the same for its within-floor HP forecasts,
whose error tracks play more closely (\cref{tab:calibration-horizon}). The act-survival forecast barely
sharpens as the act nears its end, while the forecast of HP at the end of a fight sharpens steeply as the
fight ends: calibration, like play, degrades with the distance to the outcome, which matters for an agent
that must decide how much risk to take on a long task.

\providecommand{\spireicon}[1]{\raisebox{-0.25ex}{\includegraphics[height=1.7ex]{icons/#1.pdf}}}
\begin{table}[htbp]
\centering
\caption{\textbf{How well the models predict their own game.} At the end of every floor each model is asked for its chance of surviving the current act and of winning the run; when ending a turn it may also predict its HP at the start of the next turn and at the end of the fight. Same runs as \cref{tab:main-results}. Pred.\ and Obs.\ are the mean predicted and observed rates (a floor's act counts as survived if the run went on to clear it; a run is won at the Heart). A prediction is easier the closer it is made to the moment it resolves, and stronger models make more of theirs close to an act's end, so the Brier scores (0 is perfect; always answering 50\% scores 0.25), the AUC (0.5 is chance; -- where every act had the same outcome) and the fight-end HP error are adjusted for that distance: each is scored over the same mix of distances, pooled over all models (\cref{fig:prediction-horizon}). HP: mean absolute error and signed bias (predicted minus actual; positive is optimistic) in hit points, with a prediction made in the fight that killed the player scored against 0. Next-turn HP is always one turn ahead, and close to arithmetic since enemy intents show the incoming damage. \best{Best} and \runnerup{second-best} scores among models with all 10 seeds are highlighted.}
\label{tab:calibration}
\begin{adjustbox}{max width=\linewidth}
\small
\begin{tabular}{@{}c@{\hspace{3pt}}l r rrrr rrr rr rr@{}}
\toprule
 & & & \multicolumn{4}{c}{Act survival} & \multicolumn{3}{c}{Run win} & \multicolumn{2}{c}{HP next turn} & \multicolumn{2}{c}{HP at fight end} \\
\cmidrule(lr){4-7} \cmidrule(lr){8-10} \cmidrule(lr){11-12} \cmidrule(lr){13-14}
 & Agent & Runs & Pred.\ \% & Obs.\ \% & Brier & AUC & Pred.\ \% & Obs.\ \% & Brier & MAE & Bias & MAE & Bias \\
\midrule
\spireicon{qwen} & Qwen3.8 27B {\scriptsize\textcolor{gray}{(xhigh)}} & 10 & 51 & 23 & 0.24 & 0.69 & 21 & 0 & 0.05 & 1.1 & $-$0.1 & 10.3 & +4.5 \\
\spireicon{qwen} & Qwen3.6 35B-A3B & 10 & 55 & 0 & 0.32 & -- & 31 & 0 & 0.11 & 3.1 & $-$1.2 & 12.3 & +8.2 \\
\spireicon{qwen} & Qwen3.5 122B-A10B {\scriptsize\textcolor{gray}{(xhigh)}} & 10 & 63 & 0 & 0.42 & -- & 33 & 0 & 0.12 & 2.8 & +0.6 & 17.7 & +17.2 \\
\spireicon{claude} & Claude Fable 5.1 {\scriptsize\textcolor{gray}{(max)}} & 10 & 59 & 76 & 0.18 & 0.80 & 16 & 0 & 0.05 & \runnerup{0.1} & 0.0 & 7.3 & +0.3 \\
\spireicon{claude} & Claude Opus 5 {\scriptsize\textcolor{gray}{(max)}} & 10 & 62 & 68 & \best{0.15} & \runnerup{0.85} & 14 & 0 & 0.04 & 0.2 & $-$0.2 & 8.6 & +0.1 \\
\spireicon{claude} & Claude Sonnet 5 {\scriptsize\textcolor{gray}{(max)}} & 10 & 68 & 45 & 0.30 & 0.58 & 34 & 0 & 0.12 & 0.4 & $-$0.1 & 9.5 & $-$2.9 \\
\spireicon{claude} & Claude Haiku 4.5 {\scriptsize\textcolor{gray}{(12k)}} & 10 & 42 & 15 & 0.25 & 0.39 & 19 & 0 & 0.05 & 3.0 & $-$0.6 & 8.9 & +1.1 \\
\spireicon{deepseek} & DeepSeek V4.1 Flash {\scriptsize\textcolor{gray}{(max)}} & 10 & 37 & 30 & 0.25 & 0.39 & 9 & 0 & \best{0.01} & 0.6 & $-$0.1 & 9.9 & +3.0 \\
\spireicon{gemini} & Gemini 3.8 Flash {\scriptsize\textcolor{gray}{(high)}} & 10 & 86 & 44 & 0.43 & 0.49 & 49 & 0 & 0.26 & \runnerup{0.1} & 0.0 & 10.9 & +10.6 \\
\spireicon{gemini} & Gemma 4 E4B & 10 & 76 & 0 & 0.59 & -- & 61 & 0 & 0.39 & 4.5 & $-$0.3 & 13.4 & +11.7 \\
\spireicon{meta} & Muse Spark 1.3 {\scriptsize\textcolor{gray}{(xhigh)}} & 10 & 54 & 54 & 0.23 & 0.63 & 13 & 0 & \runnerup{0.02} & 0.6 & $-$0.2 & 8.3 & +1.1 \\
\spireicon{nvidia} & Nemotron 3 Nano Omni & 10 & 72 & 0 & 0.53 & -- & 55 & 0 & 0.33 & 4.0 & +0.1 & 13.6 & +10.6 \\
\spireicon{openai} & GPT-6 Astra {\scriptsize\textcolor{gray}{(max)}} & 10 & 74 & 69 & \runnerup{0.16} & \best{0.88} & 22 & 0 & 0.07 & \best{0.0} & 0.0 & \best{6.5} & +3.7 \\
\spireicon{openai} & GPT-5.6 Sol {\scriptsize\textcolor{gray}{(max)}} & 10 & 71 & 63 & 0.22 & 0.68 & 20 & 20 & 0.10 & \best{0.0} & 0.0 & 14.0 & +13.1 \\
\spireicon{openai} & GPT-5.6 Terra {\scriptsize\textcolor{gray}{(max)}} & 10 & 50 & 47 & 0.28 & 0.49 & 12 & 0 & \runnerup{0.02} & 0.4 & $-$0.4 & 13.8 & +13.7 \\
\spireicon{openai} & GPT-5.6 Luna {\scriptsize\textcolor{gray}{(max)}} & 10 & 64 & 12 & 0.37 & 0.64 & 21 & 0 & 0.05 & 0.2 & $-$0.1 & 11.3 & +6.9 \\
\spireicon{openai} & gpt-oss-120b {\scriptsize\textcolor{gray}{(high)}} & 10 & 65 & 17 & 0.40 & 0.56 & 33 & 0 & 0.13 & 1.2 & +0.1 & 16.1 & +14.8 \\
\spireicon{grok} & Grok 4.6 {\scriptsize\textcolor{gray}{(xhigh)}} & 10 & 41 & 33 & 0.24 & 0.54 & 9 & 0 & \best{0.01} & 0.3 & $-$0.1 & \runnerup{7.0} & +2.1 \\
\spireicon{zai} & GLM-5.3 {\scriptsize\textcolor{gray}{(max)}} & 10 & 64 & 46 & 0.27 & 0.59 & 25 & 0 & 0.07 & 0.5 & 0.0 & 9.0 & +3.9 \\
\spireicon{zai} & GLM-5.3 Flash {\scriptsize\textcolor{gray}{(max)}} & 10 & 58 & 46 & 0.26 & 0.54 & 24 & 0 & 0.06 & 1.0 & $-$0.4 & 8.7 & $-$0.1 \\
\bottomrule
\end{tabular}
\end{adjustbox}
\end{table}

\begin{figure}[t]
\centering
\includegraphics[width=0.9\linewidth]{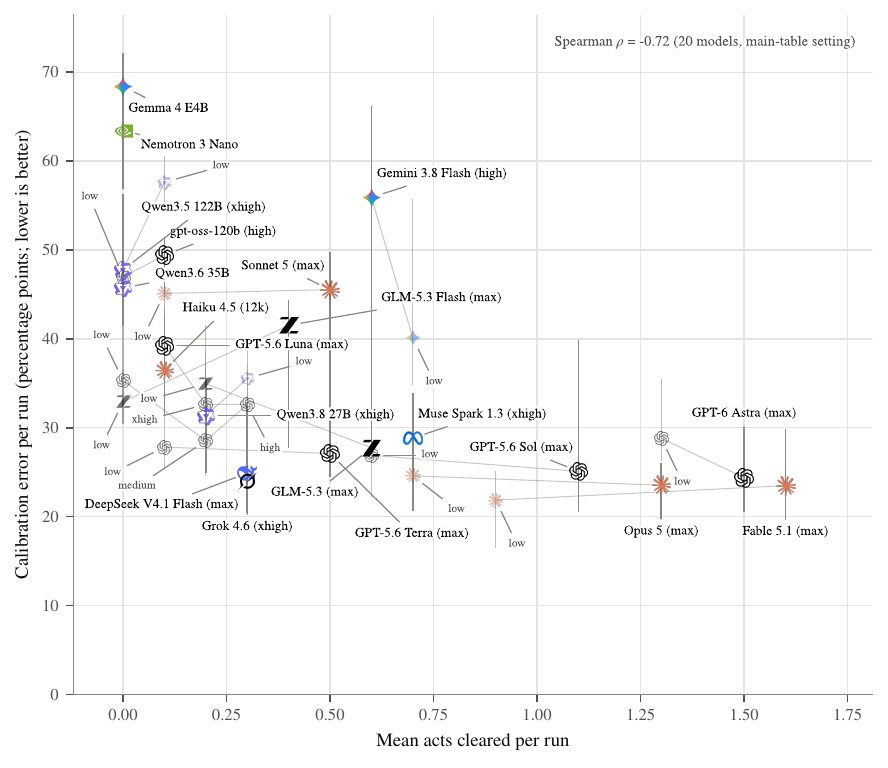}
\caption{\textbf{Do better forecasters play better?} Calibration against play, for each LLM agent in \cref{tab:main-results}. Calibration error is the
reliability term~\citep{murphy1973vector} of the Brier score~\citep{brier1950verification}, in percentage points,
over ten bins of stated probability, for the two forecasts made at the end of every floor (surviving the act,
winning the run), scored separately and averaged. It is computed per run, not adjusted for how far ahead a
prediction was made (\cref{tab:calibration}); points are the median over a model's runs and
bars the interquartile range. Lower effort levels are drawn smaller and paler, joined to the same model. The
Spearman correlation is over the levels in \cref{tab:main-results}, one per model.}
\label{fig:calibration-vs-play}
\end{figure}

\begin{figure}[t]
\centering
\includegraphics[width=\linewidth]{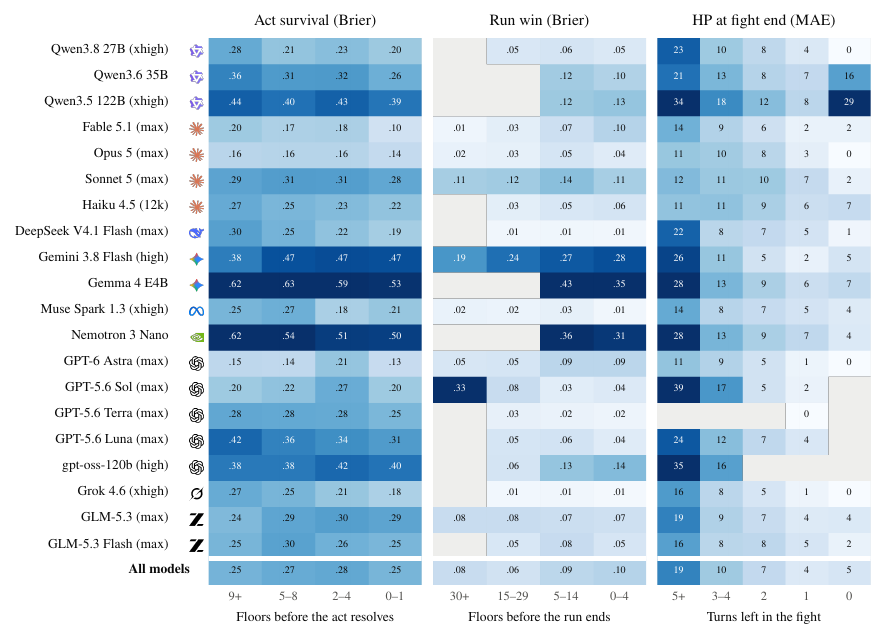}
\caption{\textbf{Prediction error against distance from the outcome, for the runs of \cref{tab:main-results}.}
Left: Brier score of the act-survival prediction against the number of floors before the act resolves
(at its end if the run cleared it, otherwise where the run died). Middle: the same for the run-win
prediction, against floors before the run ends. Right: mean absolute error, in HP, of the predicted HP at
the end of the fight against the number of turns still to be played in it. Darker is worse, on a scale
per panel; blank cells have fewer than 5 predictions. The bottom row pools all models, and its mix of
distances is the one each model is scored over in \cref{tab:calibration}. Fight-end predictions sharpen
steeply as the fight ends; act-survival predictions barely sharpen before the act resolves.}
\label{fig:prediction-horizon}
\end{figure}

\begin{figure}[t]
\centering
\includegraphics[width=0.9\linewidth]{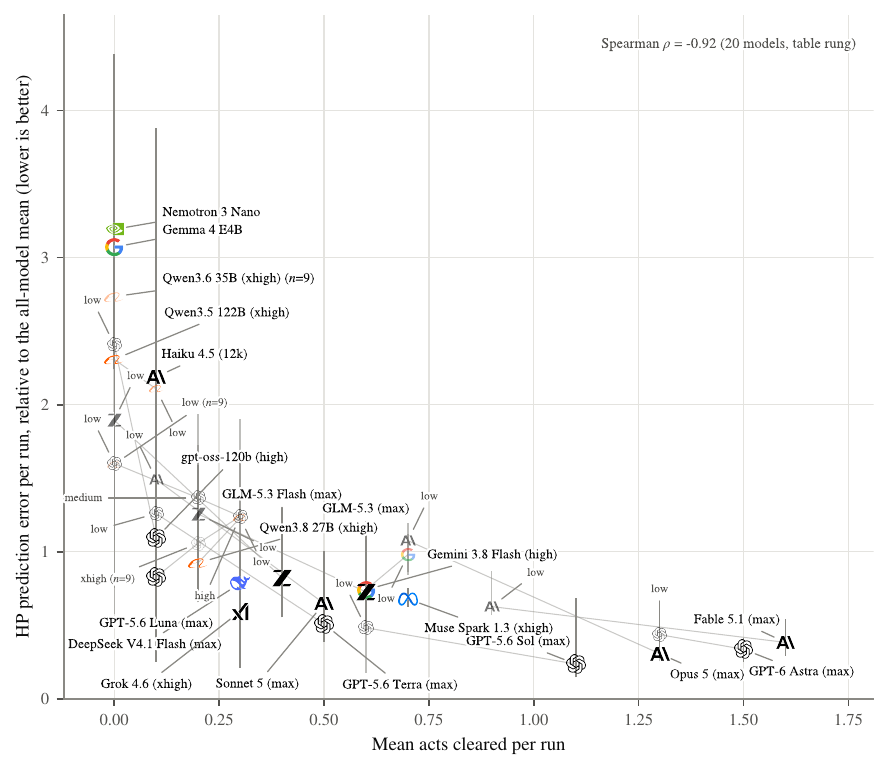}
\caption{\textbf{Do better HP forecasters play better?} As \cref{fig:calibration-vs-play}, for the forecasts made within a floor: the HP a model expects at
the start of its next turn and at the end of the fight, both optional when ending a turn. Each is scored by
its mean absolute error in HP, divided by that forecast's mean error over the predictions of
every model in \cref{tab:main-results} (so 1 is average), and the two are averaged for each run. The division
matters: fight-end errors are about ten times next-turn ones, so without it a run that made only next-turn
predictions would look excellent. A run with no HP predictions has no score. Within-floor error tracks play
more closely than end-of-floor calibration (\cref{tab:calibration-horizon}).}
\label{fig:calibration-vs-play-within-floor}
\end{figure}

\begin{table}[t]
\centering
\caption{\textbf{Which forecasts track play.} The Spearman correlation, over the 20 models of \cref{tab:main-results}, between mean acts cleared and the error of each forecast, pooled over each model's runs: mean absolute error for the two HP forecasts, made when ending a turn, and the calibration error of \cref{fig:calibration-vs-play} for the two probability forecasts, made at the end of every floor. Negative means the models that forecast better play better. Every model plays Act~1, so the Act~1 column compares all of them on forecasts about the same part of the game, while the first column also counts later acts, which only the stronger models reach and where hits are larger. Next-turn HP is close to arithmetic once enemy intents are shown, so its error measures how well a model reads the fight rather than what it knows of its own play. Almost no run is won (1.5\% of run-win forecasts), so the run-win error mostly measures how much chance a model gives itself. The bottom rows correlate the scores plotted in the two figures, which combine two forecasts each and are medians over runs.}
\label{tab:calibration-horizon}
\begin{adjustbox}{max width=\textwidth}
\small
\begin{tabular}{llrr}
\toprule
Forecast & Made at & All predictions & Act~1 only \\
\midrule
HP at the start of the next turn & ending a turn & $-0.84$ & $-0.86$ \\
HP at the end of the fight & ending a turn & $-0.70$ & $-0.82$ \\
Surviving the act & end of a floor & $-0.84$ & $-0.77$ \\
Winning the run & end of a floor & $-0.45$ & $-0.58$ \\
\midrule
Both HP forecasts (\cref{fig:calibration-vs-play-within-floor}) & & $-0.92$ & -- \\
Both probabilities (\cref{fig:calibration-vs-play}) & & $-0.72$ & -- \\
\bottomrule
\end{tabular}
\end{adjustbox}
\end{table}

\FloatBarrier
\section{Further analyses}
\label{app:further-analyses}
The analyses below support the Results and Discussion (\cref{sec:results,sec:discussion}). Unless stated otherwise, all are over the runs of
\cref{tab:main-results}, with the tests of \cref{app:methods-metrics}. %

\subsection{Time horizon}\label{app:further-time-horizon}
\providecommand{\spireicon}[1]{\raisebox{-0.25ex}{\includegraphics[height=1.7ex]{icons/#1.pdf}}}
\begin{table}[t]
\centering
\caption{\textbf{Success at each decision scale.} Scales of \cref{fig:game-overview}, for the LLM agents of \cref{tab:main-results} (10 runs each). Turn is the share of combat
turns without a rejected action, format error, or energy mismatch (a lower bound, since every error is charged to a different turn).
Combat is the share of fights won; A1 hallway is the share of ordinary Act~1 fights (not elites or the boss) won, and Boss the share of boss fights won (superscript: boss fights played). Act is the share of acts cleared among
acts entered, pooled over Acts~1--3. Run is the share of runs that cleared Act~3. $h_{50}$ is the
human time at which a logistic in $\log$ time, fitted with each scale weighted equally, crosses 50\%,
with each scale placed at a human time from Mega Crit's telemetry: 10\,s for a turn, 10\,s per turn
of an average fight, the mean of the median times to complete Acts~1--3 (8.7, 12.2, and 20.6\,min, taken
from players who completed each act), and the median run that kills the Heart (50\,min). Human rows come from \cref{app:human-baselines} and have no turn or combat data.}
\label{tab:time-horizon}
\begin{adjustbox}{max width=\linewidth}
\small
\begin{tabular}{@{}l@{\hspace{3pt}}lrrrrrrrr@{}}
\toprule
 & & \multicolumn{6}{c}{Success (\%)} & & \\
\cmidrule(lr){3-8}
 & Agent & Turn & Combat & A1 hallway & Boss & Act & Run & $h_{50}$ & Acts \\
\midrule
\spireicon{claude} & Claude Fable 5.1 {\scriptsize\textcolor{gray}{(max)}} & 98.6 & 95 & 100 & 80\rlap{$^{20}$} & 67 & 20 & 19\,min & 1.60 \\
\spireicon{openai} & GPT-6 Astra {\scriptsize\textcolor{gray}{(max)}} & 98.3 & 95 & 98 & 75\rlap{$^{20}$} & 62 & 10 & 14\,min & 1.50 \\
\spireicon{claude} & Claude Opus 5 {\scriptsize\textcolor{gray}{(max)}} & 99.3 & 94 & 100 & 81\rlap{$^{16}$} & 57 & 0 & 11\,min & 1.30 \\
\spireicon{openai} & GPT-5.6 Sol {\scriptsize\textcolor{gray}{(max)}} & 98.9 & 94 & 100 & 73\rlap{$^{15}$} & 53 & 10 & 12\,min & 1.10 \\
\spireicon{meta} & Muse Spark 1.3 {\scriptsize\textcolor{gray}{(xhigh)}} & 99.6 & 91 & 96 & 78\rlap{$^{9}$} & 41 & 0 & 7\,min & 0.70 \\
\spireicon{gemini} & Gemini 3.8 Flash {\scriptsize\textcolor{gray}{(high)}} & 99.8 & 92 & 100 & 50\rlap{$^{12}$} & 38 & 0 & 7\,min & 0.60 \\
\spireicon{zai} & GLM-5.3 {\scriptsize\textcolor{gray}{(max)}} & 99.1 & 91 & 96 & 55\rlap{$^{11}$} & 38 & 0 & 6\,min & 0.60 \\
\spireicon{openai} & GPT-5.6 Terra {\scriptsize\textcolor{gray}{(max)}} & 99.3 & 90 & 100 & 56\rlap{$^{9}$} & 33 & 0 & 6\,min & 0.50 \\
\spireicon{claude} & Claude Sonnet 5 {\scriptsize\textcolor{gray}{(max)}} & 99.4 & 90 & 100 & 50\rlap{$^{10}$} & 33 & 0 & 6\,min & 0.50 \\
\spireicon{zai} & GLM-5.3 Flash {\scriptsize\textcolor{gray}{(max)}} & 99.1 & 88 & 98 & 67\rlap{$^{6}$} & 29 & 0 & 5\,min & 0.40 \\
\spireicon{deepseek} & DeepSeek V4.1 Flash {\scriptsize\textcolor{gray}{(max)}} & 99.4 & 89 & 100 & 33\rlap{$^{9}$} & 23 & 0 & 4\,min & 0.30 \\
\spireicon{grok} & Grok 4.6 {\scriptsize\textcolor{gray}{(xhigh)}} & 95.7 & 88 & 98 & 50\rlap{$^{6}$} & 23 & 0 & 4\,min & 0.30 \\
\spireicon{qwen} & Qwen3.8 27B {\scriptsize\textcolor{gray}{(xhigh)}} & 96.6 & 88 & 98 & 29\rlap{$^{7}$} & 17 & 0 & 3\,min & 0.20 \\
\spireicon{openai} & gpt-oss-120b {\scriptsize\textcolor{gray}{(high)}} & 87.5 & 83 & 96 & 33\rlap{$^{3}$} & 9 & 0 & 2\,min & 0.10 \\
\spireicon{openai} & GPT-5.6 Luna {\scriptsize\textcolor{gray}{(max)}} & 99.4 & 87 & 98 & 14\rlap{$^{7}$} & 9 & 0 & 3\,min & 0.10 \\
\spireicon{claude} & Claude Haiku 4.5 {\scriptsize\textcolor{gray}{(12k)}} & 96.4 & 84 & 98 & 25\rlap{$^{4}$} & 9 & 0 & 2\,min & 0.10 \\
\spireicon{qwen} & Qwen3.5 122B-A10B {\scriptsize\textcolor{gray}{(xhigh)}} & 94.0 & 81 & 93 & 0\rlap{$^{1}$} & 0 & 0 & 86\,s & 0.00 \\
\spireicon{qwen} & Qwen3.6 35B-A3B & 93.9 & 84 & 94 & 0\rlap{$^{2}$} & 0 & 0 & 2\,min & 0.00 \\
\spireicon{nvidia} & Nemotron 3 Nano Omni & 94.6 & 77 & 78 & -- & 0 & 0 & 77\,s & 0.00 \\
\spireicon{gemini} & Gemma 4 E4B & 87.1 & 78 & 89 & 0\rlap{$^{2}$} & 0 & 0 & 75\,s & 0.00 \\
\midrule
 & Human, all A20 Ironclad (telemetry) & -- & -- & -- & -- & 43 & 7 & -- & 0.72 \\
 & Human, 1000+ h players (telemetry) & -- & -- & -- & -- & 47 & 10 & -- & 0.87 \\
 & Human, Baalorlord, complete log & -- & -- & -- & -- & 89 & 70 & -- & 2.98 \\
\bottomrule
\end{tabular}
\end{adjustbox}
\end{table}

\emph{Slay the Spire} interleaves four kinds of decision, on four time scales, over one
irreversible pool of resources: tactics within a turn (seconds), pricing HP within a fight (minutes),
building a deck within an act (10--20 minutes) and setting a risk posture over a run (an hour). Playing it
over such spans also needs memory management: deciding what to keep across floors, where each floor is a
fresh conversation, and across runs, where the environment starts again. The time scales are the ones on
which agent ability is now charted: \citet{kwa2025horizon} measure an agent by the length of task, in human
working time, that it completes half the time, and find it doubling about every seven months on software
tasks. A run here is an hour of human play, of the order of the horizons they report for frontier models,
but it is one task made of decisions on every shorter scale, so the benchmark can show which scale a model
fails on rather than only how long a task it survives. Measured this way (\cref{tab:time-horizon}), every
model succeeds on at least 87\% of its turns and 77\% of its fights, and falls below 50\% between a fight and
a run: for all but the top four at the act, for those four (53--67\% of acts) only at the run. The fitted 50\% horizons run from about one minute of human
play (Gemma~4 E4B) to about 19 minutes (Claude Fable~5.1), on act and run times taken from Mega
Crit's telemetry (players who complete Acts~1, 2, and 3 take about 9, 12, and 21 minutes over
them, and a run that kills the Heart takes a median of 50 minutes); they order the models as acts cleared
does, largely by construction, since the act is the only scale near 50\%. What the measure does show is
that the fall is a step rather than a slope. Of the 20 models, 18 win 93--100\% of their ordinary Act~1
fights, but boss fights, won 0--81\% of the time, decide where each model stops, and human players at this
level show the same step, clearing 43\% of the acts they enter. The difficulty is therefore not a limit on
how long a model can keep going, as a time horizon on a smooth task distribution would suggest, but rare,
high-stakes decisions inside a long task, whose preparation has to begin floors earlier.

\subsection{Public benchmarks and reasoning effort}\label{app:further-benchmarks}
The 30 model and effort points of \cref{fig:benchmarks}
are 20 models at their highest reasoning effort and 10 of them again at low effort. Acts cleared tracks the
Intelligence Index about as closely within each effort level as across both ($r = 0.80$ at the highest effort,
$0.91$ at low). Within a model the Index is a poorer guide. Lowering the effort costs acts for 8 of the 10
models with an Artificial Analysis entry at both levels (\cref{tab:ablation-effort}), but the size of the loss does not follow the size of the
drop on the Index (Spearman $\rho = 0.06$): Gemini~3.8 Flash and Qwen3.8 27B each lose about 7.5 Index
points and clear slightly more acts at low effort, while Claude Fable~5.1 loses 6.5 and clears 0.7 fewer
acts. SpireBench ranks models much as the public suite does, but measures something the suite does not
already capture. It is what the long-horizon literature would predict: if success on a long task compounds
per-step reliability~\citep{sinha2026illusion}, small differences on short evaluations should widen into
large ones over a run, and the steep rise of acts cleared with the Index (\cref{fig:benchmarks}) is
consistent with that.

\begin{table}[t]
\centering
\caption{\textbf{SpireBench against public benchmarks.} Artificial Analysis's benchmarks, over the 20 LLM agents of \cref{tab:main-results} with all 10 seeds at their highest reasoning effort, and 10 of them again at low effort (30 points), each matched to the Artificial Analysis entry for the reasoning setting it ran at (scores retrieved 2026-09-24). A low rung is included where it has all 10 seeds and Artificial Analysis lists that setting. \emph{Left}: correlation of each benchmark with mean acts cleared, over both rungs together and within each rung ($n$ is the number of points with that evaluation). \emph{Right}: how substitutable each benchmark is by the rest of the suite, with SpireBench added as one more row: its strongest correlation with any other single evaluation, its mean correlation with the others, and the share of its variance across points that the first principal component of the others predicts for a model left out of the fit, both rungs of it together ($R^2_{\mathrm{LOO}}$; negative is worse than predicting the mean). The Intelligence Index is a composite of evaluations like these, so it is left out of the right-hand block, which uses the points with every evaluation (25).}
\label{tab:benchmarks}
\begin{adjustbox}{max width=\linewidth}
\small
\begin{tabular}{@{}l r rr rr rrr@{}}
\toprule
 & \multicolumn{5}{c}{Correlation with SpireBench} & \multicolumn{3}{c}{Substitutability} \\
\cmidrule(lr){2-6} \cmidrule(lr){7-9}
 & \multicolumn{3}{c}{Both rungs} & \multicolumn{2}{c}{Pearson $r$ by rung} & & & \\
\cmidrule(lr){2-4} \cmidrule(lr){5-6}
Benchmark & $n$ & Pearson $r$ & Spearman $\rho$ & Highest & Low & max $r$ & mean $r$ & $R^2_{\mathrm{LOO}}$ \\
\midrule
AA Intelligence Index & 30 & 0.82 & 0.89 & 0.80 & 0.91 & -- & -- & -- \\
GPQA Diamond & 28 & 0.58 & 0.80 & 0.56 & 0.74 & 0.86 & 0.70 & 0.51 \\
Humanity's Last Exam & 30 & 0.82 & 0.89 & 0.81 & 0.87 & 0.92 & 0.83 & 0.85 \\
SciCode & 27 & 0.70 & 0.81 & 0.73 & 0.60 & 0.87 & 0.79 & 0.75 \\
Long-context reasoning & 30 & 0.52 & 0.66 & 0.47 & 0.70 & 0.74 & 0.64 & 0.29 \\
CritPt & 30 & 0.82 & 0.83 & 0.84 & 0.82 & 0.92 & 0.78 & 0.70 \\
AA-Omniscience & 30 & 0.83 & 0.91 & 0.82 & 0.87 & 0.86 & 0.77 & 0.70 \\
\midrule
\textbf{SpireBench} (acts cleared) & 25 & -- & -- & -- & -- & 0.83 & 0.71 & 0.50 \\
\bottomrule
\end{tabular}
\end{adjustbox}
\end{table}

\subsection{Agentic residuals}\label{app:further-agentic-residuals}
\providecommand{\spireicon}[1]{\raisebox{-0.25ex}{\includegraphics[height=1.7ex]{icons/#1.pdf}}}
\begin{table}[t]
\centering
\caption{\textbf{Which models play below, or above, what their public agentic scores predict.} For each Artificial Analysis evaluation, mean acts cleared (\cref{tab:main-results}, highest reasoning effort, all 10 seeds) is regressed on the evaluation score across models, one point per model; cells are internally studentized residuals ($z$; negative means fewer acts than the evaluation predicts). The top rows give each evaluation's correlation with acts cleared, with two-sided permutation $p$ (200{,}000 permutations) and $n$ models. Models are sorted by their mean residual over the agentic evaluations (left block). Artificial Analysis scores retrieved 2026-09-24.
Terminal:~Terminal-Bench.
HLE:~Humanity's Last Exam.
}
\label{tab:agentic-residuals}
\begin{adjustbox}{max width=\linewidth}
\small
\begin{tabular}{@{}c@{\hspace{3pt}}l r rrrr r rr@{}}
\toprule
 & & & \multicolumn{4}{c}{Agentic} & & \multicolumn{2}{c}{Reasoning} \\
\cmidrule(lr){4-7} \cmidrule(lr){9-10}
 & Model & Acts & $\tau$-Bench & Terminal 2.1 & Terminal 4.0 & GDPval (normalized) & Mean $z$ & HLE & GPQA Diamond \\
\midrule
 & \textit{Pearson $r$} & & 0.59 & 0.65 & 0.89 & 0.67 & & 0.81 & 0.56 \\
 & \textit{$p$} & & 0.013 & 0.004 & <0.001 & 0.002 & & <0.001 & 0.014 \\
 & \textit{$n$} & & 17 & 19 & 18 & 20 & & 20 & 19 \\
\midrule
\spireicon{deepseek} & DeepSeek V4.1 Flash & 0.30 & -- & -- & $-$1.41 & $-$1.16 & $-$1.29 & $-$0.91 & -- \\
\spireicon{openai} & GPT-5.6 Luna & 0.10 & $-$0.87 & $-$1.41 & $-$0.77 & $-$1.36 & $-$1.10 & $-$1.59 & $-$1.21 \\
\spireicon{zai} & GLM-5.3 Flash & 0.40 & $-$0.97 & $-$0.75 & $-$1.60 & $-$0.99 & $-$1.07 & $-$0.63 & $-$0.52 \\
\spireicon{grok} & Grok 4.6 & 0.30 & $-$1.00 & $-$1.12 & $-$0.45 & $-$1.24 & $-$0.95 & $-$1.32 & $-$0.87 \\
\spireicon{qwen} & Qwen3.8 27B & 0.20 & $-$1.50 & $-$1.12 & 0.29 & $-$1.02 & $-$0.84 & $-$0.80 & $-$0.95 \\
\spireicon{zai} & GLM-5.3 & 0.60 & $-$0.65 & $-$0.23 & $-$1.68 & $-$0.46 & $-$0.75 & $-$0.16 & $-$0.08 \\
\spireicon{openai} & GPT-5.6 Terra & 0.50 & $-$0.37 & $-$0.61 & $-$1.42 & $-$0.27 & $-$0.67 & $-$0.55 & $-$0.35 \\
\spireicon{qwen} & Qwen3.5 122B-A10B & 0.00 & $-$0.34 & $-$0.70 & $-$0.03 & $-$0.31 & $-$0.34 & $-$0.75 & $-$1.17 \\
\spireicon{qwen} & Qwen3.6 35B-A3B & 0.00 & $-$0.02 & $-$0.63 & $-$0.03 & $-$0.47 & $-$0.29 & $-$0.50 & $-$1.09 \\
\spireicon{claude} & Claude Sonnet 5 & 0.50 & $-$0.22 & $-$0.38 & 0.72 & $-$0.31 & $-$0.05 & $-$0.41 & $-$0.28 \\
\spireicon{gemini} & Gemini 3.8 Flash & 0.60 & $-$0.37 & $-$0.34 & 0.59 & 0.03 & $-$0.02 & $-$0.63 & $-$0.25 \\
\spireicon{claude} & Claude Haiku 4.5 & 0.10 & 0.25 & $-$0.35 & 0.42 & 0.15 & 0.12 & 0.89 & $-$0.03 \\
\spireicon{meta} & Muse Spark 1.3 & 0.70 & $-$0.24 & $-$0.02 & 1.33 & $-$0.15 & 0.23 & $-$0.26 & 0.04 \\
\spireicon{openai} & gpt-oss-120b & 0.10 & 0.05 & 0.19 & 0.42 & 0.42 & 0.27 & 0.06 & $-$0.57 \\
\spireicon{nvidia} & Nemotron 3 Nano Omni & 0.00 & -- & 0.56 & -- & 0.36 & 0.46 & 1.07 & 1.00 \\
\spireicon{gemini} & Gemma 4 E4B & 0.00 & -- & 0.75 & -- & 0.36 & 0.55 & 1.18 & 0.24 \\
\spireicon{openai} & GPT-5.6 Sol & 1.10 & 0.87 & 0.92 & 0.74 & 1.01 & 0.88 & 0.92 & 0.97 \\
\spireicon{claude} & Claude Opus 5 & 1.30 & 1.45 & 1.40 & 0.71 & 1.31 & 1.22 & 1.16 & 1.48 \\
\spireicon{openai} & GPT-6 Astra & 1.50 & 1.96 & 1.94 & 0.59 & 2.17 & 1.66 & 1.87 & 1.82 \\
\spireicon{claude} & Claude Fable 5.1 & 1.60 & 1.94 & 2.11 & 1.79 & 2.07 & 1.98 & 1.87 & 2.16 \\
\bottomrule
\end{tabular}
\end{adjustbox}
\end{table}

Do models that score well on public agentic evaluations nonetheless fail at
the game? Regressing mean acts cleared on each agentic evaluation Artificial Analysis reports, one point
per model at its highest effort (\cref{tab:agentic-residuals}), Grok~4.6, GLM-5.3 Flash, and GPT-5.6 Luna
clear fewer acts than all four predict, and Claude Fable~5.1, GPT-6 Astra, and Claude Opus~5 more. The
shortfall is not specific to agentic skill, however: the same models fall short of their scores on
Humanity's Last Exam and GPQA Diamond by about as much (the mean residuals correlate at
$r = 0.92$). Only one agentic evaluation, Terminal-Bench~4.0, predicts the game better than
Humanity's Last Exam ($r = 0.89$ against $0.81$) and adds information once it is controlled
for (partial $r = 0.64$, permutation $p = 0.006$, $n = 18$); $\tau$-Bench Banking, Terminal-Bench~2.1, and
GDPval predict it less well and add nothing.

\subsection{Runs ended by the harness}\label{app:further-harness-ended}
The Orak comparison runs Orak at each
model's own reasoning setting; its $85\%$ saving in tokens per floor, $31\%$ in cost once our cached
prefix is priced at its own rate, comes largely from its rule bot playing every non-combat floor without a
model call and from its shorter prompts, which is why we compare models on cost fronts
(\cref{fig:pareto}) rather than on score alone. A harness can also end runs itself, and those endings are
charged to the model. A harness that supports most of a game but not all of it stops a run when it meets
what it cannot handle, and every per-run figure then records that as the model dying there. On these
seeds Orak's harness ended 22 of its launches itself, by crashing on an empty model reply or
looping without progress, and its own failure ended 11 of the 70 seeds; ours ended none of its launches on
these arms, and 6 of the 2{,}432 it made across every arm. %
We count a run a harness ended as a finished run lost by that harness, and report
these endings separately, since a benchmark that did not would score the harness's gaps as the
model's mistakes. That is the
practical case for a harness that plays the whole game.

\subsection{One conversation per floor}\label{app:further-conversation}
The continuous conversation, which keeps one conversation for the whole run (\cref{tab:harness-compare}),
is compared on the same floors; a fresh conversation per floor, with the run's notes and
history carried forward, avoids both a long context used poorly~\citep{liu2024lost} and earlier mistakes
kept in view~\citep{sinha2026illusion}. Vending-Bench finds long-running agents derailing in ways
unrelated to when the context fills~\citep{backlund2025vendingbench}; our comparison is narrower, holding
the task fixed and changing only how much of its history the model is shown.

\subsection{What the memory banks hold}\label{app:further-memory-banks}
\providecommand{\spireicon}[1]{\raisebox{-0.25ex}{\includegraphics[height=1.7ex]{icons/#1.pdf}}}
\begin{table}[t]
\centering
\caption{\textbf{What the models keep in their memory bank after the tenth seed of \cref{tab:main-results}, classified by hand.} Bank is the final number of memories; Created, Rewritten, and Deleted count consolidation operations over the ten runs (Rewritten excludes verbatim re-saves). Composition is the share of the final bank in each category (a random 30 per model where the bank is larger): a game fact describes how the game works, a heuristic advises, an absolute rule prescribes without condition, an observation recounts one run; Other is route plans and notes about the harness. Wrong is the share stating a false game fact or a rule that fails in common situations, and H.\ how many of the coded wrong memories trace to harness text that was itself wrong (a rest-site key label, a state diff across the two final bosses, and debuff expiry shown as an enemy's action); Absolute the share phrased as ALWAYS, NEVER, or MUST anywhere; One run the share drawn from a single run. Models are ordered by mean acts cleared. Operations are read from the full run logs.}
\label{tab:memory-classification}
\begin{adjustbox}{max width=\linewidth}
\small
\begin{tabular}{@{}c@{\hspace{3pt}}l r rrrr rrrrr rrrr@{}}
\toprule
 & & & \multicolumn{4}{c}{Memories} & \multicolumn{5}{c}{Composition (\%)} & & & & \\
\cmidrule(lr){4-7} \cmidrule(lr){8-12}
 & Agent & Acts & Bank & Created & Rewritten & Deleted & Fact & Heur. & Rule & Obs. & Other & Wrong (\%) & H. & Absolute (\%) & One run (\%) \\
\midrule
\spireicon{claude} & Claude Fable 5.1 & 1.60 & 50 & 50 & 158 & 0 & 60 & 27 & 0 & 13 & 0 & 13 & 4 & 63 & 23 \\
\spireicon{openai} & GPT-6 Astra & 1.50 & 149 & 149 & 110 & 0 & 87 & 13 & 0 & 0 & 0 & 0 & -- & 3 & 53 \\
\spireicon{claude} & Claude Opus 5 & 1.30 & 82 & 83 & 228 & 1 & 33 & 63 & 3 & 0 & 0 & 17 & 1 & 60 & 33 \\
\spireicon{openai} & GPT-5.6 Sol & 1.10 & 106 & 106 & 76 & 0 & 100 & 0 & 0 & 0 & 0 & 0 & -- & 7 & 67 \\
\spireicon{meta} & Muse Spark 1.3 & 0.70 & 103 & 103 & 64 & 0 & 60 & 30 & 7 & 0 & 3 & 7 & -- & 37 & 90 \\
\spireicon{zai} & GLM-5.3 & 0.60 & 44 & 44 & 57 & 0 & 63 & 20 & 10 & 0 & 7 & 20 & -- & 30 & 47 \\
\spireicon{gemini} & Gemini 3.8 Flash & 0.60 & 79 & 79 & 35 & 0 & 93 & 7 & 0 & 0 & 0 & 3 & -- & 10 & 63 \\
\spireicon{claude} & Claude Sonnet 5 & 0.50 & 108 & 108 & 54 & 0 & 73 & 27 & 0 & 0 & 0 & 3 & -- & 10 & 57 \\
\spireicon{openai} & GPT-5.6 Terra & 0.50 & 29 & 29 & 15 & 0 & 83 & 17 & 0 & 0 & 0 & 0 & -- & 3 & 83 \\
\spireicon{zai} & GLM-5.3 Flash & 0.40 & 74 & 76 & 67 & 2 & 87 & 10 & 3 & 0 & 0 & 20 & 1 & 20 & 63 \\
\spireicon{grok} & Grok 4.6 & 0.30 & 98 & 98 & 81 & 0 & 70 & 23 & 0 & 0 & 7 & 0 & -- & 20 & 93 \\
\spireicon{deepseek} & DeepSeek V4.1 Flash & 0.30 & 34 & 34 & 16 & 0 & 93 & 0 & 7 & 0 & 0 & 13 & 1 & 30 & 60 \\
\spireicon{qwen} & Qwen3.8 27B & 0.20 & 79 & 79 & 56 & 0 & 63 & 27 & 3 & 0 & 7 & 10 & -- & 20 & 57 \\
\spireicon{claude} & Claude Haiku 4.5 & 0.10 & 69 & 80 & 3 & 11 & 7 & 37 & 50 & 7 & 0 & 60 & -- & 60 & 100 \\
\spireicon{openai} & GPT-5.6 Luna & 0.10 & 32 & 32 & 17 & 0 & 60 & 33 & 0 & 3 & 3 & 7 & -- & 13 & 93 \\
\spireicon{openai} & gpt-oss-120b & 0.10 & 79 & 79 & 2 & 0 & 27 & 70 & 3 & 0 & 0 & 23 & -- & 3 & 3 \\
\spireicon{nvidia} & Nemotron 3 Nano Omni & 0.00 & 66 & 66 & 4 & 0 & 20 & 43 & 30 & 0 & 7 & 57 & -- & 27 & 10 \\
\spireicon{gemini} & Gemma 4 E4B & 0.00 & 24 & 24 & 7 & 0 & 0 & 92 & 8 & 0 & 0 & 12 & -- & 8 & 4 \\
\spireicon{qwen} & Qwen3.6 35B-A3B & 0.00 & 13 & 27 & 25 & 14 & 62 & 15 & 23 & 0 & 0 & 54 & -- & 46 & 38 \\
\spireicon{qwen} & Qwen3.5 122B-A10B & 0.00 & 56 & 58 & 5 & 2 & 27 & 30 & 37 & 7 & 0 & 40 & -- & 53 & 87 \\
\bottomrule
\end{tabular}
\end{adjustbox}
\end{table}

Consolidating notes into a memory bank does not raise the acts a model clears (\cref{sec:results}). Of the
memory arms, only carrying notes forward verbatim does better than no memory, gaining $+0.03$ acts, $+1.1$
floors, and $+3$ points of Act~1 win rate (\cref{tab:ablation-memory}), which is suggestive rather than
established. This agrees with the line of work on unverified self-correction and memory
propagation~\citep{huang2024selfcorrect,xiong2025memory} rather than contradicting the reports of
improvement, and the problem it exposes only shows where episodes are long, stochastic, and repeated.
Reading the banks shows why (\cref{tab:memory-classification}). Of the memories the models carry out of
their tenth run, 59\% describe how the game works, 29\% are advice, 9\% are rules stated without condition,
and 2\% recount a single run, and 17\% are false.
Most
are the model's own generalizations from one run. Claude Opus~5 (incorrectly) records that another boss
``NEVER SUMMONS AGAIN'' and generalizes it to ``Most summon exactly ONCE''; Muse Spark (incorrectly) decides on its
second seed that an area attack ``does NOT scale with Strength'', misreading a damage figure that already
included its Strength, and never revisits it. Rules are the error-prone kind, 78\% of them false against
9\% of statements of fact, and models whose banks hold more errors of their own clear fewer acts
(Spearman $\rho = -0.67$, permutation $p = 0.002$, $n = 20$). How emphatically a memory is phrased does not
matter ($\rho = -0.03$): Claude Fable~5.1 and Opus~5 write in capitals as often as Claude Haiku~4.5, but
what they write is largely true. Little prunes the errors. Over the 200 runs only 5 of the 20 models
delete any memory, 30 in all, and 27 of the deletions are by three of the weakest models (Qwen3.6, 14;
Claude Haiku~4.5, 11; Qwen3.5, 2). The deletions remove correct memories nearly as often as false ones (11 against
17 of the 30). Wrong memories are rewritten about as often as correct ones without being corrected.
DeepSeek~V4.1 Flash records correctly that the act transition restores most of the HP it has lost,
overwrites it two seeds later with ``the post-boss heal is small'', and then drops the subject. Without a
check against outcomes, consolidation keeps whatever a single run suggested. Whether the models learn from
their banks across seeds at all is in \cref{app:play-learning}.

\subsection{Reskin cost by model family}\label{app:further-reskin-family}
We find no evidence that the reskin's cost follows
model family, and so cannot tie it to particular training corpora: a permutation test on developer labels
finds no more variation between families than within them (11 models from 7 developers; $p = 0.29$ for
acts, $p = 0.12$ for floors), and within a family the cost differs sharply, with GLM-5.3 Flash losing 7.5
floors against GLM-5.3's 2.3, and GPT-5.6 Luna 4.3 against none for gpt-oss-120b. The cost
appears only in models that clear 0.6--0.8 acts with the real names (Spearman $\rho = -0.67$, permutation
$p = 0.029$, between acts lost and acts cleared with real names), which is partly mechanical, since a model
that clears no acts has none to lose. The reskin removes knowledge that only the stronger players were
using. Where nonsense predicates alone collapse planning in Mystery
Blocksworld~\citep{valmeekam2023planbench}, here nonsense names cost least; a familiar name pulls a model
towards what that name usually means, which helps when it is right and misleads when it belongs to
something else~\citep{mccoy2024embers}.

\subsection{Recognition of the game under the reskin}\label{app:further-reskin-recognition}
\providecommand{\spireicon}[1]{\raisebox{-0.25ex}{\includegraphics[height=1.7ex]{icons/#1.pdf}}}
\begin{table}[t]
\centering
\caption{\textbf{Recognition of the reskinned game.} From each run's own text (reasoning, prose, and tool-call arguments). Per arm: \emph{Rec.} runs whose text names the real game or a real entity name the arm does not display; \emph{Game} runs naming ``Slay the Spire''/``StS''; \emph{Names} median distinct real entity names per run. Themed also gives the first seed (floor) with recognition. The Permuted arm displays the title and real names (for the wrong entities), so there only undisplayed names and ``StS'' count. \emph{Memory, None}: runs whose newly written memories use real names when the arm shows them (a sanity check; the real-names arm has no local logs).}
\label{tab:reskin-recognition}
\begin{adjustbox}{max width=\linewidth}
\small
\begin{tabular}{@{}c@{\hspace{3pt}}l rrrr rrr rrr rrr r@{}}
\toprule
 & & \multicolumn{4}{c}{Themed} & \multicolumn{3}{c}{Random} & \multicolumn{3}{c}{Permuted} & \multicolumn{3}{c}{Perm.+themed} & Memory \\
\cmidrule(lr){3-6} \cmidrule(lr){7-9} \cmidrule(lr){10-12} \cmidrule(lr){13-15} \cmidrule(lr){16-16}
 & Agent & Rec. & Game & Names & 1st & Rec. & Game & Names & Rec. & Game & Names & Rec. & Game & Names & None \\
\midrule
\spireicon{qwen} & Qwen3.8 27B & 10/10 & 10/10 & 0 & S1 (F0) & 10/10 & 10/10 & 0 & 10/10 & 10/10 & 0 & 3/3 & 3/3 & 0 & -- \\
\spireicon{qwen} & Qwen3.6 35B-A3B & 1/10 & 1/10 & 0 & S9 (F3) & 0/10 & 0/10 & 0 & 0/10 & 0/10 & 0 & 1/10 & 1/10 & 0 & -- \\
\spireicon{qwen} & Qwen3.5 122B-A10B & 2/10 & 2/10 & 0 & S2 (F9) & 1/10 & 1/10 & 0 & 0/10 & 0/10 & 0 & 4/10 & 4/10 & 0 & 10/10 \\
\spireicon{claude} & Claude Fable 5.1 & 10/10 & 10/10 & 22 & S1 (F0) & -- & -- & -- & -- & -- & -- & -- & -- & -- & -- \\
\spireicon{claude} & Claude Opus 5 & 9/10 & 7/10 & 2 & S1 (F6) & -- & -- & -- & -- & -- & -- & -- & -- & -- & -- \\
\spireicon{claude} & Claude Sonnet 5 & 8/10 & 8/10 & 0 & S1 (F3) & -- & -- & -- & -- & -- & -- & -- & -- & -- & -- \\
\spireicon{claude} & Claude Haiku 4.5 & 1/10 & 1/10 & 0 & S1 (F0) & -- & -- & -- & -- & -- & -- & -- & -- & -- & 10/10 \\
\spireicon{deepseek} & DeepSeek V4.1 Flash & 10/10 & 10/10 & 1 & S1 (F6) & -- & -- & -- & -- & -- & -- & -- & -- & -- & 10/10 \\
\spireicon{gemini} & Gemini 3.8 Flash & 8/10 & 7/10 & 2 & S1 (F7) & -- & -- & -- & -- & -- & -- & -- & -- & -- & 2/2 \\
\spireicon{gemini} & Gemini 3.8 Flash (low) & 8/10 & 8/10 & 1 & S1 (F0) & 10/10 & 8/10 & 46 & 2/10 & 1/10 & 0 & 4/10 & 2/10 & 0 & -- \\
\spireicon{gemini} & Gemma 4 E4B & 0/10 & 0/10 & 0 & -- & -- & -- & -- & -- & -- & -- & -- & -- & -- & -- \\
\spireicon{meta} & Muse Spark 1.3 & 0/10 & 0/10 & 0 & -- & -- & -- & -- & -- & -- & -- & -- & -- & -- & -- \\
\spireicon{nvidia} & Nemotron 3 Nano Omni & 10/10 & 10/10 & 0 & S1 (F5) & 1/9 & 1/9 & 0 & 1/10 & 1/10 & 0 & 8/10 & 8/10 & 0 & -- \\
\spireicon{openai} & GPT-6 Astra & 5/10 & 2/10 & 0 & S1 (F18) & -- & -- & -- & -- & -- & -- & -- & -- & -- & -- \\
\spireicon{openai} & GPT-5.6 Sol & 7/10 & 5/10 & 0 & S1 (F0) & -- & -- & -- & -- & -- & -- & -- & -- & -- & -- \\
\spireicon{openai} & GPT-5.6 Terra & 6/10 & 5/10 & 0 & S3 (F12) & -- & -- & -- & -- & -- & -- & -- & -- & -- & -- \\
\spireicon{openai} & GPT-5.6 Luna & 5/10 & 5/10 & 0 & S1 (F9) & 2/10 & 2/10 & 0 & 2/10 & 2/10 & 0 & 6/10 & 6/10 & 0 & 9/10 \\
\spireicon{openai} & gpt-oss-120b & 10/10 & 10/10 & 0 & S1 (F1) & 10/10 & 10/10 & 0 & 6/10 & 5/10 & 0 & 10/10 & 10/10 & 1 & 10/10 \\
\spireicon{grok} & Grok 4.6 & 3/10 & 3/10 & 0 & S1 (F22) & -- & -- & -- & -- & -- & -- & -- & -- & -- & -- \\
\spireicon{zai} & GLM-5.3 & 10/10 & 10/10 & 2 & S1 (F2) & -- & -- & -- & -- & -- & -- & -- & -- & -- & 10/10 \\
\spireicon{zai} & GLM-5.3 Flash & 10/10 & 10/10 & 1 & S1 (F0) & 10/10 & 10/10 & 6 & 10/10 & 10/10 & 0 & 9/10 & 9/10 & 0 & 10/10 \\
\bottomrule
\end{tabular}
\end{adjustbox}
\end{table}

Since a model's own text is never
renamed (\cref{sec:methods-reskin}), we can measure when it recognizes what it is playing
(\cref{tab:reskin-recognition}). %
In 125 of the 200 themed runs of \cref{tab:main-results}, from 18 of
the 20 models, the
model names \emph{Slay the Spire} or a real entity that the reskin replaced, almost always in its reasoning and
usually on the first seed. This ranges from borrowing the real game's rules, as Nemotron~3
Nano Omni does in every run (``Usually in Slay the Spire, `Coiled' is a buff that gives guard when you
take damage''), to identifying entities from their behaviour, as Claude Fable~5.1 does with the Act~1
boss: ``This 250 HP figure actually matches Slay the Spire's Guardian at Ascension 9+ \ldots\
I'm recalling its exact moveset''. Only Muse Spark and Gemma~4 E4B never do. Recognition rarely reaches the
memory bank (15 of 200 themed runs, 10 of them Fable's), and models recognize the game on
their first seed, before any memory exists, about as often as later. Nor does it buy back what the reskin
costs: across the ten models with both arms, the share of runs that recognize the game is unrelated to
the cost (Spearman $\rho = 0.18$, permutation $p = 0.62$), and within a model, runs that recognize it in
their first six floors clear no more acts than those that do not ($+0.06$, $p = 0.85$, 170 runs). Knowing
which game this is does not recover knowledge that was keyed to its names. Changing the vocabulary barely
changes recognition. For the eight models run under every mapping, 56 of 80 themed runs recognize the
game, against 44 of 79 with nonsense names and 45 of 73 with the permuted themed ones. Gemini~3.8 Flash
at low effort goes further under nonsense names, decoding them back to the real ones in every run and
correctly about nine times in ten, and it scores its best result there. When real names are swapped
between entities, six of the eight models conclude they are playing a modded \emph{Slay the Spire} (GLM-5.3
Flash: ``The mod card-identity trap: I played \dag{}Defend thinking it gave channel [block] (base-StS
instinct)''), and their appeals to the real game mostly correct the borrowed meaning rather than act on
it; runs that make such appeals clear no fewer acts ($+0.05$, $p = 0.71$, 80 runs).

\subsection{Deck building does not fail by accumulation}\label{app:further-reward-policy}
\providecommand{\spireicon}[1]{\raisebox{-0.25ex}{\includegraphics[height=1.7ex]{icons/#1.pdf}}}
\begin{table}[t]
\centering
\caption{\textbf{Card-reward skip rate by act on the main-table cohorts (seeds 1--10, A20).} Acts cleared: mean
number of acts cleared per run. A decision is a
fight survived; it is a skip when no card joined the deck. $n$ is the number of decisions in the
act. Skip rates are percentages of the act's decisions. $\Delta$ is the model's Act~2+ skip rate minus its
Act~1 rate (pooled over its decisions), in percentage points,
shown where it has at least 10 Act~2+ decisions. Humans are Mega Crit telemetry, A20 Ironclad, where
a Singing Bowl pick also counts as a skip. The first- and random-option baselines never skip by
construction; Bottled AI skips every card not on its strategy's wish list.}
\label{tab:reward-policy}
\begin{adjustbox}{max width=\linewidth}
\small
\begin{tabular}{@{}c@{\hspace{3pt}}l r rr rr rr r@{}}
\toprule
 & & \multirow[b]{2}{*}{\makecell{Acts\\cleared}} & \multicolumn{2}{c}{Act 1} & \multicolumn{2}{c}{Act 2} & \multicolumn{2}{c}{Act 3} & \\
\cmidrule(lr){4-5} \cmidrule(lr){6-7} \cmidrule(lr){8-9}
 & Agent & & skip (\%) & $n$ & skip (\%) & $n$ & skip (\%) & $n$ & $\Delta$ (pp) \\
\midrule
\multicolumn{10}{l}{\textit{LLM agents, sorted by performance}} \\
\spireicon{claude} & Claude Fable 5.1 & 1.60 & 0 & 74 & 8 & 52 & 12 & 24 & +9 \\
\spireicon{openai} & GPT-6 Astra & 1.50 & 5 & 77 & 32 & 60 & 57 & 35 & +36 \\
\spireicon{claude} & Claude Opus 5 & 1.30 & 1 & 81 & 20 & 51 & 20 & 15 & +18 \\
\spireicon{openai} & GPT-5.6 Sol & 1.10 & 4 & 79 & 44 & 43 & 73 & 11 & +47 \\
\spireicon{meta} & Muse Spark 1.3 & 0.70 & 4 & 74 & 14 & 29 & 0 & 3 & +8 \\
\spireicon{zai} & GLM-5.3 & 0.60 & 3 & 71 & 16 & 25 & -- & 0 & +13 \\
\spireicon{gemini} & Gemini 3.8 Flash & 0.60 & 14 & 85 & 57 & 21 & 86 & 7 & +50 \\
\spireicon{claude} & Claude Sonnet 5 & 0.50 & 1 & 73 & 0 & 17 & 0 & 1 & $-$1 \\
\spireicon{openai} & GPT-5.6 Terra & 0.50 & 4 & 80 & 0 & 10 & -- & 0 & $-$4 \\
\spireicon{zai} & GLM-5.3 Flash & 0.40 & 2 & 63 & 25 & 12 & -- & 0 & +23 \\
\spireicon{grok} & Grok 4.6 & 0.30 & 59 & 61 & 100 & 11 & -- & 0 & +41 \\
\spireicon{deepseek} & DeepSeek V4.1 Flash & 0.30 & 0 & 72 & 0 & 6 & -- & 0 & -- \\
\spireicon{qwen} & Qwen3.8 27B & 0.20 & 0 & 64 & 25 & 8 & -- & 0 & -- \\
\spireicon{claude} & Claude Haiku 4.5 & 0.10 & 54 & 50 & 0 & 2 & -- & 0 & -- \\
\spireicon{openai} & GPT-5.6 Luna & 0.10 & 6 & 64 & 0 & 3 & -- & 0 & -- \\
\spireicon{openai} & gpt-oss-120b & 0.10 & 2 & 47 & 0 & 1 & -- & 0 & -- \\
\spireicon{nvidia} & Nemotron 3 Nano Omni & 0.00 & 0 & 33 & -- & 0 & -- & 0 & -- \\
\spireicon{gemini} & Gemma 4 E4B & 0.00 & 0 & 36 & -- & 0 & -- & 0 & -- \\
\spireicon{qwen} & Qwen3.6 35B-A3B & 0.00 & 27 & 52 & -- & 0 & -- & 0 & -- \\
\spireicon{qwen} & Qwen3.5 122B-A10B & 0.00 & 11 & 44 & -- & 0 & -- & 0 & -- \\
\midrule
 & All LLM runs (pooled) & 0.49 & 9 & 1,280 & 25 & 351 & 42 & 96 & +20 \\
 & 5 deepest LLMs (pooled) & -- & 3 & 385 & 24 & 235 & 39 & 88 & +25 \\
\midrule
\multicolumn{10}{l}{\textit{Human players (telemetry, own seeds)}} \\
\spireicon{crowd} & all A20 Ironclad & -- & 21 & 2,957,700 & 39 & 1,080,279 & 52 & 423,758 & +22 \\
\spireicon{crowd} & 1000+ hours & -- & 19 & 123,030 & 36 & 50,199 & 49 & 21,973 & +22 \\
\bottomrule
\end{tabular}
\end{adjustbox}
\end{table}

In Act~1 almost any card improves a starting
deck, but by Act~3 a marginal card dilutes the deck that has to beat the boss, so the right policy for
card rewards reverses during a run. The models do reverse it on average (\cref{tab:reward-policy}): their
skip rate rises from 0.09 in Act~1 to 0.25 and 0.42, as steeply as human players' (0.21,
0.39, and 0.52 over every Ascension~20 Ironclad run in the telemetry), and within runs as well as between
them (of the runs whose rate changes, 35 rise and 8 fall; exact sign test $p = 4 \times 10^{-5}$). But
41\% of the model runs that reach a second act never skip a card at all, against 10\% of human runs, and
the deepest models are the most reluctant to skip early (3\% in Act~1, against 19\% for players with over
1000 hours). Neither deck size at a matched floor, within a model, nor how much a model adapts its skip
rate predicts how many acts it clears (Spearman $\rho = 0.04$, $p = 0.91$, over the 11 models with enough
decisions after Act~1). Models do not lose by filling their decks; many never revise the policy they
start with, and among those that do, revising it more does not carry them further.

\subsection{Where models go wrong}\label{app:further-failures}
\providecommand{\spireicon}[1]{\raisebox{-0.25ex}{\includegraphics[height=1.7ex]{icons/#1.pdf}}}
\begin{table}[t]
\centering
\caption{\textbf{Tool-call errors per 100 model calls, for the LLM agents of \cref{tab:main-results}.} \emph{No usable call}: a response in which no tool call could be executed, because every call was malformed or refused or none could be parsed; the harness asks again. \emph{Refused call}: one call refused before it reached the game, as illegal on the current screen or with invalid arguments; the model is told why. \emph{Energy mismatch}: a turn ended with a different amount of energy left than the model declared when ending it, a sign that it lost track of its own turn. \emph{Recovered from prose}: a call the model wrote as text rather than through the tool interface, which the harness recovered and executed. \emph{Errors} sums the first three, which are the ones the model is told about. \emph{Batch reminders}: each time the model sends one action where it could have sent several, the harness reminds it to send all the actions of a turn in one response; not counted in \emph{Errors}. Rates are totals over the cohort divided by its calls. \best{Lowest} and \runnerup{second lowest} in each column, among agents with all 10 seeds.}
\label{tab:tool-errors}
\begin{adjustbox}{max width=\linewidth}
\small
\begin{tabular}{@{}c@{\hspace{3pt}}l r rrrr r r@{}}
\toprule
 & Agent & Calls/run & No usable call & Refused call & Energy mismatch & Recovered from prose & Errors & Batch reminders \\
\midrule
\spireicon{qwen} & Qwen3.8 27B {\scriptsize\textcolor{gray}{(xhigh)}} & 124 & 0.24 & \best{0.00} & 0.73 & \best{0.00} & 0.97 & 1.5 \\
\spireicon{qwen} & Qwen3.6 35B-A3B & 104 & 0.48 & 0.38 & 0.58 & 0.38 & 1.44 & 21.1 \\
\spireicon{qwen} & Qwen3.5 122B-A10B {\scriptsize\textcolor{gray}{(xhigh)}} & 77 & 0.39 & 0.39 & 1.29 & 0.13 & 2.07 & 0.4 \\
\spireicon{claude} & Claude Fable 5.1 {\scriptsize\textcolor{gray}{(max)}} & 268 & \best{0.00} & 0.15 & 0.22 & \best{0.00} & 0.37 & 3.1 \\
\spireicon{claude} & Claude Opus 5 {\scriptsize\textcolor{gray}{(max)}} & 301 & 0.03 & 0.10 & \best{0.00} & \best{0.00} & 0.13 & 3.1 \\
\spireicon{claude} & Claude Sonnet 5 {\scriptsize\textcolor{gray}{(max)}} & 159 & 0.06 & \runnerup{0.06} & 0.06 & \best{0.00} & 0.19 & 3.1 \\
\spireicon{claude} & Claude Haiku 4.5 {\scriptsize\textcolor{gray}{(12k)}} & 101 & \best{0.00} & \best{0.00} & 1.09 & \best{0.00} & 1.09 & \best{0.1} \\
\spireicon{deepseek} & DeepSeek V4.1 Flash {\scriptsize\textcolor{gray}{(max)}} & 119 & 0.08 & 0.08 & \best{0.00} & \best{0.00} & 0.17 & 3.1 \\
\spireicon{gemini} & Gemini 3.8 Flash {\scriptsize\textcolor{gray}{(high)}} & 208 & 0.05 & \best{0.00} & \best{0.00} & \best{0.00} & \best{0.05} & 4.4 \\
\spireicon{gemini} & Gemma 4 E4B & 82 & 0.85 & 1.83 & 1.10 & \runnerup{0.12} & 3.78 & 10.5 \\
\spireicon{meta} & Muse Spark 1.3 {\scriptsize\textcolor{gray}{(xhigh)}} & 228 & 0.04 & \best{0.00} & 0.04 & \best{0.00} & \runnerup{0.09} & 18.7 \\
\spireicon{nvidia} & Nemotron 3 Nano Omni & 71 & 0.28 & 0.42 & 0.99 & 4.38 & 1.69 & \runnerup{0.3} \\
\spireicon{openai} & GPT-6 Astra {\scriptsize\textcolor{gray}{(max)}} & 546 & \runnerup{0.02} & 0.22 & \runnerup{0.02} & \best{0.00} & 0.26 & 37.4 \\
\spireicon{openai} & GPT-5.6 Sol {\scriptsize\textcolor{gray}{(max)}} & 331 & 0.06 & 0.15 & \best{0.00} & \best{0.00} & 0.21 & 29.7 \\
\spireicon{openai} & GPT-5.6 Terra {\scriptsize\textcolor{gray}{(max)}} & 230 & \best{0.00} & 0.13 & \best{0.00} & \best{0.00} & 0.13 & 41.1 \\
\spireicon{openai} & GPT-5.6 Luna {\scriptsize\textcolor{gray}{(max)}} & 168 & \best{0.00} & \runnerup{0.06} & 0.06 & \best{0.00} & 0.12 & 27.2 \\
\spireicon{openai} & gpt-oss-120b {\scriptsize\textcolor{gray}{(high)}} & 175 & 0.23 & \best{0.00} & 1.77 & 8.64 & 2.00 & 34.2 \\
\spireicon{grok} & Grok 4.6 {\scriptsize\textcolor{gray}{(xhigh)}} & 171 & 1.29 & \best{0.00} & 0.06 & \best{0.00} & 1.34 & 3.3 \\
\spireicon{zai} & GLM-5.3 {\scriptsize\textcolor{gray}{(max)}} & 160 & \best{0.00} & \best{0.00} & 0.25 & \best{0.00} & 0.25 & 2.3 \\
\spireicon{zai} & GLM-5.3 Flash {\scriptsize\textcolor{gray}{(max)}} & 107 & 0.09 & \best{0.00} & 0.19 & \best{0.00} & 0.28 & 2.2 \\
\bottomrule
\end{tabular}
\end{adjustbox}
\end{table}

\begin{table}[t]
\centering\small
\caption{\textbf{Excerpts from the runs of \cref{tab:main-results}, verbatim.} Names are as the model saw them under the
reskin, with the real game term after it in grey brackets. Excerpts are taken from the full run logs and cut at the
ellipsis.}
\label{tab:trace-excerpts}
\begin{adjustbox}{max width=\textwidth}
\small
\begin{tabular}{@{}>{\raggedright\arraybackslash}p{2.5cm}>{\raggedright\arraybackslash}p{2.4cm}>{\raggedright\arraybackslash}p{8.6cm}@{}}
\toprule
Failure & Model (seed, floor) & Excerpt \\
\midrule
Deliberating over a routine forecast & Nemotron 3 Nano Omni (S3, 6) & ``We need to estimate probabilities. The player is at camp after sleeping, HP 24/75 (32\%). Actually after sleeping they healed 22 HP, so HP should be 24+22 = 46? Wait initial HP 24/75, sleeping heals 22 \ldots'' \\
\addlinespace
Reasoning and action disagree & Grok 4.6 (S9, 10) & ``\ldots\ I think Blood Temper \textcolor{gray}{[Armaments]} is the better pick here\ldots\ I'll go with Blood Temper.'' Then skips the reward, writing beside the call: ``Skip Pulse and Smash as filler\ldots'' \\
\addlinespace
Miscounted turn & GLM-5.3 (S5, 13) & ``I miscounted energy! \ldots\ My error: I planned Trip(2) + Slash(1) + Slash(1) = 4E but only have 3E! Oops.'' \\
Blow-by-blow note & Claude Fable 5.1 (S10, 18) & ``CRIMSON WARLOCK \textcolor{gray}{[Chosen]} solo 100 HP (Act 2 R18 opener), run 10 script: T1 Strong Debuff = Jinx \textcolor{gray}{[Hex]} \ldots\ $\to$ T2 Debuff = Fatigue \textcolor{gray}{[Weak]} 3 on ME + Str 3 to itself \ldots'' (280 words) \\
\addlinespace
Unsure of a stated rule & Nemotron 3 Nano Omni (S3, 2) & ``\ldots\ player is Exposed \textcolor{gray}{[Vulnerable]}, taking 50\% more damage, so effective damage = 7 * 1.5 = 10.5, rounded down? Probably 10 or 11.'' \\
\addlinespace
Misattributed act heal & GPT-5.6 Terra (S5, end) & ``Wanderer's Pack \textcolor{gray}{[Tiny House]} is an excellent low-HP boss reward: \ldots\ +5 Max HP/current HP \ldots\ In this run it converted 16/75 to 65/80 before the next Act-2 fight \ldots'' \\
\addlinespace
Correct memory overwritten & DeepSeek V4.1 Flash (S8, end) & ``KEY NEW LESSON (this run): \ldots\ the post-boss heal is small --- I entered the act at 11/75'' (its log that run: +44 HP, 16$\to$60) \\
\addlinespace
Rule from one run & Claude Haiku 4.5 (S9, end) & ``Arriving at any combat-potential room (R13+) with $<$35\% HP on thin deck without full Guard \textcolor{gray}{[Block]} infrastructure is terminal. This run: R13 reached at 27/75 HP (36\%) \ldots'' \\
\bottomrule
\end{tabular}
\end{adjustbox}
\end{table}

Reading the runs of \cref{tab:main-results} shows recurring failures
(excerpts in \cref{tab:trace-excerpts}). Weaker models spend reasoning on routine calls: the end-of-floor
forecast asks for two probabilities, yet Claude Haiku~4.5, gpt-oss-120b, and Grok~4.6 spend more than
1{,}000 reasoning tokens on 58--83\% of theirs, against at most 9\% for the strongest models (Spearman
$\rho = -0.49$ between the median spent and mean acts cleared). Grok~4.6's reasoning often
settles on one action while its tool call takes another: in 20 of the 68 decisions where its reasoning
ends by naming a card or an option (29\%), the call differs, seven times skipping a card reward it had
just said it would take, and the prose it writes beside the call agrees with the call rather than the
reasoning. gpt-oss-120b does so in 2 of 31 such decisions, and none of the other 17 models in 1{,}405.
Most models sometimes lose track of their own turn: 15 of the 20 end a turn with a different amount of
energy from the one they declared (\cref{tab:tool-errors}). Within a run it is the strongest models that
write too much in their notes: Claude Opus~5 writes 143 notes per run at a median of 192 words, and 23--42\% of
the notes of Gemini~3.8 Flash, GPT-6 Astra, Claude Fable~5.1, Muse Spark, and GLM-5.3 are turn-by-turn
accounts of a single fight, while the weakest write a few short, generic notes.
A rule the prompt states is not always taken in. The prompt gives Vulnerable's 50\% and says
that ``percentage-based calculations always round down at the end'', yet gpt-oss-120b and Nemotron~3 Nano Omni work the rounding out afresh in all
ten of their runs, Qwen3.6 35B-A3B and Claude Haiku~4.5 in nine, and none of them ever writes it down, while
eleven models keep it in their memory bank. What the harness leaves unstated is learnt unevenly: of the 16
models that clear Act~1, nine end with the fact
that the act transition restores most lost HP in their bank; Muse Spark and Grok~4.6 never record it and
weigh their low HP when choosing the boss relic, GPT-5.6 Terra credits the heal to the relic it chose,
and DeepSeek~V4.1 Flash records it and later overwrites it with the claim that the heal is small.
End-of-run consolidation accumulates rather than curates: over the 200 runs the models create 1{,}404
memories (two of them left by Gemma~4 E4B's crashed first attempt), rewrite 1{,}080, re-save another 863 unchanged, and delete 30, and 15 of the 20 never delete one;
what the banks end up holding is classified in \cref{tab:memory-classification}. The miscounted turns
echo what is known of long executions: a slip early in a turn is carried into the rest of it, as errors
in context make further errors more likely~\citep{sinha2026illusion}. The notes and memories are the
memory failure of \cref{app:further-memory-banks}, seen from inside a run: nothing tells the model which of its observations were
worth keeping, so it keeps the wrong ones~\citep{xiong2025memory}.

\FloatBarrier
\section[\texorpdfstring{Reading the reasoning}{Reading the reasoning}]{Reading the reasoning}
\label{app:traces}
The analyses above measure what the models did. Here we read what they wrote while doing it: their
reasoning, tool calls, notes, and memories. We ask whether the stronger models reason from more correct
facts, why the models rest where humans upgrade (\cref{fig:hp-resource}), whether they use and correct
what they carry between runs, and how they explain a loss.

\subsection{Method}
\label{app:traces-method}

We had an LLM read the runs of 16 of the models of \cref{tab:main-results}, all ten seeds of each (160
runs), and tag every passage that fits one of a fixed set of categories: for example, stating a rule
wrongly, citing a memory as the reason for a decision, or planning several actions and sending only one.
The annotator was Claude Opus~5.5, which is not one of the models played.
Whereas models played \emph{Slay the Spire} under a reskin, the annotator was aware of the reskin and some properties of the codebase, and thus could better generalize from its knowledge of \emph{Slay the Spire} to know the facts of the game being played.
Each instance read either a
stretch of play or one end-of-run consolidation, together with the
system prompt, the reskin mapping, and the memories the run started with. For every tag it quoted the
model verbatim and gave the line in the log.
A statement is marked wrong only if the annotator cites the line of the log or of the system prompt that
contradicts it.
99.6\% of the 53{,}456 quotes are where the annotator said.
Tags in the forecasts, reflections, and consolidations the harness requires are kept apart
from those in the model's own decisions.

An author ruled afterwards on five facts the annotators lacked or judged from their own knowledge of the game:
how much a rest heals, how much the heal between acts restores, that an unknown room never holds an elite,
when the harness drops the rest of a batch of actions, and that three keys are needed to reach the final
boss. The 533 tags that turn on these were re-judged.

\providecommand{\spireicon}[1]{\raisebox{-0.25ex}{\includegraphics[height=1.7ex]{icons/#1.pdf}}}
\begin{table}[t]
\centering
\caption{%
\textbf{What the models state about the game.} From the reasoning of ten seeds per model, as tagged by the annotators (\cref{app:traces-method}). \emph{Rules}: statements of a rule or of what an upgrade does that the annotator could judge against the log, and the share judged wrong; \emph{told} restricts this to facts the system prompt states. \emph{Shown numbers}: statements, right/wrong, of whether the number on a card in hand, or on an enemy's intent, already includes modifiers. \emph{Rest sites}: all rest sites in the runs, how many the model rested at, and at how many its reasoning weighed upgrading, mentioned the heal at the end of the act, or projected its HP into a named coming fight.
}
\label{tab:trace-facts}
\begin{adjustbox}{max width=\linewidth}
\small
\begin{tabular}{@{}c@{\hspace{3pt}}l rrr cc rrrrr@{}}
\toprule
 & & \multicolumn{3}{c}{Rules} & \multicolumn{2}{c}{Shown numbers} & \multicolumn{5}{c}{Rest sites} \\
\cmidrule(lr){3-5} \cmidrule(lr){6-7} \cmidrule(lr){8-12}
 & Agent & Judged & Wrong & Wrong, told & Cards & Intents & \# & Rested & Upgrade & Act heal & HP \\
\midrule
\spireicon{qwen} & Qwen3.8 27B & 1279 & 28\% & 23\% & 37/44 & 17/6 & 27 & 18 & 27 & 1 & 18 \\
\spireicon{qwen} & Qwen3.6 35B-A3B & 479 & 66\% & 62\% & 8/22 & 5/19 & 15 & 9 & 11 & 0 & 1 \\
\spireicon{qwen} & Qwen3.5 122B-A10B & 456 & 57\% & 49\% & 14/44 & 13/22 & 14 & 10 & 13 & 0 & 6 \\
\spireicon{claude} & Claude Fable 5.1 & 1566 & 7\% & 4\% & 32/7 & 41/0 & 70 & 44 & 56 & 19 & 47 \\
\spireicon{claude} & Claude Opus 5 & 2189 & 11\% & 8\% & 101/0 & 76/8 & 58 & 44 & 48 & 17 & 45 \\
\spireicon{claude} & Claude Sonnet 5 & 928 & 12\% & 7\% & 49/4 & 61/6 & 33 & 30 & 28 & 0 & 18 \\
\spireicon{claude} & Claude Haiku 4.5 & 734 & 58\% & 57\% & 11/66 & 12/40 & 16 & 11 & 16 & 0 & 6 \\
\spireicon{deepseek} & DeepSeek V4.1 Flash & 1020 & 14\% & 11\% & 34/25 & 15/18 & 27 & 23 & 25 & 0 & 22 \\
\spireicon{gemini} & Gemini 3.8 Flash & 910 & 19\% & 14\% & 20/2 & 15/8 & 42 & 26 & 30 & 4 & 18 \\
\spireicon{openai} & GPT-6 Astra & 274 & 6\% & 10\% & 1/0 & 2/0 & 63 & 34 & 30 & 0 & 5 \\
\spireicon{openai} & GPT-5.6 Sol & 370 & 6\% & 7\% & 7/0 & 4/1 & 47 & 35 & 13 & 0 & 5 \\
\spireicon{openai} & GPT-5.6 Terra & 298 & 10\% & 8\% & 7/0 & 2/5 & 31 & 26 & 15 & 0 & 4 \\
\spireicon{openai} & GPT-5.6 Luna & 263 & 21\% & 18\% & 6/1 & 2/6 & 20 & 17 & 9 & 0 & 3 \\
\spireicon{openai} & gpt-oss-120b & 961 & 43\% & 41\% & 29/13 & 26/42 & 12 & 6 & 12 & 0 & 3 \\
\spireicon{zai} & GLM-5.3 & 1513 & 18\% & 10\% & 115/8 & 81/4 & 38 & 29 & 33 & 6 & 32 \\
\spireicon{zai} & GLM-5.3 Flash & 1100 & 23\% & 14\% & 63/16 & 21/12 & 23 & 14 & 23 & 0 & 14 \\
\bottomrule
\end{tabular}
\end{adjustbox}
\end{table}

\subsection{What the models believe about the game}
\label{app:traces-facts}

The stronger models reason from more correct facts (\cref{tab:trace-facts}). Of the statements about a
rule or an upgrade that could be checked against the log, 22\% are wrong, from 6--7\% for Astra, Sol, and
Claude Fable~5.1 to 57--66\% for Claude Haiku~4.5, Qwen3.5, and Qwen3.6. Missing information does not
explain this: on rules the system prompt states outright, Qwen3.6 is wrong 62\% of the time and Haiku~4.5
57\%, against 4\% for Fable~5.1. Within a family the stronger model is usually the more accurate.

The models rarely misstate what an upgrade does (6\% of such statements are wrong), which bears on the
choice between resting and upgrading (\cref{fig:hp-resource}). The weaker models fail instead on the
numbers the game displays. The value shown on a card or an enemy's intent already
includes the modifiers in force, which the harness does not say, and Qwen3.8 27B, Qwen3.5, and Haiku~4.5
(for cards) and gpt-oss-120b (for intents) apply a modifier a second time more often than not. Holding a
card shown as 9 Block, already reduced by Frail, Qwen3.8 27B writes ``Unwanton Ward [Ghostly Armor]: 9
Guard [Block] $\to$ with Brittle [Frail] 4: 6.75 $\to$ 6?'', plans its turn on 6, and gains 9. Claude
Opus~5 gets this right in all 101 of its cases.

Knowledge of the game from outside the run is judged correct in half of the 797 decisions that use it. In
86 the prompt already held the answer, and 32 of these are gpt-oss-120b, which, with the act map in front
of it, estimates what is left of the act from ``generic slay stats''. Recognising the game
(\cref{app:further-reskin-recognition}) supplies rules that the weaker models do not check against what
they are shown.

\begin{table}[t]
\centering
\caption{\textbf{Recurring misconceptions in the models' reasoning.} Statements judged wrong (\cref{app:traces-facts}), grouped by keyword rules on the statement and the annotator's note, so the counts are approximate. \emph{Calls}: LLM calls with such a statement; \emph{Models}: of 16; \emph{Per 1000 calls}: the rate in the top four models by acts cleared, the six annotated models of the next eight and the six of the bottom eight (\cref{tab:trace-tiers}). Examples are verbatim, with the real name in brackets after a renamed one.}
\label{tab:trace-misconceptions}
\begin{adjustbox}{max width=\linewidth}
\small
\begin{tabular}{p{4.0cm} rr rrr p{3.2cm} p{5.8cm}}
\toprule
 & & & \multicolumn{3}{c}{Per 1000 calls} & & \\
\cmidrule(lr){4-6}
Misconception & Calls & Models & Top & Mid & Bot & Most often & Example \\
\midrule
Applies a modifier that the number shown on a card or an enemy's intent already includes & 360 & 15 & 1.0 & 9.3 & 33.9 & Claude Haiku 4.5 (73); Qwen3.5 122B-A10B (58) & ``Exposed [Vulnerable] 1 on player means damage is increased by 50\%, so likely 36 damage incoming (24 * 1.5)'' (gpt-oss-120b) \\
\addlinespace
Plans to remove Ascender's Bane, whose text says it cannot be removed & 244 & 15 & 0.3 & 2.7 & 28.3 & Qwen3.6 35B-A3B (113); Qwen3.5 122B-A10B (48) & ``So my plan: Remove Sins of the Past [Ascender's Bane] for 75g. This is critical because Sins blocks the Bloodlust [Clash] strategy'' (Qwen3.6 35B-A3B) \\
\addlinespace
Misjudges how much a rest or the end of an act heals & 240 & 15 & 1.6 & 13.5 & 11.2 & GLM-5.3 (68); GLM-5.3 Flash (42) & ``Since the boss heals to full anyway, HP only matters for survival'' (Claude Fable 5.1) \\
\addlinespace
Relies on a card, potion, or effect it does not have & 168 & 16 & 0.9 & 4.5 & 14.8 & gpt-oss-120b (36); Claude Haiku 4.5 (23) & ``wait no, Purge [Burning Pact] is in MY HAND right now'' (GLM-5.3) \\
\addlinespace
Expects Block to last into a later turn & 147 & 13 & 0.3 & 4.7 & 12.8 & gpt-oss-120b (48); Claude Haiku 4.5 (21) & ``Guard [Block] does NOT wear off, which is important'' (Gemini 3.8 Flash) \\
\addlinespace
Mispredicts an enemy's moves or splitting & 147 & 15 & 1.6 & 5.8 & 8.9 & Qwen3.8 27B (22); Qwen3.6 35B-A3B (20) & ``Corrosive Ooze [Acid Slime (M)] at 32 HP (need to be careful not to drop below 16 HP or it splits)'' (Qwen3.6 35B-A3B) \\
\addlinespace
Wrong price or limit for removing a card at a shop & 111 & 11 & 0.6 & 0.8 & 12.7 & Qwen3.5 122B-A10B (42); Qwen3.6 35B-A3B (29) & ``The first removal costs 50g, second costs 75g, totaling 125g'' (Qwen3.5 122B-A10B) \\
\addlinespace
Reads stacks of Weak, Frail, or Vulnerable as strength rather than turns & 88 & 14 & 0.3 & 1.9 & 8.7 & Qwen3.6 35B-A3B (28); Claude Haiku 4.5 (13) & ``So 2 Fatigue [Weak] stacks = 75\% damage reduction'' (Qwen3.6 35B-A3B) \\
\bottomrule
\end{tabular}
\end{adjustbox}
\end{table}

The wrong beliefs fall into a few recurring kinds (\cref{tab:trace-misconceptions}). The most common is applying
a modifier a second time to a number the game already shows with it (360 calls, 15 models), most often by
Haiku~4.5, Qwen3.5, and gpt-oss-120b. Next come plans to remove Ascender's Bane, a curse whose card
text says it cannot be removed (244 calls, 113 of them by Qwen3.6, which keeps the plan in its memory from seed to
seed), and misjudging how much a rest or the end of an act heals (240). Others are rules of the turn: that Block
lasts into a later turn (147 calls, a third of them by gpt-oss-120b) and that stacks of Weak, Frail, or Vulnerable
add strength rather than turns (88). In 168 calls a model relies on a card, potion, or effect it does not have.
Two more concern what the harness shows only in part: an enemy's moves (147 calls), of which it
shows only the next intent, such as expecting a medium slime to split as the large ones do, and the price of
removing a card (111), which it shows only at a shop. Several models take that price to start at 50 gold, some with
the shop's price of 75 in front of them.
The prompt does not say either that an upgrade made at a rest site lasts the rest of the run, but the models
almost never doubt it: across the 160 annotated runs, gpt-oss-120b twice takes such an upgrade to last only one
fight (once at a rest site, where it rests anyway), and Haiku~4.5 once stops to check, concluding correctly that
it lasts. The recurring error is the reverse: taking an upgrade made during a fight, which the card text says lasts
``for the rest of combat'', to be permanent. Qwen3.5 counts its Trip Attack [Bash] as upgraded in later fights
although its deck lists it plain, Gemini~3.8 Flash takes Enchanter's Hammer [Warped Tongs] to
upgrade cards for good, and even Claude Fable~5.1 once calls Blood Temper [Armaments] a permanent upgrade. Qwen3.6
carries ``Blood Temper enhances ALL cards permanently'' in its memory bank from seed to seed, and keeps the line
after seeing the card upgrade a single card.

\begin{table}[t]
\centering
\caption{\textbf{Which rule errors persist in the strongest models (\cref{app:traces-facts}).} For each kind of rule statement, the share judged wrong, and in brackets the statements judged, for the top four models of \cref{tab:main-results} by acts cleared, the six annotated models of the next eight, and the six of the bottom eight. Rows are ordered by the ratio of the bottom tier's share to the top tier's; the last row is from the tracked mechanics. GPT-6 Astra and GPT-5.6 Sol log little reasoning, so the top tier's counts are mostly Claude Fable~5.1's and Claude Opus~5's.}
\label{tab:trace-tiers}
\begin{adjustbox}{max width=\linewidth}
\small
\begin{tabular}{l rrr}
\toprule
Statements of & Top 4 & Middle & Bottom \\
\midrule
What an enemy will do & 15\% (371) & 21\% (494) & 34\% (343) \\
What a relic, potion, power, or card does & 10\% (582) & 18\% (701) & 51\% (614) \\
Damage arithmetic & 7\% (380) & 18\% (532) & 54\% (452) \\
Shop, removal, and reward rules & 4\% (215) & 8\% (289) & 48\% (289) \\
Turn mechanics (Block, energy, hand) & 4\% (224) & 12\% (408) & 41\% (346) \\
How much healing restores & 3\% (564) & 25\% (475) & 37\% (226) \\
When a debuff ticks or expires & 3\% (302) & 9\% (448) & 36\% (254) \\
\midrule
Rounding after a multiplier & 6\% (286) & 4\% (557) & 4\% (344) \\
\bottomrule
\end{tabular}
\end{adjustbox}
\end{table}

Most of these are errors of the weaker models (\cref{tab:trace-tiers}). Applying a modifier twice occurs in 3.4\%
of calls in the bottom tier, 0.9\% in the middle and 0.1\% in the top four, and the rules of the turn, of the
shop, and of debuff timing are misstated in 3--4\% of the top four's statements against 36--48\% of
the bottom tier's. Two kinds of error persist in the strongest models, and make up more than half
of their wrong statements: what an enemy will do, wrong in 15\% of the top four's statements
against 34\% of the bottom tier's, and what a relic, potion, power, or card does (10\% against
51\%). The game shows only an enemy's next intent, so its pattern has to be learnt: Claude Opus~5
expects Giant Head (``the turtle'') to keep ``only dealing about 17 per turn'', and its next attack is 66. Item
effects are mostly stated in the prompt, yet Opus~5 decides that ``Healing Locket [Blood Vial] gives +2 per ROOM
entered'' and keeps the rule in its notes into the next seed. Rounding after a multiplier is the one tracked rule on
which the top tier is no better (6\% of its statements wrong, against 4\%), although the prompt states it.
What separates the tiers, then, is the general rules, which come up again and again and most of which the prompt
states; what the strongest models still get wrong is
knowledge a player builds up over many runs, the behaviour of each enemy and the effect of each of hundreds of
items. That is what the memory bank is for, and it can hold the error instead: Opus~5's belief about
Giant Head came from its own notes, which record that it debuffs for two turns and hits for little.
Rule knowledge and long-horizon judgement can thus be told apart. A poor decision can come from misreading the
turn in front of the model (where the intents shown determine the next HP exactly, the bottom tier mispredicts it
31\% of the time and the top four 0.5\%; \cref{sec:results}), from a wrong belief about the game (this section), or
from misjudging the decision with the facts right. The weaker models already fail at the first two. The strongest
rarely do, except on knowledge built up over many runs, so their gap to expert players lies there and in
judgement: Fable~5.1 and Opus~5 weigh the end-of-act heal at late-act rest sites, and still rest in most of those
decisions (\cref{app:traces-rest}).

\subsection{Why the models rest}
\label{app:traces-rest}

The models rest at 70\% of rest sites and upgrade at 23\% (\cref{tab:trace-facts}). They usually consider
upgrading and value upgrades correctly. Many also project their HP into a named coming fight at most of their
rest sites: Fable~5.1 at 47 of 70, Opus~5 at 45 of 58, GLM-5.3 at 32 of 38, and DeepSeek~V4.1 Flash at 22 of
27. The four GPT-5.6 and GPT-6 models are seen to do so at only 3--5 rest sites each, but they log little
reasoning. In the annotated reasoning only five models ever take into account the heal at the end of an act,
which makes resting before a boss worth less. Opus~5, choosing to upgrade, writes: ``I recall the
post-boss treasure room heals roughly 75\% of missing HP''. Sol, at a third of its HP: ``Since survival is
a top priority at 33\%, sleeping is essential''.

The caution does not come from the prompt's tips, which a model gives as its reason only
three times. When a model declines a risk with a long-term payoff (an elite, a key, a gamble at an event),
it most often cites its current HP or a lesson in its memory. Declining is often right for a model that
fights badly (\cref{app:play-risk}); what the reasoning adds is that projecting HP ahead does not change the
choice: the models that do it most still rest at most of their rest sites (Fable~5.1 at 44 of 70, Opus~5 at 44 of
58, GLM-5.3 at 29 of 38).

\providecommand{\spireicon}[1]{\raisebox{-0.25ex}{\includegraphics[height=1.7ex]{icons/#1.pdf}}}
\begin{table}[t]
\centering
\caption{%
\textbf{Notes, memories, and their use.} Over ten seeds per model (\cref{app:traces-method}). \emph{Run notes}: notes held at the end of the runs, and the share of which nothing reaches a memory. \emph{Memories saved}: by kind (facts about enemies, evaluations of cards and relics, rules of the game, rules of strategy; the few of other kinds are in the total), and those holding a claim the run's own record contradicts. \emph{In play}: calls that cite a memory from an earlier seed as the reason for a decision, calls that act against one, and mechanics the model discovers that it also writes into a note. \emph{Missed}: runs whose review never mentions a large problem visible in the room history.
}
\label{tab:trace-memory}
\begin{adjustbox}{max width=\linewidth}
\small
\begin{tabular}{@{}c@{\hspace{3pt}}l rr rrrrrr rrc c@{}}
\toprule
 & & \multicolumn{2}{c}{Run notes} & \multicolumn{6}{c}{Memories saved} & \multicolumn{3}{c}{In play} & \\
\cmidrule(lr){3-4} \cmidrule(lr){5-10} \cmidrule(lr){11-13}
 & Agent & Notes & Dropped & All & Enemy & Card & Game & Strategy & False & Cites & Against & Noted & Missed \\
\midrule
\spireicon{qwen} & Qwen3.8 27B & 252 & 42\% & 134 & 36 & 31 & 23 & 39 & 25 & 224 & 33 & 59/84 & 6/10 \\
\spireicon{qwen} & Qwen3.6 35B-A3B & 135 & 50\% & 58 & 13 & 6 & 6 & 30 & 35 & 20 & 11 & 12/29 & 9/10 \\
\spireicon{qwen} & Qwen3.5 122B-A10B & 165 & 69\% & 63 & 8 & 16 & 3 & 33 & 25 & 51 & 15 & 4/17 & 9/10 \\
\spireicon{claude} & Claude Fable 5.1 & 468 & 3\% & 208 & 55 & 36 & 40 & 58 & 20 & 278 & 28 & 172/172 & 1/10 \\
\spireicon{claude} & Claude Opus 5 & 561 & 19\% & 311 & 59 & 24 & 35 & 161 & 26 & 550 & 59 & 154/156 & 3/10 \\
\spireicon{claude} & Claude Sonnet 5 & 340 & 57\% & 162 & 34 & 43 & 26 & 56 & 11 & 210 & 16 & 100/108 & 5/10 \\
\spireicon{claude} & Claude Haiku 4.5 & 332 & 64\% & 80 & 7 & 5 & 4 & 53 & 40 & 66 & 17 & 11/35 & 8/10 \\
\spireicon{deepseek} & DeepSeek V4.1 Flash & 261 & 79\% & 50 & 16 & 11 & 9 & 14 & 8 & 195 & 10 & 67/75 & 7/10 \\
\spireicon{gemini} & Gemini 3.8 Flash & 191 & 44\% & 114 & 44 & 44 & 8 & 18 & 12 & 146 & 5 & 46/57 & 8/10 \\
\spireicon{openai} & GPT-6 Astra & 239 & 5\% & 309 & 165 & 81 & 31 & 30 & 1 & 30 & 7 & 81/82 & 2/10 \\
\spireicon{openai} & GPT-5.6 Sol & 183 & 13\% & 230 & 99 & 64 & 32 & 35 & 4 & 67 & 9 & 73/75 & 1/10 \\
\spireicon{openai} & GPT-5.6 Terra & 127 & 84\% & 87 & 18 & 47 & 0 & 22 & 2 & 34 & 9 & 23/26 & 5/10 \\
\spireicon{openai} & GPT-5.6 Luna & 153 & 82\% & 56 & 12 & 26 & 2 & 15 & 9 & 39 & 9 & 18/20 & 5/10 \\
\spireicon{openai} & gpt-oss-120b & 389 & 75\% & 81 & 5 & 24 & 13 & 39 & 14 & 25 & 38 & 19/40 & 10/10 \\
\spireicon{zai} & GLM-5.3 & 370 & 51\% & 101 & 35 & 15 & 22 & 25 & 13 & 293 & 22 & 127/136 & 3/10 \\
\spireicon{zai} & GLM-5.3 Flash & 264 & 33\% & 129 & 43 & 30 & 22 & 28 & 25 & 267 & 17 & 106/111 & 2/10 \\
\bottomrule
\end{tabular}
\end{adjustbox}
\end{table}

\subsection{Memories in play}
\label{app:traces-memory}

\begin{table}[t]
\centering
\caption{%
\textbf{Rule errors and memory use by seed.} Per hundred LLM calls, over the 16 models read on all ten seeds (\cref{app:traces-memory}). \emph{Wrong}: statements of a rule or of an upgrade's effect judged wrong. \emph{Cites}: a memory from an earlier seed cited as the reason for a decision. \emph{Against}: an applicable lesson in the model's memories or notes left unused or contradicted. \emph{Doubts}: a note or memory doubted, overridden, or corrected. \emph{Discovers}: a mechanic found out during play. $\tau$: Kendall's $\tau$ of each model's rate against seed order, averaged over the models; \emph{Models rising}: how many have $\tau>0$; $p$: two-sided, for the average, when each model's seed order is shuffled independently (20{,}000 permutations). Seed 1 is left out of the trends because the memory bank is empty then. Each seed is a different game, so a column mixes learning with the difficulty of the seed.
}
\label{tab:trace-seeds}
\begin{adjustbox}{max width=\linewidth}
\small
\begin{tabular}{l r rrrrr}
\toprule
Seed & Calls & Wrong & Cites & Against & Doubts & Discovers \\
\midrule
1 & 3{,}313 & 9.2 & 0.0 & 0.4 & 1.2 & 5.4 \\
2 & 3{,}495 & 13.6 & 7.0 & 0.7 & 4.0 & 4.5 \\
3 & 3{,}333 & 9.4 & 7.7 & 1.1 & 2.9 & 4.4 \\
4 & 1{,}419 & 12.6 & 8.4 & 1.1 & 4.4 & 3.2 \\
5 & 3{,}924 & 10.1 & 7.6 & 0.9 & 4.0 & 3.9 \\
6 & 2{,}471 & 10.2 & 11.3 & 1.1 & 5.0 & 4.0 \\
7 & 2{,}508 & 9.9 & 9.7 & 1.0 & 3.8 & 3.4 \\
8 & 4{,}243 & 9.5 & 8.6 & 0.7 & 4.2 & 3.9 \\
9 & 4{,}112 & 8.7 & 8.8 & 1.1 & 4.0 & 3.8 \\
10 & 2{,}967 & 8.7 & 11.1 & 1.7 & 5.2 & 3.8 \\
\midrule
$\tau$, seeds 2--10 &  & $-$0.07 & +0.31 & +0.19 & +0.13 & $-$0.04 \\
Models rising &  & 5/16 & 15/16 & 11/16 & 11/16 & 7/16 \\
$p$ &  & 0.273 & $<$0.001 & 0.004 & 0.046 & 0.547 \\
\bottomrule
\end{tabular}
\end{adjustbox}
\end{table}

The models use their memory banks, and more as the banks grow
(\cref{tab:trace-memory,tab:trace-seeds}). A memory cited as the reason for a decision appears in 7\% of
calls on seed 2 and 11\% on seed 10, and the rate rises across seeds in 15 of the 16 models
($p<0.001$). The models differ more than the seeds do: on seed 10 Opus~5 cites a memory in 28\% of
calls, while the four GPT-5.6 and GPT-6 models, gpt-oss-120b, and Qwen3.6 never exceed 7\% on any
seed. Where it could be checked, a cited memory was correct more than ten times as often as it was false, but false
memories persist: GLM-5.3 carries from seed to seed the (incorrect) rule that the heal between acts is 55\% of max HP.
Acting against an applicable lesson also becomes more frequent (11 of 16 models, $p=0.004$), as does
doubting or correcting a memory ($p=0.046$). More use of memory does not reduce errors: the rate of wrong
statements shows no trend ($p=0.27$). Each seed is a different game, so these rates mix learning with the
difficulty of the seed.

The models usually write down what they work out during a run (88\% of their discoveries of every kind;
Fable~5.1 every time), but the weakest do so in a third of cases or fewer. gpt-oss-120b, seeing a Strike shown as 4 instead
of 6 while it is Weak, concludes ``So the UI shows damage after Fatigued [Weakened] and maybe other
debuffs'' and does not note it.

\begin{table}[t]
\centering
\caption{\textbf{Rules the models observed but did not keep (\cref{app:traces-memory}).} For each mechanic the annotators tracked: \emph{Discovered}, the times a model worked the rule out from what had just happened, and \emph{Noted}, how many of those reached a note; \emph{Wrong after}, the statements of the rule judged wrong that a model made after it had discovered the rule itself, in the same run or a later seed, and \emph{Models}, how many models made them. Below, four of them: a discovery and a later wrong statement by the same model.}
\label{tab:trace-relapse}
\begin{adjustbox}{max width=\linewidth}
\small
\begin{tabular}{l rrrr}
\toprule
Rule & Discovered & Noted & Wrong after & Models \\
\midrule
Card numbers include modifiers & 152 & 51 & 241 & 12 \\
Ascender's Bane cannot be removed & 24 & 15 & 129 & 7 \\
Block expires at the start of the next turn & 77 & 51 & 116 & 10 \\
Intent numbers include modifiers & 79 & 55 & 105 & 11 \\
Rounding down after a multiplier & 60 & 37 & 32 & 9 \\
How much the end of an act heals & 43 & 42 & 56 & 7 \\
\bottomrule
\end{tabular}
\end{adjustbox}

\vspace{0.6em}
\begin{adjustbox}{max width=\linewidth}
\small
\begin{tabular}{p{2.6cm} p{6.0cm} p{6.0cm}}
\toprule
Model & Discovers & Later \\
\midrule
Claude Haiku 4.5 (seed 1 $\to$ seed 2) & ``now it shows 11, so Fatigue [Weak] must already be factored into the hand display'' & ``with Fatigue 1 still active, Trip Attack [Bash] deals: 6 * (1 - 0.25) = 4.5 = 4 damage'' \\
\addlinespace
Qwen3.6 35B-A3B (same run) & ``It says `Cannot be removed from your deck'!'' & ``Remove Sins of the Past [Ascender's Bane] (cost will be 100g now: 50 + 25 + 25 = 100g)'' \\
\addlinespace
gpt-oss-120b (same run) & ``Guard [Block] is currently 0, as guard from previous turn was consumed at start of our turn'' & ``Actually Guard persists until it's consumed by damage'' \\
\addlinespace
Claude Fable 5.1 (seed 1 $\to$ seed 3) & ``Picking up boss artifact also healed 32 HP (HP 32\textrightarrow{}64)'' & ``Since the boss heals to full anyway'' \\
\bottomrule
\end{tabular}
\end{adjustbox}
\end{table}

Many errors are of rules the model had already seen for itself (\cref{tab:trace-relapse}). The annotators
followed six mechanics through every run and recorded each time a model worked one out from what had just
happened. A discovery about the numbers on cards reaches a note in only a third of cases, the others in 62--98\%,
but noting it does not keep it: after discovering a rule, models state it wrongly again 32 to 241 times per rule,
mostly in later seeds. Claude Haiku~4.5 works out on seed 1 that the hand already shows damage after Weak, and on
seed 2 applies Weak to it again. Qwen3.6 reads that Ascender's Bane cannot be removed and one floor
later budgets gold to remove it. gpt-oss-120b observes five times that its Block is gone at the start of its turn,
and states that Block lasts into the next turn in eight of its ten runs. The models weigh what is
worth recording almost only when the harness asks them to, but they forecast fights unasked: 1{,}454
unprompted numeric forecasts in decisions, over half of them by Fable~5.1 and Opus~5.

\subsection{What consolidation keeps}
\label{app:traces-consolidation}

At the end of each run the model turns its run notes into permanent memories
(\cref{sec:methods-memory}). Of 4{,}430 notes, 44\% leave no trace in any memory, and the models differ
widely (\cref{tab:trace-memory}): Fable~5.1 and Astra drop 3--5\% of their notes, while GPT-5.6 Terra and
Luna, DeepSeek~V4.1 Flash, and gpt-oss-120b drop 75--84\%. A model seldom says why it keeps or drops a
note. What it keeps is characteristic: about half of Astra's memories are facts about
enemies and half of Opus~5's are rules of strategy. One memory in eight contains a claim
the run's own record contradicts, and half of Haiku~4.5's do (a narrower
test than \cref{tab:memory-classification}: only what the model could have checked in its own prompt).
Consolidation is not limited to the notes: in all but one run the model saves a lesson that appears in
none of them (e.g.\ a lesson learned from reviewing the floor history).

\subsection{How the models explain a loss}
\label{app:traces-loss}

Of the 160 runs, 157 are defeats; the others are Sol's Act 4 win (Heart kill) and two runs in which
Fable~5.1 beats both Act~3 bosses without all three keys, which ends the game without progressing to the final Act.
The models nearly always state a cause and consult the room history, but seldom pin the loss on a decision: a routing
mistake is named in 75 runs, a card wrongly taken in 40, a card wrongly skipped in 28, and a wrong upgrade
or rest in 9. The models blame the final fight in 87 runs; the Opus~5.5 annotator, reading the same history, in only 27.

The account is often wrong. In 67 runs it contradicts the run's own room history,
including all ten of Haiku~4.5's. Explaining a death with 498 gold unspent, Haiku~4.5
writes ``no shop appeared in R12-R14 Unknowns (R12 Unknown/event, R13 Shop, R14'', naming the shop in
the sentence that denies it. In 84 runs the model misses a large problem the history shows, such as a run
with no upgrades or a death with gold unspent: all ten runs of gpt-oss-120b, against one each of
Fable~5.1 and Sol.

Even Sol's win goes unexplained. Its memories after killing the Heart record what the
enemies of Act~4 do, not what made the run unusual (\cref{app:traces-seed5}), so the bank it carries into
seed 6 does not say how it got there. Fable~5.1, after each of its Act~3 clears, does name the relics and
cards that carried the run and why it did not reach Act~4.

\subsection{One seed, four models}
\label{app:traces-seed5}

\begin{table}[t]
\centering
\caption{%
\textbf{Seed 5 for the four strongest models.} From each run's room history and annotated reasoning
(\cref{app:traces-method}). \emph{Free} elites are those fought while the opening blessing set enemies to
1 HP. \emph{Starter cards} are the Strikes and Defends left in the final deck. The reasoning rows count
LLM calls: all calls of the run, calls giving an unasked numeric forecast of a fight, calls choosing a
card, upgrade, or relic for a named boss, and calls citing a memory from an earlier seed.
}
\label{tab:trace-seed5}
\begin{adjustbox}{max width=\linewidth}
\small
\begin{tabular}{@{}l llll@{}}
\toprule
 & GPT-5.6 Sol & GPT-6 Astra & Claude Opus 5 & Claude Fable 5.1 \\
\midrule
Result & Won (Heart, room 56) & Died, room 50 & Died, room 51 & Died, room 24 \\
Died to & -- & First Act 3 boss & Second Act 3 boss & Act 2 elite \\
Opening blessing & 1-HP enemies, 3 fights & 1-HP enemies, 3 fights & 250 gold and a curse & 1-HP enemies, 3 fights \\
Elites fought (free) & 8 (2) & 6 (2) & 2 (0) & 3 (1) \\
Relics & 21 & 16 & 15 & 8 \\
Keys & 3 & 3 & 0 & 1 \\
Rest sites: rest / upgrade / key & 6 / 1 / 1 & 6 / 0 / 1 & 5 / 4 / 0 & 0 / 2 / 1 \\
Act 2 boss relic & Pandora's Box & Slaver's Collar & Slaver's Collar & -- \\
HP entering the Act 3 bosses & 118/120 & 69/101 & 74/96 & -- \\
Final deck (starter cards) & 30 (0) & 32 (8) & 39 (9) & 25 (9) \\
Maximum HP at the end & 122 & 101 & 96 & 82 \\
\midrule
LLM calls & 725 & 793 & 441 & 203 \\
Unasked forecasts of a fight & 2 & 1 & 49 & 33 \\
Choices made for a named boss & 8 & 0 & 26 & 13 \\
Memories cited for a decision & 8 & 6 & 91 & 24 \\
\bottomrule
\end{tabular}
\end{adjustbox}
\end{table}

Seed 5 is where GPT-5.6 Sol kills the Heart{}, on the fifth run of its chain and in the main-table
configuration (themed reskin, xml notes, read-write memory bank); the other three of the top four
fail on the same seed (\cref{tab:trace-seed5}). It is close to a controlled comparison. Sol, GPT-6 Astra,
and Claude Fable~5.1 take the same opening blessing on the first floor and the same path: Sol and Astra visit the same 46
rooms in the same order, and Fable follows them, apart from one branch, until it dies in room 24. Claude
Opus~5 takes its own route. Each model starts with its own memories of seeds 1 to 4, and Sol and Astra
expose little of their reasoning through the API, so for those two what happened is firmer than estimations from their thinking traces.

\paragraph{Fable.} The Neow's Lament starting blessing sets enemies to 1~HP for three fights.
The map enables Sol and Astra to path through two elites in their first three combats, and so acquire two relics and additional gold without spending any hit points.
Fable turns aside to take a key at a rest site, spends
its third fight on ordinary enemies, and later pays 19 HP for an elite the others had for free. All three then meet the same Act~2 elite, a
group of three enemies, in room 24. Sol rests before it and Astra takes a key; both enter nearly full,
use two potions, and lose 7--10 HP. Fable upgrades instead, enters with no potions after spending its gold
on two cards, and loses all 70 of its HP in four turns: it has no multi-enemy attack to take them down efficiently, while Sol
and Astra had both bought one (Whirlwind) from the shop. Fable is the most deliberate of the three, forecasting fights
and planning for named bosses, and its review of the death is accurate. Its reasoning was sound at each
step; the run was lost on the route.

\paragraph{Sol and Astra.} The two are level through Act~2, and Astra is ahead after the Act~2 boss, at
which Sol falls to 23 of 96 HP. Four choices then separate them. In the Act~2 shop Sol buys a relic that
discounts later shops, and Astra one that gives Block on the first turn. From the boss chest Sol takes
Pandora's Box, which replaces all nine of its Strikes and Defends with random cards: ``The
Iridescent Capsule [Pandora's Box] could remix my weak basics into powerful Blood Mage
[Ironclad] cards, though the randomness might lead to losing reliable blocks''. Astra takes
Slaver's Collar, which gives energy in elite and boss fights. In the first Act~3 shop,
with about 500 gold each, Sol buys two relics among other things and Astra none. In Act~3, Sol kills the
first elite in one turn for 9 HP, where Astra takes seven turns and loses 46. Sol reaches the Act~3 bosses
at 118 of 120 HP and Astra at 69 of 101, and Astra dies to the first. They rest equally often (6 rest sites
each): Sol's fights are short, so resting keeps it full, and Astra's are
long, so resting only keeps it alive.

\paragraph{Opus~5.} Opus~5 reasons the most of the four, with 49 unasked forecasts and 91 citations of its
memories in this run, and its memories set its course. At the first map it writes: ``Given my rough
history with Act 1 elites but a decently funded deck by R6 thanks to 349 gold and two shops,
I'm inclined to lean toward Path A for safety''. Planning Act~3 it cites two more lessons,
``Per my own notes I've died gold-rich twice, and `artifact quantity is not
power.'\,'', and avoids every Act~3 elite. It fights two elites in the whole run and ends with a 39-card
deck that still holds nine starter cards. It reaches the two Act~3 bosses at 74 of 96 HP and dies to the
second. Its review blames the size of the deck and not knowing that the act ends with two bosses; it does
not mention the 27 HP it chose to spend at two events shortly before, and keeps one of those choices in
memory as correct.

\paragraph{What separates them.} The winner is not the model that reasons most, by any measure we have.
Fable and Opus~5 forecast, plan, and consult their memories far more often than Sol and Astra, and none
outlasts Astra by more than a room. By Act~3 Sol has a deck with no starter cards and five more relics
than any of the others, and both come from choices that carried risk: fighting elites, and a relic with a
random result. Opus~5 declines such choices because of lessons from earlier deaths, each true of the run
that taught it. Chance played its part: Pandora's Box could have gone badly. One seed
does not show that taking risks wins. It shows that on the one occasion a model killed the Heart, the
margin was not in the quality of its reasoning as we can read it, and that the model did not record where
the margin came from.

\subsection{Beliefs about the interface}
\label{app:traces-interface}

Some apparently poor planning is a wrong belief about the harness, which accepts several actions in one
response (\cref{app:harness-features}) and, whenever a model sends one action where it could have sent
several, reminds it to send them all at once. The belief survives the reminders. In 1{,}244 calls a model
plans several plays and sends only the first, most often gpt-oss-120b, the four GPT-5.6 and GPT-6 models,
and Qwen3.6. Sol gives its reason: ``the tool system only allows one tool call per message''.
The reminders are frequent and mostly ignored (\cref{tab:tool-errors}). GPT-6 Astra is reminded 204 times
per run, GPT-5.6 Sol and Terra about 95 times, and gpt-oss-120b 60 times, and the reply that follows a
reminder is again a single action 93--99\% of the time; the Claude models are reminded 5--9 times per run.
Two of the strongest LLMs, Astra and Sol, thus hold a belief about the interface that an instruction
repeated about a hundred times or more per run does not change.

\FloatBarrier
\section{Ablation results in full}
\label{app:ablations-by-model}
\Cref{fig:conditions} summarizes each ablation model by model. The subsections below then give each
ablation in turn, pooled over models and model by model: the harness, reasoning effort, memory, the
reskin, and the knowledge given to the model. How each ablation was run is described in
\cref{sec:methods-ablations}.

\begin{figure}[htbp]
\centering
\includegraphics[width=\linewidth]{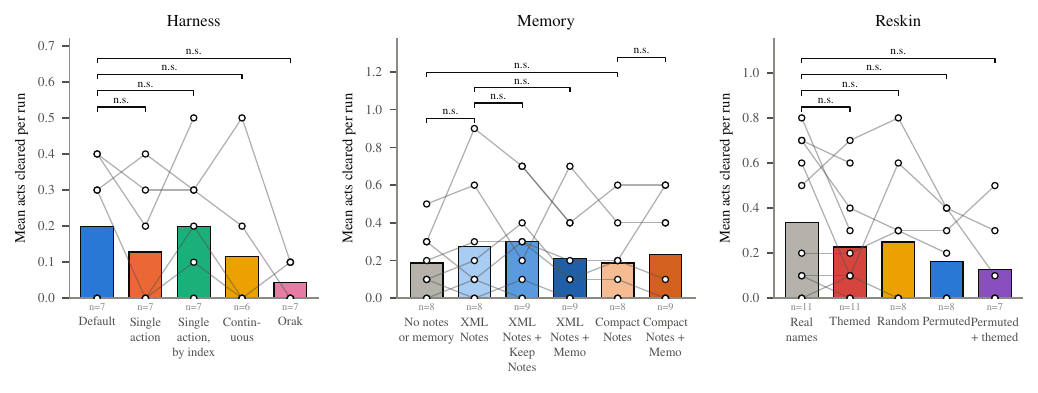}
\caption{\textbf{Each ablation, model by model.} Each grey line is one LLM agent: its mean over the 10 seeds in each
condition, shown only where it has all 10 seeds in both that condition and the condition's baseline, so every
comparison is paired. Bars are the mean over those agents, and $n$ beneath each bar is how many there are.
Baselines are the first bar of each panel, except in the memory panel, where XML Notes and Compact Notes are read
against No notes or memory, and each memory bank against the arm that writes the same kind of notes without one. %
 In the memory panel \emph{No notes or memory} writes nothing within a run. Every other arm takes notes during each run, as keyed xml notes (blue) or, in the two compact arms, as one summary the model rewrites (orange); they differ in what reaches the next run: \emph{XML Notes} and \emph{Compact Notes} discard the notes, \emph{+ Keep Notes} carries them over verbatim, and \emph{+ Memo} first rewrites them into a memory bank (\cref{app:methods-memory}). Panels show mean acts cleared per run; the pooled tables are %
\cref{tab:harness-compare,tab:ablation-memory,tab:ablation-reskin}. Brackets test each condition
against its baseline with a paired sign-flip permutation test over agents (* $p<0.05$; n.s.\ not
significant). Orak is shown for the four models with all 10 of its seeds scored; its full paired comparison is in
\cref{tab:harness-compare}.}
\label{fig:conditions}
\end{figure}

\FloatBarrier
\subsection{Harness}
\label{app:ablation-harness}
\providecommand{\spireicon}[1]{\raisebox{-0.25ex}{\includegraphics[height=1.7ex]{icons/#1.pdf}}}
\begin{table}[t]
\centering
\caption{\textbf{Effect of the harness, per model.} Supplementary to \cref{tab:harness-compare}. Orak cells score runs killed by Orak's harness as a run lost at the floor it had reached; all harness variants were run without the reskin or a cross-run memory bank. All cells are on A20, with each model at its main-table reasoning setting (grey). Act~1: Act~1 win rate (\%). Acts: mean number of acts cleared per run. Floor: mean final floor. \best{Best} and \runnerup{second best} are per row, among cells with all 10 seeds.}
\label{tab:ablation-harness-by-model}
\begin{adjustbox}{max width=\linewidth}
\small
\begin{tabular}{@{}c@{\hspace{3pt}}l rrr rrr rrr rrr rrr@{}}
\toprule
  &   & \multicolumn{3}{c}{Default} & \multicolumn{3}{c}{Single action} & \multicolumn{3}{c}{Single action, by index} & \multicolumn{3}{c}{Continuous} & \multicolumn{3}{c}{Orak} \\
\cmidrule(lr){3-5} \cmidrule(lr){6-8} \cmidrule(lr){9-11} \cmidrule(lr){12-14} \cmidrule(lr){15-17}
 & Agent & Act~1 & Acts & Floor & Act~1 & Acts & Floor & Act~1 & Acts & Floor & Act~1 & Acts & Floor & Act~1 & Acts & Floor \\
\midrule
\spireicon{qwen} & Qwen3.8 27B {\scriptsize\textcolor{gray}{(xhigh)}} & \best{40} & \best{0.40} & \runnerup{15.7} & \runnerup{30} & \runnerup{0.30} & 13.6 & \runnerup{30} & \runnerup{0.30} & 14.8 & 20 & 0.20 & \best{16.3} & 0 & 0.00 & 10.0 \\
\spireicon{qwen} & Qwen3.6 35B-A3B & 0 & 0.00 & \runnerup{8.6} & 0 & 0.00 & 8.3 & 0 & 0.00 & 8.5 & 0 & 0.00 & 7.5 & 0 & 0.00 & \best{9.4} \\
\spireicon{qwen} & Qwen3.5 122B-A10B {\scriptsize\textcolor{gray}{(xhigh)}} & 0 & 0.00 & \best{10.6} & 0 & 0.00 & 8.5 & 10 & 0.10 & 8.8 & 0 & 0.00 & 8.9 & 0 & 0.00 & \runnerup{9.1} \\
\spireicon{gemini} & Gemini 3.8 Flash {\scriptsize\textcolor{gray}{(low)}} & \runnerup{30} & 0.30 & 16.0 & \runnerup{30} & \runnerup{0.40} & 17.0 & 20 & 0.30 & \runnerup{17.9} & \best{40} & \best{0.50} & \best{19.6} & 10 & 0.10 & 13.5 \\
\spireicon{nvidia} & Nemotron 3 Nano Omni & 0 & 0.00 & \best{9.2} & 0 & 0.00 & 5.2 & 0 & 0.00 & 6.8 & 0 & 0.00 & \runnerup{7.2} & 0 & 0.00 & 4.2 \\
\spireicon{openai} & gpt-oss-120b {\scriptsize\textcolor{gray}{(high)}} & \best{30} & \best{0.30} & \best{15.3} & 0 & 0.00 & 12.9 & \runnerup{20} & \runnerup{0.20} & \runnerup{13.7} & 0 & 0.00 & 9.6 & 10 & 0.10 & 10.0 \\
\midrule
 & Mean (6 agents) & \best{17} & \best{0.17} & \best{12.6} & 10 & 0.12 & 10.9 & \runnerup{13} & \runnerup{0.15} & \runnerup{11.8} & 10 & 0.12 & 11.5 & 3 & 0.03 & 9.4 \\
\bottomrule
\end{tabular}
\end{adjustbox}
\end{table}

\Cref{tab:harness-compare} in the main text pools each harness variant over the models run with it;
\cref{tab:ablation-harness-by-model} gives the six models that have all ten seeds in every arm. These are mostly
weaker models: with our default harness they clear 0.17 acts per run and reach floor 12.6 on average, so a single
cleared act moves a model's mean by 0.1, and most differences in acts rest on one or two runs.
Floors are the more sensitive measure here.

Our default harness has the highest mean on all three measures, but no variant is worse for every model. Allowing
one action per response costs four of the six models 2.1 to 4.0 floors, Nemotron~3 Nano Omni the most (9.2 to 5.2);
making every choice by number as well recovers about half of that on average (11.8 floors against 10.9), and with it
gpt-oss-120b clears Act~1 in two seeds rather than none. Gemini~3.8 Flash at low effort is the exception: it reaches
more floors under every variant of our harness, and most with one conversation for the whole run (19.6 against 16.0,
and 0.50 acts against 0.30). The continuous conversation hurts most the two models whose context window it most often
overflowed: gpt-oss-120b and Nemotron~3 Nano Omni, whose oldest turns were dropped on 35\% and 38\% of their calls
(\cref{sec:methods-ablations}), lose 5.7 and 2.0 floors, and gpt-oss-120b loses all three of its Act~1 clears. Under
Orak's harness, whose rule bot plays every non-combat floor, five of the six models reach fewer
floors, and only Gemini~3.8 Flash and gpt-oss-120b clear Act~1, once each; Qwen3.6 35B-A3B, which clears no act under
any harness, reaches its deepest floors there (9.4 against 8.6).

\FloatBarrier
\subsection{Reasoning effort}
\label{app:ablation-effort}
\label{app:frontiers-thinking}
\providecommand{\spireicon}[1]{\raisebox{-0.25ex}{\includegraphics[height=1.7ex]{icons/#1.pdf}}}
\begin{table}[t]
\centering
\caption{\textbf{Effect of reasoning effort.} Each model at every effort level it was run at, A20 on the held-out seeds with the main-table configuration (themed reskin, xml notes, read-write memory, generic guidance). Columns as in \cref{tab:main-results}, with Acts cleared the mean per run; Act~4 (a Heart kill) is not shown but counts in Acts cleared. Only levels with all 10 seeds completed are shown, and only models with two or more such levels. Bold marks the \best{best reasoning level per model} on each column. $^{\dagger}$Self-hosted; priced at Artificial Analysis headline rates.}
\label{tab:ablation-effort}
\begin{adjustbox}{max width=\linewidth}
\small
\begin{tabular}{@{}c@{\hspace{3pt}}ll rrrr r rrc r rr@{}}
\toprule
 & & & & \multicolumn{3}{c}{Act win rate (\%)} & \multirow[b]{2}{*}{\makecell{Acts\\cleared}} & \multicolumn{3}{c}{Floor} & \multirow[b]{2}{*}{\makecell{Thinking\\(k/floor)}} & \multicolumn{2}{c}{Cost (\$)} \\
\cmidrule(lr){5-7} \cmidrule(lr){9-11} \cmidrule(lr){13-14}
 & Agent & Effort & Seeds & 1 & 2 & 3 & & \multicolumn{1}{c}{mean} & \multicolumn{1}{c}{median} & \multicolumn{1}{c}{[min--max]} & & \multicolumn{1}{c}{/floor} & \multicolumn{1}{c}{/run} \\
\midrule
\spireicon{qwen} & Qwen3.8 27B & low & 10 & \best{30} & 0 & 0 & \best{0.30} & \best{15.2} & \best{16.0} & [7--24] & \best{7.8} & \best{0.045}\rlap{$^{\dagger}$} & \best{0.74}\rlap{$^{\dagger}$} \\
 &  & medium & 10 & 20 & 0 & 0 & 0.20 & 12.8 & 10.5 & [6--24] & 9.3 & 0.055\rlap{$^{\dagger}$} & 0.76\rlap{$^{\dagger}$} \\
 &  & xhigh & 10 & 20 & 0 & 0 & 0.20 & 14.9 & \best{16.0} & [5--29] & 18.1 & 0.091\rlap{$^{\dagger}$} & 1.45\rlap{$^{\dagger}$} \\
\midrule
\spireicon{qwen} & Qwen3.5 122B-A10B & low & 10 & 10 & 0 & 0 & 0.10 & 9.4 & \best{9.0} & [4--18] & 3.8 & 0.074\rlap{$^{\dagger}$} & 0.78\rlap{$^{\dagger}$} \\
 &  & medium & 10 & 0 & 0 & 0 & 0.00 & 9.5 & 7.0 & [4--16] & 3.8 & 0.077\rlap{$^{\dagger}$} & \best{0.76}\rlap{$^{\dagger}$} \\
 &  & xhigh & 10 & 0 & 0 & 0 & 0.00 & \best{9.8} & \best{9.0} & [7--16] & \best{3.7} & \best{0.072}\rlap{$^{\dagger}$} & 0.77\rlap{$^{\dagger}$} \\
\midrule
\spireicon{claude} & Claude Fable 5.1 & low & 10 & 70 & 20 & 0 & 0.90 & 27.5 & 27.5 & [7--51] & \best{1.0} & \best{0.275} & \best{8.12} \\
 &  & max & 10 & \best{90} & \best{50} & 20 & \best{1.60} & \best{36.2} & \best{39.5} & [7--52] & 17.7 & 1.664 & 61.91 \\
\midrule
\spireicon{claude} & Claude Opus 5 & low & 10 & 70 & 0 & 0 & 0.70 & 22.7 & 22.0 & [10--33] & \best{0.9} & \best{0.204} & \best{4.84} \\
 &  & max & 10 & \best{90} & 40 & 0 & \best{1.30} & \best{32.5} & \best{33.0} & [7--51] & 21.3 & 1.743 & 60.62 \\
\midrule
\spireicon{claude} & Claude Sonnet 5 & low & 10 & 10 & 0 & 0 & 0.10 & 12.3 & 13.5 & [4--24] & \best{0.8} & \best{0.074} & \best{1.03} \\
 &  & max & 10 & \best{40} & 10 & 0 & \best{0.50} & \best{19.3} & \best{16.0} & [7--36] & 23.2 & 0.509 & 10.70 \\
\midrule
\spireicon{gemini} & Gemini 3.8 Flash & low & 10 & \best{60} & \best{10} & 0 & \best{0.70} & 22.6 & \best{22.0} & [5--50] & \best{2.5} & \best{0.063} & \best{1.56} \\
 &  & high & 10 & 50 & \best{10} & 0 & 0.60 & \best{22.9} & 18.0 & [16--47] & 13.7 & 0.117 & 2.89 \\
\midrule
\spireicon{openai} & GPT-6 Astra & low & 10 & 70 & \best{60} & 0 & 1.30 & 34.3 & \best{40.0} & [7--51] & \best{0.1} & \best{0.383} & \best{13.66} \\
 &  & max & 10 & \best{90} & 50 & 10 & \best{1.50} & \best{37.0} & \best{40.0} & [5--55] & 3.3 & 0.834 & 33.09 \\
\midrule
\spireicon{openai} & GPT-5.6 Sol & low & 10 & 50 & 10 & 0 & 0.60 & 21.8 & 18.5 & [13--50] & \best{0.5} & \best{0.133} & \best{3.15} \\
 &  & max & 10 & \best{70} & \best{20} & 10 & \best{1.10} & \best{27.8} & \best{26.0} & [7--57] & 4.0 & 0.272 & 8.39 \\
\midrule
\spireicon{openai} & GPT-5.6 Terra & low & 10 & 10 & 0 & 0 & 0.10 & 12.3 & 13.0 & [5--20] & \best{0.6} & \best{0.060} & \best{0.81} \\
 &  & max & 10 & \best{50} & 0 & 0 & \best{0.50} & \best{18.2} & \best{18.0} & [10--25] & 5.0 & 0.149 & 2.91 \\
\midrule
\spireicon{openai} & GPT-5.6 Luna & low & 10 & 0 & 0 & 0 & 0.00 & 10.8 & 8.5 & [6--16] & \best{0.4} & \best{0.005} & \best{0.06} \\
 &  & medium & 10 & 20 & 0 & 0 & 0.20 & 11.9 & 10.5 & [5--21] & 0.8 & 0.007 & 0.09 \\
 &  & high & 10 & \best{30} & 0 & 0 & \best{0.30} & \best{14.5} & \best{16.0} & [6--25] & 1.6 & 0.008 & 0.14 \\
 &  & xhigh & 10 & 20 & 0 & 0 & 0.20 & 14.4 & 14.5 & [5--24] & 2.7 & 0.010 & 0.16 \\
 &  & max & 10 & 10 & 0 & 0 & 0.10 & 14.2 & \best{16.0} & [7--24] & 5.0 & 0.015 & 0.24 \\
\midrule
\spireicon{openai} & gpt-oss-120b & low & 10 & 0 & 0 & 0 & 0.00 & 7.3 & 6.5 & [1--16] & \best{1.5} & \best{0.042}\rlap{$^{\dagger}$} & \best{0.36}\rlap{$^{\dagger}$} \\
 &  & medium & 10 & 0 & 0 & 0 & 0.00 & 9.2 & \best{8.5} & [2--16] & 4.3 & 0.047\rlap{$^{\dagger}$} & 0.47\rlap{$^{\dagger}$} \\
 &  & high & 10 & 10 & 0 & 0 & 0.10 & \best{10.4} & 8.0 & [5--21] & 17.0 & 0.085\rlap{$^{\dagger}$} & 0.91\rlap{$^{\dagger}$} \\
\midrule
\spireicon{zai} & GLM-5.3 & low & 10 & 20 & 0 & 0 & 0.20 & 14.3 & 16.0 & [5--24] & \best{0.8} & \best{0.046} & \best{0.72} \\
 &  & max & 10 & \best{60} & 0 & 0 & \best{0.60} & \best{22.1} & \best{22.0} & [5--33] & 13.4 & 0.167 & 3.92 \\
\midrule
\spireicon{zai} & GLM-5.3 Flash & low & 10 & 0 & 0 & 0 & 0.00 & 11.1 & 10.0 & [6--16] & \best{0.7} & \best{0.005} & \best{0.07} \\
 &  & medium & 10 & 20 & 0 & 0 & 0.20 & 12.9 & 13.5 & [7--20] & 2.7 & 0.010 & 0.14 \\
 &  & max & 10 & \best{40} & 0 & 0 & \best{0.40} & \best{15.6} & \best{16.0} & [5--27] & 14.4 & 0.024 & 0.40 \\
\bottomrule
\end{tabular}
\end{adjustbox}
\end{table}

\begin{figure}[tbhp]
\centering
\includegraphics[width=0.9\linewidth]{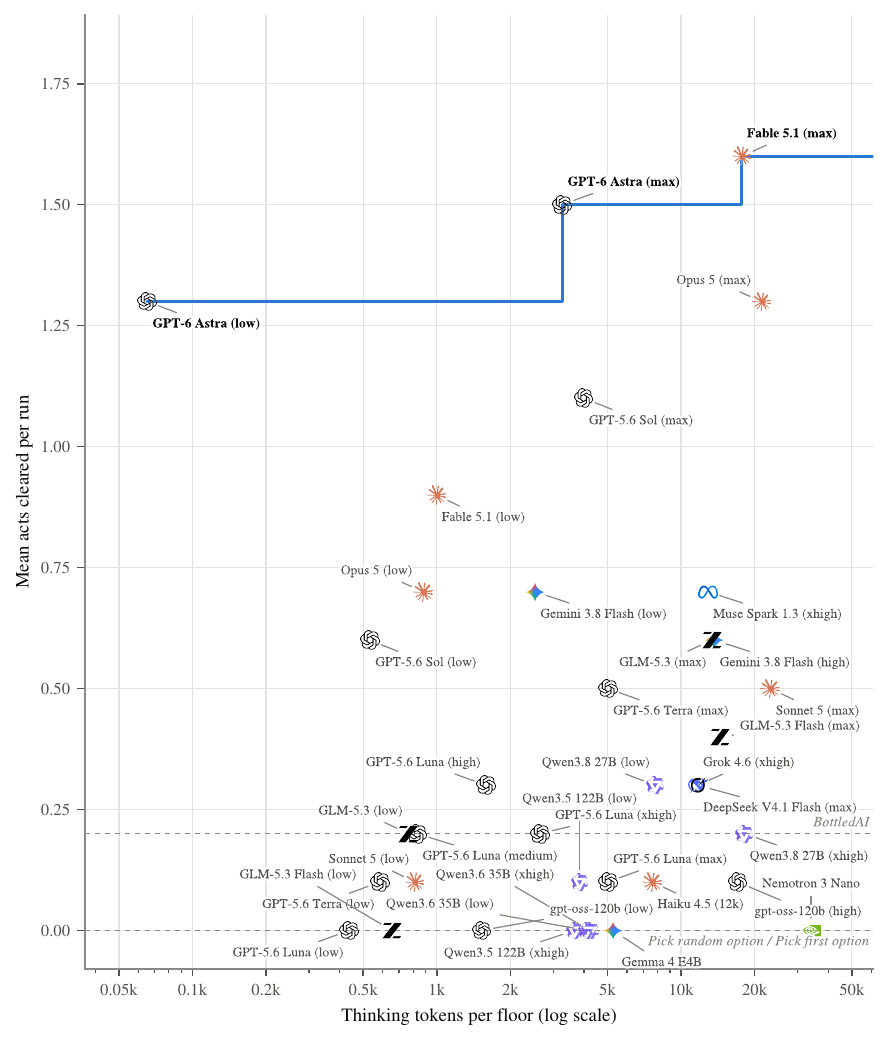}
\caption{\textbf{Performance against thinking.} Mean acts cleared per run against thinking tokens per floor (log
scale), for the same points as \cref{fig:pareto}.}
\label{fig:pareto-thinking}
\end{figure}

One would expect models to play better when they can reason more before acting.
To test this, we ran all models at both the lowest and the highest effort level available, and some at levels in between.

\Cref{tab:ablation-effort} gives the full results.
Within a model, play generally improves with reasoning effort, but not always (e.g.\ GPT-5.6 Luna peaks at high effort and does worse at xhigh and max).

Across models, \cref{fig:pareto-thinking} shows no clear relationship between the number of thinking tokens and performance.
This is because stronger models use their reasoning tokens more efficiently: the best model at a small thinking budget is GPT-6 Astra at low effort.

\FloatBarrier
\subsection{Memory}
\label{app:ablation-memory}
\begin{table}[t]
\centering
\caption{\textbf{Effect of notes within a run and of memory across runs.} \emph{No notes or memory}: nothing is written within a run and no memory bank is kept. \emph{XML Notes}: keyed xml notes within each run, discarded at its end; \emph{Compact Notes}: a single compact summary, rewritten at the end of every floor and discarded at the end of the run. \emph{+ Keep Notes}: the xml notes are kept verbatim as the memory bank; \emph{+ Memo}: at the end of each run the model rewrites its notes into the bank (xml memories, or one compact summary of every run). The first group compares writing notes with No notes or memory; the other two compare each bank with the arm that writes the same kind of notes without one. All with the themed reskin and generic guidance. Pooled over models: each method is compared with its baseline only on matched runs (the same model on the same seed; Pairs, over Agents models), and the Act~1 win rate, mean acts cleared, and mean floor reached are over those runs. Usage is compared on like-for-like floors only: the same model, seed, act, and room type, and the same occurrence of that room type within the act, kept only where every method run on that model and seed completed it (Floors), so a method is neither credited with cheap early floors nor charged for expensive late ones. Model calls and k~tokens (input, including cached, plus output, in thousands) are means per floor; the token difference is the ratio of totals. Each $\Delta$ is the method minus its baseline. All runs are A20 on the held-out seeds, each model at its main-table reasoning setting. Models: Qwen3.8 27B, Qwen3.6 35B-A3B, Qwen3.5 122B-A10B, Claude Opus 5 (low), Gemini 3.8 Flash (low), Nemotron 3 Nano Omni, GPT-5.6 Luna, gpt-oss-120b in every comparison; GLM-5.3 Flash also in XML Notes + Keep Notes, XML Notes + Memo, and Compact Notes + Memo. Per-model results are in \cref{tab:ablation-memory-by-model}; the paired tests are in \cref{app:ablation-memory}.}
\label{tab:ablation-memory}
\begin{adjustbox}{max width=\linewidth}
\small
\begin{tabular}{@{}l rr rr rr rr rr r rr rr@{}}
\toprule
 & & & \multicolumn{6}{c}{Performance (matched runs)} & \multicolumn{5}{c}{Usage per floor (matched floors)} \\
\cmidrule(lr){4-9} \cmidrule(lr){10-14}
 & & & \multicolumn{2}{c}{Act~1 win rate (\%)} & \multicolumn{2}{c}{Mean acts cleared} & \multicolumn{2}{c}{Mean floor reached} & & \multicolumn{2}{c}{Model calls} & \multicolumn{2}{c}{k tokens} \\
\cmidrule(lr){4-5} \cmidrule(lr){6-7} \cmidrule(lr){8-9} \cmidrule(lr){11-12} \cmidrule(lr){13-14}
Method & Agents & Pairs & Value & $\Delta$ & Value & $\Delta$ & Value & $\Delta$ & Floors & Value & $\Delta$ & Value & $\Delta$ \\
\midrule
\multicolumn{14}{l}{\textit{Notes within a run (no memory bank)}} \\
No notes or memory (baseline) & 8 & 80 & 19 & -- & 0.19 & -- & 13.9 & -- & 775 & 6.8 & -- & 144 & -- \\
XML Notes & 8 & 80 & 25 & \dup{$+6$} & 0.28 & \dup{$+0.09$} & 13.8 & \ddown{$-0.1$} & 775 & 8.4 & \ddown{$+1.6$} & 186 & \ddown{$+29$\%} \\
Compact Notes & 8 & 80 & 19 & \dzero{$\pm0$} & 0.19 & \dzero{$\pm0.00$} & 13.0 & \ddown{$-1.0$} & 775 & 7.7 & \ddown{$+1.0$} & 169 & \ddown{$+17$\%} \\
\midrule
\multicolumn{14}{l}{\textit{XML Notes carried across runs}} \\
XML Notes (baseline) & 9 & 90 & 24 & -- & 0.27 & -- & 13.9 & -- & 758 & 7.9 & -- & 172 & -- \\
XML Notes + Keep Notes & 9 & 90 & 28 & \dup{$+3$} & 0.30 & \dup{$+0.03$} & 15.0 & \dup{$+1.1$} & 758 & 8.2 & \ddown{$+0.3$} & 288 & \ddown{$+67$\%} \\
XML Notes + Memo & 9 & 90 & 20 & \ddown{$-4$} & 0.21 & \ddown{$-0.06$} & 13.5 & \ddown{$-0.4$} & 758 & 8.4 & \ddown{$+0.5$} & 233 & \ddown{$+35$\%} \\
\midrule
\multicolumn{14}{l}{\textit{Compact Notes carried across runs}} \\
Compact Notes (baseline) & 9 & 90 & 22 & -- & 0.22 & -- & 13.5 & -- & 758 & 7.3 & -- & 157 & -- \\
Compact Notes + Memo & 9 & 90 & 22 & \dzero{$\pm0$} & 0.23 & \dup{$+0.01$} & 14.2 & \dup{$+0.7$} & 758 & 7.6 & \ddown{$+0.3$} & 170 & \ddown{$+8$\%} \\
\bottomrule
\end{tabular}
\end{adjustbox}
\end{table}

\Cref{tab:ablation-memory} pools the memory arms over models, and \cref{tab:ablation-memory-by-model} gives
them model by model. \Cref{fig:memory-growth} shows how the bank grows over a chain, and whether play
changes along it is in \cref{app:play-learning}.

Writing notes within a run raises the acts a model clears by about half. Against \emph{No notes or memory}, which writes nothing and
keeps no bank (8 models), \emph{XML Notes} (keyed notes discarded at the end of each run) raises mean acts cleared from 0.19 to 0.28 per run
($+0.09$; $p = 0.50$; 95\% interval $-0.07$ to $+0.30$, from inverting a two-sided exact sign-flip test over the models' mean paired
differences) and the share of runs that clear Act~1 from 19\% to 25\%. At ten seeds per model this gain is not significant, and floors do
not change ($-0.1$; $p = 0.98$; $-2.6$ to $+3.4$). \emph{Compact Notes} changes neither acts ($\pm 0.00$;
$-0.10$ to $+0.10$) nor floors ($-1.0$; $p = 0.40$; $-2.8$ to $+2.1$). Against the arm that writes the
same kind of notes without a bank, no bank raises acts by more than 0.03: \emph{XML Notes + Keep Notes} $+0.03$ ($-0.17$ to $+0.23$),
\emph{XML Notes + Memo} $-0.06$ ($-0.20$ to $+0.07$) and \emph{Compact Notes + Memo} $+0.01$ ($-0.07$ to $+0.13$), with $p \ge 0.62$,
and none changes floors ($p \ge 0.47$).

\Cref{fig:learning} compares each memory cohort's mean acts
cleared over seeds 1--5 with seeds 6--10. The two halves are different games, so the same model's
no-memory cohort on the same seeds is shown alongside as a control for the halves' difficulty:
learning appears as a larger improvement than the control's.
The arrows point in different directions, some strong models' backwards, mostly because five runs per
half are too few: a single win moves a half's mean by 0.8 acts. The paired difference by position in the chain,
below, removes each seed's difficulty and shows no trend. Where a bank does change play, it need not help: models
whose banks hold more errors clear fewer acts (\cref{app:further-memory-banks}), and the banks advise caution far
more often than risk (\cref{app:play-risk}).

\begin{figure}[ht]
\centering
\vspace{-3.5em}
\iftoggle{arxiv}{
\includegraphics[width=0.94\linewidth]{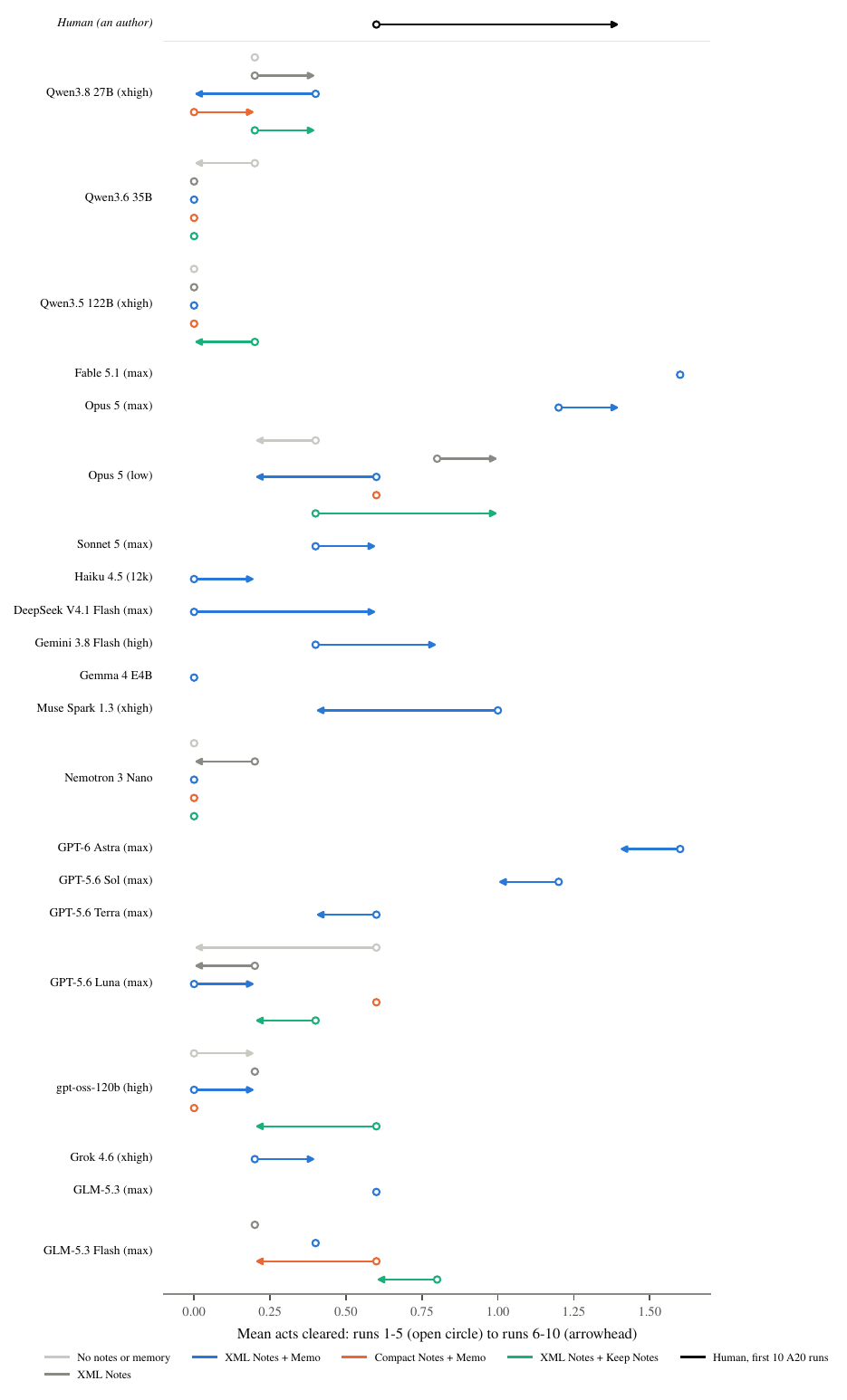}
}{
\includegraphics[width=0.97\linewidth]{figures/fig_learning_acts.pdf}
}
\vspace{-1.5em}%
\caption{%
\footnotesize
\textbf{Do models learn from their memories?} Mean acts cleared per run over the first five runs of each chain
(open circle) against the last five (arrowhead), by memory mechanism; a circle alone means no change.
The two cohorts without a memory bank on the same seeds (greys) are controls for how hard each half is:
learning shows as an arrow further right than the grey ones.
\emph{XML Notes} still writes xml notes within each run and discards them at its end, and \emph{No notes or memory} writes none; the
other three are the mechanisms of \cref{tab:ablation-memory}. For reference, the
top row is one of the authors (a non-professional player) over their first ten A20 Ironclad runs,
playing the unmodified game. There are no consistent trends across models.
}
\label{fig:learning}
\end{figure}

If a model learned from what it carries forward, the gain over the no-memory baseline should grow along the
chain, as the bank fills. \Cref{fig:memory-trend,tab:ablation-memory-trend} split each paired difference of
\cref{tab:ablation-memory} by the run's position in the chain. No method shows a trend: the fitted change
per run is within $\pm 0.021$ acts and between $-0.22$ and $+0.03$ floors, with $p \ge 0.29$ throughout, and
for every method the mean floor difference over runs 6--10 is smaller than over runs 1--5.
A single paired difference has a standard deviation of 7.0 to 8.7 floors, and already 4.8 to 6.9 floors at run~1, where every bank is still empty, so a trend of less than about half a floor per run would not be detected here.

\begin{figure}[htbp]
\centering
\includegraphics[width=\linewidth]{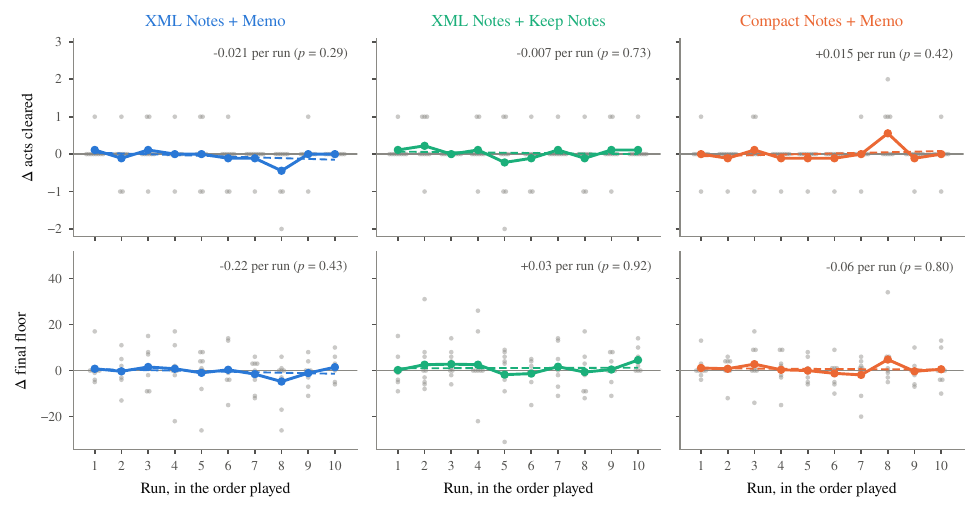}
\caption{
\textbf{Does the effect of memory grow along the memory formation chain?} Each memory method is compared against its no-memory baseline on
the same seed, by the run's position in the 10-run chain of memory-forming runs.
We show the difference in acts cleared (top) and in final floor reached (bottom).
Grey: individual pairs (9 models per position).
Solid line: their mean.
Dashed line: linear trend (slope and two-sided permutation $p$ (\cref{tab:ablation-memory-trend} annotated).
At run~1 every bank is empty and so the two conditions are identical (zero delta is expected for run~1).
}
\label{fig:memory-trend}
\end{figure}

\begin{table}[t]
\centering
\caption{\textbf{Does the effect of memory grow along the chain?} Each memory method minus its no-memory baseline, paired by model and seed and averaged over the 9 models of \cref{tab:ablation-memory} at each position in the 10-run chain (\cref{fig:memory-trend} shows every pair). At run~1 the bank is empty, so that row shows the spread of a paired difference when nothing separates the arms. $\Delta$Acts is the difference in acts cleared and $\Delta$Floor in final floor. Slope is the change in the difference per run, fitted with a separate intercept for each model; $p$ is two-sided, from permuting the differences over positions within each model (20{,}000 draws).
}
\label{tab:ablation-memory-trend}
\begin{adjustbox}{max width=\linewidth}
\small
\begin{tabular}{l rr rr rr}
\toprule
 & \multicolumn{2}{c}{XML Notes + Memo} & \multicolumn{2}{c}{XML Notes + Keep Notes} & \multicolumn{2}{c}{Compact Notes + Memo} \\
\cmidrule(lr){2-3} \cmidrule(lr){4-5} \cmidrule(lr){6-7}
Run & $\Delta$Acts & $\Delta$Floor & $\Delta$Acts & $\Delta$Floor & $\Delta$Acts & $\Delta$Floor \\
\midrule
1 & \dup{$+0.11$} & \dup{$+0.8$} & \dup{$+0.11$} & \dup{$+0.2$} & \dzero{$\pm0.00$} & \dup{$+1.1$} \\
2 & \ddown{$-0.11$} & \ddown{$-0.3$} & \dup{$+0.22$} & \dup{$+2.6$} & \ddown{$-0.11$} & \dup{$+0.8$} \\
3 & \dup{$+0.11$} & \dup{$+1.6$} & \dzero{$\pm0.00$} & \dup{$+2.8$} & \dup{$+0.11$} & \dup{$+2.8$} \\
4 & \dzero{$\pm0.00$} & \dup{$+0.8$} & \dup{$+0.11$} & \dup{$+2.6$} & \ddown{$-0.11$} & \dup{$+0.3$} \\
5 & \dzero{$\pm0.00$} & \ddown{$-1.0$} & \ddown{$-0.22$} & \ddown{$-1.8$} & \ddown{$-0.11$} & \dzero{$\pm0.0$} \\
6 & \ddown{$-0.11$} & \dup{$+0.3$} & \ddown{$-0.11$} & \ddown{$-1.3$} & \ddown{$-0.11$} & \ddown{$-1.2$} \\
7 & \ddown{$-0.11$} & \ddown{$-1.6$} & \dup{$+0.11$} & \dup{$+1.7$} & \dzero{$\pm0.00$} & \ddown{$-1.9$} \\
8 & \ddown{$-0.44$} & \ddown{$-4.8$} & \ddown{$-0.11$} & \ddown{$-0.7$} & \dup{$+0.56$} & \dup{$+4.8$} \\
9 & \dzero{$\pm0.00$} & \ddown{$-1.1$} & \dup{$+0.11$} & \dup{$+0.4$} & \ddown{$-0.11$} & \ddown{$-0.3$} \\
10 & \dzero{$\pm0.00$} & \dup{$+1.4$} & \dup{$+0.11$} & \dup{$+4.6$} & \dzero{$\pm0.00$} & \dup{$+0.6$} \\
\midrule
Mean & \ddown{$-0.06$} & \ddown{$-0.4$} & \dup{$+0.03$} & \dup{$+1.1$} & \dup{$+0.01$} & \dup{$+0.7$} \\
Slope per run & \ddown{$-0.021$} & \ddown{$-0.22$} & \ddown{$-0.007$} & \dup{$+0.03$} & \dup{$+0.015$} & \ddown{$-0.06$} \\
$p$ & 0.29 & 0.43 & 0.73 & 0.92 & 0.42 & 0.80 \\
\bottomrule
\end{tabular}
\end{adjustbox}
\end{table}

\begin{figure}[htbp]
\centering
\includegraphics[width=\linewidth]{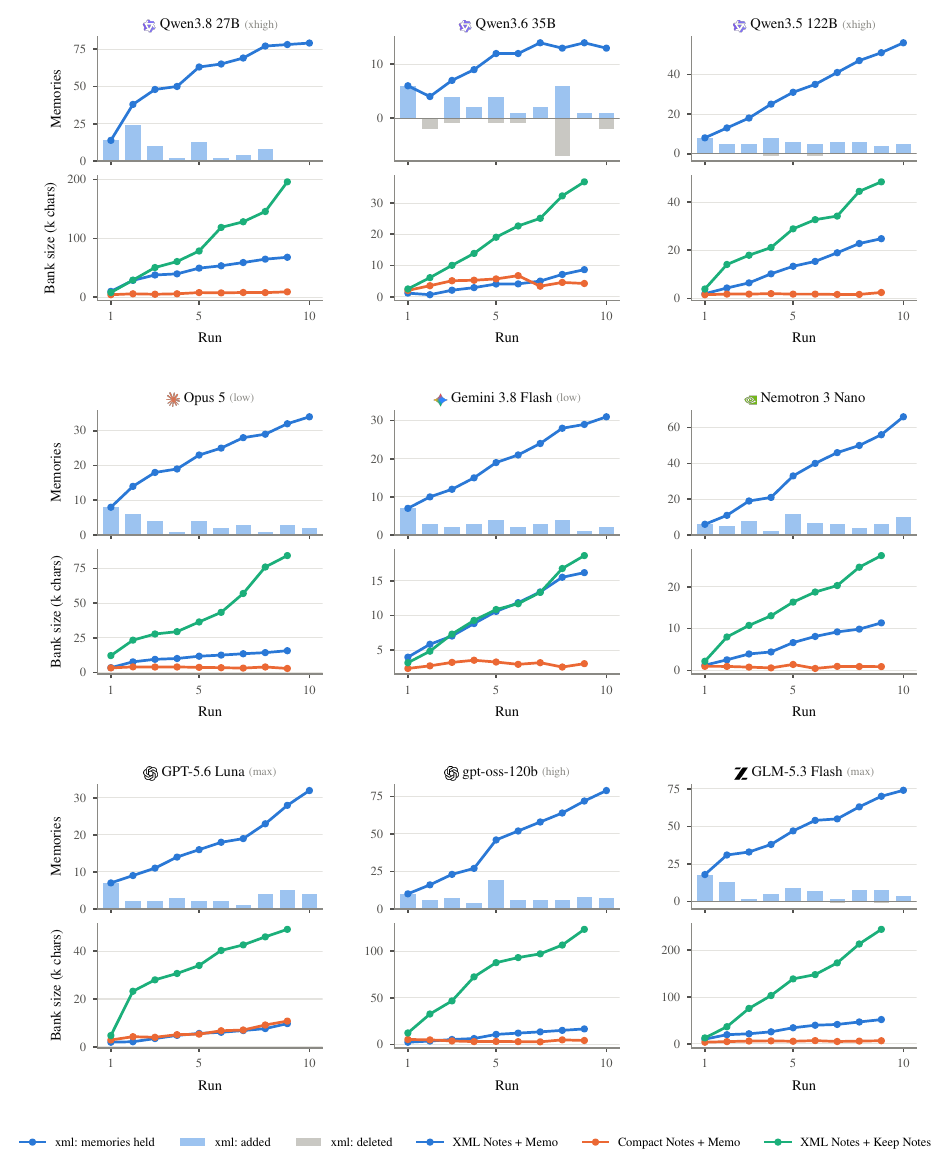}
\caption{\textbf{How the memory bank grows over a 10-run chain.} For each model, top: the number of xml memories held
after each run (line), with the memories added (blue, up) and deleted (grey, down) in that run's
consolidation. Bottom: the size of the bank carried into the next run, in thousands of characters, for
XML Notes + Memo, Compact Notes + Memo, and XML Notes + Keep Notes. Size is measured exactly, from the
system prompt of the following run, relative to the first run's empty bank; the bank after run 10 is never
read, so that curve ends at run 9.}
\label{fig:memory-growth}
\end{figure}

\providecommand{\spireicon}[1]{\raisebox{-0.25ex}{\includegraphics[height=1.7ex]{icons/#1.pdf}}}
\begin{table}[t]
\centering
\caption{\textbf{Effect of notes within a run and of memory across runs, per model.} Supplementary to \cref{tab:ablation-memory}. \emph{No notes or memory}: nothing is written within a run and no memory bank is kept. \emph{XML Notes}: keyed xml notes within each run, discarded at its end; \emph{Compact Notes}: a single compact summary, rewritten at the end of every floor and discarded at the end of the run. \emph{+ Keep Notes}: the xml notes are kept verbatim as the memory bank; \emph{+ Memo}: at the end of each run the model rewrites its notes into the bank (xml memories, or one compact summary of every run). All with the themed reskin and generic guidance. All cells are A20 on the held-out seeds, with each model at its main-table reasoning setting (grey). Act~1 is the Act~1 win rate (\%), Acts the mean number of acts cleared per run, and Floor the mean final floor. \best{Best} and \runnerup{second best} are per row, among cells with all 10 seeds. -- marks a No notes or memory cohort still being played (GLM-5.3 Flash).}
\label{tab:ablation-memory-by-model}
\begin{adjustbox}{max width=\linewidth}
\small
\begin{tabular}{@{}c@{\hspace{3pt}}l rrr rrr rrr rrr rrr rrr@{}}
\toprule
 &  & \multicolumn{3}{c}{Control} & \multicolumn{9}{c}{XML Notes} & \multicolumn{6}{c}{Compact Notes} \\
\cmidrule(lr){3-5} \cmidrule(lr){6-14} \cmidrule(lr){15-20}
  &   & \multicolumn{3}{c}{No notes or memory} & \multicolumn{3}{c}{XML Notes} & \multicolumn{3}{c}{XML Notes + Keep Notes} & \multicolumn{3}{c}{XML Notes + Memo} & \multicolumn{3}{c}{Compact Notes} & \multicolumn{3}{c}{Compact Notes + Memo} \\
\cmidrule(lr){3-5} \cmidrule(lr){6-8} \cmidrule(lr){9-11} \cmidrule(lr){12-14} \cmidrule(lr){15-17} \cmidrule(lr){18-20}
 & Agent & Act~1 & Acts & Floor & Act~1 & Acts & Floor & Act~1 & Acts & Floor & Act~1 & Acts & Floor & Act~1 & Acts & Floor & Act~1 & Acts & Floor \\
\midrule
\spireicon{qwen} & Qwen3.8 27B {\scriptsize\textcolor{gray}{(xhigh)}} & \runnerup{20} & \runnerup{0.20} & \best{16.0} & \best{30} & \best{0.30} & 13.6 & \best{30} & \best{0.30} & 14.2 & \runnerup{20} & \runnerup{0.20} & 14.9 & \runnerup{20} & \runnerup{0.20} & \runnerup{15.0} & 10 & 0.10 & 13.2 \\
\spireicon{qwen} & Qwen3.6 35B-A3B & 10 & 0.10 & 10.0 & 0 & 0.00 & 8.6 & 0 & 0.00 & \runnerup{10.1} & 0 & 0.00 & \best{10.2} & 0 & 0.00 & 8.0 & 0 & 0.00 & 8.4 \\
\spireicon{qwen} & Qwen3.5 122B-A10B {\scriptsize\textcolor{gray}{(xhigh)}} & 0 & 0.00 & \best{11.2} & 0 & 0.00 & 8.3 & 10 & 0.10 & 9.6 & 0 & 0.00 & \runnerup{9.8} & 0 & 0.00 & 6.8 & 0 & 0.00 & 7.5 \\
\spireicon{claude} & Claude Opus 5 {\scriptsize\textcolor{gray}{(low)}} & 30 & 0.30 & 16.5 & \best{70} & \best{0.90} & \best{24.9} & \runnerup{60} & \runnerup{0.70} & 22.2 & 40 & 0.40 & 17.1 & \runnerup{60} & 0.60 & 22.8 & 50 & 0.60 & \runnerup{23.9} \\
\spireicon{gemini} & Gemini 3.8 Flash {\scriptsize\textcolor{gray}{(low)}} & \runnerup{50} & 0.50 & 19.1 & \best{60} & \runnerup{0.60} & \runnerup{22.3} & 20 & 0.20 & 15.3 & \best{60} & \best{0.70} & \best{22.6} & 40 & 0.40 & 17.1 & 40 & 0.40 & 17.6 \\
\spireicon{nvidia} & Nemotron 3 Nano Omni & 0 & 0.00 & \best{9.4} & 10 & 0.10 & \runnerup{8.1} & 0 & 0.00 & 7.7 & 0 & 0.00 & 6.8 & 0 & 0.00 & \runnerup{8.1} & 0 & 0.00 & \runnerup{8.1} \\
\spireicon{openai} & GPT-5.6 Luna {\scriptsize\textcolor{gray}{(max)}} & \runnerup{30} & \runnerup{0.30} & 15.9 & 10 & 0.10 & 12.9 & 20 & \runnerup{0.30} & \runnerup{17.4} & 10 & 0.10 & 14.2 & 20 & 0.20 & 14.5 & \best{60} & \best{0.60} & \best{21.1} \\
\spireicon{openai} & gpt-oss-120b {\scriptsize\textcolor{gray}{(high)}} & 10 & 0.10 & \runnerup{13.5} & \runnerup{20} & \runnerup{0.20} & 12.0 & \best{40} & \best{0.40} & \best{16.3} & 10 & 0.10 & 10.4 & 10 & 0.10 & 11.4 & 0 & 0.00 & 11.8 \\
\spireicon{zai} & GLM-5.3 Flash {\scriptsize\textcolor{gray}{(max)}} & -- & -- & -- & 20 & 0.20 & 14.4 & \best{70} & \best{0.70} & \best{22.2} & 40 & 0.40 & 15.6 & \runnerup{50} & \runnerup{0.50} & \runnerup{17.6} & 40 & 0.40 & 15.9 \\
\midrule
 & Mean (8 agents) & 19 & 0.19 & \runnerup{13.9} & \best{25} & \best{0.28} & 13.8 & \runnerup{22} & \runnerup{0.25} & \best{14.1} & 18 & 0.19 & 13.2 & 19 & 0.19 & 13.0 & 20 & 0.21 & \runnerup{14.0} \\
\bottomrule
\end{tabular}
\end{adjustbox}
\end{table}

\FloatBarrier
\clearpage
\subsection{Knowledge reskin}
\label{app:ablation-reskin}
\begin{table}[th]
\centering
\caption{\textbf{Effect of the knowledge reskin.} \emph{None} shows the real game names and is the baseline; \emph{Themed} is our default Lich's Lair reskin; \emph{Random} replaces every name with pronounceable nonsense; \emph{Permuted} swaps each name for another real game name of the same kind; \emph{Perm.+themed} applies the same permutation in the themed vocabulary. All with xml notes, a read-write memory bank, and generic guidance. Pooled over models: each method is compared with its baseline only on matched runs (the same model on the same seed; Pairs, over Agents models), and the Act~1 win rate, mean acts cleared, and mean floor reached are over those runs. Usage is compared on like-for-like floors only: the same model, seed, act, and room type, and the same occurrence of that room type within the act, kept only where every method run on that model and seed completed it (Floors), so a method is neither credited with cheap early floors nor charged for expensive late ones. Model calls and k~tokens (input, including cached, plus output, in thousands) are means per floor; the token difference is the ratio of totals. Each $\Delta$ is the method minus its baseline. All runs are A20 on the held-out seeds, each model at its main-table reasoning setting. Models: Qwen3.6 35B-A3B, Qwen3.5 122B-A10B, Gemini 3.8 Flash (low), Nemotron 3 Nano Omni, GPT-5.6 Luna, gpt-oss-120b, GLM-5.3 Flash. Per-model results are in \cref{tab:ablation-reskin-by-model}.}
\label{tab:ablation-reskin}
\begin{adjustbox}{max width=\linewidth}
\small
\begin{tabular}{@{}l rr rr rr rr rr r rr rr@{}}
\toprule
 & & & \multicolumn{6}{c}{Performance (matched runs)} & \multicolumn{5}{c}{Usage per floor (matched floors)} \\
\cmidrule(lr){4-9} \cmidrule(lr){10-14}
 & & & \multicolumn{2}{c}{Act~1 win rate (\%)} & \multicolumn{2}{c}{Mean acts cleared} & \multicolumn{2}{c}{Mean floor reached} & & \multicolumn{2}{c}{Model calls} & \multicolumn{2}{c}{k tokens} \\
\cmidrule(lr){4-5} \cmidrule(lr){6-7} \cmidrule(lr){8-9} \cmidrule(lr){11-12} \cmidrule(lr){13-14}
Method & Agents & Pairs & Value & $\Delta$ & Value & $\Delta$ & Value & $\Delta$ & Floors & Value & $\Delta$ & Value & $\Delta$ \\
\midrule
None (baseline) & 7 & 70 & 26 & -- & 0.27 & -- & 14.0 & -- & 540 & 8.5 & -- & 209 & -- \\
Themed & 7 & 70 & 17 & \ddown{$-9$} & 0.19 & \ddown{$-0.09$} & 12.8 & \ddown{$-1.2$} & 540 & 8.7 & \ddown{$+0.2$} & 226 & \ddown{$+8$\%} \\
Random & 7 & 70 & 20 & \ddown{$-6$} & 0.24 & \ddown{$-0.03$} & 13.6 & \ddown{$-0.4$} & 540 & 8.8 & \ddown{$+0.2$} & 231 & \ddown{$+10$\%} \\
Permuted & 7 & 70 & 14 & \ddown{$-11$} & 0.16 & \ddown{$-0.11$} & 12.0 & \ddown{$-2.0$} & 540 & 8.8 & \ddown{$+0.3$} & 236 & \ddown{$+13$\%} \\
Perm.+themed & 7 & 70 & 13 & \ddown{$-13$} & 0.13 & \ddown{$-0.14$} & 11.9 & \ddown{$-2.1$} & 540 & 8.9 & \ddown{$+0.4$} & 227 & \ddown{$+8$\%} \\
\bottomrule
\end{tabular}
\end{adjustbox}
\end{table}

\Cref{tab:ablation-reskin} pools the five mappings over models (examples in \cref{tab:reskin-examples}),
and \cref{tab:ablation-reskin-by-model} gives them model by model. The themed reskin was run on all 12 model
configurations and the three control mappings on eight of them. So that every mapping is compared on the same
models, the pooled table and the last row of the per-model table cover the seven with all ten seeds in every
mapping (70 paired runs each); the themed reskin on all 11 models with ten seeds is in \cref{tab:reskin-compact}. %

In floors the mappings fall into the same two groups as in acts (\cref{sec:results}): against the real names, the
themed reskin costs 1.2 floors and the two permutations 2.0 and 2.1, and nonsense names 0.4 (real names 14.0
floors, nonsense 13.6, themed 12.8, and the two permutations 12.0 and 11.9). Every mapping uses more tokens per
floor than real names: 13\% more for the permuted names, 10\% for nonsense names, and 8\% for both the themed
reskin and the permutation in the themed vocabulary.

The cost of the themed reskin does not depend on the memory bank. With cross-run memory off (xml notes within a run,
no bank), on the seven models with all ten seeds under both names, the themed reskin lowers the Act~1 win rate from
20\% to 19\% and the mean floor from 13.1 to 12.3; on the same seven models with memory on, it lowers the Act~1 win
rate from 19\% to 14\% and leaves the mean floor at 12.7. The models it affects most do the same either way:
GPT-5.6 Luna loses 30 points of Act~1 clears and 3.3 floors without the bank and 50 points and 4.3 floors with it, and
Gemini~3.8 Flash at low effort gains 30 points and 6.3 floors without it and 20 points and 2.4 floors with it.

Model by model, the cost of the themed reskin falls on the models that do well with the real names. Every model that
clears 0.6 acts or more with real names reaches fewer floors under the themed reskin, by 2.3 (GLM-5.3) to 7.5
(GLM-5.3 Flash), or 8.1 for Gemini~3.8 Flash at high effort, of whose real-name seeds five finished. Of the seven
that clear at most 0.5 acts with real names, six reach as many floors or more under the themed reskin, and only
Qwen3.5 122B-A10B reaches fewer (1.1 floors). Gemini~3.8 Flash at low effort shows the split between the mappings
most clearly: it does best with nonsense names (0.80 acts and 24.1 floors, against 0.50 and 20.2 with real names)
and worst under the two permutations (17.6 and 16.4 floors), so what misleads it is a name that means something
else, not a name it does not know. GLM-5.3 Flash is the exception, losing 5 to 8 floors under every mapping,
nonsense included.

\providecommand{\spireicon}[1]{\raisebox{-0.25ex}{\includegraphics[height=1.7ex]{icons/#1.pdf}}}
\begin{table}[th]
\centering
\caption{\textbf{Effect of the knowledge reskin, per model.} Supplementary to \cref{tab:ablation-reskin}. \emph{None} shows the real game names; \emph{Themed} is our default Lich's Lair reskin; \emph{Random} replaces every name with pronounceable nonsense; \emph{Permuted} swaps each name for another real game name of the same kind; \emph{Perm.+themed} applies the same permutation in the themed vocabulary. All with xml notes, a read-write memory bank, and generic guidance. Gemini 3.8 Flash (high) is shown on the 5 seeds its arms share, so its cells compare the same games. All cells are A20 on the held-out seeds, with each model at its main-table reasoning setting (grey). Act~1 is the Act~1 win rate (\%), Acts the mean number of acts cleared per run, and Floor the mean final floor. A superscript gives the seed count where fewer than 10 seeds completed. \best{Best} and \runnerup{second best} are per row, among cells with all 10 seeds.}
\label{tab:ablation-reskin-by-model}
\begin{adjustbox}{max width=\linewidth}
\small
\begin{tabular}{@{}c@{\hspace{3pt}}l rrr rrr rrr rrr rrr@{}}
\toprule
  &   & \multicolumn{3}{c}{None} & \multicolumn{3}{c}{Themed} & \multicolumn{3}{c}{Random} & \multicolumn{3}{c}{Permuted} & \multicolumn{3}{c}{Perm.+themed} \\
\cmidrule(lr){3-5} \cmidrule(lr){6-8} \cmidrule(lr){9-11} \cmidrule(lr){12-14} \cmidrule(lr){15-17}
 & Agent & Act~1 & Acts & Floor & Act~1 & Acts & Floor & Act~1 & Acts & Floor & Act~1 & Acts & Floor & Act~1 & Acts & Floor \\
\midrule
\spireicon{qwen} & Qwen3.8 27B {\scriptsize\textcolor{gray}{(xhigh)}} & \runnerup{20} & \runnerup{0.20} & 14.0 & \runnerup{20} & \runnerup{0.20} & \runnerup{14.9} & \best{30} & \best{0.30} & \best{15.9} & \runnerup{20} & \runnerup{0.20} & 14.7 & 43\rlap{$^{7}$} & 0.43\rlap{$^{7}$} & 17.9\rlap{$^{7}$} \\
\spireicon{qwen} & Qwen3.6 35B-A3B & 0 & 0.00 & 9.7 & 0 & 0.00 & \runnerup{10.2} & 0 & 0.00 & 9.5 & 0 & 0.00 & 9.5 & 0 & 0.00 & \best{11.0} \\
\spireicon{qwen} & Qwen3.5 122B-A10B {\scriptsize\textcolor{gray}{(xhigh)}} & 10 & 0.10 & \best{10.9} & 0 & 0.00 & \runnerup{9.8} & 0 & 0.00 & 7.6 & 0 & 0.00 & 9.2 & 0 & 0.00 & 8.0 \\
\spireicon{claude} & Claude Haiku 4.5 {\scriptsize\textcolor{gray}{(12k)}} & \best{10} & \best{0.10} & \runnerup{11.4} & \best{10} & \best{0.10} & \best{11.9} & -- & -- & -- & -- & -- & -- & -- & -- & -- \\
\spireicon{deepseek} & DeepSeek V4.1 Flash {\scriptsize\textcolor{gray}{(max)}} & \best{70} & \best{0.80} & \best{22.0} & \runnerup{30} & \runnerup{0.30} & \runnerup{16.9} & -- & -- & -- & -- & -- & -- & -- & -- & -- \\
\spireicon{gemini} & Gemini 3.8 Flash {\scriptsize\textcolor{gray}{(high)}} & 80\rlap{$^{5}$} & 1.20\rlap{$^{5}$} & 31.0\rlap{$^{5}$} & 20\rlap{$^{5}$} & 0.40\rlap{$^{5}$} & 22.2\rlap{$^{5}$} & -- & -- & -- & -- & -- & -- & -- & -- & -- \\
\spireicon{gemini} & Gemini 3.8 Flash {\scriptsize\textcolor{gray}{(low)}} & \runnerup{40} & 0.50 & 20.2 & \best{60} & \runnerup{0.70} & \runnerup{22.6} & \best{60} & \best{0.80} & \best{24.1} & 30 & 0.40 & 17.6 & 30 & 0.30 & 16.4 \\
\spireicon{nvidia} & Nemotron 3 Nano Omni & 0 & 0.00 & 5.3 & 0 & 0.00 & 6.8 & 0 & 0.00 & \best{8.3} & 0 & 0.00 & \runnerup{7.4} & 0 & 0.00 & 7.0 \\
\spireicon{openai} & GPT-5.6 Luna {\scriptsize\textcolor{gray}{(max)}} & \best{60} & \best{0.60} & \runnerup{18.5} & 10 & 0.10 & 14.2 & \runnerup{50} & \best{0.60} & \best{19.6} & 40 & \runnerup{0.40} & 15.0 & 10 & 0.10 & 12.9 \\
\spireicon{openai} & gpt-oss-120b {\scriptsize\textcolor{gray}{(high)}} & 0 & 0.00 & \runnerup{10.4} & 10 & 0.10 & \runnerup{10.4} & 0 & 0.00 & \best{10.7} & 0 & 0.00 & 9.2 & 0 & 0.00 & 9.8 \\
\spireicon{zai} & GLM-5.3 {\scriptsize\textcolor{gray}{(max)}} & \best{60} & \best{0.70} & \best{24.4} & \best{60} & \runnerup{0.60} & \runnerup{22.1} & -- & -- & -- & -- & -- & -- & -- & -- & -- \\
\spireicon{zai} & GLM-5.3 Flash {\scriptsize\textcolor{gray}{(max)}} & \best{70} & \best{0.70} & \best{23.1} & 40 & 0.40 & 15.6 & 30 & 0.30 & 15.3 & 30 & 0.30 & 16.1 & \runnerup{50} & \runnerup{0.50} & \runnerup{18.0} \\
\midrule
 & Mean (7 agents) & \best{26} & \best{0.27} & \best{14.0} & 17 & 0.19 & 12.8 & \runnerup{20} & \runnerup{0.24} & \runnerup{13.6} & 14 & 0.16 & 12.0 & 13 & 0.13 & 11.9 \\
\bottomrule
\end{tabular}
\end{adjustbox}
\end{table}

\FloatBarrier
\subsection[\texorpdfstring{Knowledge given to the model}{Knowledge given to the model}]{Knowledge given to the model}
\label{app:ablation-prompt}
\FloatBarrier
\begin{table}[tbh]
\centering
\caption{\textbf{Effect of changing the knowledge base given to the model.} \emph{Reskin} is the themed reskin with nothing added and is the baseline; \emph{Real names} shows the real game names instead; \emph{No compendium} removes the card, potion, and relic compendium from the reskin; \emph{Bestiary} adds the enemy compendium and \emph{Reddit guide} the Reddit high-ascension strategy guide, each renamed by the reskin (\cref{app:prompt-material}); \emph{Own} adds a read-only copy of the memory bank the model's own 10-seed chain ended with, and \emph{Fable 5.1} and \emph{GPT-6 Astra} one of the bank that model's chain in \cref{tab:main-results} ended with. All with xml notes and generic guidance, and no run writes a memory bank. Pooled over models: each method is compared with its baseline only on matched runs (the same model on the same seed; Pairs, over Agents models), and the Act~1 win rate, mean acts cleared, and mean floor reached are over those runs. Usage is compared on like-for-like floors only: the same model, seed, act, and room type, and the same occurrence of that room type within the act, kept only where every method run on that model and seed completed it (Floors), so a method is neither credited with cheap early floors nor charged for expensive late ones. Model calls and k~thinking tokens (the reasoning the model generated, in thousands; where a provider reports no separate count, estimated from the reasoning text's share of the output) are means per floor; the thinking difference is the ratio of totals. Each $\Delta$ is the method minus its baseline. All runs are A20 on the 10 evaluation seeds, each model at its main-table reasoning setting. *: $p\!<\!0.05$, by a two-sided exact sign-flip test over the agents' mean paired differences (for thinking, over their log ratios of thinking, method over baseline).
Per-model results are in \cref{tab:ablation-prompt-by-model}.}
\label{tab:ablation-prompt}
\begin{adjustbox}{max width=\linewidth}
\small
\begin{tabular}{@{}l rr rr rr rr rr r rr rr@{}}
\toprule
 & & & \multicolumn{6}{c}{Performance (matched runs)} & \multicolumn{5}{c}{Usage per floor (matched floors)} \\
\cmidrule(lr){4-9} \cmidrule(lr){10-14}
 & & & \multicolumn{2}{c}{Act~1 (\%)} & \multicolumn{2}{c}{Mean acts cleared} & \multicolumn{2}{c}{Mean floor} & & \multicolumn{2}{c}{Model calls} & \multicolumn{2}{c}{k thinking tokens} \\
\cmidrule(lr){4-5} \cmidrule(lr){6-7} \cmidrule(lr){8-9} \cmidrule(lr){11-12} \cmidrule(lr){13-14}
Method & Agents & Pairs & Value & $\Delta$ & Value & $\Delta$ & Value & $\Delta$ & Floors & Value & $\Delta$ & Value & $\Delta$ \\
\midrule
\multicolumn{14}{l}{\textit{Game knowledge}} \\
Reskin (baseline) & 8 & 80 & 26 & -- & 0.26 & -- & 15.3 & -- & 563 & 7.5 & -- & 6.8 & -- \\
Real names & 8 & 80 & 29 & \dup{$+2$} & 0.38 & \dup{$+0.11$} & 16.1 & \dup{$+0.8$} & 563 & 7.6 & \ddown{$+0.1$} & 7.8 & \dup{$+15$\%} \\
No compendium & 8 & 80 & 19 & \ddown{$-8$} & 0.19 & \ddown{$-0.07$} & 14.7 & \ddown{$-0.7$} & 563 & 7.5 & \ddown{$+0.1$} & 6.6 & \ddown{$-2$\%} \\
Bestiary & 8 & 80 & 22 & \ddown{$-4$} & 0.24 & \ddown{$-0.03$} & 15.3 & \dzero{$\pm0.0$} & 563 & 7.5 & \ddown{$+0.1$} & 7.3 & \dup{$+8$\%} \\
Reddit guide & 8 & 80 & 18 & \ddown{$-9$} & 0.21 & \ddown{$-0.05$} & 13.4 & \ddown{$-1.9$}\rlap{*} & 563 & 7.3 & \dup{$-0.2$} & 6.5 & \ddown{$-4$\%} \\
\midrule
\multicolumn{14}{l}{\textit{Frozen memory bank}} \\
Own & 8 & 80 & 35 & \dup{$+9$} & 0.38 & \dup{$+0.11$} & 17.4 & \dup{$+2.1$} & 563 & 7.5 & \dzero{$\pm0.0$} & 7.8 & \dup{$+16$\%} \\
Fable 5.1 & 8 & 80 & 32 & \dup{$+6$} & 0.34 & \dup{$+0.07$} & 17.4 & \dup{$+2.0$} & 563 & 7.5 & \ddown{$+0.1$} & 8.2 & \dup{$+22$\%} \\
GPT-6 Astra & 8 & 80 & 30 & \dup{$+4$} & 0.34 & \dup{$+0.07$} & 17.0 & \dup{$+1.6$} & 563 & 7.5 & \dzero{$\pm0.0$} & 8.3 & \dup{$+23$\%} \\
\bottomrule
\end{tabular}
\end{adjustbox}
\end{table}

\Cref{tab:ablation-prompt} pools the changes to the knowledge given to the model over models (described in
\cref{app:prompt-material}), and \cref{tab:ablation-prompt-by-model,tab:ablation-prompt-by-model-floor,tab:ablation-prompt-by-model-thinking} give it model by model, in acts, in floors, and in thinking per floor.

\providecommand{\spireicon}[1]{\raisebox{-0.25ex}{\includegraphics[height=1.7ex]{icons/#1.pdf}}}
\begin{table}[tbh]
\centering
\caption{\textbf{Effect of changing the knowledge base given to the model, per model.} On the 10 evaluation seeds (disjoint from those of \cref{tab:main-results}): mean acts cleared per run (\cref{tab:ablation-prompt-by-model-floor} gives the mean final floor). The baseline is our themed reskin (\emph{Reskin}); \emph{Real names} shows the real game names instead, \emph{No compendium} removes the card, potion, and relic compendium from the reskin, and each other arm adds one source of knowledge under the reskin: the enemy compendium (\emph{Bestiary}), the Reddit high-ascension strategy guide (\cref{app:prompt-material}), or a read-only copy of the memory bank that the model's own 10-seed chain ended with (its chain in \cref{tab:main-results}, or its low-effort chain for the entries marked low) or that Claude Fable~5.1's or GPT-6 Astra's chain in \cref{tab:main-results} ended with, each renamed by the reskin. All arms take xml notes within a run, write no memory bank, and use generic guidance, with each model at its main-table reasoning setting (grey). Beside each value is its change against the Reskin baseline, \textcolor{ForestGreen}{green} where it rises and \textcolor{BrickRed}{red} where it falls, shown where both cells have all 10 seeds. \best{Best} and \runnerup{second best} are per row, among cells with all 10 seeds. Pooled over models in \cref{tab:ablation-prompt}.}
\label{tab:ablation-prompt-by-model}
\begin{adjustbox}{max width=\linewidth}
\small
\begin{tabular}{@{}c@{\hspace{3pt}}l r r@{}l r@{}l r@{}l r@{}l r@{}l r@{}l r@{}l@{}}
\toprule
 &  & \multicolumn{1}{c}{Baseline} & \multicolumn{8}{c}{Game knowledge} & \multicolumn{6}{c}{Frozen memory bank} \\
\cmidrule(lr){3-3} \cmidrule(lr){4-11} \cmidrule(lr){12-17}
 & Agent & Reskin & \multicolumn{2}{c}{Real names} & \multicolumn{2}{c}{No compendium} & \multicolumn{2}{c}{Bestiary} & \multicolumn{2}{c}{Reddit guide} & \multicolumn{2}{c}{Own} & \multicolumn{2}{c}{Fable 5.1} & \multicolumn{2}{c}{GPT-6 Astra} \\
\midrule
\spireicon{qwen} & Qwen3.8 27B {\scriptsize\textcolor{gray}{(xhigh)}} & 0.30 & 0.30 & \,{\scriptsize\dzero{$\pm0.00$}} & 0.20 & \,{\scriptsize\ddown{$-0.10$}} & \runnerup{0.50} & \,{\scriptsize\dup{$+0.20$}} & 0.10 & \,{\scriptsize\ddown{$-0.20$}} & \best{0.70} & \,{\scriptsize\dup{$+0.40$}} & \runnerup{0.50} & \,{\scriptsize\dup{$+0.20$}} & 0.40 & \,{\scriptsize\dup{$+0.10$}} \\
\spireicon{qwen} & Qwen3.6 35B-A3B & 0.00 & 0.00 & \,{\scriptsize\dzero{$\pm0.00$}} & 0.00 & \,{\scriptsize\dzero{$\pm0.00$}} & 0.00 & \,{\scriptsize\dzero{$\pm0.00$}} & 0.10 & \,{\scriptsize\dup{$+0.10$}} & 0.00 & \,{\scriptsize\dzero{$\pm0.00$}} & 0.00 & \,{\scriptsize\dzero{$\pm0.00$}} & 0.00 & \,{\scriptsize\dzero{$\pm0.00$}} \\
\spireicon{claude} & Claude Opus 5 {\scriptsize\textcolor{gray}{(low)}} & 0.30 & \runnerup{0.60} & \,{\scriptsize\dup{$+0.30$}} & 0.40 & \,{\scriptsize\dup{$+0.10$}} & 0.50 & \,{\scriptsize\dup{$+0.20$}} & 0.30 & \,{\scriptsize\dzero{$\pm0.00$}} & \runnerup{0.60} & \,{\scriptsize\dup{$+0.30$}} & \best{0.80} & \,{\scriptsize\dup{$+0.50$}} & \best{0.80} & \,{\scriptsize\dup{$+0.50$}} \\
\spireicon{gemini} & Gemini 3.8 Flash {\scriptsize\textcolor{gray}{(low)}} & \runnerup{0.90} & \best{1.30} & \,{\scriptsize\dup{$+0.40$}} & 0.10 & \,{\scriptsize\ddown{$-0.80$}} & 0.50 & \,{\scriptsize\ddown{$-0.40$}} & 0.80 & \,{\scriptsize\ddown{$-0.10$}} & 0.70 & \,{\scriptsize\ddown{$-0.20$}} & 0.70 & \,{\scriptsize\ddown{$-0.20$}} & 0.80 & \,{\scriptsize\ddown{$-0.10$}} \\
\spireicon{nvidia} & Nemotron 3 Nano Omni & 0.00 & 0.00 & \,{\scriptsize\dzero{$\pm0.00$}} & 0.00 & \,{\scriptsize\dzero{$\pm0.00$}} & 0.00 & \,{\scriptsize\dzero{$\pm0.00$}} & 0.00 & \,{\scriptsize\dzero{$\pm0.00$}} & 0.00 & \,{\scriptsize\dzero{$\pm0.00$}} & 0.00 & \,{\scriptsize\dzero{$\pm0.00$}} & 0.00 & \,{\scriptsize\dzero{$\pm0.00$}} \\
\spireicon{openai} & GPT-5.6 Luna {\scriptsize\textcolor{gray}{(max)}} & 0.10 & \runnerup{0.30} & \,{\scriptsize\dup{$+0.20$}} & \best{0.50} & \,{\scriptsize\dup{$+0.40$}} & 0.10 & \,{\scriptsize\dzero{$\pm0.00$}} & \runnerup{0.30} & \,{\scriptsize\dup{$+0.20$}} & \runnerup{0.30} & \,{\scriptsize\dup{$+0.20$}} & \runnerup{0.30} & \,{\scriptsize\dup{$+0.20$}} & 0.10 & \,{\scriptsize\dzero{$\pm0.00$}} \\
\spireicon{openai} & gpt-oss-120b {\scriptsize\textcolor{gray}{(high)}} & \runnerup{0.10} & \best{0.20} & \,{\scriptsize\dup{$+0.10$}} & 0.00 & \,{\scriptsize\ddown{$-0.10$}} & 0.00 & \,{\scriptsize\ddown{$-0.10$}} & 0.00 & \,{\scriptsize\ddown{$-0.10$}} & 0.00 & \,{\scriptsize\ddown{$-0.10$}} & 0.00 & \,{\scriptsize\ddown{$-0.10$}} & 0.00 & \,{\scriptsize\ddown{$-0.10$}} \\
\spireicon{zai} & GLM-5.3 Flash {\scriptsize\textcolor{gray}{(max)}} & 0.40 & 0.30 & \,{\scriptsize\ddown{$-0.10$}} & 0.30 & \,{\scriptsize\ddown{$-0.10$}} & 0.30 & \,{\scriptsize\ddown{$-0.10$}} & 0.10 & \,{\scriptsize\ddown{$-0.30$}} & \best{0.70} & \,{\scriptsize\dup{$+0.30$}} & 0.40 & \,{\scriptsize\dzero{$\pm0.00$}} & \runnerup{0.60} & \,{\scriptsize\dup{$+0.20$}} \\
\midrule
 & Mean (8 agents) & 0.26 & \best{0.38} & \,{\scriptsize\dup{$+0.11$}} & 0.19 & \,{\scriptsize\ddown{$-0.08$}} & 0.24 & \,{\scriptsize\ddown{$-0.03$}} & 0.21 & \,{\scriptsize\ddown{$-0.05$}} & \best{0.38} & \,{\scriptsize\dup{$+0.11$}} & \runnerup{0.34} & \,{\scriptsize\dup{$+0.08$}} & \runnerup{0.34} & \,{\scriptsize\dup{$+0.08$}} \\
\bottomrule
\end{tabular}
\end{adjustbox}
\end{table}

\providecommand{\spireicon}[1]{\raisebox{-0.25ex}{\includegraphics[height=1.7ex]{icons/#1.pdf}}}
\begin{table}[tbh]
\centering
\caption{\textbf{Effect of changing the knowledge base given to the model on the final floor, per model.} On the 10 evaluation seeds: mean final floor. Arms, changes against the Reskin baseline, and marks as in \cref{tab:ablation-prompt-by-model}.}
\label{tab:ablation-prompt-by-model-floor}
\begin{adjustbox}{max width=\linewidth}
\small
\begin{tabular}{@{}c@{\hspace{3pt}}l r r@{}l r@{}l r@{}l r@{}l r@{}l r@{}l r@{}l@{}}
\toprule
 &  & \multicolumn{1}{c}{Baseline} & \multicolumn{8}{c}{Game knowledge} & \multicolumn{6}{c}{Frozen memory bank} \\
\cmidrule(lr){3-3} \cmidrule(lr){4-11} \cmidrule(lr){12-17}
 & Agent & Reskin & \multicolumn{2}{c}{Real names} & \multicolumn{2}{c}{No compendium} & \multicolumn{2}{c}{Bestiary} & \multicolumn{2}{c}{Reddit guide} & \multicolumn{2}{c}{Own} & \multicolumn{2}{c}{Fable 5.1} & \multicolumn{2}{c}{GPT-6 Astra} \\
\midrule
\spireicon{qwen} & Qwen3.8 27B {\scriptsize\textcolor{gray}{(xhigh)}} & 15.1 & 16.4 & \,{\scriptsize\dup{$+1.3$}} & 16.4 & \,{\scriptsize\dup{$+1.3$}} & 19.2 & \,{\scriptsize\dup{$+4.1$}} & 13.3 & \,{\scriptsize\ddown{$-1.8$}} & \best{21.0} & \,{\scriptsize\dup{$+5.9$}} & \runnerup{20.0} & \,{\scriptsize\dup{$+4.9$}} & 19.0 & \,{\scriptsize\dup{$+3.9$}} \\
\spireicon{qwen} & Qwen3.6 35B-A3B & 11.0 & 10.7 & \,{\scriptsize\ddown{$-0.3$}} & 11.7 & \,{\scriptsize\dup{$+0.7$}} & 10.6 & \,{\scriptsize\ddown{$-0.4$}} & 9.6 & \,{\scriptsize\ddown{$-1.4$}} & \runnerup{12.0} & \,{\scriptsize\dup{$+1.0$}} & \best{13.1} & \,{\scriptsize\dup{$+2.1$}} & 10.1 & \,{\scriptsize\ddown{$-0.9$}} \\
\spireicon{claude} & Claude Opus 5 {\scriptsize\textcolor{gray}{(low)}} & 19.1 & 22.0 & \,{\scriptsize\dup{$+2.9$}} & 20.7 & \,{\scriptsize\dup{$+1.6$}} & 21.2 & \,{\scriptsize\dup{$+2.1$}} & 16.8 & \,{\scriptsize\ddown{$-2.3$}} & 22.9 & \,{\scriptsize\dup{$+3.8$}} & \best{23.9} & \,{\scriptsize\dup{$+4.8$}} & \runnerup{23.6} & \,{\scriptsize\dup{$+4.5$}} \\
\spireicon{gemini} & Gemini 3.8 Flash {\scriptsize\textcolor{gray}{(low)}} & \runnerup{26.2} & \best{30.1} & \,{\scriptsize\dup{$+3.9$}} & 14.0 & \,{\scriptsize\ddown{$-12.2$}} & 21.3 & \,{\scriptsize\ddown{$-4.9$}} & 22.9 & \,{\scriptsize\ddown{$-3.3$}} & 24.8 & \,{\scriptsize\ddown{$-1.4$}} & 23.9 & \,{\scriptsize\ddown{$-2.3$}} & 25.5 & \,{\scriptsize\ddown{$-0.7$}} \\
\spireicon{nvidia} & Nemotron 3 Nano Omni & 6.3 & 6.5 & \,{\scriptsize\dup{$+0.2$}} & 7.1 & \,{\scriptsize\dup{$+0.8$}} & 5.8 & \,{\scriptsize\ddown{$-0.5$}} & 5.7 & \,{\scriptsize\ddown{$-0.6$}} & \runnerup{7.5} & \,{\scriptsize\dup{$+1.2$}} & 6.0 & \,{\scriptsize\ddown{$-0.3$}} & \best{9.4} & \,{\scriptsize\dup{$+3.1$}} \\
\spireicon{openai} & GPT-5.6 Luna {\scriptsize\textcolor{gray}{(max)}} & 15.2 & 13.9 & \,{\scriptsize\ddown{$-1.3$}} & \runnerup{17.9} & \,{\scriptsize\dup{$+2.7$}} & 15.2 & \,{\scriptsize\dzero{$\pm0.0$}} & 15.6 & \,{\scriptsize\dup{$+0.4$}} & 17.6 & \,{\scriptsize\dup{$+2.4$}} & \best{19.0} & \,{\scriptsize\dup{$+3.8$}} & 15.2 & \,{\scriptsize\dzero{$\pm0.0$}} \\
\spireicon{openai} & gpt-oss-120b {\scriptsize\textcolor{gray}{(high)}} & \runnerup{12.4} & \best{13.1} & \,{\scriptsize\dup{$+0.7$}} & 11.7 & \,{\scriptsize\ddown{$-0.7$}} & 11.7 & \,{\scriptsize\ddown{$-0.7$}} & 10.8 & \,{\scriptsize\ddown{$-1.6$}} & 11.7 & \,{\scriptsize\ddown{$-0.7$}} & 11.9 & \,{\scriptsize\ddown{$-0.5$}} & 11.7 & \,{\scriptsize\ddown{$-0.7$}} \\
\spireicon{zai} & GLM-5.3 Flash {\scriptsize\textcolor{gray}{(max)}} & 17.5 & 16.4 & \,{\scriptsize\ddown{$-1.1$}} & 17.7 & \,{\scriptsize\dup{$+0.2$}} & 17.7 & \,{\scriptsize\dup{$+0.2$}} & 12.7 & \,{\scriptsize\ddown{$-4.8$}} & \best{22.1} & \,{\scriptsize\dup{$+4.6$}} & 21.0 & \,{\scriptsize\dup{$+3.5$}} & \runnerup{21.2} & \,{\scriptsize\dup{$+3.7$}} \\
\midrule
 & Mean (8 agents) & 15.4 & 16.1 & \,{\scriptsize\dup{$+0.8$}} & 14.6 & \,{\scriptsize\ddown{$-0.7$}} & 15.3 & \,{\scriptsize\dzero{$\pm0.0$}} & 13.4 & \,{\scriptsize\ddown{$-1.9$}} & \best{17.4} & \,{\scriptsize\dup{$+2.1$}} & \best{17.4} & \,{\scriptsize\dup{$+2.0$}} & \runnerup{17.0} & \,{\scriptsize\dup{$+1.6$}} \\
\bottomrule
\end{tabular}
\end{adjustbox}
\end{table}

\providecommand{\spireicon}[1]{\raisebox{-0.25ex}{\includegraphics[height=1.7ex]{icons/#1.pdf}}}
\begin{table}[tbh]
\centering
\caption{\textbf{Effect of changing the knowledge base given to the model on thinking, per model.} On the 10 evaluation seeds: thousands of thinking tokens per floor (the reasoning the model generated; where a provider reports no separate count, estimated from the reasoning text's share of the output), on the matched floors of \cref{tab:ablation-prompt}: for each model and seed, the floors that every arm completed. Beside each value is its change against the Reskin baseline as a percentage of the model's Reskin thinking, \textcolor{ForestGreen}{green} where it rises and \textcolor{BrickRed}{red} where it falls; it is shown where both cells have all 10 seeds. Arms as in \cref{tab:ablation-prompt-by-model}.}
\label{tab:ablation-prompt-by-model-thinking}
\begin{adjustbox}{max width=\linewidth}
\small
\begin{tabular}{@{}c@{\hspace{3pt}}l r r@{}r r@{}r r@{}r r@{}r r@{}r r@{}r r@{}r@{}}
\toprule
 &  & \multicolumn{1}{c}{Baseline} & \multicolumn{8}{c}{Game knowledge} & \multicolumn{6}{c}{Frozen memory bank} \\
\cmidrule(lr){3-3} \cmidrule(lr){4-11} \cmidrule(lr){12-17}
 & Agent & Reskin & \multicolumn{2}{c}{Real names} & \multicolumn{2}{c}{No compendium} & \multicolumn{2}{c}{Bestiary} & \multicolumn{2}{c}{Reddit guide} & \multicolumn{2}{c}{Own} & \multicolumn{2}{c}{Fable 5.1} & \multicolumn{2}{c}{GPT-6 Astra} \\
\midrule
\spireicon{qwen} & Qwen3.8 27B {\scriptsize\textcolor{gray}{(xhigh)}} & 7.2 & 9.0 & \,{\scriptsize \dup{$+25$\%}} & 8.6 & \,{\scriptsize \dup{$+20$\%}} & 7.6 & \,{\scriptsize \dup{$+6$\%}} & 8.2 & \,{\scriptsize \dup{$+13$\%}} & 11.2 & \,{\scriptsize \dup{$+56$\%}} & 13.6 & \,{\scriptsize \dup{$+89$\%}} & 10.7 & \,{\scriptsize \dup{$+49$\%}} \\
\spireicon{qwen} & Qwen3.6 35B-A3B & 5.0 & 4.0 & \,{\scriptsize \ddown{$-19$\%}} & 4.5 & \,{\scriptsize \ddown{$-10$\%}} & 4.4 & \,{\scriptsize \ddown{$-12$\%}} & 5.2 & \,{\scriptsize \dup{$+4$\%}} & 3.7 & \,{\scriptsize \ddown{$-26$\%}} & 4.6 & \,{\scriptsize \ddown{$-8$\%}} & 5.6 & \,{\scriptsize \dup{$+12$\%}} \\
\spireicon{claude} & Claude Opus 5 {\scriptsize\textcolor{gray}{(low)}} & 0.6 & 0.7 & \,{\scriptsize \dup{$+10$\%}} & 0.6 & \,{\scriptsize \dzero{$\pm0$\%}} & 0.7 & \,{\scriptsize \dup{$+7$\%}} & 0.7 & \,{\scriptsize \dup{$+5$\%}} & 0.6 & \,{\scriptsize \ddown{$-1$\%}} & 0.7 & \,{\scriptsize \dup{$+19$\%}} & 0.7 & \,{\scriptsize \dup{$+19$\%}} \\
\spireicon{gemini} & Gemini 3.8 Flash {\scriptsize\textcolor{gray}{(low)}} & 2.6 & 2.6 & \,{\scriptsize \dzero{$\pm0$\%}} & 0.0 & \,{\scriptsize \ddown{$-100$\%}} & 2.6 & \,{\scriptsize \dup{$+1$\%}} & 2.6 & \,{\scriptsize \dup{$+4$\%}} & 1.9 & \,{\scriptsize \ddown{$-27$\%}} & 2.1 & \,{\scriptsize \ddown{$-17$\%}} & 3.1 & \,{\scriptsize \dup{$+21$\%}} \\
\spireicon{nvidia} & Nemotron 3 Nano Omni & 28.5 & 25.0 & \,{\scriptsize \ddown{$-12$\%}} & 18.6 & \,{\scriptsize \ddown{$-35$\%}} & 31.2 & \,{\scriptsize \dup{$+9$\%}} & 21.1 & \,{\scriptsize \ddown{$-26$\%}} & 25.7 & \,{\scriptsize \ddown{$-10$\%}} & 21.4 & \,{\scriptsize \ddown{$-25$\%}} & 23.9 & \,{\scriptsize \ddown{$-16$\%}} \\
\spireicon{openai} & GPT-5.6 Luna {\scriptsize\textcolor{gray}{(max)}} & 3.2 & 3.4 & \,{\scriptsize \dup{$+6$\%}} & 3.0 & \,{\scriptsize \ddown{$-6$\%}} & 3.6 & \,{\scriptsize \dup{$+13$\%}} & 3.0 & \,{\scriptsize \ddown{$-5$\%}} & 3.7 & \,{\scriptsize \dup{$+14$\%}} & 3.9 & \,{\scriptsize \dup{$+22$\%}} & 3.9 & \,{\scriptsize \dup{$+22$\%}} \\
\spireicon{openai} & gpt-oss-120b {\scriptsize\textcolor{gray}{(high)}} & 16.7 & 16.4 & \,{\scriptsize \ddown{$-2$\%}} & 17.5 & \,{\scriptsize \dup{$+4$\%}} & 14.3 & \,{\scriptsize \ddown{$-15$\%}} & 14.7 & \,{\scriptsize \ddown{$-12$\%}} & 14.4 & \,{\scriptsize \ddown{$-14$\%}} & 9.4 & \,{\scriptsize \ddown{$-44$\%}} & 12.1 & \,{\scriptsize \ddown{$-28$\%}} \\
\spireicon{zai} & GLM-5.3 Flash {\scriptsize\textcolor{gray}{(max)}} & 7.3 & 15.0 & \,{\scriptsize \dup{$+104$\%}} & 11.7 & \,{\scriptsize \dup{$+60$\%}} & 11.8 & \,{\scriptsize \dup{$+61$\%}} & 8.7 & \,{\scriptsize \dup{$+19$\%}} & 14.9 & \,{\scriptsize \dup{$+103$\%}} & 19.8 & \,{\scriptsize \dup{$+170$\%}} & 18.7 & \,{\scriptsize \dup{$+156$\%}} \\
\midrule
 & Mean (8 agents) & 8.9 & 9.5 & \,{\scriptsize \dup{$+7$\%}} & 8.1 & \,{\scriptsize \ddown{$-9$\%}} & 9.5 & \,{\scriptsize \dup{$+7$\%}} & 8.0 & \,{\scriptsize \ddown{$-10$\%}} & 9.5 & \,{\scriptsize \dup{$+7$\%}} & 9.4 & \,{\scriptsize \dup{$+6$\%}} & 9.9 & \,{\scriptsize \dup{$+11$\%}} \\
\bottomrule
\end{tabular}
\end{adjustbox}
\end{table}

Each arm is compared with the themed reskin on the same model and seed, by a two-sided exact sign-flip test
over the models' mean paired differences, and by the 95\% interval for the mean difference that
inverting this test gives. The frozen memory banks have a positive effect. Each adds about two floors: $+2.1$ for
the model's own bank ($p = 0.062$; interval $-0.2$ to $+4.3$; 6 of 8 models higher), $+2.0$ for
Claude Fable~5.1's ($p = 0.094$; $-0.4$ to $+4.3$) and $+1.6$ for GPT-6 Astra's
($p = 0.12$; $-0.7$ to $+3.8$). They also raise the share of runs that clear Act~1, by 9, 6, and 4 points, and acts
cleared by $0.07$ to $0.11$: a gain of about the size of removing the reskin altogether, which is worth 1.3 floors
and 9 points of Act~1 clears (\cref{tab:reskin-compact}). Every estimate
points to a benefit, though at ten seeds per model none reaches conventional significance.
The banks of two of the strongest models of \cref{tab:main-results} carry no more than the model's own.
No other knowledge given helps.
The enemy compendium changes neither acts ($-0.03$; $-0.20$ to $+0.13$) nor floors ($0.0$; $-2.3$ to $+2.1$), and the Reddit guide costs $1.9$ floors ($p = 0.016$; $-3.3$ to $-0.6$; 7 of 8 models lower) without changing acts ($-0.05$; $-0.20$ to $+0.10$).
Removing the card, potion, and relic compendium changes neither acts ($-0.07$; $-0.40$ to $+0.17$) nor floors ($-0.7$; $-5.4$ to $+1.7$) on average, but costs Gemini~3.8 Flash at low effort 12.2 floors.
Real names add $0.11$ acts and $0.8$ floors on these seeds ($p = 0.19$ and $0.30$), in the direction of the reskin ablation (\cref{tab:ablation-reskin}).
The strongest agent here, Gemini~3.8 Flash at low effort, loses floors to every addition (\cref{tab:ablation-prompt-by-model-floor}).

More knowledge does not make the models think less (\cref{tab:ablation-prompt-by-model-thinking}).
Pooled, the thinking per floor rises by $15\%$ with real names and by $16\%$, $22\%$, and $23\%$ with the three banks, but the direction depends on the model, and no change is significant ($p = 0.20$ to $0.89$, by the same test over the models' log ratios of thinking).
GLM-5.3 Flash and Qwen3.8 27B think much more with a bank ($+103\%$ to $+170\%$ and $+49\%$ to $+89\%$), while gpt-oss-120b and Nemotron~3 Nano Omni think less ($-14\%$ to $-44\%$ and $-10\%$ to $-25\%$).
Gemini~3.8 Flash at low effort, which chooses on each call whether to think, stops thinking without the compendium: 28 thinking tokens over its ten runs, against 2.6 thousand per floor with the compendium.

\providecommand{\spireicon}[1]{\raisebox{-0.25ex}{\includegraphics[height=1.7ex]{icons/#1.pdf}}}
\begin{table}[tbh]
\centering
\caption{\textbf{Effect of withholding what things in the game do, per model.} With the real game names and under our themed reskin, on the 10 evaluation seeds (those of \cref{tab:ablation-prompt-by-model}): mean acts cleared per run (\cref{tab:ablation-names-blind-floor} gives the mean final floor).
\emph{Full}: the model sees the card, potion, and relic compendium in the system prompt (our default in other tables, unless specified otherwise), and the text of each card, relic, potion, and power wherever the game shows it (also our default).
\emph{Blind}: the rules are the same, but no descriptions are \emph{ever} given to the model across any prompt (system or turn).
We expect models to rely on their (pre)training knowledge about \emph{Slay the Spire} to know what cards and relics do.
All arms take notes within a run (xml), with no cross-run memory bank, so do have the opportunity to deduce what cards do in one combat, save it, and recall it later.
\textcolor{ForestGreen}{Green}: knowledge helps. \textcolor{BrickRed}{Red}: knowledge hinders.}
\label{tab:ablation-names-blind}
\begin{adjustbox}{max width=\linewidth}
\small
\begin{tabular}{c@{\hspace{3pt}}l rrr rrr rr}
\toprule
 & & \multicolumn{3}{c}{Real names} & \multicolumn{3}{c}{Themed reskin} & \multicolumn{2}{c}{Real names $-$ reskin} \\
\cmidrule(lr){3-5} \cmidrule(lr){6-8} \cmidrule(lr){9-10}
 & Agent & Full & Blind & $\Delta$ & Full & Blind & $\Delta$ & \makebox[4.2em][r]{Full} & \makebox[4.2em][r]{Blind} \\
\midrule
\spireicon{qwen} & Qwen3.8 27B {\scriptsize\textcolor{gray}{(xhigh)}} & 0.30 & 0.00 & {\scriptsize \dup{$+0.30$}} & 0.30 & 0.10 & {\scriptsize \dup{$+0.20$}} & {\scriptsize \dzero{$\pm0.00$}} & {\scriptsize \ddown{$-0.10$}} \\
\spireicon{qwen} & Qwen3.6 35B-A3B & 0.00 & 0.00 & {\scriptsize \dzero{$\pm0.00$}} & 0.00 & 0.00 & {\scriptsize \dzero{$\pm0.00$}} & {\scriptsize \dzero{$\pm0.00$}} & {\scriptsize \dzero{$\pm0.00$}} \\
\spireicon{claude} & Claude Opus 5 {\scriptsize\textcolor{gray}{(low)}} & 0.60 & 0.50 & {\scriptsize \dup{$+0.10$}} & 0.30 & 0.30 & {\scriptsize \dzero{$\pm0.00$}} & {\scriptsize \dup{$+0.30$}} & {\scriptsize \dup{$+0.20$}} \\
\spireicon{gemini} & Gemini 3.8 Flash {\scriptsize\textcolor{gray}{(low)}} & 1.30 & 0.30 & {\scriptsize \dup{$+1.00$}} & 0.90 & 0.20 & {\scriptsize \dup{$+0.70$}} & {\scriptsize \dup{$+0.40$}} & {\scriptsize \dup{$+0.10$}} \\
\spireicon{nvidia} & Nemotron 3 Nano Omni & 0.00 & 0.00 & {\scriptsize \dzero{$\pm0.00$}} & 0.00 & 0.00 & {\scriptsize \dzero{$\pm0.00$}} & {\scriptsize \dzero{$\pm0.00$}} & {\scriptsize \dzero{$\pm0.00$}} \\
\spireicon{openai} & GPT-5.6 Luna {\scriptsize\textcolor{gray}{(max)}} & 0.30 & 0.40 & {\scriptsize \ddown{$-0.10$}} & 0.10 & 0.00 & {\scriptsize \dup{$+0.10$}} & {\scriptsize \dup{$+0.20$}} & {\scriptsize \dup{$+0.40$}} \\
\spireicon{openai} & gpt-oss-120b {\scriptsize\textcolor{gray}{(high)}} & 0.20 & 0.00 & {\scriptsize \dup{$+0.20$}} & 0.10 & 0.10 & {\scriptsize \dzero{$\pm0.00$}} & {\scriptsize \dup{$+0.10$}} & {\scriptsize \ddown{$-0.10$}} \\
\spireicon{zai} & GLM-5.3 Flash {\scriptsize\textcolor{gray}{(max)}} & 0.30 & 0.20 & {\scriptsize \dup{$+0.10$}} & 0.40 & 0.00 & {\scriptsize \dup{$+0.40$}} & {\scriptsize \ddown{$-0.10$}} & {\scriptsize \dup{$+0.20$}} \\
\midrule
 & Mean (8 agents) & 0.38 & 0.18 & {\scriptsize \dup{$+0.20$}} & 0.26 & 0.09 & {\scriptsize \dup{$+0.17$}} & {\scriptsize \dup{$+0.12$}} & {\scriptsize \dup{$+0.09$}} \\
\bottomrule
\end{tabular}
\end{adjustbox}
\end{table}

\providecommand{\spireicon}[1]{\raisebox{-0.25ex}{\includegraphics[height=1.7ex]{icons/#1.pdf}}}
\begin{table}[tbh]
\centering
\caption{\textbf{Effect of withholding what things in the game do on the final floor, per model.} With the real game names and under our themed reskin, on the 10 evaluation seeds: mean final floor. Arms, differences, and marks as in \cref{tab:ablation-names-blind}.}
\label{tab:ablation-names-blind-floor}
\begin{adjustbox}{max width=\linewidth}
\small
\begin{tabular}{c@{\hspace{3pt}}l rrr rrr rr}
\toprule
 & & \multicolumn{3}{c}{Real names} & \multicolumn{3}{c}{Themed reskin} & \multicolumn{2}{c}{Real names $-$ reskin} \\
\cmidrule(lr){3-5} \cmidrule(lr){6-8} \cmidrule(lr){9-10}
 & Agent & Full & Blind & $\Delta$ & Full & Blind & $\Delta$ & \makebox[4.2em][r]{Full} & \makebox[4.2em][r]{Blind} \\
\midrule
\spireicon{qwen} & Qwen3.8 27B {\scriptsize\textcolor{gray}{(xhigh)}} & 16.4 & 13.0 & {\scriptsize \dup{$+3.4$}} & 15.1 & 10.9 & {\scriptsize \dup{$+4.2$}} & {\scriptsize \dup{$+1.3$}} & {\scriptsize \dup{$+2.1$}} \\
\spireicon{qwen} & Qwen3.6 35B-A3B & 10.7 & 9.0 & {\scriptsize \dup{$+1.7$}} & 11.0 & 7.6 & {\scriptsize \dup{$+3.4$}} & {\scriptsize \ddown{$-0.3$}} & {\scriptsize \dup{$+1.4$}} \\
\spireicon{claude} & Claude Opus 5 {\scriptsize\textcolor{gray}{(low)}} & 22.0 & 19.8 & {\scriptsize \dup{$+2.2$}} & 19.1 & 17.3 & {\scriptsize \dup{$+1.8$}} & {\scriptsize \dup{$+2.9$}} & {\scriptsize \dup{$+2.5$}} \\
\spireicon{gemini} & Gemini 3.8 Flash {\scriptsize\textcolor{gray}{(low)}} & 30.1 & 14.8 & {\scriptsize \dup{$+15.3$}} & 26.2 & 15.2 & {\scriptsize \dup{$+11.0$}} & {\scriptsize \dup{$+3.9$}} & {\scriptsize \ddown{$-0.4$}} \\
\spireicon{nvidia} & Nemotron 3 Nano Omni & 6.5 & 8.0 & {\scriptsize \ddown{$-1.5$}} & 6.3 & 5.9 & {\scriptsize \dup{$+0.4$}} & {\scriptsize \dup{$+0.2$}} & {\scriptsize \dup{$+2.1$}} \\
\spireicon{openai} & GPT-5.6 Luna {\scriptsize\textcolor{gray}{(max)}} & 13.9 & 19.4 & {\scriptsize \ddown{$-5.5$}} & 15.2 & 13.1 & {\scriptsize \dup{$+2.1$}} & {\scriptsize \ddown{$-1.3$}} & {\scriptsize \dup{$+6.3$}} \\
\spireicon{openai} & gpt-oss-120b {\scriptsize\textcolor{gray}{(high)}} & 13.1 & 7.5 & {\scriptsize \dup{$+5.6$}} & 12.4 & 13.1 & {\scriptsize \ddown{$-0.7$}} & {\scriptsize \dup{$+0.7$}} & {\scriptsize \ddown{$-5.6$}} \\
\spireicon{zai} & GLM-5.3 Flash {\scriptsize\textcolor{gray}{(max)}} & 16.4 & 14.3 & {\scriptsize \dup{$+2.1$}} & 17.5 & 9.0 & {\scriptsize \dup{$+8.5$}} & {\scriptsize \ddown{$-1.1$}} & {\scriptsize \dup{$+5.3$}} \\
\midrule
 & Mean (8 agents) & 16.1 & 13.2 & {\scriptsize \dup{$+2.9$}} & 15.4 & 11.5 & {\scriptsize \dup{$+3.9$}} & {\scriptsize \dup{$+0.7$}} & {\scriptsize \dup{$+1.7$}} \\
\bottomrule
\end{tabular}
\end{adjustbox}
\end{table}

To expand on this, we explored how the model's performance was impacted when the card and relic descriptions were removed from the model's input entirely (\emph{blinded}).
In this case, we expect the model to rely on its pretraining knowledge about \emph{Slay the Spire} to know what cards and relics do---knowledge which should be ablated entirely by reskinning the game.
As shown in \cref{tab:ablation-names-blind,tab:ablation-names-blind-floor}, the models overall perform worse when they are blinded to the effects of the cards and relics in the game, as expected.
But surprisingly, the impact of blinding was very similar when the models saw the real item names or the reskinned names, suggesting models were not able to fall back on their pretraining to know what individual game items do.
This suggests the difference in performance when using real names versus reskinned names comes from heuristics about the game in pretraining data, but the model has not memorized the effects of each game item.
Some models were more impacted by the blinding than others, for example Gemini 3.8 Flash fell by 0.7 mean acts cleared per run (11.0 floors) when blinded with reskinned names, yet its performance fell \emph{even more} when the real names were available (1.0 mean acts; 15.3 floors), suggesting it heavily referenced the description text even when items were listed with their real names.

\FloatBarrier
\section{Cost and reasoning}
\label{app:frontiers}
This appendix gives the cost frontiers in full, then what a run costs in time and money, as a guide to budgeting a new cohort. How costs are priced is described in
\cref{sec:methods-cost}.
For performance against reasoning budget, see \cref{fig:pareto-thinking}.

\subsection{Cost}
\label{app:frontiers-cost}
\Cref{fig:pareto} (in the main text) names only the points on its front. \Cref{fig:pareto-full} repeats it with every point
named, and \cref{fig:pareto-cost-floor} gives the front against cost per floor, which ranks models on
efficiency rather than on what a whole run costs.

\begin{figure}[tbhp]
\centering
\includegraphics[width=0.9\linewidth]{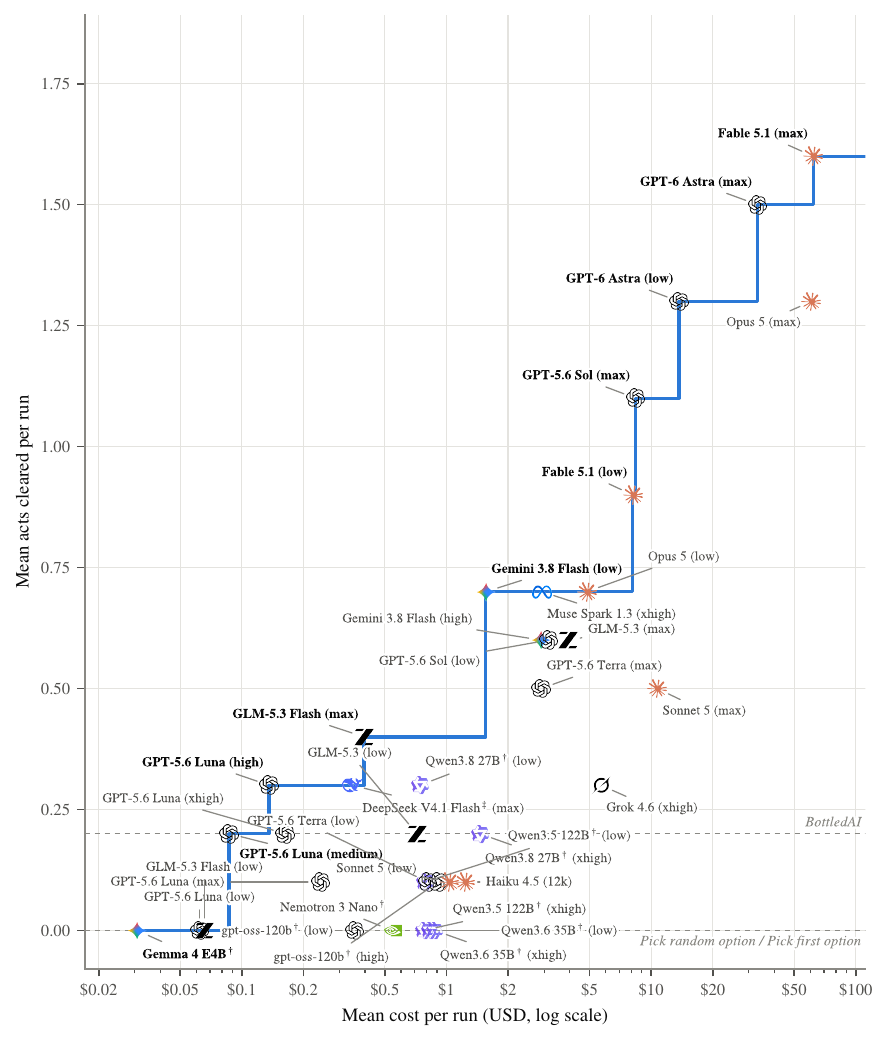}
\caption{\textbf{Acts cleared against cost per run, in full.} \Cref{fig:pareto} with every point named, with its reasoning-effort level and, where it has fewer
than 10 seeds, its seed count.}
\label{fig:pareto-full}
\end{figure}

\begin{figure}[tbhp]
\centering
\includegraphics[width=0.9\linewidth]{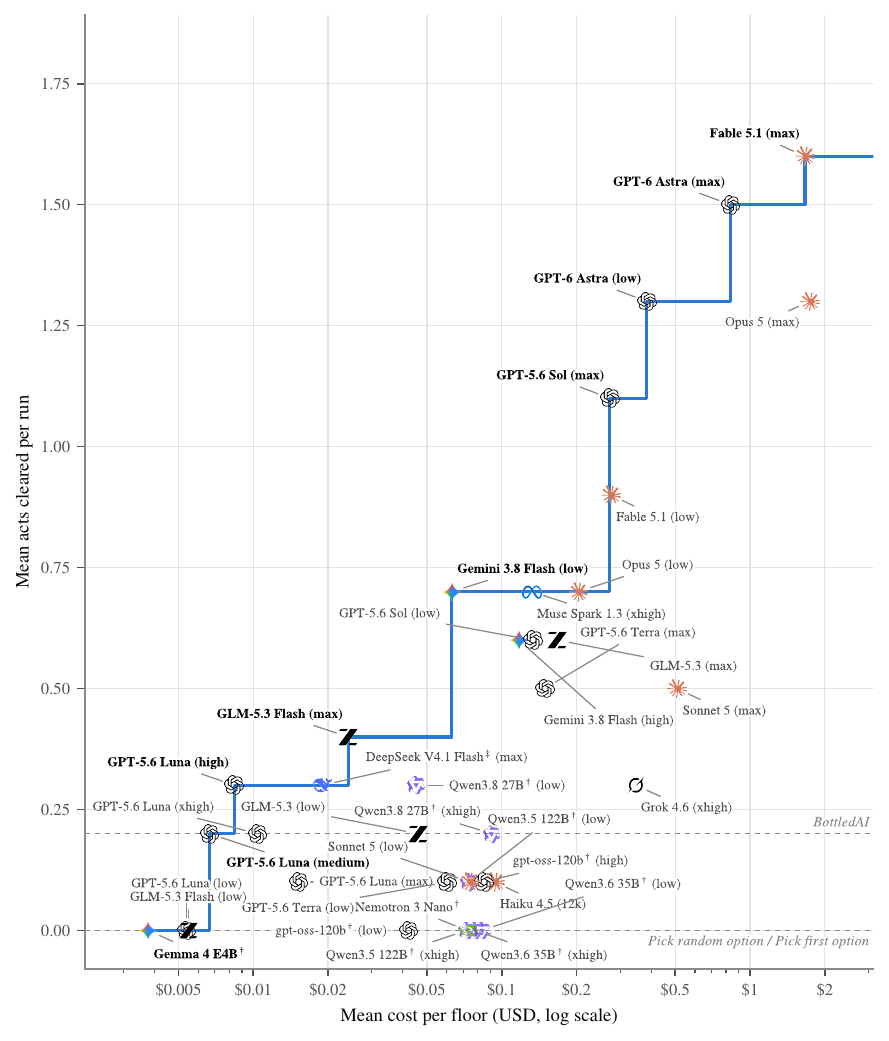}
\caption{\textbf{Performance against cost per floor.} Mean acts cleared per run against mean cost per floor (log
scale), for the same points as \cref{fig:pareto}. $^{\dagger}$Self-hosted; priced at Artificial Analysis
headline rates. $^{\ddagger}$Peak-rate list price, an upper bound.}
\label{fig:pareto-cost-floor}
\end{figure}

\FloatBarrier
\clearpage
\subsection{Time and cost of a run}
\label{app:budget}
\Cref{tab:budget} gives what a run costs in wall-clock time, money, tokens, and mechanical errors, as a
guide to budgeting a new cohort. A cohort's seeds are played in series against one game
instance, so its wall-clock time is the sum of its runs'.

The non-LLM baselines finish a seed in about two minutes. The LLM agents take from a median of 13 minutes per seed
(Qwen3.5 122B-A10B) to 231 (Claude Fable~5.1 and Claude Opus~5), so a ten-seed cohort of either Claude model takes
about a day and a half. Time is not cost: Nemotron~3 Nano Omni, self-hosted, takes 135 minutes per seed on average
and costs \$0.54 per run. For some models the mean is well above the median (Grok~4.6, 205 against 58 minutes;
Gemma~4 E4B, 61 against 17), because a few runs took much longer, so the mean is the better guide to budgeting time.

The 20 main-table cohorts cost about \$2{,}000 in all, and three account for 78\% of it: Claude Fable~5.1 (\$619),
Claude Opus~5 (\$606), and GPT-6 Astra (\$331). Every other cohort costs \$107 or less, and 13 of those 17 under
\$30. A run makes from 71 (Nemotron~3 Nano Omni) to 546 (GPT-6 Astra) model calls and uses from 1.3 million
(Gemma~4 E4B) to 44.5 million tokens (Claude Opus~5).

\providecommand{\spireicon}[1]{\raisebox{-0.25ex}{\includegraphics[height=1.7ex]{icons/#1.pdf}}}
\begin{table}[htbp]
\centering
\caption{\textbf{What a run costs.} For the agents of \cref{tab:main-results}: wall-clock minutes per seed (one game instance, seeds in series), dollars per run and per 10-seed cohort (priced as in \cref{tab:main-results}), model calls and tokens per run (input including cached, plus output, in thousands); tool-call errors are in \cref{tab:tool-errors}. Time depends on the provider's throughput on the day as well as on the model.}
\label{tab:budget}
\begin{adjustbox}{max width=\linewidth}
\small
\begin{tabular}{c@{\hspace{3pt}}l r rr rr rr}
\toprule
 & & & \multicolumn{2}{c}{Minutes/seed} & \multicolumn{2}{c}{Cost (\$)} & \multicolumn{2}{c}{Per run} \\
\cmidrule(lr){4-5} \cmidrule(lr){6-7} \cmidrule(lr){8-9}
 & Agent & Runs & median & mean & /run & /10 seeds & Calls & k tokens \\
\midrule
\spireicon{left} & Pick first option & 10 & 2 & 2 & -- & -- & -- & -- \\
\spireicon{random} & Pick random option & 10 & 2 & 2 & -- & -- & -- & -- \\
\midrule
\spireicon{bottled_ai} & BottledAI w/o Heart & 10 & 2 & 2 & -- & -- & -- & -- \\
\spireicon{bottled_ai} & BottledAI w/ Heart & 10 & 2 & 2 & -- & -- & -- & -- \\
\midrule
\spireicon{qwen} & Qwen3.8 27B {\scriptsize\textcolor{gray}{(xhigh)}} & 10 & 99 & 107 & 1.45 & 15 & 124 & 4,641 \\
\spireicon{qwen} & Qwen3.6 35B-A3B & 10 & 30 & 31 & 0.87 & 9 & 104 & 2,036 \\
\spireicon{qwen} & Qwen3.5 122B-A10B {\scriptsize\textcolor{gray}{(xhigh)}} & 10 & 13 & 14 & 0.77 & 8 & 77 & 1,601 \\
\spireicon{claude} & Claude Fable 5.1 {\scriptsize\textcolor{gray}{(max)}} & 10 & 231 & 229 & 61.91 & 619 & 268 & 24,604 \\
\spireicon{claude} & Claude Opus 5 {\scriptsize\textcolor{gray}{(max)}} & 10 & 231 & 214 & 60.62 & 606 & 301 & 44,517 \\
\spireicon{claude} & Claude Sonnet 5 {\scriptsize\textcolor{gray}{(max)}} & 10 & 107 & 157 & 10.70 & 107 & 159 & 14,122 \\
\spireicon{claude} & Claude Haiku 4.5 {\scriptsize\textcolor{gray}{(12k)}} & 10 & 21 & 31 & 1.23 & 12 & 101 & 3,395 \\
\spireicon{deepseek} & DeepSeek V4.1 Flash {\scriptsize\textcolor{gray}{(max)}} & 10 & 23 & 24 & 0.35 & 3 & 119 & 4,256 \\
\spireicon{gemini} & Gemini 3.8 Flash {\scriptsize\textcolor{gray}{(high)}} & 10 & 44 & 62 & 2.89 & 29 & 208 & 8,382 \\
\spireicon{gemini} & Gemma 4 E4B & 10 & 17 & 61 & 0.03 & 0 & 82 & 1,349 \\
\spireicon{meta} & Muse Spark 1.3 {\scriptsize\textcolor{gray}{(xhigh)}} & 10 & 52 & 50 & 2.93 & 29 & 228 & 6,972 \\
\spireicon{nvidia} & Nemotron 3 Nano Omni & 10 & 130 & 135 & 0.54 & 5 & 71 & 1,595 \\
\spireicon{openai} & GPT-6 Astra {\scriptsize\textcolor{gray}{(max)}} & 10 & 108 & 165 & 33.09 & 331 & 546 & 20,032 \\
\spireicon{openai} & GPT-5.6 Sol {\scriptsize\textcolor{gray}{(max)}} & 10 & 55 & 58 & 8.39 & 84 & 331 & 10,435 \\
\spireicon{openai} & GPT-5.6 Terra {\scriptsize\textcolor{gray}{(max)}} & 10 & 35 & 34 & 2.91 & 29 & 230 & 5,667 \\
\spireicon{openai} & GPT-5.6 Luna {\scriptsize\textcolor{gray}{(max)}} & 10 & 21 & 49 & 0.24 & 2 & 168 & 4,384 \\
\spireicon{openai} & gpt-oss-120b {\scriptsize\textcolor{gray}{(high)}} & 10 & 84 & 87 & 0.91 & 9 & 175 & 6,498 \\
\spireicon{grok} & Grok 4.6 {\scriptsize\textcolor{gray}{(xhigh)}} & 10 & 58 & 205 & 5.68 & 57 & 171 & 4,649 \\
\spireicon{zai} & GLM-5.3 {\scriptsize\textcolor{gray}{(max)}} & 10 & 119 & 121 & 3.92 & 39 & 160 & 7,194 \\
\spireicon{zai} & GLM-5.3 Flash {\scriptsize\textcolor{gray}{(max)}} & 10 & 103 & 118 & 0.40 & 4 & 107 & 4,641 \\
\bottomrule
\end{tabular}
\end{adjustbox}
\end{table}

\FloatBarrier
\FloatBarrier
\section[\texorpdfstring{Puzzle benchmark}{Puzzle benchmark}]{Puzzle benchmark}
\label{app:puzzles}

A run of \textit{Slay the Spire} mixes many decisions, and its outcome alone says little about which of them an agent gets wrong. This makes it difficult to dissect where different agents fail, or even if specific game rules are understood correctly. 
To better understand specific decision making and rules understanding we developed an additional puzzle benchmark.  Each puzzle isolates single decisions, and evaluates the model's performance at an isolated task.
A single puzzle is one authored room with a \emph{known} best answer with a fixed deck, hand, HP, relics, potions, and enemies as input. An agent plays each puzzle once, and the result is whether it found the answer. The puzzles cover only decisions that have an objectively best option, which is mostly how to
resolve a turn or a short fight, and some reward and shop choices where the deck fixes which option is right.

\subsection{Implementation}
\label{app:puzzles-implementation}

\paragraph{Engine.}
The puzzles are played in the released game, through the same interface and harness as the runs
(\cref{app:methods-harness}). For this we developed the Puzzle Suite mod, in which a single room is presented to the player, built from a
puzzle file: this defines the draw pile, hand, discard pile, and exhaust pile directly, sets the player's HP, energy,
relics, potions, and gold, and places the enemies with their HP, powers, and a fixed attack that they repeat every
turn. The first state the agent sees is the board. A puzzle ends when the room is left or the player dies. A reward
or shop puzzle continues past the fight to the reward screen, or opens in a shop.

\paragraph{What the agent is given.}
The agent plays through the same harness as in the main results.
No notes or memories from earlier play are loaded, so each puzzle is met fresh and nothing is carried from one puzzle to another.
A puzzle is a single floor, and notes are written between floors, so the agent has nothing to write during one.
A puzzle is played under the game's real names and under the thematic reskin of the main results (\cref{app:methods-reskin}).
An agent's score on a puzzle is the mean of the arms it has played, and its accuracy is the mean of those scores over the puzzles: the arms are averaged first because what makes a puzzle hard is mostly the puzzle, not the naming.
Every agent has played all 123 graded puzzles in both arms.
It is not told what a puzzle tests or how it is graded.
After the room it is asked once, through a tool named \texttt{rate\_puzzle}, to rate the difficulty of what it faced and its confidence in its line; the rating plays no part in the grade, and the name of that tool is the only sign the agent is given that the room is a puzzle.
Each puzzle is a separate process with its own game, so a failure in one cannot affect another.

\paragraph{Grading.}
A puzzle is graded as correct or not, by fixed rules (recorded in the puzzle file):
\begin{itemize}[leftmargin=*,nosep]
\item A fight puzzle is correct if the room is won. Where the puzzle sets a target (HP remaining, maximum HP, a
relic counter, a potion made, or a value a card keeps), a win that falls short of the target is not correct. 71 of the 124 puzzles set one, so that a line
that survives but wastes what the best line keeps is told apart from the answer.
\item A reward or shop puzzle is correct if the agent leaves the room owning what the puzzle requires (a specific
card or relic, or the absence of one).
\item One puzzle has no solution. It is played to record how long an agent searches, and is not graded.
\end{itemize}
Before a row is recorded, the first state in the room is checked against the puzzle (the player's maximum HP and
each enemy's name and maximum HP). A game that opened on anything else is recorded as an infrastructure failure,
which is neither correct nor incorrect, and the puzzle is played again.
One play is the exception: GLM-5.3, under the reskin, never answered one puzzle (every request ran to its 30-minute limit, seven times over), and that play is counted as incorrect.

\paragraph{Authoring and verification.}
The puzzles were proposed by the authors and by four LLMs, each working as a coding agent that built its puzzles and played them in the live game: of the 124, 55 were first written by Claude Opus 5, 46 by Claude Fable 5.2, 14 by GLM-5.3 Flash, and 9 by GPT-5.6 Sol.
The authors' requests and corrections shaped many of them, and one author's review of the first suites, and later playtest of them all, changed or removed many more (see below).
Each puzzle was written around one decision, such that there is a deterministically optimal play (which is \emph{not} typically the case for in-game \emph{Slay the Spire} decisions).
A tempting wrong line is usually dealt first in the hand so that playing the hand from left to right does not trivially win the fight.
Each puzzle was verified by playing in a live game, confirming the the intended line works, and each wrong line the puzzle was built around does not work.

\paragraph{Human playtest.}
One of the authors, an experienced player of the game, then played every puzzle blind: 164 at the time, under anonymous names and in a shuffled order, rating each for difficulty and for whether it was sound.
The playtest changed the catalogue in three ways.
Twenty puzzles were dropped: eleven that the pick-first policy and all five seeds of the random policy solved, and nine reward, relic, or shop choices in which a second option presented was defensible (no clear best choice).
The target value (i.e. final HP total for the fight) was raised on eight puzzles where a better line existed than the one the puzzle had been written around: on five to the line the playtester found, and on three, whose outcome depends on a shuffle, to the best finish that can be guaranteed whatever the shuffle, found by an exhaustive search over the lines of play.

\paragraph{Final puzzle catalogue.}
We finalized a suite of 124 puzzles. 89 are played as the Ironclad character (though most do not use Ironclad-specific mechanics) which was used in the main results, 13 as the Silent, 12 as the Defect, and 10 as the Watcher (the other three player characters in the game).
110 are fights and 14 are reward or shop decisions.
The fights are short: 58 are decided within one turn, 25 within two, and the longest allow eight. Most have a single enemy. The enemies
repeat one attack, so no puzzle asks the agent to read a pattern of intents, and none cover rest sites, events, or
the map; since those decisions have no objectively good outcomes suitable for puzzles.

\subsection{Time horizons and skills}
\label{app:puzzles-skills}

To measure what an agent is good at, we categorize every graded puzzle in two ways (\cref{tab:puzzle-taxonomy}).
The first classification is the time horizon of its decision, following the scales of \cref{fig:game-overview}.
\begin{itemize}[leftmargin=*,nosep]
\item \emph{One turn.} The answer is the play of a single turn, with every card it needs in view.
\item \emph{One fight.} The fight is won by a plan that spans turns: a turn of set-up, or of waiting, pays in a later one.
\item \emph{Rest of the run.} The decision pays after the fight. Either the fight is won whatever the agent does and the decision is what is carried out of it (maximum HP gained from a kill, a relic left charged for the next fight, or HP that will not be healed), or it is a reward or shop choice, which shapes the deck for every fight that follows.
\end{itemize}
The second is the skill the answer relies on, within that horizon.
Puzzles often touch more than one skill; they are categorized according to whose absence loses it.
No table pools puzzles across horizons.
The classification was made from the design notes of each puzzle, not from the results, and is released with the puzzles.

\begin{table}[t]
\centering\small
\caption{\textbf{How the puzzles are classified.} Every graded puzzle has one time horizon, which is how far ahead its decision looks, and within it one skill, the decision its answer turns on. \emph{Solved} is the mean over the agents of \cref{tab:puzzle-skills} of the percentage of those puzzles answered correctly.}
\label{tab:puzzle-taxonomy}
\begin{adjustbox}{max width=\linewidth}
\small
\begin{tabular}{l p{0.62\linewidth} rr}
\toprule
Skill & What the puzzle asks & Puzzles & Solved (\%) \\
\midrule
\rowcolor{vlightgray}
\textit{One turn} & \textit{The answer is the play of a single turn, with every card it needs in view.} & \textit{58} & \textit{72} \\
\addlinespace
\quad Pay & Paying HP, or a turn's attack, for something worth more this turn; or seeing that the price cannot be met. & 4 & 85 \\
\addlinespace
\quad Order & The cards to play are plain; the turn is won or lost by the order they are played in. & 19 & 75 \\
\addlinespace
\quad Engine & Finding an infinite loop that pays for itself within the turn, often by first removing (exhausting) the cards that would break it. & 13 & 58 \\
\addlinespace
\quad Rules & The answer follows from what one relic, potion, or enemy power does, and the usual best card is wrong. & 10 & 78 \\
\addlinespace
\quad Count & Several ways to play the hand are open; the right line is found by adding up what each would do. & 8 & 70 \\
\addlinespace
\quad Enemy & More than one enemy: which to aim at, or when an attack on all of them beats one on one. & 4 & 76 \\
\midrule
\rowcolor{vlightgray}
\textit{One fight} & \textit{The fight is won by a plan that spans turns: a turn of set-up, or of waiting, pays in a later one.} & \textit{40} & \textit{62} \\
\addlinespace
\quad Tank & Taking damage, spending HP, or declining to block because the line that looks safe loses the fight. & 7 & 68 \\
\addlinespace
\quad Plan & A turn spent on set-up or held back so that a later turn pays; relic counters and timed effects inside the fight. & 22 & 66 \\
\addlinespace
\quad Engine & Assembling an infinite loop over more than one turn: a turn of slimming the deck, then the loop. & 3 & 28 \\
\addlinespace
\quad Rules & Two effects that only pay together, set up on the first turn and collected over the fight. & 5 & 48 \\
\addlinespace
\quad Hail-Mary & The only winning line depends on a random outcome and has to be taken anyway; the ``locally best'' line certainly loses. & 3 & 83 \\
\midrule
\rowcolor{vlightgray}
\textit{Rest of the run} & \textit{The decision pays after the fight: what is carried out of a fight that is won either way, or a reward or shop choice that shapes the deck.} & \textit{25} & \textit{48} \\
\addlinespace
\quad Stall & A kill postponed, an attack withheld, or a hit taken, for maximum HP, a charged relic, or HP that will not be healed. & 11 & 26 \\
\addlinespace
\quad Pick-up & A relic and a card on one reward screen, where the order they are taken in changes what is owned. & 3 & 62 \\
\addlinespace
\quad Card & Which offered card fits the deck that has just been shown in a fight. & 3 & 72 \\
\addlinespace
\quad Shop & What to buy, or remove, with a purse that covers only some of the stock; or whether an offered relic's permanent drawback is unbreakable because it breaks the deck. & 8 & 64 \\
\bottomrule
\end{tabular}
\end{adjustbox}
\end{table}

Within one turn, \emph{Order} and \emph{Count} puzzles ask for a correct turn from a hand whose cards are all in view: the right cards in the right order, or, among several ways to play the hand, the one whose damage or Block adds up to the most or reaches an exact number.
\emph{Rules} puzzles turn on what one relic, potion, or enemy power does, and are built so that a card that would be best in most situations is wrong here.
\emph{Enemy} puzzles have more than one target.
\emph{Engine} puzzles ask the agent to see that a small set of cards can be repeated infinitely (winning the fight), and usually require the removal of specific cards that would break the loop first.
\emph{Pay} puzzles ask the agent to spend HP for something worth more within the turn.

Across one fight, \emph{Plan} puzzles require a turn of set-up or of waiting, and the \emph{Engine} puzzles of this horizon need a turn spent slimming the deck before the loop can run.
\emph{Tank} puzzles are those in which the line that looks safe (blocking or healing) loses the fight, and the hit has to be taken.
The five \emph{Rules} puzzles of this horizon pair two effects on the first turn for something collected over the whole fight (two potions, or a card and the relic that cancels its drawback), and the three \emph{Hail-Mary} puzzles have a single winning line whose success depends on a random outcome.

In the last horizon, the eleven \emph{Stall} puzzles are fights in which winning is not the question.
Each can be won by the obvious line, and each sets a target that the obvious line misses: a kill postponed so that \emph{Feed} lands it, an attack withheld so that a relic's counter is left one short of paying out, a hit taken now because a later heal will be larger, or a turn of waiting for a card that makes a potion or strengthens itself for the rest of the run.
The other three skills of this horizon are the reward and shop decisions.

\subsection{Results}
\label{app:puzzles-results}

\begin{table}[t]
\centering\small
\caption{\textbf{Puzzle accuracy by the time horizon of the decision.} Horizons as in \cref{tab:puzzle-taxonomy}: the play of one turn, a plan across the turns of one fight, and a decision that pays after the fight, for the rest of the run. Each cell is the percentage of that column's puzzles the agent answered correctly, shaded from white (the random policy's score in that column, or below) to blue (100); the grey row gives the number of puzzles in the column. A puzzle is played under the game's real names and under the thematic reskin; an agent's score on a puzzle is the mean of the arms it has played, and a cell is the mean of those scores over the column's puzzles (\cref{tab:puzzle-reskin} sets the arms apart). \emph{Overall} is over every puzzle in the table. Agents are in the order of \cref{tab:main-results}, with the reasoning setting the puzzles were played at in grey. The baseline policies read no names and are played once, under the real names; the random policy is the mean over its seeds.}
\label{tab:puzzle-horizons}
\begin{adjustbox}{max width=\linewidth}
\small
\begin{tabular}{c@{\hspace{3pt}}l r rrr}
\toprule
 & & & \multicolumn{3}{c}{Time horizon of the decision} \\
\cmidrule(lr){4-6}
 & & Overall & One turn & One fight & Rest of run \\
 & Agent & \textcolor{gray}{123} & \textcolor{gray}{58} & \textcolor{gray}{40} & \textcolor{gray}{25} \\
\midrule
\multicolumn{6}{@{}l}{\textit{Baseline policies}} \\
\spireicon{left} & Pick first option & 13 & 9 & 8 & \cellcolor{tblue!9}32 \\
\spireicon{random} & Pick random option {\scriptsize\textcolor{gray}{(mean of 5)}} & 13 & 12 & 8 & 20 \\
\midrule
\multicolumn{6}{@{}l}{\textit{Symbolic agent (Bottled AI)}} \\
\spireicon{bottled_ai} & BottledAI & \cellcolor{tblue!19}41 & \cellcolor{tblue!30}57 & \cellcolor{tblue!19}38 & 8 \\
\midrule
\multicolumn{6}{@{}l}{\textit{LLM agents}} \\
\spireicon{qwen} & Qwen3.8 27B {\scriptsize\textcolor{gray}{(xhigh)}} & \cellcolor{tblue!28}53 & \cellcolor{tblue!32}59 & \cellcolor{tblue!31}55 & \cellcolor{tblue!12}36 \\
\spireicon{qwen} & Qwen3.6 35B-A3B & \cellcolor{tblue!11}29 & \cellcolor{tblue!16}35 & \cellcolor{tblue!11}25 & \cellcolor{tblue!2}22 \\
\spireicon{qwen} & Qwen3.5 122B-A10B & \cellcolor{tblue!6}21 & \cellcolor{tblue!6}21 & \cellcolor{tblue!10}24 & 18 \\
\spireicon{claude} & Claude Fable 5.1 {\scriptsize\textcolor{gray}{(max)}} & \cellcolor{tblue!51}86 & \cellcolor{tblue!58}\runnerup{97} & \cellcolor{tblue!53}\runnerup{89} & \cellcolor{tblue!27}56 \\
\spireicon{claude} & Claude Opus 5 {\scriptsize\textcolor{gray}{(max)}} & \cellcolor{tblue!50}85 & \cellcolor{tblue!58}\runnerup{97} & \cellcolor{tblue!51}86 & \cellcolor{tblue!27}56 \\
\spireicon{claude} & Claude Sonnet 5 {\scriptsize\textcolor{gray}{(max)}} & \cellcolor{tblue!40}72 & \cellcolor{tblue!48}83 & \cellcolor{tblue!39}68 & \cellcolor{tblue!24}52 \\
\spireicon{claude} & Claude Haiku 4.5 {\scriptsize\textcolor{gray}{(12k)}} & \cellcolor{tblue!17}37 & \cellcolor{tblue!20}41 & \cellcolor{tblue!21}40 & \cellcolor{tblue!3}24 \\
\spireicon{deepseek} & DeepSeek V4.1 Flash {\scriptsize\textcolor{gray}{(max)}} & \cellcolor{tblue!39}70 & \cellcolor{tblue!50}85 & \cellcolor{tblue!36}64 & \cellcolor{tblue!18}44 \\
\spireicon{gemini} & Gemini 3.8 Flash {\scriptsize\textcolor{gray}{(high)}} & \cellcolor{tblue!53}\runnerup{89} & \cellcolor{tblue!58}\runnerup{97} & \cellcolor{tblue!52}88 & \cellcolor{tblue!40}\runnerup{74} \\
\spireicon{gemini} & Gemma 4 E4B & \cellcolor{tblue!3}16 & \cellcolor{tblue!1}14 & \cellcolor{tblue!3}12 & \cellcolor{tblue!6}28 \\
\spireicon{meta} & Muse Spark 1.3 {\scriptsize\textcolor{gray}{(xhigh)}} & \cellcolor{tblue!46}79 & \cellcolor{tblue!53}90 & \cellcolor{tblue!45}78 & \cellcolor{tblue!27}56 \\
\spireicon{nvidia} & Nemotron 3 Nano Omni & \cellcolor{tblue!9}26 & \cellcolor{tblue!12}29 & \cellcolor{tblue!9}22 & \cellcolor{tblue!2}22 \\
\spireicon{openai} & GPT-6 Astra {\scriptsize\textcolor{gray}{(max)}} & \cellcolor{tblue!57}\best{96} & \cellcolor{tblue!59}\best{99} & \cellcolor{tblue!58}\best{96} & \cellcolor{tblue!51}\best{88} \\
\spireicon{openai} & GPT-5.6 Sol {\scriptsize\textcolor{gray}{(max)}} & \cellcolor{tblue!51}87 & \cellcolor{tblue!55}93 & \cellcolor{tblue!52}88 & \cellcolor{tblue!38}70 \\
\spireicon{openai} & GPT-5.6 Terra {\scriptsize\textcolor{gray}{(max)}} & \cellcolor{tblue!49}84 & \cellcolor{tblue!55}92 & \cellcolor{tblue!49}84 & \cellcolor{tblue!34}66 \\
\spireicon{openai} & GPT-5.6 Luna {\scriptsize\textcolor{gray}{(max)}} & \cellcolor{tblue!34}61 & \cellcolor{tblue!42}73 & \cellcolor{tblue!31}56 & \cellcolor{tblue!16}42 \\
\spireicon{openai} & gpt-oss-120b {\scriptsize\textcolor{gray}{(high)}} & \cellcolor{tblue!26}51 & \cellcolor{tblue!35}63 & \cellcolor{tblue!22}42 & \cellcolor{tblue!12}36 \\
\spireicon{grok} & Grok 4.6 {\scriptsize\textcolor{gray}{(xhigh)}} & \cellcolor{tblue!46}80 & \cellcolor{tblue!53}90 & \cellcolor{tblue!46}79 & \cellcolor{tblue!28}58 \\
\spireicon{zai} & GLM-5.3 {\scriptsize\textcolor{gray}{(max)}} & \cellcolor{tblue!47}82 & \cellcolor{tblue!54}91 & \cellcolor{tblue!49}82 & \cellcolor{tblue!28}58 \\
\spireicon{zai} & GLM-5.3 Flash {\scriptsize\textcolor{gray}{(max)}} & \cellcolor{tblue!41}73 & \cellcolor{tblue!49}84 & \cellcolor{tblue!40}69 & \cellcolor{tblue!24}52 \\
\midrule
 & Mean over LLM agents & \cellcolor{tblue!35}64 & \cellcolor{tblue!41}72 & \cellcolor{tblue!35}62 & \cellcolor{tblue!21}48 \\
\bottomrule
\end{tabular}
\end{adjustbox}
\end{table}

\Cref{tab:puzzle-horizons} gives each agent's accuracy by time horizon and \cref{tab:puzzle-skills} by skill within each horizon, for the 20 agents that have played every puzzle in at least one arm.
Gemma~4 E4B played the puzzles on a different deployment of the model (\texttt{Gemma-4-E4B-it}) than the one used for its runs in \cref{tab:main-results} (\texttt{gemma4:e4b-highcontext}); we treat the two as the same model.
Three things stand out.

First, the puzzles separate the agents.
Accuracy over all puzzles ranges from 16\% to 96\%, with four open-weight models at the bottom.
The two baseline policies that do not look at the game answer 13\% (picking the first option) and 13\% (picking at random, the mean of five seeds), and every agent is above them.
They do best on the reward and shop decisions, where picking the first option is right at least half the time on each of the three skills, and the tables are shaded from the random policy's score for that reason.
Bottled AI, the symbolic agent of \cref{tab:main-results}, answers 41\%, which is more than the five weakest LLM agents.
Its profile is the reverse of the LLM agents' on the last horizon: it answers 57\% of the one-turn puzzles and 8\% of those whose decision pays after the fight, which its fixed rules do not consider.

Second, accuracy falls as the horizon lengthens.
The agents answer 72\% of the one-turn puzzles on average, 62\% of the puzzles that need a plan across a fight, and 48\% of those whose decision pays after the fight.
The order holds for 17 of the 20 agents; the other three, Gemma~4 E4B, Qwen3.5 122B-A10B, and Nemotron 3 Nano Omni, are the weakest, and answer between 12\% and 29\% at every horizon.
Four agents answer 97\% or more of the one-turn puzzles, so that at this horizon there is little left to measure.
The \emph{Stall} puzzles are answered 26\% of the time: all but one of the 20 agents answer 55\% of them or less, and sixteen answer a third or less.
The exception is GPT-6 Astra, at 86\%.

Third, within a horizon the skills differ.
Two agents make no error at all on \emph{Order}, and \emph{Pay} is the easiest skill (85\%), ahead of \emph{Hail-Mary} (83\%): the agents spend HP readily when the return comes in the same turn.
\emph{Engine} is the hardest skill at both of its horizons, at 58\% within a turn and 28\% across a fight, where five agents answer more than half; only \emph{Stall} (26\%) is harder.
In five of the fifteen skills no agent answers every puzzle: \emph{Count}, \emph{Plan}, the fight-horizon \emph{Rules}, \emph{Stall}, and \emph{Shop}; the best score on them is 94\%, 95\%, 90\%, 86\%, and 94\%.

A cell is at most two attempts at each puzzle, so a cell over a handful of puzzles moves in large steps: \emph{Hail-Mary}, \emph{Card}, \emph{Pick-up}, and the fight-horizon \emph{Engine} are three puzzles each, \emph{Enemy} and \emph{Pay} are four, and the fight-horizon \emph{Rules} is five.
Differences between neighbouring agents within such a column are not evidence of a difference in skill; the pattern across a row is.

\begin{table}[t]
\centering\small
\caption{\textbf{Puzzle accuracy by skill, within each time horizon (\cref{tab:puzzle-taxonomy}).} Each cell is the percentage of that column's puzzles the agent answered correctly, shaded from white (the random policy's score in that column, or below) to blue (100); the grey row gives the number of puzzles in the column. A puzzle is played under the game's real names and under the thematic reskin; an agent's score on a puzzle is the mean of the arms it has played, and a cell is the mean of those scores over the column's puzzles (\cref{tab:puzzle-reskin} sets the arms apart). \emph{Overall} is over every puzzle in the table. Agents are in the order of \cref{tab:main-results}, with the reasoning setting the puzzles were played at in grey. The baseline policies read no names and are played once, under the real names; the random policy is the mean over its seeds.}
\label{tab:puzzle-skills}
\begin{adjustbox}{max width=\linewidth}
\small
\begin{tabular}{@{}c@{\hspace{3pt}}l r rrrrrr rrrrr rrrr@{}}
\toprule
 & & & \multicolumn{6}{c}{One turn} & \multicolumn{5}{c}{One fight} & \multicolumn{4}{c}{Rest of the run} \\
\cmidrule(lr){4-9} \cmidrule(lr){10-14} \cmidrule(lr){15-18}
 & & Overall & Pay & Order & Engine & Rules & Count & Enemy & Tank & Plan & Engine & Rules & \makecell[b]{Hail-\\Mary} & Stall & Pick-up & Card & Shop \\
 & Agent & \textcolor{gray}{123} & \textcolor{gray}{4} & \textcolor{gray}{19} & \textcolor{gray}{13} & \textcolor{gray}{10} & \textcolor{gray}{8} & \textcolor{gray}{4} & \textcolor{gray}{7} & \textcolor{gray}{22} & \textcolor{gray}{3} & \textcolor{gray}{5} & \textcolor{gray}{3} & \textcolor{gray}{11} & \textcolor{gray}{3} & \textcolor{gray}{3} & \textcolor{gray}{8} \\
\midrule
\multicolumn{18}{@{}l}{\textit{Baseline policies}} \\
\spireicon{left} & Pick first option & 13 & \cellcolor{tblue!22}50 & 11 & 8 & 0 & 0 & 0 & 0 & \cellcolor{tblue!2}9 & 0 & 0 & \cellcolor{tblue!5}33 & 0 & \cellcolor{tblue!10}67 & \cellcolor{tblue!30}67 & \cellcolor{tblue!20}50 \\
\spireicon{random} & Pick random option {\scriptsize\textcolor{gray}{(mean of 5)}} & 13 & 20 & 12 & 11 & 20 & 5 & 10 & 11 & 5 & 0 & 8 & 27 & 2 & 60 & 33 & 25 \\
\midrule
\multicolumn{18}{@{}l}{\textit{Symbolic agent (Bottled AI)}} \\
\spireicon{bottled_ai} & BottledAI & \cellcolor{tblue!19}41 & \cellcolor{tblue!22}50 & \cellcolor{tblue!31}58 & \cellcolor{tblue!24}46 & \cellcolor{tblue!30}60 & \cellcolor{tblue!36}62 & \cellcolor{tblue!43}75 & \cellcolor{tblue!21}43 & \cellcolor{tblue!20}36 & 0 & \cellcolor{tblue!8}20 & \cellcolor{tblue!60}100 & \cellcolor{tblue!4}9 & 0 & 33 & 0 \\
\midrule
\multicolumn{18}{@{}l}{\textit{LLM agents}} \\
\spireicon{qwen} & Qwen3.8 27B {\scriptsize\textcolor{gray}{(xhigh)}} & \cellcolor{tblue!28}53 & \cellcolor{tblue!51}\runnerup{88} & \cellcolor{tblue!33}61 & \cellcolor{tblue!13}31 & \cellcolor{tblue!41}75 & \cellcolor{tblue!36}62 & \cellcolor{tblue!43}75 & \cellcolor{tblue!26}50 & \cellcolor{tblue!33}57 & 0 & \cellcolor{tblue!34}60 & \cellcolor{tblue!60}\best{100} & \cellcolor{tblue!4}9 & 50 & 33 & \cellcolor{tblue!35}69 \\
\spireicon{qwen} & Qwen3.6 35B-A3B & \cellcolor{tblue!11}29 & \cellcolor{tblue!32}62 & \cellcolor{tblue!21}42 & \cellcolor{tblue!1}12 & \cellcolor{tblue!11}35 & \cellcolor{tblue!24}44 & \cellcolor{tblue!18}38 & \cellcolor{tblue!2}14 & \cellcolor{tblue!11}23 & 0 & \cellcolor{tblue!21}40 & \cellcolor{tblue!33}67 & \cellcolor{tblue!7}14 & 0 & 33 & \cellcolor{tblue!10}38 \\
\spireicon{qwen} & Qwen3.5 122B-A10B & \cellcolor{tblue!6}21 & \cellcolor{tblue!13}38 & \cellcolor{tblue!15}34 & 4 & 15 & \cellcolor{tblue!5}12 & \cellcolor{tblue!10}25 & \cellcolor{tblue!12}29 & \cellcolor{tblue!11}23 & 0 & \cellcolor{tblue!21}40 & 17 & \cellcolor{tblue!4}9 & 33 & 33 & 19 \\
\spireicon{claude} & Claude Fable 5.1 {\scriptsize\textcolor{gray}{(max)}} & \cellcolor{tblue!51}86 & \cellcolor{tblue!60}\best{100} & \cellcolor{tblue!60}\best{100} & \cellcolor{tblue!55}92 & \cellcolor{tblue!60}\best{100} & \cellcolor{tblue!56}\best{94} & \cellcolor{tblue!60}\best{100} & \cellcolor{tblue!60}\best{100} & \cellcolor{tblue!57}\best{95} & \cellcolor{tblue!40}67 & \cellcolor{tblue!27}50 & \cellcolor{tblue!60}\best{100} & \cellcolor{tblue!16}27 & \cellcolor{tblue!35}\runnerup{83} & \cellcolor{tblue!60}\best{100} & \cellcolor{tblue!35}69 \\
\spireicon{claude} & Claude Opus 5 {\scriptsize\textcolor{gray}{(max)}} & \cellcolor{tblue!50}85 & \cellcolor{tblue!60}\best{100} & \cellcolor{tblue!58}\runnerup{97} & \cellcolor{tblue!55}92 & \cellcolor{tblue!60}\best{100} & \cellcolor{tblue!56}\best{94} & \cellcolor{tblue!60}\best{100} & \cellcolor{tblue!60}\best{100} & \cellcolor{tblue!57}\best{95} & \cellcolor{tblue!20}33 & \cellcolor{tblue!27}50 & \cellcolor{tblue!60}\best{100} & \cellcolor{tblue!18}32 & \cellcolor{tblue!60}\best{100} & \cellcolor{tblue!30}67 & \cellcolor{tblue!35}69 \\
\spireicon{claude} & Claude Sonnet 5 {\scriptsize\textcolor{gray}{(max)}} & \cellcolor{tblue!40}72 & \cellcolor{tblue!60}\best{100} & \cellcolor{tblue!48}82 & \cellcolor{tblue!37}65 & \cellcolor{tblue!56}\runnerup{95} & \cellcolor{tblue!56}\best{94} & \cellcolor{tblue!43}75 & \cellcolor{tblue!45}79 & \cellcolor{tblue!46}77 & 0 & \cellcolor{tblue!14}30 & \cellcolor{tblue!60}\best{100} & \cellcolor{tblue!16}27 & \cellcolor{tblue!35}\runnerup{83} & \cellcolor{tblue!60}\best{100} & \cellcolor{tblue!25}56 \\
\spireicon{claude} & Claude Haiku 4.5 {\scriptsize\textcolor{gray}{(12k)}} & \cellcolor{tblue!17}37 & \cellcolor{tblue!32}62 & \cellcolor{tblue!19}39 & \cellcolor{tblue!1}12 & \cellcolor{tblue!26}55 & \cellcolor{tblue!32}56 & \cellcolor{tblue!35}62 & \cellcolor{tblue!26}50 & \cellcolor{tblue!22}41 & 0 & \cellcolor{tblue!27}50 & \cellcolor{tblue!5}33 & \cellcolor{tblue!7}14 & 33 & \cellcolor{tblue!15}50 & 25 \\
\spireicon{deepseek} & DeepSeek V4.1 Flash {\scriptsize\textcolor{gray}{(max)}} & \cellcolor{tblue!39}70 & \cellcolor{tblue!51}\runnerup{88} & \cellcolor{tblue!53}89 & \cellcolor{tblue!39}69 & \cellcolor{tblue!52}90 & \cellcolor{tblue!56}\best{94} & \cellcolor{tblue!52}\runnerup{88} & \cellcolor{tblue!45}79 & \cellcolor{tblue!40}68 & 0 & \cellcolor{tblue!21}40 & \cellcolor{tblue!60}\best{100} & \cellcolor{tblue!10}18 & 50 & \cellcolor{tblue!30}67 & \cellcolor{tblue!35}69 \\
\spireicon{gemini} & Gemini 3.8 Flash {\scriptsize\textcolor{gray}{(high)}} & \cellcolor{tblue!53}\runnerup{89} & \cellcolor{tblue!60}\best{100} & \cellcolor{tblue!58}\runnerup{97} & \cellcolor{tblue!57}\runnerup{96} & \cellcolor{tblue!60}\best{100} & \cellcolor{tblue!56}\best{94} & \cellcolor{tblue!60}\best{100} & \cellcolor{tblue!55}\runnerup{93} & \cellcolor{tblue!53}89 & \cellcolor{tblue!50}\runnerup{83} & \cellcolor{tblue!40}\runnerup{70} & \cellcolor{tblue!60}\best{100} & \cellcolor{tblue!32}\runnerup{55} & \cellcolor{tblue!60}\best{100} & \cellcolor{tblue!60}\best{100} & \cellcolor{tblue!45}81 \\
\spireicon{gemini} & Gemma 4 E4B & \cellcolor{tblue!3}16 & \cellcolor{tblue!13}38 & 8 & \cellcolor{tblue!3}15 & \cellcolor{tblue!4}25 & \cellcolor{tblue!1}6 & 0 & \cellcolor{tblue!7}21 & \cellcolor{tblue!2}9 & 0 & \cellcolor{tblue!14}30 & 0 & \cellcolor{tblue!2}5 & 33 & \cellcolor{tblue!30}67 & \cellcolor{tblue!15}44 \\
\spireicon{meta} & Muse Spark 1.3 {\scriptsize\textcolor{gray}{(xhigh)}} & \cellcolor{tblue!46}79 & \cellcolor{tblue!60}\best{100} & \cellcolor{tblue!55}92 & \cellcolor{tblue!47}81 & \cellcolor{tblue!56}\runnerup{95} & \cellcolor{tblue!48}81 & \cellcolor{tblue!60}\best{100} & \cellcolor{tblue!41}71 & \cellcolor{tblue!51}86 & \cellcolor{tblue!20}33 & \cellcolor{tblue!34}60 & \cellcolor{tblue!60}\best{100} & \cellcolor{tblue!13}23 & \cellcolor{tblue!10}67 & \cellcolor{tblue!60}\best{100} & \cellcolor{tblue!45}81 \\
\spireicon{nvidia} & Nemotron 3 Nano Omni & \cellcolor{tblue!9}26 & \cellcolor{tblue!22}50 & \cellcolor{tblue!14}32 & 4 & \cellcolor{tblue!26}55 & \cellcolor{tblue!13}25 & \cellcolor{tblue!10}25 & \cellcolor{tblue!16}36 & \cellcolor{tblue!8}18 & 0 & 0 & \cellcolor{tblue!46}\runnerup{83} & \cellcolor{tblue!4}9 & 33 & 17 & \cellcolor{tblue!10}38 \\
\spireicon{openai} & GPT-6 Astra {\scriptsize\textcolor{gray}{(max)}} & \cellcolor{tblue!57}\best{96} & \cellcolor{tblue!60}\best{100} & \cellcolor{tblue!60}\best{100} & \cellcolor{tblue!60}\best{100} & \cellcolor{tblue!60}\best{100} & \cellcolor{tblue!56}\best{94} & \cellcolor{tblue!60}\best{100} & \cellcolor{tblue!60}\best{100} & \cellcolor{tblue!57}\best{95} & \cellcolor{tblue!60}\best{100} & \cellcolor{tblue!53}\best{90} & \cellcolor{tblue!60}\best{100} & \cellcolor{tblue!52}\best{86} & \cellcolor{tblue!60}\best{100} & \cellcolor{tblue!60}\best{100} & \cellcolor{tblue!45}81 \\
\spireicon{openai} & GPT-5.6 Sol {\scriptsize\textcolor{gray}{(max)}} & \cellcolor{tblue!51}87 & \cellcolor{tblue!60}\best{100} & \cellcolor{tblue!58}\runnerup{97} & \cellcolor{tblue!47}81 & \cellcolor{tblue!56}\runnerup{95} & \cellcolor{tblue!56}\best{94} & \cellcolor{tblue!60}\best{100} & \cellcolor{tblue!55}\runnerup{93} & \cellcolor{tblue!54}\runnerup{91} & \cellcolor{tblue!40}67 & \cellcolor{tblue!40}\runnerup{70} & \cellcolor{tblue!60}\best{100} & \cellcolor{tblue!27}45 & \cellcolor{tblue!35}\runnerup{83} & \cellcolor{tblue!60}\best{100} & \cellcolor{tblue!50}\runnerup{88} \\
\spireicon{openai} & GPT-5.6 Terra {\scriptsize\textcolor{gray}{(max)}} & \cellcolor{tblue!49}84 & \cellcolor{tblue!60}\best{100} & \cellcolor{tblue!56}95 & \cellcolor{tblue!47}81 & \cellcolor{tblue!60}\best{100} & \cellcolor{tblue!52}\runnerup{88} & \cellcolor{tblue!60}\best{100} & \cellcolor{tblue!55}\runnerup{93} & \cellcolor{tblue!53}89 & \cellcolor{tblue!30}50 & \cellcolor{tblue!34}60 & \cellcolor{tblue!60}\best{100} & \cellcolor{tblue!18}32 & \cellcolor{tblue!35}\runnerup{83} & \cellcolor{tblue!60}\best{100} & \cellcolor{tblue!55}\best{94} \\
\spireicon{openai} & GPT-5.6 Luna {\scriptsize\textcolor{gray}{(max)}} & \cellcolor{tblue!34}61 & \cellcolor{tblue!60}\best{100} & \cellcolor{tblue!48}82 & \cellcolor{tblue!34}62 & \cellcolor{tblue!38}70 & \cellcolor{tblue!36}62 & \cellcolor{tblue!43}75 & \cellcolor{tblue!16}36 & \cellcolor{tblue!38}66 & \cellcolor{tblue!10}17 & \cellcolor{tblue!21}40 & \cellcolor{tblue!60}\best{100} & \cellcolor{tblue!7}14 & 33 & \cellcolor{tblue!15}50 & \cellcolor{tblue!45}81 \\
\spireicon{openai} & gpt-oss-120b {\scriptsize\textcolor{gray}{(high)}} & \cellcolor{tblue!26}51 & \cellcolor{tblue!41}75 & \cellcolor{tblue!40}71 & \cellcolor{tblue!16}35 & \cellcolor{tblue!34}65 & \cellcolor{tblue!40}69 & \cellcolor{tblue!52}\runnerup{88} & \cellcolor{tblue!26}50 & \cellcolor{tblue!24}43 & 0 & \cellcolor{tblue!21}40 & \cellcolor{tblue!33}67 & \cellcolor{tblue!7}14 & 33 & \cellcolor{tblue!15}50 & \cellcolor{tblue!30}62 \\
\spireicon{grok} & Grok 4.6 {\scriptsize\textcolor{gray}{(xhigh)}} & \cellcolor{tblue!46}80 & \cellcolor{tblue!60}\best{100} & \cellcolor{tblue!56}95 & \cellcolor{tblue!50}85 & \cellcolor{tblue!52}90 & \cellcolor{tblue!44}75 & \cellcolor{tblue!60}\best{100} & \cellcolor{tblue!41}71 & \cellcolor{tblue!51}86 & \cellcolor{tblue!40}67 & \cellcolor{tblue!27}50 & \cellcolor{tblue!60}\best{100} & \cellcolor{tblue!16}27 & \cellcolor{tblue!35}\runnerup{83} & \cellcolor{tblue!60}\best{100} & \cellcolor{tblue!40}75 \\
\spireicon{zai} & GLM-5.3 {\scriptsize\textcolor{gray}{(max)}} & \cellcolor{tblue!47}82 & \cellcolor{tblue!60}\best{100} & \cellcolor{tblue!58}\runnerup{97} & \cellcolor{tblue!44}77 & \cellcolor{tblue!60}\best{100} & \cellcolor{tblue!52}\runnerup{88} & \cellcolor{tblue!52}\runnerup{88} & \cellcolor{tblue!60}\best{100} & \cellcolor{tblue!54}\runnerup{91} & \cellcolor{tblue!20}33 & \cellcolor{tblue!21}40 & \cellcolor{tblue!60}\best{100} & \cellcolor{tblue!21}36 & \cellcolor{tblue!35}\runnerup{83} & \cellcolor{tblue!60}\best{100} & \cellcolor{tblue!30}62 \\
\spireicon{zai} & GLM-5.3 Flash {\scriptsize\textcolor{gray}{(max)}} & \cellcolor{tblue!41}73 & \cellcolor{tblue!60}\best{100} & \cellcolor{tblue!51}87 & \cellcolor{tblue!42}73 & \cellcolor{tblue!52}90 & \cellcolor{tblue!48}81 & \cellcolor{tblue!52}\runnerup{88} & \cellcolor{tblue!50}86 & \cellcolor{tblue!44}75 & 0 & \cellcolor{tblue!21}40 & \cellcolor{tblue!60}\best{100} & \cellcolor{tblue!10}18 & \cellcolor{tblue!10}67 & \cellcolor{tblue!45}\runnerup{83} & \cellcolor{tblue!45}81 \\
\midrule
 & Mean over LLM agents & \cellcolor{tblue!35}64 & \cellcolor{tblue!49}85 & \cellcolor{tblue!43}75 & \cellcolor{tblue!32}58 & \cellcolor{tblue!43}78 & \cellcolor{tblue!41}70 & \cellcolor{tblue!44}76 & \cellcolor{tblue!38}68 & \cellcolor{tblue!38}66 & \cellcolor{tblue!16}28 & \cellcolor{tblue!26}48 & \cellcolor{tblue!46}83 & \cellcolor{tblue!15}26 & \cellcolor{tblue!2}62 & \cellcolor{tblue!35}72 & \cellcolor{tblue!31}64 \\
\bottomrule
\end{tabular}
\end{adjustbox}
\end{table}

\subsection{The effect of the reskin}
\label{app:puzzles-reskin}

\begin{table}[t]
\centering\small
\caption{\textbf{The effect of the reskin on puzzle accuracy.} Over the puzzles an agent has played under both namings, which is all of them for every agent. \emph{Real} and \emph{Reskin} are the percentage of those puzzles answered correctly when the game's cards, relics, potions, and powers keep their names and when they are renamed. The remaining columns are the reskin's accuracy minus the real-name accuracy, in percentage points, overall and by time horizon (\cref{tab:puzzle-taxonomy}): red where the reskin is lower, blue where it is higher. The grey row gives the number of puzzles in the column.}
\label{tab:puzzle-reskin}
\begin{adjustbox}{max width=\linewidth}
\small
\begin{tabular}{c@{\hspace{3pt}}l rr rrrr}
\toprule
 & & \multicolumn{2}{c}{Accuracy (\%)} & \multicolumn{4}{c}{Reskin $-$ Real} \\
\cmidrule(lr){3-4} \cmidrule(lr){5-8}
 & & Real & Reskin & Overall & One turn & One fight & Rest of the run \\
 & Agent & \textcolor{gray}{123} & \textcolor{gray}{123} & \textcolor{gray}{123} & \textcolor{gray}{58} & \textcolor{gray}{40} & \textcolor{gray}{25} \\
\midrule
\spireicon{qwen} & Qwen3.8 27B {\scriptsize\textcolor{gray}{(xhigh)}} & 53 & 54 & \cellcolor{tblue!2}$+1$ & \cellcolor{tred!3}$-2$ & \cellcolor{tblue!20}$+10$ & \cellcolor{tred!16}$-8$ \\
\spireicon{qwen} & Qwen3.6 35B-A3B & 33 & 26 & \cellcolor{tred!13}$-7$ & \cellcolor{tred!3}$-2$ & \cellcolor{tred!10}$-5$ & \cellcolor{tred!40}$-20$ \\
\spireicon{qwen} & Qwen3.5 122B-A10B & 20 & 22 & \cellcolor{tblue!3}$+2$ & 0 & \cellcolor{tred!5}$-2$ & \cellcolor{tblue!24}$+12$ \\
\spireicon{claude} & Claude Fable 5.1 {\scriptsize\textcolor{gray}{(max)}} & 87 & \runnerup{85} & \cellcolor{tred!3}$-2$ & \cellcolor{tred!3}$-2$ & \cellcolor{tred!5}$-2$ & 0 \\
\spireicon{claude} & Claude Opus 5 {\scriptsize\textcolor{gray}{(max)}} & 88 & 82 & \cellcolor{tred!11}$-6$ & \cellcolor{tred!7}$-3$ & \cellcolor{tred!5}$-2$ & \cellcolor{tred!32}$-16$ \\
\spireicon{claude} & Claude Sonnet 5 {\scriptsize\textcolor{gray}{(max)}} & 72 & 71 & \cellcolor{tred!3}$-2$ & 0 & \cellcolor{tred!20}$-10$ & \cellcolor{tblue!16}$+8$ \\
\spireicon{claude} & Claude Haiku 4.5 {\scriptsize\textcolor{gray}{(12k)}} & 36 & 39 & \cellcolor{tblue!7}$+3$ & \cellcolor{tblue!21}$+10$ & \cellcolor{tred!10}$-5$ & 0 \\
\spireicon{deepseek} & DeepSeek V4.1 Flash {\scriptsize\textcolor{gray}{(max)}} & 74 & 66 & \cellcolor{tred!16}$-8$ & \cellcolor{tred!17}$-9$ & \cellcolor{tred!5}$-2$ & \cellcolor{tred!32}$-16$ \\
\spireicon{gemini} & Gemini 3.8 Flash {\scriptsize\textcolor{gray}{(high)}} & \runnerup{94} & \runnerup{85} & \cellcolor{tred!20}$-10$ & \cellcolor{tred!10}$-5$ & \cellcolor{tred!20}$-10$ & \cellcolor{tred!40}$-20$ \\
\spireicon{gemini} & Gemma 4 E4B & 16 & 16 & 0 & \cellcolor{tblue!7}$+3$ & \cellcolor{tred!10}$-5$ & 0 \\
\spireicon{meta} & Muse Spark 1.3 {\scriptsize\textcolor{gray}{(xhigh)}} & 81 & 76 & \cellcolor{tred!10}$-5$ & \cellcolor{tred!7}$-3$ & \cellcolor{tred!10}$-5$ & \cellcolor{tred!16}$-8$ \\
\spireicon{nvidia} & Nemotron 3 Nano Omni & 25 & 26 & \cellcolor{tblue!2}$+1$ & \cellcolor{tblue!7}$+3$ & \cellcolor{tred!10}$-5$ & \cellcolor{tblue!8}$+4$ \\
\spireicon{openai} & GPT-6 Astra {\scriptsize\textcolor{gray}{(max)}} & \best{97} & \best{95} & \cellcolor{tred!3}$-2$ & \cellcolor{tred!3}$-2$ & \cellcolor{tred!5}$-2$ & 0 \\
\spireicon{openai} & GPT-5.6 Sol {\scriptsize\textcolor{gray}{(max)}} & 88 & \runnerup{85} & \cellcolor{tred!5}$-2$ & \cellcolor{tblue!7}$+3$ & \cellcolor{tred!20}$-10$ & \cellcolor{tred!8}$-4$ \\
\spireicon{openai} & GPT-5.6 Terra {\scriptsize\textcolor{gray}{(max)}} & 86 & 82 & \cellcolor{tred!8}$-4$ & \cellcolor{tred!3}$-2$ & \cellcolor{tred!5}$-2$ & \cellcolor{tred!24}$-12$ \\
\spireicon{openai} & GPT-5.6 Luna {\scriptsize\textcolor{gray}{(max)}} & 70 & 53 & \cellcolor{tred!34}$-17$ & \cellcolor{tred!38}$-19$ & \cellcolor{tred!35}$-18$ & \cellcolor{tred!24}$-12$ \\
\spireicon{openai} & gpt-oss-120b {\scriptsize\textcolor{gray}{(high)}} & 50 & 51 & \cellcolor{tblue!2}$+1$ & \cellcolor{tred!10}$-5$ & \cellcolor{tblue!20}$+10$ & 0 \\
\spireicon{grok} & Grok 4.6 {\scriptsize\textcolor{gray}{(xhigh)}} & 83 & 76 & \cellcolor{tred!13}$-7$ & \cellcolor{tred!7}$-3$ & \cellcolor{tred!25}$-12$ & \cellcolor{tred!8}$-4$ \\
\spireicon{zai} & GLM-5.3 {\scriptsize\textcolor{gray}{(max)}} & 85 & 79 & \cellcolor{tred!11}$-6$ & 0 & \cellcolor{tred!20}$-10$ & \cellcolor{tred!24}$-12$ \\
\spireicon{zai} & GLM-5.3 Flash {\scriptsize\textcolor{gray}{(max)}} & 73 & 72 & \cellcolor{tred!2}$-1$ & \cellcolor{tblue!7}$+3$ & \cellcolor{tred!5}$-2$ & \cellcolor{tred!16}$-8$ \\
\midrule
 & Mean over agents & 66 & 62 & \cellcolor{tred!7}$-3$ & \cellcolor{tred!3}$-2$ & \cellcolor{tred!9}$-5$ & \cellcolor{tred!12}$-6$ \\
\bottomrule
\end{tabular}
\end{adjustbox}
\end{table}

The reskin renames every card, relic, potion, and power, so that an agent must work from the text in front of it and cannot draw on what it remembers of the game.
\Cref{tab:puzzle-reskin} sets the two arms side by side, over all 123 puzzles for every agent.

Renaming costs a little accuracy for most agents.
The mean over the agents falls from 66\% to 62\%, and 14 of the 20 agents score lower under the reskin.
The cost is uneven: GPT-5.6 Luna loses 17 points, Gemini 3.8 Flash 10, and DeepSeek V4.1 Flash 8, while GLM-5.3 Flash loses a single point, Gemma~4 E4B scores the same, and Claude Haiku 4.5, Qwen3.8 27B, Qwen3.5 122B-A10B, Nemotron 3 Nano Omni, and gpt-oss-120b score slightly higher.

The cost grows with the time horizon of the decision.
Under the reskin the agents lose 2 points on the one-turn puzzles, 5 on those that need a plan across a fight, and 6 on those whose decision pays after the fight.
The difference is not broken down by skill: most skills hold between three and eight puzzles, so one play that changes outcome moves a cell by 13 to 33 points.
A one-turn puzzle states everything it needs: the cards are in hand, their text is on them, and the answer is a calculation.
The further ahead a decision looks, the more it rests on judging what a card or relic is worth, and there an agent that knows the game can lean on what is known about that card; renamed, it has only the description.
The cost of the reskin over whole runs is in \cref{app:ablation-reskin}.

\paragraph{A rule that is not on the card.}
Two puzzles turn on a behaviour of the game that its text does not state.
The relic \emph{Sundial} gives 2 energy every third time the draw pile is shuffled.
When a card draws two cards from a draw pile that holds one, the game shuffles twice, once for each draw that finds the pile empty, so a deck of two or three cards advances \emph{Sundial} faster than its description suggests, and the loop that this makes possible is the answer.
A player of the game may know this; nothing shown to the agent says it.
Under the game's real names, then, a correct answer may be recalled and not worked out, and these two puzzles are the ones where that concern is sharpest.

The reskin arm is the control, and here it does not show a reliance on recall.
The harder of the two puzzles is answered by 13 of the 20 agents under the real names and by 13 under the reskin.
What the reskinned agents do instead is experiment: the relic's counter is part of the state the agent is shown, and an agent can play one card and read it.
One agent wrote, under the reskin, that the line ``hinged on an implementation detail I couldn't know up front'', namely whether a draw from an empty pile advances the counter, and that it settled the matter by playing one card of each kind before committing.
The second puzzle is answered by 15 agents under the real names and by 15 under the reskin.
A puzzle of this kind therefore measures recall only in an agent that cannot, or does not, test the rule in play.

\subsection{How the agents fail}
\label{app:puzzles-unsolved}

\begin{table}[t]
\centering\small
\caption{\textbf{How each agent's puzzles ended.} As a number of its plays (each of the 123 puzzles under the game's real names, under the reskin, or both; \emph{Plays} is how many that is). \emph{Below target}: the fight was won, but short of the HP, maximum HP, or relic counters that the best play keeps. \emph{Wrong pick}: a reward or shop was left alive with the wrong card or relic. \emph{Died}: the player was killed. The last column is the share of the agent's failures that were deaths.}
\label{tab:puzzle-outcomes}
\begin{adjustbox}{max width=\linewidth}
\small
\begin{tabular}{c@{\hspace{3pt}}l r r rrr r}
\toprule
 & & & & \multicolumn{3}{c}{Failed} & \\
\cmidrule(lr){5-7}
 & Agent & Plays & Correct & Below target & Wrong pick & Died & Deaths (\% of failures) \\
\midrule
\spireicon{qwen} & Qwen3.8 27B {\scriptsize\textcolor{gray}{(xhigh)}} & 246 & 131 & 41 & 12 & 62 & 54 \\
\spireicon{qwen} & Qwen3.6 35B-A3B & 246 & 72 & 39 & 19 & 116 & 67 \\
\spireicon{qwen} & Qwen3.5 122B-A10B & 246 & 52 & 40 & 20 & 134 & 69 \\
\spireicon{claude} & Claude Fable 5.1 {\scriptsize\textcolor{gray}{(max)}} & 246 & 212 & 22 & 6 & 6 & 18 \\
\spireicon{claude} & Claude Opus 5 {\scriptsize\textcolor{gray}{(max)}} & 246 & 209 & 21 & 7 & 9 & 24 \\
\spireicon{claude} & Claude Sonnet 5 {\scriptsize\textcolor{gray}{(max)}} & 246 & 176 & 29 & 8 & 33 & 47 \\
\spireicon{claude} & Claude Haiku 4.5 {\scriptsize\textcolor{gray}{(12k)}} & 246 & 92 & 39 & 19 & 96 & 62 \\
\spireicon{deepseek} & DeepSeek V4.1 Flash {\scriptsize\textcolor{gray}{(max)}} & 246 & 172 & 29 & 10 & 35 & 47 \\
\spireicon{gemini} & Gemini 3.8 Flash {\scriptsize\textcolor{gray}{(high)}} & 246 & \runnerup{220} & 16 & 3 & 7 & 27 \\
\spireicon{gemini} & Gemma 4 E4B & 246 & 40 & 34 & 13 & 159 & 77 \\
\spireicon{meta} & Muse Spark 1.3 {\scriptsize\textcolor{gray}{(xhigh)}} & 246 & 194 & 27 & 5 & 20 & 38 \\
\spireicon{nvidia} & Nemotron 3 Nano Omni & 246 & 63 & 38 & 16 & 129 & 70 \\
\spireicon{openai} & GPT-6 Astra {\scriptsize\textcolor{gray}{(max)}} & 246 & \best{236} & 4 & 3 & 3 & 30 \\
\spireicon{openai} & GPT-5.6 Sol {\scriptsize\textcolor{gray}{(max)}} & 246 & 213 & 21 & 3 & 9 & 27 \\
\spireicon{openai} & GPT-5.6 Terra {\scriptsize\textcolor{gray}{(max)}} & 246 & 207 & 26 & 2 & 11 & 28 \\
\spireicon{openai} & GPT-5.6 Luna {\scriptsize\textcolor{gray}{(max)}} & 246 & 151 & 41 & 10 & 44 & 46 \\
\spireicon{openai} & gpt-oss-120b {\scriptsize\textcolor{gray}{(high)}} & 246 & 125 & 38 & 12 & 71 & 59 \\
\spireicon{grok} & Grok 4.6 {\scriptsize\textcolor{gray}{(xhigh)}} & 246 & 196 & 30 & 5 & 15 & 30 \\
\spireicon{zai} & GLM-5.3 {\scriptsize\textcolor{gray}{(max)}} & 246 & 201 & 22 & 7 & 16 & 36 \\
\spireicon{zai} & GLM-5.3 Flash {\scriptsize\textcolor{gray}{(max)}} & 246 & 179 & 33 & 6 & 28 & 42 \\
\midrule
 & Mean over agents & 246.0 & 157.1 & 29.5 & 9.3 & 50.1 & 56 \\
\bottomrule
\end{tabular}
\end{adjustbox}
\end{table}

\paragraph{Dying and winning badly.}
A puzzle can be failed in three ways, and they are not equally bad (\cref{tab:puzzle-outcomes}).
The player can die, which in a run ends it.
It can win the fight but finish short of what the best play keeps, which in a run costs HP or a later payout.
Or, in a reward or shop puzzle, it can leave with the wrong card or relic.

The weak agents mostly die: deaths are 67\% to 77\% of the failures of the four lowest-scoring agents, and Gemma~4 E4B dies in 159 of its 246 plays.
The strong agents rarely do.
GPT-6 Astra dies three times, Claude Fable 5.1 six, and Gemini 3.8 Flash seven, and deaths are 18\% to 30\% of the failures of the five highest-scoring agents.
The number of fights won below target changes less from the bottom of the table to the top than the number of deaths does: it is 34 to 40 for the four weakest agents and 4 to 22 for the five strongest, where deaths fall from between 116 and 159 to between 3 and 9.
A stronger agent therefore mostly converts deaths into wins, and less often converts a win into the best win.
Over the five agents with the highest accuracy, 84 of 140 failures are fights won below target, 34 are deaths, and 22 are a wrong pick.

\paragraph{Value carried to the next fight.}
Many of these targets are about what a fight leaves behind, and the clearest case is a relic with a counter.
These are the \emph{Stall} puzzles, on which the agents answer 26\% (\cref{tab:puzzle-skills}).
Several relics count events and pay out at a fixed count: \emph{Pen Nib} doubles the damage of every tenth attack, \emph{Nunchaku} gives one energy every tenth attack, \emph{Happy Flower} gives one energy every third turn, and \emph{Incense Burner} makes the player nearly invulnerable for one turn in every six.
The count is kept from one fight to the next.
Where a fight ends therefore decides what the next one starts with: a \emph{Pen Nib} left on nine makes the first attack of the next fight a doubled one, and a \emph{Pen Nib} that fires on a blow that would have killed anyway is spent on nothing.
A strong human player manages these counters deliberately, and will play one attack fewer, or let an easy fight run a turn longer, to end it with the relics charged.
The same reasoning applies to HP that is not healed after the fight, and to maximum HP gained from a kill.
A puzzle is a single room, so nothing in it uses what is carried out; the agent is not told this, and in a run there is always a next fight.

\paragraph{Puzzles almost no agent solves.}
Each puzzle is played 40 times, twice by each of 20 agents.
No puzzle is answered correctly on every play, and none is failed on every play, but two are answered correctly on three plays or fewer, and GPT-6 Astra is the only agent to answer either of them in both arms.
In both, every agent wins the fight; what almost no agent does is wait.

In the first, the player holds \emph{Reaper}, an attack that heals for the damage it deals, and \emph{Demon Form+}, which adds 3 Strength every turn.
The kill is on offer by the second turn.
Taking it then leaves the player badly hurt, because the heal is still small, whereas absorbing the enemy's hits for three more turns lets one swing of \emph{Reaper} kill the enemy and restore more HP than the wait cost.
Thirty-seven of the 40 plays kill on the first or second turn (17 on the first, 20 on the second) and finish on 16 to 21 HP, against 24 for the patient line.
GPT-6 Astra waits until the fifth turn in both arms, and GLM-5.3 does under the real names.
Nothing in this puzzle is hidden or uncertain: the enemy repeats one attack, the growth is printed on the card, and the arithmetic is a few additions.
The agents can do that arithmetic, since they solve harder sums elsewhere in the suite.
We read the failure as one of not asking the question: a kill that is available ends the search.

The second is a counter puzzle.
The player has all four relics described above for the \emph{Stall} puzzles, and a deck of five cards, so that every turn deals the same hand, and the enemy can be killed on the second turn with the \emph{Bludgeon} that is always in hand.
Ending the fight on the fifth turn with nine attacks played leaves all four relics one short of paying out.
Thirty-six of the 40 plays end it on the third turn, at full HP, with \emph{Pen Nib} and \emph{Nunchaku} well short of nine; GPT-6 Astra plays the five-turn line in one of its two attempts, the only correct play of this puzzle.

\paragraph{Puzzles few agents solve.}
Several more puzzles are answered correctly by few of the 20 agents.
\begin{itemize}[leftmargin=*,nosep]
\item \emph{A second counter puzzle.}
Eight attacks and then a \emph{Bludgeon} kill an 80-HP enemy exactly and leave \emph{Pen Nib} on nine.
In 35 of the 40 plays the agent plays a ninth attack, so that the \emph{Bludgeon} is the doubled tenth: it wins on the same turn, deals twice what was needed, and resets the counter.
GPT-6 Astra meets the target in both arms.
The only other agents to meet it, once each, are three of the weakest, and by accident: two played one attack fewer than they could have and afterwards rated the fight as simple, one of them as having ``no meaningful strategic decisions'', and the third described its play as ``maximizing damage immediately''.
One of the strongest saw the choice and did not take it, writing after the fight that it was ``slightly unsure whether holding a Strike back on turn 2 to manipulate counters \ldots{} would have been meaningfully better''.
\item \emph{A puzzle that destroys the deck.}
The hand must be filled to its limit of ten cards, in the only order of draw effects that reaches ten, so that one potion exhausts everything except the cards of a loop that then repeats without end.
It is answered on 9 of its 40 plays; the other 31 die attempting the loop.
\item \emph{A kill worth postponing.}
Three \emph{Strikes} kill the enemy on the first turn at full HP.
Blocking that turn instead costs nothing and brings \emph{Feed}, which raises maximum HP for the rest of the run when it lands the killing blow, into the next hand.
Four agents wait for it in at least one arm; 25 of the 40 plays take the kill on the first turn.
\item \emph{A fight with no heal after it.}
One \emph{Strike} and two \emph{Defends} every turn wins without taking damage; attacking harder wins sooner and takes damage that, in this puzzle, is not healed afterwards.
The patient line is played on 5 of the 40 plays, three of them by two of the weakest agents.
\item \emph{Two potions that only pay together.}
A \emph{Speed Potion} gives Dexterity that is taken back at the end of the turn, unless an \emph{Ancient Potion} is drunk with it, whose Artifact cancels the taking back.
With both, every later turn can be blocked in full; without the pair the fight cannot be won at full HP.
Three agents find it, on 4 of the 40 plays.
\item \emph{A curse worth keeping.}
A shop offers the removal of a curse or a relic that draws two more cards at the start of each fight, with gold for one of them.
The curse is removed on 27 of the 40 plays.
\item \emph{A turn spent for something kept.}
In two puzzles the enemy can be killed on the first turn, no damage is at stake, and one more turn brings a card that leaves something behind: a potion, or a permanent gain on the card itself.
Thirty of the 40 plays of one and 33 of the 40 of the other take the kill at once; five and six plays wait long enough to meet the target.
\end{itemize}

\paragraph{What this says about the agents.}
These are failures of planning inside a single room, with the whole future in view, and they match what the agents do over a whole run.
There, they fight fewer elites and rest more than strong human players (\cref{app:play-act1}), and they under-invest whenever the payoff is delayed (\cref{sec:discussion}).
The puzzles show that this is not only a response to the uncertainty of a run: the same preference for the immediate, safe outcome appears when the delayed payoff is certain and can be computed.
What the strongest agents still get wrong is rarely how to win the fight in front of them.
It is declining a win that is on the table for something worth more: a later payout, HP kept for the next fight, or a relic left ready.
The one agent that does so consistently, GPT-6 Astra, is also the agent that reaches the highest mean floor in \cref{tab:main-results}.

\subsection{Which skills go with success in runs}
\label{app:puzzles-runs}

\begin{figure}[t]
\centering
\includegraphics[width=\linewidth]{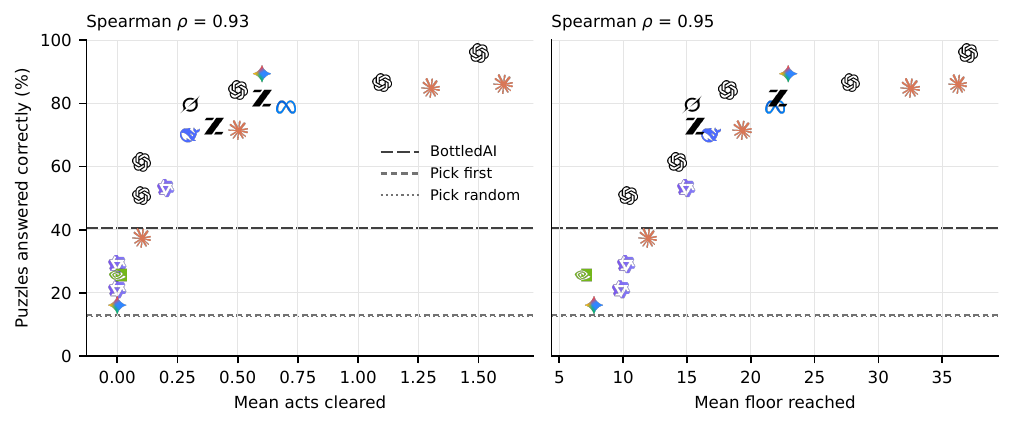}
\caption{\textbf{Puzzle accuracy against performance in runs.}
Each point is one agent, drawn as its provider's icon: its mean acts cleared (left) and mean floor reached (right) in \cref{tab:main-results} against the percentage of all puzzles it answered correctly (each puzzle scored by the mean of the arms played).
The dashed lines are the puzzle accuracy of BottledAI and of the two baseline policies.}
\label{fig:puzzle-runs}
\end{figure}

\begin{table}[t]
\centering\small
\caption{\textbf{Puzzle accuracy against performance in runs.} Over the 20 agents of \cref{tab:puzzle-skills}. Each entry is Spearman's $\rho$ between the agents' accuracy on a set of puzzles and their mean floor reached or mean acts cleared in \cref{tab:main-results}; an entry in grey is not distinguishable from zero (permutation test, $p \geq 0.05$). \emph{Partial} is the $\rho$ with mean floor that remains when accuracy on all the other puzzles is held fixed.}
\label{tab:puzzle-correlation}
\begin{adjustbox}{max width=\linewidth}
\small
\begin{tabular}{l r rr r}
\toprule
 & & \multicolumn{2}{c}{$\rho$ with} & \\
\cmidrule(lr){3-4}
Puzzles & $n$ & Mean floor & Acts cleared & Partial \\
\midrule
All puzzles & 123 & 0.95 & 0.93 & -- \\
\midrule
\rowcolor{vlightgray}
\textit{One turn} & 58 & 0.95 & 0.94 & 0.58 \\
\quad Pay & 4 & 0.83 & 0.83 & 0.02 \\
\quad Order & 19 & 0.95 & 0.93 & 0.37 \\
\quad Engine & 13 & 0.90 & 0.90 & 0.14 \\
\quad Rules & 10 & 0.93 & 0.93 & 0.42 \\
\quad Count & 8 & 0.92 & 0.89 & 0.58 \\
\quad Enemy & 4 & 0.85 & 0.87 & $-0.19$ \\
\midrule
\rowcolor{vlightgray}
\textit{One fight} & 40 & 0.96 & 0.95 & 0.63 \\
\quad Tank & 7 & 0.93 & 0.93 & 0.50 \\
\quad Plan & 22 & 0.97 & 0.97 & 0.72 \\
\quad Engine & 3 & 0.77 & 0.77 & $-0.47$ \\
\quad Rules & 5 & 0.65 & 0.65 & $-0.12$ \\
\quad Hail-Mary & 3 & 0.78 & 0.77 & 0.12 \\
\midrule
\rowcolor{vlightgray}
\textit{Rest of the run} & 25 & 0.87 & 0.86 & $-0.18$ \\
\quad Stall & 11 & 0.90 & 0.87 & 0.03 \\
\quad Pick-up & 3 & 0.90 & 0.89 & 0.19 \\
\quad Card & 3 & 0.79 & 0.80 & 0.10 \\
\quad Shop & 8 & 0.62 & 0.63 & $-0.33$ \\
\bottomrule
\end{tabular}
\end{adjustbox}
\end{table}

A puzzle tests one decision with everything in view; a run asks for hundreds of them, under uncertainty, with consequences that carry forward.
\Cref{fig:puzzle-runs} and \cref{tab:puzzle-correlation} ask how far the first predicts the second, by ranking the agents on each set of puzzles and on their mean floor and mean acts cleared in \cref{tab:main-results}.

Puzzle accuracy and run performance order the agents almost identically ($\rho = 0.95$ with mean floor and $0.93$ with mean acts cleared, over all puzzles).
The relation is not a straight line (\cref{fig:puzzle-runs}).
No agent that answers fewer than 85\% of the puzzles clears more than 0.7 acts on average, and the four agents that clear more than one act all answer at least 85\%.

The two horizons inside a fight follow run performance equally closely ($\rho$ with mean floor of $0.95$ for one turn and $0.96$ for one fight), and each still says something about an agent's runs once its accuracy on all the other puzzles is held fixed (partial $\rho$ of $0.58$ and $0.63$).
The last horizon follows it less closely ($0.87$) and adds nothing once the others are held fixed.
It is not uniform: the \emph{Stall} and \emph{Pick-up} puzzles track the runs ($0.90$ each), and the shop decisions do so least ($0.62$).
This does not show that deck building matters little in a run.
The reward and shop puzzles ask for one purchase or one relic with the deck in view, and what a run asks for is a deck built over fifty floors for fights not yet seen (\cref{app:play-deck}).

By skill (\cref{fig:puzzle-skills-runs}), \emph{Plan} ($0.97$), \emph{Order} ($0.95$), the one-turn \emph{Rules} ($0.93$), and \emph{Tank} ($0.93$) follow run performance most closely.
\emph{Plan}, \emph{Count}, \emph{Tank}, and \emph{Rules} are the skills that still say most once the other puzzles are held fixed (partial $\rho$ of $0.72$, $0.58$, $0.50$, and $0.42$).
The one-turn \emph{Engine} puzzles are strongly correlated on their own ($0.90$) but add little beyond the rest (partial $\rho = 0.14$).
Among the deck-building skills, \emph{Pick-up} tracks the runs ($0.90$), and \emph{Shop} least ($0.62$).

\begin{figure}[t]
\centering
\includegraphics[width=\linewidth]{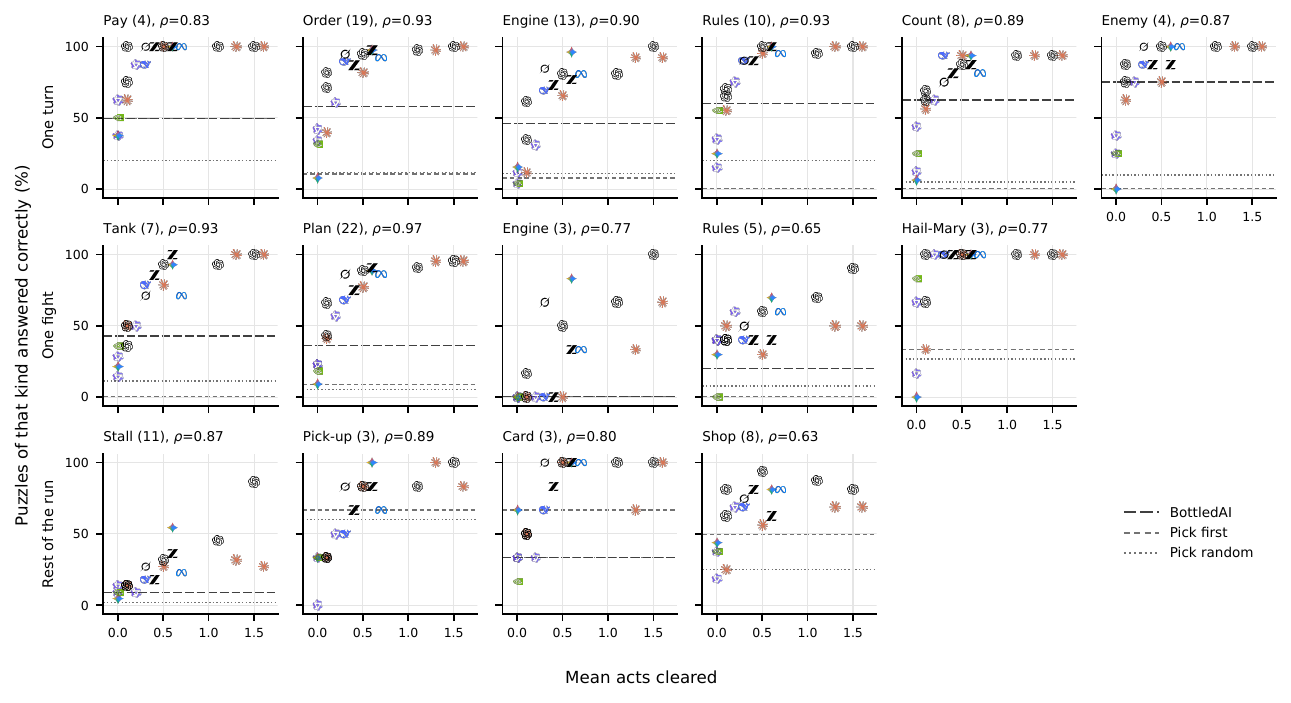}
\caption{\textbf{Puzzle accuracy by skill against mean acts cleared.}
One row per time horizon and one panel per skill of \cref{tab:puzzle-taxonomy}, with the number of puzzles in brackets; each point is one agent, drawn as its provider's icon, and $\rho$ is Spearman's rank correlation.
The dashed lines are the accuracy of BottledAI and of the two baseline policies on the puzzles of that skill.
The skills of the first two rows rise with the acts cleared; in the last row \emph{Stall} and \emph{Pick-up} do, and \emph{Shop} least.}
\label{fig:puzzle-skills-runs}
\end{figure}

Solving the puzzles is not enough.
Gemini 3.8 Flash answers 89\% of the puzzles, the second most of any agent, and reaches a mean floor of 22.9; Claude Fable 5.1 answers 86\% and reaches 36.2.
What separates the agents at the top of \cref{tab:main-results} is therefore not only the single decisions that a puzzle can isolate.

These are rank correlations over 20 agents, and the skills are correlated with one another, so the partial values in particular are unstable and the ordering among the fight skills is suggestive only.

\FloatBarrier
\FloatBarrier
\section[\texorpdfstring{Expert-decision benchmark}{Expert-decision benchmark}]{Expert-decision benchmark}
\label{app:expert-decisions}

The puzzles of \cref{app:puzzles} have definitive authored answers, but primarily span the shorter time-frame planning and decision making aspects of the game (micro).
This benchmark covers decisions affecting longer time-frames (macro) by comparison with expert play in real games.
Each item is one decision from a winning run of an expert player, recovered from run logs of their games.
An agent is presented with the same scenario seen by the expert player and scored on whether it makes the choice the expert made.
An expert's choice is not \emph{necessarily} the best one, so agreement is a measure of how closely an agent's choices follow expert play, not of whether each choice is correct.

\subsection{Items}
\label{app:expert-decisions-items}

\paragraph{Source.}
The items are taken from the run records of the four expert players of \cref{app:human-baselines}: 1{,}070 runs at the highest difficulty that end in a win over the final boss of Act~4.
We filter down to winning runs to ensure we do not compare the model against a run-losing macro mistake from an expert player, which would score it against a poor decision.
A run log record lists, floor by floor, the cards offered and taken, the purchases and removals, the event options, the rest-site choices, the relics and potions gained, and the deck and relics at the end of the run.

\paragraph{Replay.}
Each run is replayed in the released game from its seed, with the fights skipped and the recorded choice made at every decision.
At each decision the state the game reports is saved, with the history of the floors before it.
This allows us to extract the game state to show to the agent later, mirroring what the human expert saw when they made their decision.
A screen whose content the record also holds is checked against it before it is kept: the cards of a card reward, the three boss relics, a shop's stock, and the event met.

\paragraph{Repair.}
\label{app:expert-decisions-deck}
A replay can diverge from the expert's actual deck due to random effects not accounted for in the game seed which are not recorded in the run log.
Examples include events that upgrade a random card, and player choices not recorded such as which cards to remove on pick-up of the Empty Cage relic.
We therefore repair the replay by comparing it against the deck held according to the run log's record.
What the record does not name is solved from the deck at the end of the run: the cards it has no room for were the ones removed, and a card it holds one more of was the one copied, where these were not recorded.

The deck carried to the last floor must then equal the recorded final deck, and a run's items are used only where the two agree.
This holds for 1{,}058 of the 1{,}070 runs.
Of the 87{,}913 saved states, 84{,}106 already showed the recorded deck, 2{,}936 were corrected, and the remaining 871 were removed.
Filtering on the replay alone, without this step, would be biased as it would drop the later floors exclusively of runs which took relics that have effects which are not directly recorded in the run log.

\paragraph{Decision categories.}
There are nine categories: the blessing chosen at the start of a run (Neow), an event option, the next map node, a card reward, a boss's card reward, a boss relic, a rest-site option, the card upgraded at a rest site, and the purchases of a shop visit.

\subsection{The draw}
\label{app:expert-decisions-draw}

An evaluation asks 25 items of each category for each of the four characters, 900 items in all, so each category is scored on 100 items.
Within a category the acts hold shares in proportion to the number of floors on which the category occurs, so that every floor weighs alike.
Within an act the draw takes the player with the fewest items of the category so far, then the run with the fewest, then the floor with the fewest.
Each player therefore gives an equal share of a category (8 or 9 of the 25 where a character has three players, 6 or 7 where it has four), except among card upgrades, where a player with few upgrades in an act leaves them to the others.
No run gives two items of a category other than card upgrades, and the items of a character come from 106 to 122 runs.
The draw is nested: a larger number per category keeps every item of a smaller one, so the set can be extended without asking any item again.

\subsection{What the agent is given, and scoring}
\label{app:expert-decisions-scoring}

In this benchmark the agent use the harness from the main results (\cref{app:methods-harness}), with notes and cross-run memory off, under the game's real names.
It is given the saved state as the first state of a new floor, with the summary of the earlier floors that a run would carry, and it answers with the same tools as in a run.
Each item is a separate conversation, and nothing is carried from one item to another.
The agent is not told that the state comes from an expert's run.

An answer is right when it is the expert's: the same card or a skip at a card reward, the same relic, the same node, the same option.
At a rest site where the expert upgraded a card, the card is scored as a separate item on the upgrade screen.
A shop visit is right when the agent leaves with exactly the expert's purchases.
An agent's score is the mean over the nine categories of its share right, with the four characters pooled within a category.

Three baselines answer the same items without a model.
\emph{Pick random option} (uniform random choice) takes each option of a screen with equal chance, a skip or leave included where the screen offers one, and its score is the expected share right, computed exactly from the options of each item.
On the upgrade screen, which lists every copy of a card, it picks among the card names, since any copy of the expert's card is right.
For shop visits, where the expected share depends on a sequence of purchases, it is estimated from 64 simulated visits per item.
\emph{Pick first option} always takes the first.
BottledAI~\citep{bottledai} answers with its strategy for the character, in the variant that goes on to the Heart (\emph{w/ Heart} in \cref{tab:main-results}), since every expert run is a Heart win; it therefore takes the keys, recalling at rest sites for the ruby key.

\subsection{Results}
\label{app:expert-decisions-results}

\Cref{tab:expert-decisions} gives the results for 19 models and three baselines, each of which has answered all 900 items.
Gemma~4 E4B answered from a different deployment of the model (\texttt{Gemma-4-E4B-it}) than the one used for its runs in \cref{tab:main-results} (\texttt{gemma4:e4b-highcontext}); we treat the two as the same model.

\begin{table}[t]
\centering
\caption{\textbf{Expert-decision benchmark.}
Share of items (\%) on which an agent makes the expert's choice, by decision category, and the mean of the nine categories.
The grey numbers under the column headings are the items in each column.
Cells are white at the expected score of uniform random choice in that column, blue above it, and red below it, darker in proportion to the distance from it on the same scale on both sides, with blue darkest at 100; the best and second-best LLM agents of a column are marked.}
\label{tab:expert-decisions}
\providecommand{\spireicon}[1]{\raisebox{-0.25ex}{\includegraphics[height=1.7ex]{icons/#1.pdf}}}
\begin{adjustbox}{max width=\textwidth}
\small
\begin{tabular}{@{}c@{\hspace{3pt}}l r rrrrrrrrr@{}}
\toprule
 & & Mean & Neow & Event & Path & Card & Boss card & Boss relic & Rest & Upgrade & Shop \\
 & Agent & \textcolor{gray}{900} & \textcolor{gray}{100} & \textcolor{gray}{100} & \textcolor{gray}{100} & \textcolor{gray}{100} & \textcolor{gray}{100} & \textcolor{gray}{100} & \textcolor{gray}{100} & \textcolor{gray}{100} & \textcolor{gray}{100} \\
\midrule
\multicolumn{12}{@{}l}{\textit{Baseline policies}} \\
\spireicon{left} & Pick first option & \cellcolor{tred!4}22.1 & \cellcolor{tred!7}17 & \cellcolor{tblue!11}54 & \cellcolor{tred!4}41 & \cellcolor{tred!2}22 & \cellcolor{tred!2}23 & \cellcolor{tred!2}23 & \cellcolor{tred!22}15 & \cellcolor{tred!8}1 & \cellcolor{tblue!1}3 \\
\spireicon{random} & Pick random option & 26.8 & 26 & 44 & 45 & 25 & 25 & 25 & 38 & 12 & 2 \\
\midrule
\multicolumn{12}{@{}l}{\textit{Symbolic agent (Bottled AI)}} \\
\spireicon{bottled_ai} & BottledAI w/ Heart & \cellcolor{tblue!9}38.3 & \cellcolor{tblue!1}27 & \cellcolor{tblue!16}59 & \cellcolor{tblue!8}52 & \cellcolor{tblue!13}41 & \cellcolor{tblue!10}37 & \cellcolor{tblue!12}40 & \cellcolor{tblue!13}51 & \cellcolor{tblue!9}26 & \cellcolor{tblue!6}12 \\
\midrule
\multicolumn{12}{@{}l}{\textit{LLM agents}} \\
\spireicon{qwen} & Qwen3.8 27B {\scriptsize\textcolor{gray}{(xhigh)}} & \cellcolor{tblue!8}37.1 & \cellcolor{tred!7}17 & \cellcolor{tblue!20}62 & \cellcolor{tblue!9}53 & \cellcolor{tblue!7}34 & \cellcolor{tblue!9}36 & \cellcolor{tblue!16}45 & \cellcolor{tblue!10}48 & \cellcolor{tblue!13}32 & \cellcolor{tblue!3}7 \\
\spireicon{qwen} & Qwen3.6 35B-A3B & \cellcolor{tblue!4}31.1 & \cellcolor{tred!6}18 & \cellcolor{tblue!15}58 & \cellcolor{tblue!8}52 & \cellcolor{tblue!5}31 & \cellcolor{tblue!2}28 & \cellcolor{tblue!6}33 & 38 & \cellcolor{tblue!6}21 & 1 \\
\spireicon{claude} & Claude Fable 5.1 {\scriptsize\textcolor{gray}{(max)}} & \cellcolor{tblue!13}43.1 & \cellcolor{tred!2}23 & \cellcolor{tblue!20}62 & \cellcolor{tblue!5}49 & \cellcolor{tblue!24}55 & \cellcolor{tblue!24}55 & \cellcolor{tblue!21}51 & \cellcolor{tblue!4}42 & \cellcolor{tblue!20}42 & \cellcolor{tblue!4}9 \\
\spireicon{claude} & Claude Opus 5 {\scriptsize\textcolor{gray}{(max)}} & \cellcolor{tblue!10}39.3 & 26 & \cellcolor{tblue!16}59 & \cellcolor{tblue!8}52 & \cellcolor{tblue!16}45 & \cellcolor{tblue!8}35 & \cellcolor{tblue!18}47 & \cellcolor{tblue!2}40 & \cellcolor{tblue!18}39 & \cellcolor{tblue!6}11 \\
\spireicon{claude} & Claude Sonnet 5 {\scriptsize\textcolor{gray}{(max)}} & \cellcolor{tblue!9}37.2 & \cellcolor{tblue!2}28 & \cellcolor{tblue!15}58 & \cellcolor{tblue!12}56 & \cellcolor{tblue!11}39 & \cellcolor{tblue!9}36 & \cellcolor{tblue!11}39 & \cellcolor{tblue!7}45 & \cellcolor{tblue!9}26 & \cellcolor{tblue!4}8 \\
\spireicon{claude} & Claude Haiku 4.5 {\scriptsize\textcolor{gray}{(12k)}} & \cellcolor{tblue!5}32.6 & \cellcolor{tred!5}20 & \cellcolor{tblue!15}58 & \cellcolor{tred!5}40 & \cellcolor{tblue!14}43 & \cellcolor{tblue!7}34 & \cellcolor{tblue!8}35 & 38 & \cellcolor{tblue!5}20 & \cellcolor{tblue!2}5 \\
\spireicon{deepseek} & DeepSeek V4.1 Flash {\scriptsize\textcolor{gray}{(max)}} & \cellcolor{tblue!16}46.2 & \cellcolor{tblue!3}30 & \cellcolor{tblue!32}74 & \cellcolor{tblue!10}54 & \cellcolor{tblue!18}48 & \cellcolor{tblue!22}52 & \cellcolor{tblue!28}60 & \cellcolor{tblue!18}56 & \cellcolor{tblue!14}33 & \cellcolor{tblue!4}9 \\
\spireicon{gemini} & Gemini 3.8 Flash {\scriptsize\textcolor{gray}{(high)}} & \cellcolor{tblue!27}\runnerup{59.3} & \cellcolor{tblue!18}48 & \cellcolor{tblue!33}75 & \cellcolor{tblue!29}\best{71} & \cellcolor{tblue!34}\runnerup{67} & \cellcolor{tblue!35}\best{69} & \cellcolor{tblue!31}64 & \cellcolor{tblue!29}\runnerup{68} & \cellcolor{tblue!22}\runnerup{45} & \cellcolor{tblue!15}\best{27} \\
\spireicon{gemini} & Gemma 4 E4B & \cellcolor{tblue!5}33.4 & \cellcolor{tred!5}20 & \cellcolor{tblue!13}56 & \cellcolor{tblue!5}49 & \cellcolor{tblue!6}32 & \cellcolor{tblue!10}37 & \cellcolor{tblue!14}42 & \cellcolor{tblue!18}57 & \cellcolor{tred!5}5 & \cellcolor{tblue!1}3 \\
\spireicon{meta} & Muse Spark 1.3 {\scriptsize\textcolor{gray}{(xhigh)}} & \cellcolor{tblue!17}48.1 & \cellcolor{tblue!5}32 & \cellcolor{tblue!23}65 & \cellcolor{tblue!19}62 & \cellcolor{tblue!26}57 & \cellcolor{tblue!23}54 & \cellcolor{tblue!26}58 & \cellcolor{tblue!16}54 & \cellcolor{tblue!19}40 & \cellcolor{tblue!6}11 \\
\spireicon{nvidia} & Nemotron 3 Nano Omni & \cellcolor{tblue!2}29.1 & \cellcolor{tred!2}24 & \cellcolor{tblue!18}60 & \cellcolor{tred!1}44 & \cellcolor{tblue!3}29 & \cellcolor{tblue!4}30 & \cellcolor{tblue!10}38 & \cellcolor{tred!10}27 & \cellcolor{tred!2}10 & \cellcolor{tred!1}0 \\
\spireicon{openai} & GPT-6 Astra {\scriptsize\textcolor{gray}{(max)}} & \cellcolor{tblue!29}\best{62.2} & \cellcolor{tblue!22}\best{53} & \cellcolor{tblue!41}\best{82} & \cellcolor{tblue!26}69 & \cellcolor{tblue!36}\best{70} & \cellcolor{tblue!32}65 & \cellcolor{tblue!41}\best{76} & \cellcolor{tblue!30}\best{69} & \cellcolor{tblue!26}\best{50} & \cellcolor{tblue!15}\runnerup{26} \\
\spireicon{openai} & GPT-5.6 Sol {\scriptsize\textcolor{gray}{(max)}} & \cellcolor{tblue!26}57.9 & \cellcolor{tblue!19}\runnerup{50} & \cellcolor{tblue!37}\runnerup{78} & \cellcolor{tblue!28}\runnerup{70} & \cellcolor{tblue!26}57 & \cellcolor{tblue!33}\runnerup{66} & \cellcolor{tblue!36}\runnerup{70} & \cellcolor{tblue!24}63 & \cellcolor{tblue!22}\runnerup{45} & \cellcolor{tblue!12}22 \\
\spireicon{openai} & GPT-5.6 Terra {\scriptsize\textcolor{gray}{(max)}} & \cellcolor{tblue!18}48.4 & \cellcolor{tblue!9}37 & \cellcolor{tblue!32}74 & \cellcolor{tblue!21}64 & \cellcolor{tblue!23}54 & \cellcolor{tblue!27}59 & \cellcolor{tblue!23}54 & \cellcolor{tblue!7}45 & \cellcolor{tblue!20}41 & \cellcolor{tblue!4}8 \\
\spireicon{openai} & GPT-5.6 Luna {\scriptsize\textcolor{gray}{(max)}} & \cellcolor{tblue!17}47.9 & \cellcolor{tblue!3}30 & \cellcolor{tblue!24}66 & \cellcolor{tblue!13}57 & \cellcolor{tblue!26}57 & \cellcolor{tblue!26}58 & \cellcolor{tblue!27}59 & \cellcolor{tblue!22}61 & \cellcolor{tblue!16}35 & \cellcolor{tblue!4}8 \\
\spireicon{openai} & gpt-oss-120b {\scriptsize\textcolor{gray}{(high)}} & \cellcolor{tblue!4}31.1 & \cellcolor{tred!10}14 & \cellcolor{tblue!19}61 & \cellcolor{tblue!6}50 & \cellcolor{tblue!9}36 & \cellcolor{tblue!5}31 & \cellcolor{tblue!4}30 & \cellcolor{tred!1}37 & \cellcolor{tblue!5}19 & 2 \\
\spireicon{grok} & Grok 4.6 {\scriptsize\textcolor{gray}{(xhigh)}} & \cellcolor{tblue!17}48.0 & \cellcolor{tred!2}23 & \cellcolor{tblue!20}62 & \cellcolor{tblue!25}68 & \cellcolor{tblue!30}63 & \cellcolor{tblue!27}59 & \cellcolor{tblue!22}53 & \cellcolor{tblue!18}57 & \cellcolor{tblue!14}33 & \cellcolor{tblue!8}14 \\
\spireicon{zai} & GLM-5.3 {\scriptsize\textcolor{gray}{(max)}} & \cellcolor{tblue!15}44.7 & 26 & \cellcolor{tblue!22}64 & \cellcolor{tblue!22}65 & \cellcolor{tblue!18}47 & \cellcolor{tblue!25}56 & \cellcolor{tblue!17}46 & \cellcolor{tblue!13}51 & \cellcolor{tblue!16}35 & \cellcolor{tblue!6}12 \\
\spireicon{zai} & GLM-5.3 Flash {\scriptsize\textcolor{gray}{(max)}} & \cellcolor{tblue!13}42.3 & \cellcolor{tblue!3}30 & \cellcolor{tblue!27}69 & \cellcolor{tblue!9}53 & \cellcolor{tblue!19}49 & \cellcolor{tblue!19}49 & \cellcolor{tblue!15}44 & \cellcolor{tblue!8}46 & \cellcolor{tblue!16}35 & \cellcolor{tblue!3}6 \\
\bottomrule
\end{tabular}
\end{adjustbox}

\end{table}

Three things stand out.
First, every model is above the pick-random and pick-first baselines (26.8\% and 22.1\%), and the strongest follow the expert on about three items in five: GPT-6 Astra on 62.2\%, Gemini~3.8 Flash on 59.3\%, and GPT-5.6 Sol on 57.9\%.
The other hosted models lie between 32.6\% and 48.4\%, with the two GLM models at 44.7\% and 42.3\% and the four Claude models at the lower end (Fable~5.1 43.1\%, Opus~5 39.3\%, Sonnet~5 37.2\%, and Haiku~4.5 32.6\%).
BottledAI, at 38.3\%, is above the five open-weight models (29.1\% to 37.1\%), Claude Sonnet~5, and Claude Haiku~4.5, and below every other model.
Second, the categories differ far more than the agents do.
Event options and map nodes are followed most often (up to 82\% and 71\%), and the first option alone is right on 54\% of events.
Shop visits are followed least: no agent leaves with the expert's exact purchases on more than 27\% of visits, and sixteen of the 19 models do so on fewer than one in five.
Third, the choice of the starting blessing is near chance for most agents (26\% for uniform random choice, and 14\% to 53\% for the models), which suggests the experts' reasons for it are not ones the agents share.

\paragraph{Rest sites and elites.}
Two splits set the caution of the runs (\cref{sec:results}) against the experts on the same states.
At the 72 rest sites where the expert upgraded a card, the models rest instead on 22\% (GPT-6 Astra) to 81\% (Nemotron~3 Nano Omni) of them, and the top four of \cref{tab:main-results} on 22\% to 53\%; uniform random choice would rest on 34\%.
At the 10 rest sites where the expert rested, every model but Qwen3.8 27B (7 of 10), GLM-5.3, and Gemma~4 E4B (8 of 10 each) rests on at least 90\%.
The models therefore rest where the experts rest, and also at a fifth to four fifths of the sites where the experts upgrade.
On the 20 map choices where the expert went to an elite, GPT-5.6 Sol, GPT-6 Astra, and Gemini~3.8 Flash take a node on the expert's route on 75\% to 85\%, while Claude Fable~5.1 and Claude Opus~5 do so on 15\% and 25\%, below uniform random choice (47\%); a wrong answer here is almost never another elite.
These counts are small.
The agents play these items under the game's real names and with no memory bank, so neither the reskin nor the advice of a bank accounts for the caution.

\paragraph{Acts and characters.}
The categories asked differ between acts: all the Neow items are in Act~1, and there are no boss rewards in Act~3.
We therefore compare acts 1 to 3 on the six categories asked in all three (580 items), taking an act's score as the mean of its shares in those categories (\cref{tab:expert-breakdowns}).
Averaged over the models, agreement falls from 45.6\% in Act~1 to 43.8\% in Act~2 and 41.3\% in Act~3, while uniform random choice stays at 27\% to 28\%.
Seventeen of the 19 models fall from Act~1 to Act~3, but the fall, 2.2 points per act, is not significant ($p = 0.10$, permuting the acts of the items within each category).
Event options, map nodes, card rewards, upgrades, and shop visits are all followed less often in later acts, while rest sites are followed more often (34\%, 55\%, and 56\%).
By character, we take each model's score on a character (the mean over the nine categories) less its mean over the four characters.
Averaged over the models this is $+1.9$ points for the Ironclad, $+0.6$ for the Silent, $-0.3$ for the Defect, and $-2.1$ for the Watcher, and 15 models are above their own mean on the Ironclad against four on the Watcher; this is not significant either ($p = 0.40$).
The stronger models show the difference most (GPT-5.6 Sol $+5.2$ on the Ironclad and $-6.3$ on the Watcher, Grok~4.6 $+3.6$ and $-7.1$), while several of the weaker ones, among them Claude Haiku~4.5 and Claude Sonnet~5, do best on the Defect.

\begin{table}[t]
\centering
\caption{\textbf{Expert-decision benchmark by act and by character.}
Left: the share (\%) of items on which an agent makes the expert's choice in each of acts 1 to 3, over the six categories asked in all three acts (event, path, card reward, rest, upgrade, and shop), the mean of the six category shares; \emph{Per act} is the least-squares slope over the three acts, in points.
Right: an agent's score on each character (the mean over the nine categories) less its mean over the four characters, in points.
The $p$ values test the mean over the models, permuting the act or character of the items within each category.}
\label{tab:expert-breakdowns}
\providecommand{\spireicon}[1]{\raisebox{-0.25ex}{\includegraphics[height=1.7ex]{icons/#1.pdf}}}
\begin{adjustbox}{max width=\textwidth}
\small
\begin{tabular}{@{}c@{\hspace{3pt}}l rrrr c rrrr@{}}
\toprule
 & & \multicolumn{4}{c}{By act, 6 categories (\%)} & & \multicolumn{4}{c}{By character (points)} \\
\cmidrule(lr){3-6} \cmidrule(lr){8-11}
 & Agent & Act 1 & Act 2 & Act 3 & Per act & & Ironclad & Silent & Defect & Watcher \\
\midrule
\spireicon{qwen} & Qwen3.8 27B {\scriptsize\textcolor{gray}{(xhigh)}} & 40.3 & 39.1 & 38.3 & $-1.0$ &  & $-0.7$ & $-4.2$ & $+2.9$ & $+2.0$ \\
\spireicon{qwen} & Qwen3.6 35B-A3B & 37.9 & 34.8 & 28.0 & $-4.9$ &  & $-1.3$ & $+0.9$ & $-1.8$ & $+2.2$ \\
\spireicon{claude} & Claude Fable 5.1 {\scriptsize\textcolor{gray}{(max)}} & 45.6 & 45.1 & 40.2 & $-2.7$ &  & $+4.9$ & $-1.3$ & $+0.0$ & $-3.6$ \\
\spireicon{claude} & Claude Opus 5 {\scriptsize\textcolor{gray}{(max)}} & 42.7 & 43.1 & 38.1 & $-2.3$ &  & $+0.2$ & $-1.6$ & $+2.0$ & $-0.7$ \\
\spireicon{claude} & Claude Sonnet 5 {\scriptsize\textcolor{gray}{(max)}} & 42.6 & 34.3 & 38.6 & $-2.0$ &  & $-2.6$ & $-1.2$ & $+3.7$ & $+0.1$ \\
\spireicon{claude} & Claude Haiku 4.5 {\scriptsize\textcolor{gray}{(12k)}} & 31.9 & 34.7 & 34.8 & $+1.5$ &  & $-3.2$ & $-1.0$ & $+4.3$ & $-0.1$ \\
\spireicon{deepseek} & DeepSeek V4.1 Flash {\scriptsize\textcolor{gray}{(max)}} & 50.1 & 44.6 & 43.1 & $-3.5$ &  & $+1.8$ & $+5.3$ & $-2.2$ & $-4.9$ \\
\spireicon{gemini} & Gemini 3.8 Flash {\scriptsize\textcolor{gray}{(high)}} & 60.7 & 59.2 & 56.6 & $-2.1$ &  & $+3.8$ & $+3.8$ & $-2.9$ & $-4.7$ \\
\spireicon{gemini} & Gemma 4 E4B & 36.1 & 36.0 & 29.1 & $-3.5$ &  & $+2.1$ & $-2.3$ & $+0.3$ & $-0.1$ \\
\spireicon{meta} & Muse Spark 1.3 {\scriptsize\textcolor{gray}{(xhigh)}} & 47.5 & 44.6 & 49.4 & $+0.9$ &  & $+2.1$ & $+3.4$ & $-4.1$ & $-1.4$ \\
\spireicon{nvidia} & Nemotron 3 Nano Omni & 28.9 & 31.4 & 24.3 & $-2.3$ &  & $+1.1$ & $-1.1$ & $-2.0$ & $+2.0$ \\
\spireicon{openai} & GPT-6 Astra {\scriptsize\textcolor{gray}{(max)}} & 66.1 & 62.5 & 55.9 & $-5.1$ &  & $+4.0$ & $+0.4$ & $-2.7$ & $-1.8$ \\
\spireicon{openai} & GPT-5.6 Sol {\scriptsize\textcolor{gray}{(max)}} & 60.1 & 55.1 & 52.9 & $-3.6$ &  & $+5.2$ & $+2.1$ & $-1.0$ & $-6.3$ \\
\spireicon{openai} & GPT-5.6 Terra {\scriptsize\textcolor{gray}{(max)}} & 50.7 & 42.8 & 47.7 & $-1.5$ &  & $+4.0$ & $+2.7$ & $-4.0$ & $-2.7$ \\
\spireicon{openai} & GPT-5.6 Luna {\scriptsize\textcolor{gray}{(max)}} & 48.6 & 46.7 & 44.8 & $-1.9$ &  & $+5.4$ & $-0.8$ & $-0.3$ & $-4.3$ \\
\spireicon{openai} & gpt-oss-120b {\scriptsize\textcolor{gray}{(high)}} & 37.7 & 34.3 & 31.2 & $-3.2$ &  & $+2.2$ & $-1.8$ & $+0.4$ & $-0.9$ \\
\spireicon{grok} & Grok 4.6 {\scriptsize\textcolor{gray}{(xhigh)}} & 48.6 & 49.7 & 48.4 & $-0.1$ &  & $+3.6$ & $+4.9$ & $-1.3$ & $-7.1$ \\
\spireicon{zai} & GLM-5.3 {\scriptsize\textcolor{gray}{(max)}} & 46.6 & 49.3 & 41.6 & $-2.5$ &  & $+2.9$ & $+2.4$ & $-0.7$ & $-4.7$ \\
\spireicon{zai} & GLM-5.3 Flash {\scriptsize\textcolor{gray}{(max)}} & 44.1 & 44.4 & 41.4 & $-1.3$ &  & $+0.3$ & $-0.1$ & $+3.0$ & $-3.2$ \\
\midrule
 & Mean over the models & 45.6 & 43.8 & 41.3 & $-2.2$ &  & $+1.9$ & $+0.6$ & $-0.3$ & $-2.1$ \\
 & \textcolor{gray}{$p$ (permutation)} &  &  &  & \textcolor{gray}{0.10} &  &  &  & \textcolor{gray}{0.40} &  \\
\spireicon{random} & Uniform random choice (expected) & 28.1 & 27.0 & 27.2 &  &  & $+0.5$ & $-0.5$ & $-0.5$ & $+0.5$ \\
\bottomrule
\end{tabular}
\end{adjustbox}

\end{table}

\paragraph{Agreement with experts and success in runs.}
\Cref{fig:expert-runs} sets each model's agreement with the experts against its mean acts cleared and mean floor reached in \cref{tab:main-results}, for the 19 models of \cref{tab:expert-decisions}.
The two go together, but loosely: Spearman's $\rho$ is $0.69$ with mean acts cleared and $0.72$ with mean floor, against $0.93$ and $0.95$ for puzzle accuracy (\cref{app:puzzles-runs}).
Claude Fable~5.1 and Claude Opus~5, which clear the most and the third-most acts, agree with the experts no more often than models that clear far fewer.
By category (\cref{fig:expert-categories-runs}), the choice of the card to upgrade ($\rho = 0.87$) and shop visits ($0.80$) follow run performance most closely, then the three reward categories ($0.59$ to $0.66$).
Map nodes and events follow it less ($0.52$ and $0.54$), and rest sites least ($0.35$).

\paragraph{Model families.}
Within a family, agreement with the experts orders the models as their runs do: GPT-6 Astra, GPT-5.6 Sol, GPT-5.6 Terra, and GPT-5.6 Luna (62.2\%, 57.9\%, 48.4\%, and 47.9\%), and Claude Fable~5.1, Opus~5, Sonnet~5, and Haiku~4.5 (43.1\%, 39.3\%, 37.2\%, and 32.6\%).
Between the two families there is an offset.
Pairing their models in that order, the OpenAI model follows the experts more often in 34 of the 36 comparisons of a category; the other two, rest sites and shop visits for GPT-5.6 Terra against Claude Sonnet~5, are ties.
GPT-5.6 Luna, which clears 0.10 acts in \cref{tab:main-results}, follows the experts more often than Claude Fable~5.1, which clears 1.60, in seven of the nine categories.
The two Claude models among the top four therefore reach a comparable depth in runs to the two OpenAI models while choosing less like the experts, and this offset is much of why agreement follows run performance less closely ($\rho = 0.69$) than puzzle accuracy does (\cref{app:puzzles-runs}).
The gap is as large in card rewards, boss relics, and events (14, 22, and 16 points on average over the four pairs) as in the categories that turn on risk (map nodes and rest sites: 16 and 18 points), so a preference for caution does not account for all of it.
The items are played under the game's real names, so part of it may be knowledge of what expert players choose; playing the items under the reskin would test this, and we have not done so.

\begin{figure}[t]
\centering
\includegraphics[width=\linewidth]{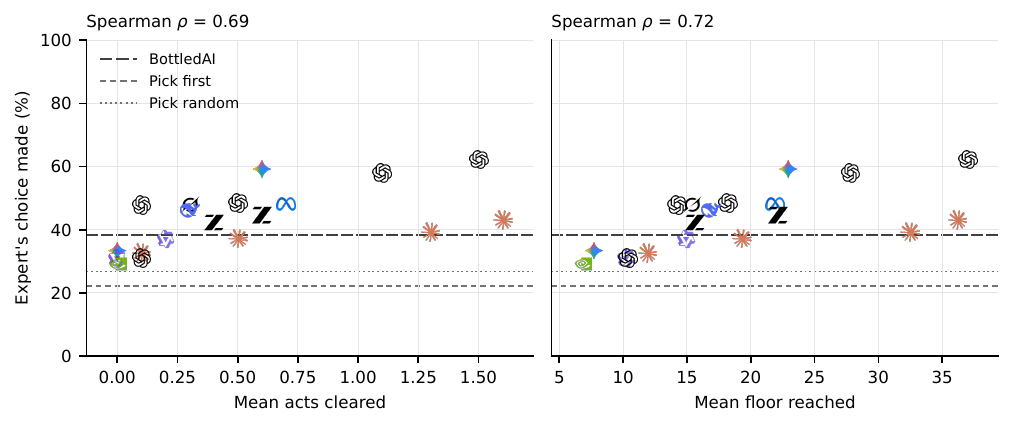}
\caption{\textbf{Agreement with the experts against performance in runs.}
Each point is one agent, drawn as its provider's icon: its mean acts cleared (left) and mean floor reached (right) in \cref{tab:main-results} against the mean over the nine categories of the share of items on which it makes the expert's choice (\cref{tab:expert-decisions}).
The dashed lines are the scores of BottledAI and of the two baseline policies.}
\label{fig:expert-runs}
\end{figure}

\begin{figure}[t]
\centering
\includegraphics[width=\linewidth]{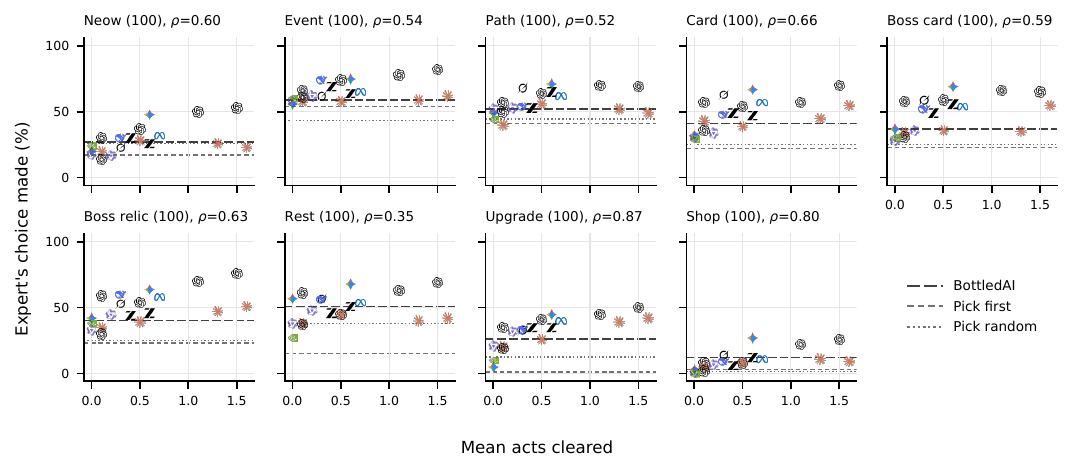}
\caption{\textbf{Agreement with the experts by decision category against mean acts cleared.}
One panel per category of \cref{tab:expert-decisions}, with the number of items in brackets; each point is one agent, drawn as its provider's icon, and $\rho$ is Spearman's rank correlation.
The dashed lines are the scores of BottledAI and of the two baseline policies in that category.}
\label{fig:expert-categories-runs}
\end{figure}

\end{document}